\documentclass[10pt,twocolumn,letterpaper]{article}

\usepackage{cvpr}              

\definecolor{cvprblue}{rgb}{0.21,0.49,0.74}
\usepackage[pagebackref,breaklinks,colorlinks,allcolors=cvprblue]{hyperref}
\usepackage{array}
\usepackage{adjustbox}
\def\paperID{10358} 
\def\confName{CVPR}
\def\confYear{2025}

\title{SCSA: A Plug-and-Play Semantic Continuous-Sparse Attention for Arbitrary Semantic Style Transfer}

\author{Chunnan Shang\\
Zhejiang University\\
{\tt\small chunnan.22@intl.zju.edu.cn}
\\
Hongwei Wang\\
Zhejiang University\\
{\tt\small hongweiwang@intl.zju.edu.cn}
\and 
Zhizhong Wang\footnotemark[2]\\
Zhejiang University\\
{\tt\small endywon@zju.edu.cn }
\\
Xiangming Meng\footnotemark[2]\\
Zhejiang University\\
{\tt\small xiangmingmeng@intl.zju.edu.cn}
}

\begin{document}
\maketitle

\renewcommand{\thefootnote}{\fnsymbol{footnote}}
\footnotetext[2]{Corresponding authors.}
\renewcommand{\thefootnote}{\fnsymbol{footnote}}

\begin{abstract}
Attention-based arbitrary style transfer methods, including CNN-based, Transformer-based, and Diffusion-based, have flourished and produced high-quality stylized images. However, they perform poorly on the content and style images with the same semantics, i.e., the style of the corresponding semantic region of the generated stylized image is inconsistent with that of the style image. We argue that the root cause lies in their failure to consider the relationship between local regions and semantic regions. To address this issue, we propose a plug-and-play semantic continuous-sparse attention, dubbed SCSA, for arbitrary semantic style transfer---each query point considers certain key points in the corresponding semantic region. Specifically, semantic continuous attention ensures each query point fully attends to all the continuous key points in the same semantic region that reflect the overall style characteristics of that region; Semantic sparse attention allows each query point to focus on the most similar sparse key point in the same semantic region that exhibits the specific stylistic texture of that region. By combining the two modules, the resulting SCSA aligns the overall style of the corresponding semantic regions while transferring the vivid textures of these regions. Qualitative and quantitative results prove that SCSA enables attention-based arbitrary style transfer methods to produce high-quality semantic stylized images. The codes can be found in 
\href{https://github.com/scn-00/SCSA}{https://github.com/scn-00/SCSA}.
\end{abstract}    
\section{Introduction}
\label{sec:intro}

\begin{figure}
\centering
\resizebox{0.415\textwidth}{!}{
\setlength{\tabcolsep}{0.05cm} 
\begin{tabular}{cccc}
\footnotesize Content Inputs & \footnotesize Style Inputs & \footnotesize SANet & \footnotesize SANet + SCSA
\\
\includegraphics[width=0.22\linewidth]{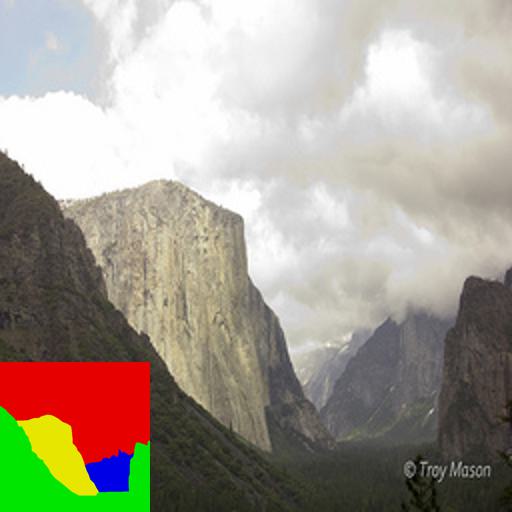}& \includegraphics[width=0.22\linewidth]{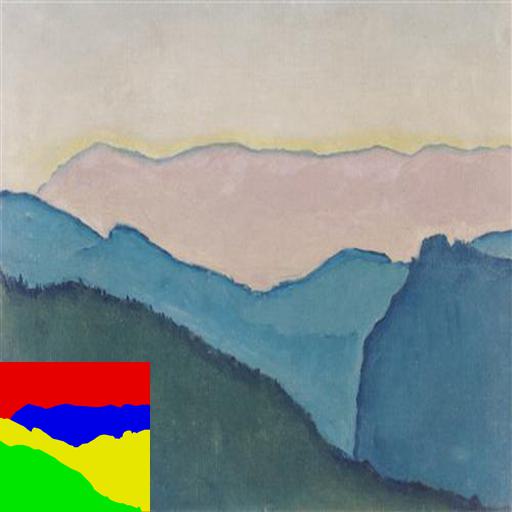}  & \includegraphics[width=0.22\linewidth]{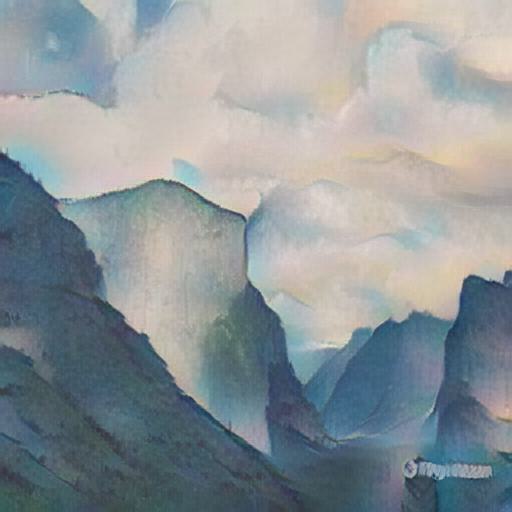} &
\includegraphics[width=0.22\linewidth]{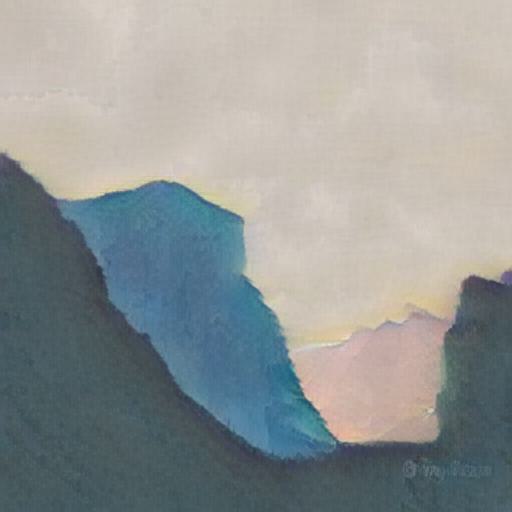} 
\\
\footnotesize Content Inputs & \footnotesize Style Inputs & \footnotesize StyTR$^2$ & \footnotesize StyTR$^2$ + SCSA \\
\includegraphics[width=0.22\linewidth]{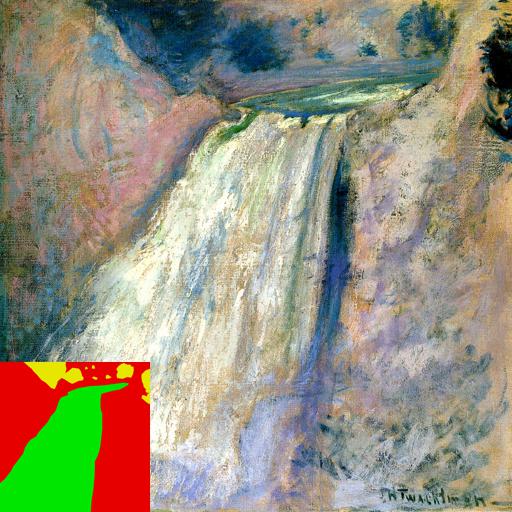}& \includegraphics[width=0.22\linewidth]{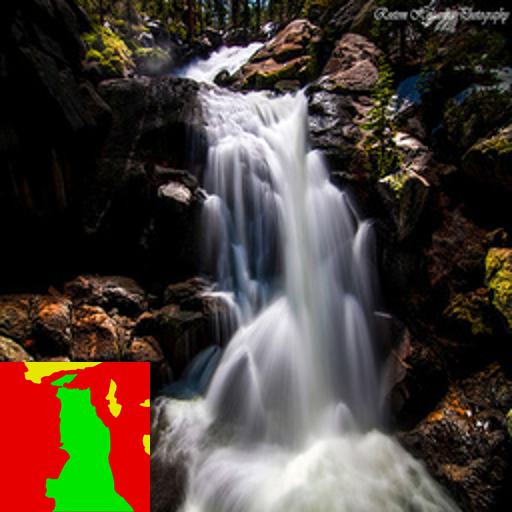}  & \includegraphics[width=0.22\linewidth]{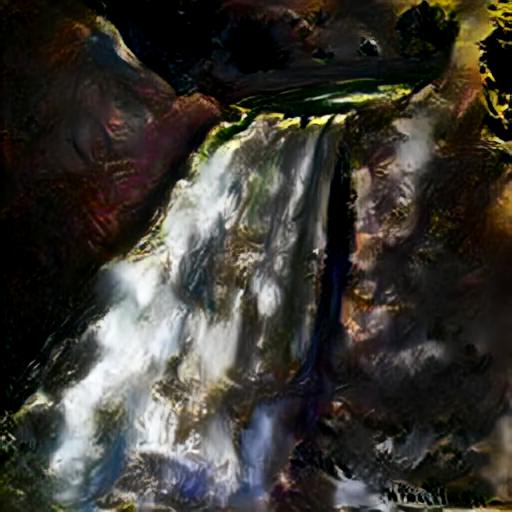} &
\includegraphics[width=0.22\linewidth]{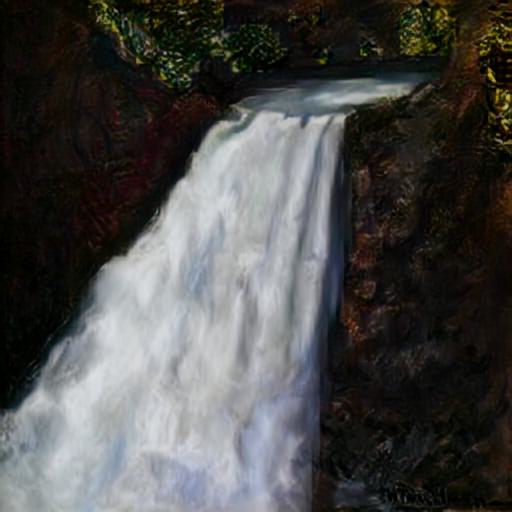} \\
\footnotesize Content Inputs & \footnotesize Style Inputs & \footnotesize StyleID & \footnotesize StyleID + SCSA \\
\includegraphics[width=0.22\linewidth]{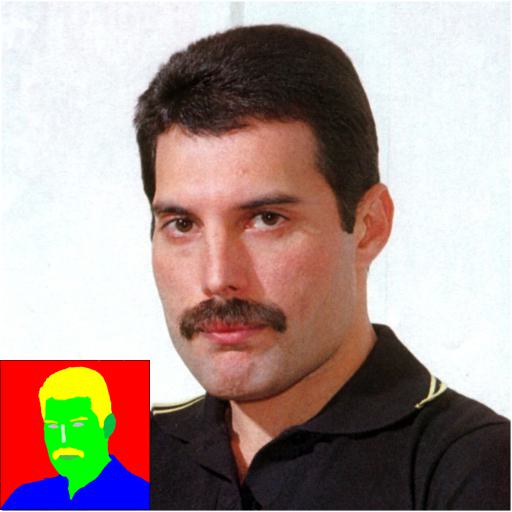}& \includegraphics[width=0.22\linewidth]{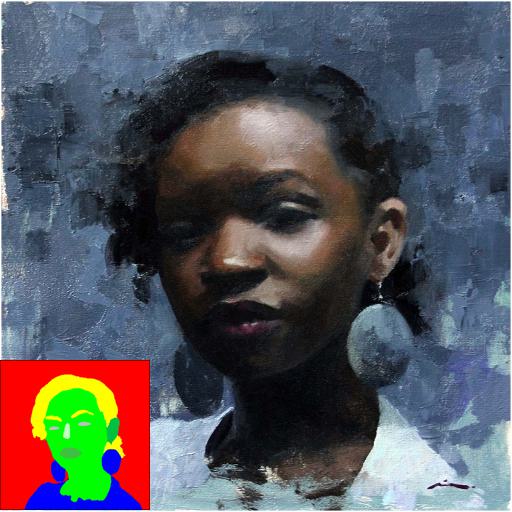}  & \includegraphics[width=0.22\linewidth]{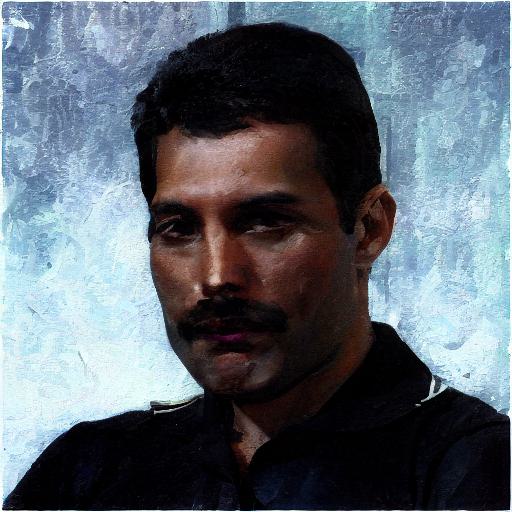} &
\includegraphics[width=0.22\linewidth]{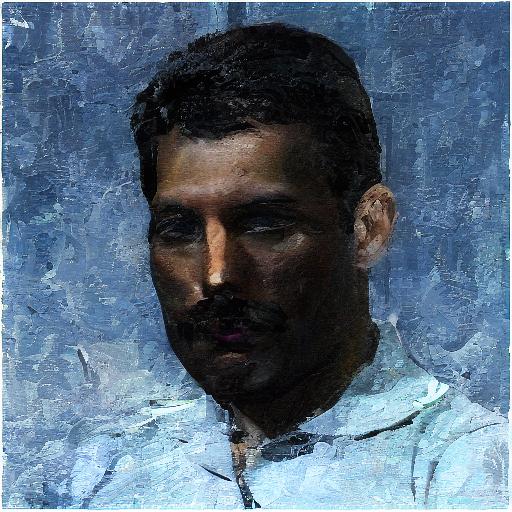} 
\\
\end{tabular}
}
\caption{Comparisons of the Attn-AST approaches--CNN-based SANet~\cite{park2019arbitrary}, Transformer-based StyTR$^2$~\cite{deng2022stytr2}, and Diffusion-based StyleID~\cite{chung2024style}--without and with our SCSA. The stylized images generated by the Attn-AST approaches exhibit style discontinuity in adjacent regions within the identical semantic regions (e.g., the background in the $1st$ row), style inconsistency between corresponding semantic (e.g., the cloth color in the $3rd$ row), and fewer textures (e.g., the block textures in the $3rd$ row).}
\label{fig:8}
\end{figure}

Initially, style transfer~\cite{gatys2016image} was proposed to preserve the content of a specified content image while transferring the style from a specified style image by iteratively optimizing the generated stylized image. However, its excessive slowness prevents practical application. Given the need for efficiency, effectiveness, and generalization, Arbitrary Style Transfer (AST)~\cite{li2018learning,an2021artflow,liu2021adaattn,chen2023artfusion,chung2024style} has become a highly prominent research focus in the style transfer field. AST seeks to generate a new image that replicates the content of an arbitrary content image while adopting the style of an arbitrary style image. In spite of many recent studies aiming to advance the AST effect by incorporating novel knowledge~\cite{wang2022aesust,huang2023quantart,tarres2024parasol} and innovative constraints~\cite{wu2022ccpl,zhang2022domain,wang2023stylediffusion}, the core of AST continues to reside in feature transformation modules, which remains indispensable for all approaches. Hence, the existing AST approaches can be classified into two categories, statistics-based methods~\cite{huang2017arbitrary,li2017universal,zhang2022exact,chen2024artadapter} and attention-based methods~\cite{park2019arbitrary,chen2021artistic,liu2021adaattn,ma2023rast,chung2024style,deng2022stytr2}.

\begin{figure*}[t]
  \centering
   \includegraphics[width=1\linewidth]{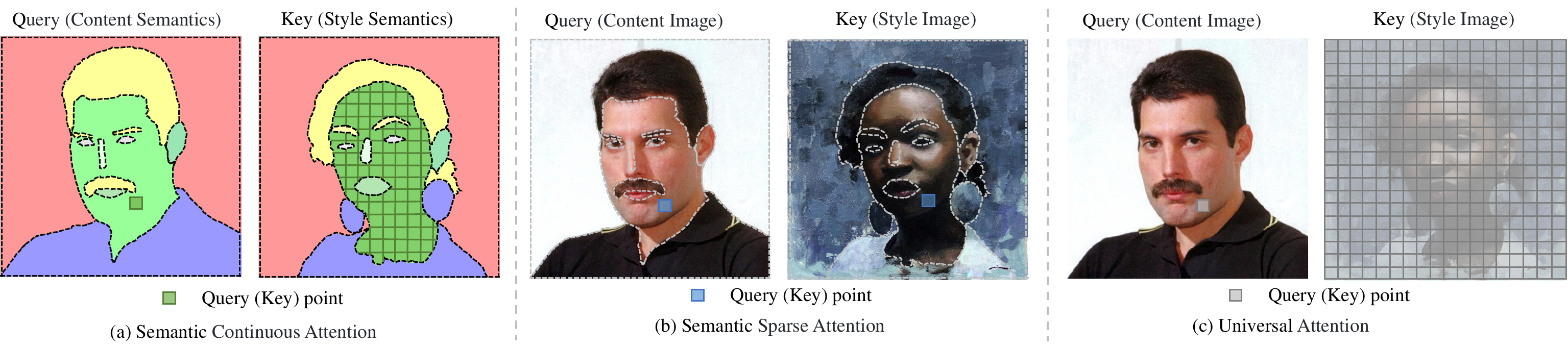}

   \caption{Comparison between Semantic Continuous-Sparse Attention (SCSA) and Universal Attention (UA). SCSA includes two parts: (a) Semantic Continuous Attention (SCA): The query point of the content semantic map features can match all continuous key points of the style semantic map features in the same semantic region. Therefore, SCA can fully account for the overall stylistic characteristics (e.g., color and texture) of regions with the same semantics; (b) Semantic Sparse Attention (SSA): The query point of the content image features can match the most similar sparse key point of the style image features in the same semantic region. Hence, SSA can intently concentrate on the specific stylistic texture of regions with the same semantics. In contrast, (c) Universal Attention (UA)~\cite{park2019arbitrary}: The query point of the content image features pays attention to all key points of the style image features. As a result,  UA fails to accurately convey the intricate overall stylistic characteristics and specific textures of regions that share identical semantics.
}
   \label{fig:1}
\end{figure*}

Attention-based Arbitrary Style Transfer (Attn-AST) approaches, leveraging the ability to capture global information and focus on the related features of the attention mechanism~\cite{bahdanau2014neural}, preserve the content of the content image while transferring the style that is relevant to the content of the content image from the style image. They encompass CNN-based~\cite{park2019arbitrary,liu2021adaattn,ma2023rast,wang2022aesust}, Transformer-based~\cite{deng2022stytr2,zhang2024s2wat,wu2021styleformer,hong2023aespa}, and Diffusion-based methods~\cite{chung2024style,xiang2024training,susladkar2024d2styler}. Commonly, Attn-AST methods first establish a holistic relationship, an attention map, between each content query point of the content image features and all style key points of the style image features. Then, Attn-AST methods get the transferred style for each content point by weighting all the style points, where the weights are the {\it softmax} function outputs of a naturally calculated attention map of content and style points.

Although Attn-AST methods have thrived and produced high-quality stylized images due to the capacity mentioned above for global capture and focus, they perform poorly on the content and style image with the same semantics. As shown in Fig.~\ref{fig:8}, the style of the corresponding semantic region of the stylized image generated by Attn-AST approaches exhibits a style inconsistent with that of the style image. We argue that the root cause lies in their failure to consider the relationship between local regions and semantic regions. (1) During the formulation of the attention map, Attn-AST methods bluntly and simplistically consider the relationship between each query point and all key points, as shown in Fig.~\ref{fig:1}~(c), without adequately accounting for the semantic regions to which these points belong. This oversimplification may cause Attn-AST methods to disproportionately emphasize structurally similar points across different semantic regions, resulting in artifacts or inaccurate style~\cite{liu2021adaattn,park2019arbitrary}, e.g., the $2nd$ row of Fig.~\ref{fig:8}. (2) Attn-AST approaches fundamentally depend on structural principles for generating a global style, as they utilize content and style image features. The subtle disparity in content structure details across adjacent regions can cause different stylization effects. Thus, the stylized images often lack stylistic continuity across adjacent regions within the semantic regions, e.g., the background in the $1st$ row of Fig.~\ref{fig:8}. (3) Attn-AST approaches acquire the transferred style by weighting all the style points. However, the most accurate representation of the original stylistic texture resides in the encoded single points. The weighted style struggles to preserve the original stylistic texture, leading to a lack of vividness, e.g., the block textures in the $3rd$ row of Fig.~\ref{fig:8}.

To maintain the benefits of Attn-AST methods while addressing their existing issues, we propose a plug-and-play semantic continuous-sparse attention, dubbed SCSA, for arbitrary semantic style transfer. The key insight of SCSA is that, based on the interpretability of the attention mechanism, each query point considers certain key points in the corresponding semantic region. Specifically, SCSA consists of two parts: Semantic Continuous Attention (SCA) and Semantic Sparse Attention (SSA). (1) SCA: Semantic continuous attention ensures each query point fully attends to {\it all the continuous key points} in the same semantic region by using content and style {\it semantic map features} as query and key, respectively. Thus, SCA can achieve the overall style characteristics of the corresponding semantic regions of the style image to generate the stylized image with stylistic continuity across adjacent regions within the semantic regions. (2) SSA: Semantic sparse attention allows each query point to focus on the {\it most similar sparse key point} in the same semantic region by using content and style {\it image features} as query and key, respectively. Hence, SSA can obtain the specific stylistic texture of the corresponding semantic regions of the style image to produce the stylized image with vivid stylistic textures. The intuitive comparison between our SCSA and universal attention can be found in Fig.~\ref{fig:1}.

By seamlessly integrating SCSA into any Attn-AST approach {\it without training}, we enable arbitrary semantic style transfer while preserving and even enhancing the original stylization effect. To our knowledge, we are the first work to extend attention-based arbitrary style transfer to arbitrary semantic style transfer in a plug-and-play way. 

In brief, our contributions are divided into three parts:
\begin{itemize}[leftmargin=2em]
    \item We reveal that Attn-AST methods struggle with content and style images with the same semantics and identify the root causes for their subpar performance.
    \item We propose a semantic continuous-sparse attention, dubbed SCSA, that can extend Attn-AST to arbitrary semantic style transfer in a plug-and-play manner.
    \item We conduct extensive experiments and comparisons to demonstrate the effectiveness and generalization of SCSA, enabling Attn-AST methods (CNN-based, Transformer-based, and Diffusion-based approaches) to perform arbitrary semantic style transfer while preserving and even enhancing the original stylization.
\end{itemize}

\section{Related Work}
\label{sec:related work}

Since the pioneering research by Gatys et al. ~\cite{gatys2016image}, style transfer has rapidly ascended to prominence as an influential topic. However, it generates images through iterative optimization, posing challenges for practical applications. To address this, Johnson et al.~\cite{johnson2016perceptual} train a separate feed-forward network for each style for rapid stylization, but it proves prohibitively expensive. To concurrently meet the demands of efficiency, quality, and generalization, Arbitrary Style Transfer (AST) emerged and rapidly flourished. The existing AST approaches can be roughly divided into two groups: statistics-based and attention-based methods.

{\bf Statistics-based Methods.} Statistics-based AST methods~\cite{zhang2022domain,wu2022ccpl,an2021artflow,li2018learning,jing2020dynamic} regard style as the statistical traits inherent in feature distribution. Specifically, they achieve stylized features by aligning the feature statistics of the content image with those of the style image, making them a global style transfer approach. The two most representative approaches are AdaIN~\cite{huang2017arbitrary} and WCT~\cite{li2017universal}, which match the first-order and second-order statistics of features, respectively. To more comprehensively capture the stylistic characteristics of features, higher-order statistical methods, such as CMD~\cite{kalischek2021light} and EFDM~\cite{zhang2022exact}, have been continually explored and refined. Expanding on the above foundation, semantic style transfer~\cite{gatys_controlling_2017,lu_decoder_2017,liao2022semantic,zhao2020automatic,kim2023bridging,lu_closed-form_2019} has been studied by aligning feature statistics within corresponding semantic regions. STROTSS~\cite{kolkin2019style} employs the Earth Movers Distance~\cite{kusner2015word} to reduce the feature distribution difference within semantic regions and self-similarity to preserve the semantics and spatial structure. MAST~\cite{huo2021manifold} aligns the manifold distributions of semantic regions to facilitate semantic style transfer while preserving the content structure.

Since feature statistics only provide a global representation, the stylized images they generate often lack fine local textures. To address this, we propose SCSA, which could yield a coherent overall style across adjacent regions by overlooking structure details, while producing vivid textures by emphasizing them via the attention mechanism.

{\bf Attention-based Methods.} Attention-based AST (Attn-AST) methods define style as the weighted convergence of all style features. Exactly, Attn-AST methods obtain the stylized feature of each content feature by weighting all the style features, where weights are derived from the correlation between the content feature and all style features obtained through the attention mechanism, making it a local structure style transfer. Attn-AST approaches mainly include CNN-based methods~\cite{park2019arbitrary,liu2021adaattn}, Transformer-based methods~\cite{deng2022stytr2}, and Diffusion-based methods~\cite{chung2024style}. SANet~\cite{park2019arbitrary} is the pioneering work to integrate attention into AST, employing content features as query and style features as key and value to obtain stylized features via the attention mechanism. StyTr$^2$~\cite{deng2022stytr2} integrates the outputs of the content and style transformer~\cite{vaswani2017attention} encoders into the transformer feature fusion module, obtaining stylized features. StyleID~\cite{chung2024style} injects the content query, style key, and style value, derived from DDIM inversion~\cite{song2020denoising}, into certain self-attention layers of U-Net~\cite{rombach2022high} to merge style, achieving stylized features through multiple inversion steps. Even though they continually improve the quality of stylized images, their performance remains poor on content and style images with identical semantics. In contrast, patch-based methods~\cite{chen2016fast,yang2023zero,champandard2016semantic,hamazaspyan2023diffusion,wang2020glstylenet} achieve semantic style transfer by swapping the styles of patches with the most similar content structures. When the patch size in patch-based methods is 1, they can be roughly considered attention-based methods. Deep-Image-Analogy (DIA)~\cite{liao2017visual} introduces nearest-neighbor field search to tackle the challenge of patch matching for images, which are visually distinct yet semantically similar. GLStyleNet~\cite{wang2020glstylenet} uses multi-level pyramid features for semantic optimization via patch matching, iteratively creating stylized images that meet semantic criteria. TR~\cite{wang2022texture} utilizes the patch match to achieve precise semantic-guided and structure-preserving texture transfer. 

As the above methods target local structural similarity, subtle changes in the content structure of adjacent regions can result in stylization discontinuity in the identical semantic region. To mitigate this concern, we introduce SCSA, which provides more stable and continuous stylization by fully comprehending semantics and transfers vivid stylistic textures by focusing on local structure. Most importantly, it can be easily integrated into existing frameworks, while other methods are either designed for specific networks or require multiple iterations for optimization.




\section{Method}

\begin{figure}[t]
  \centering
   \includegraphics[width=1\linewidth]{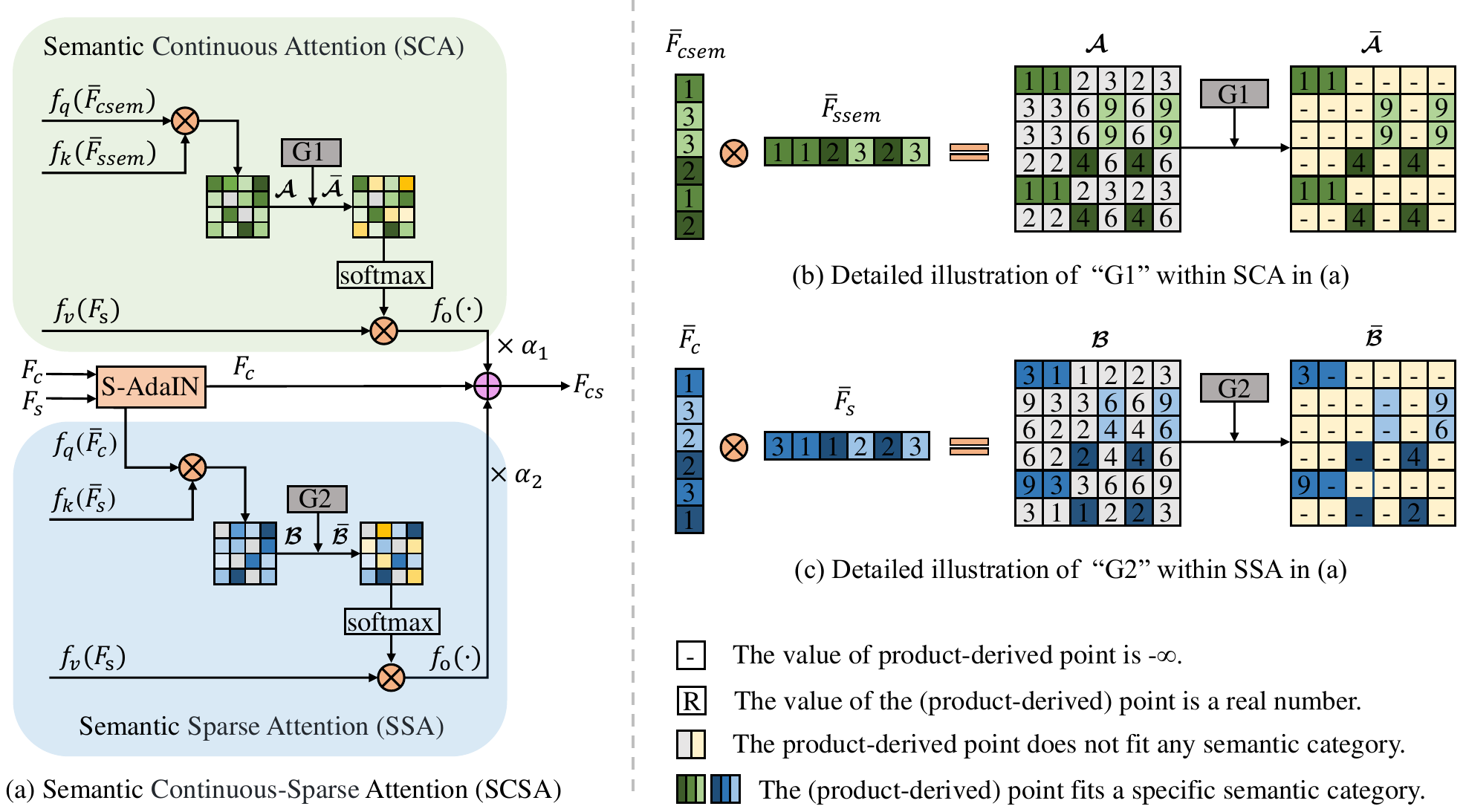}

   \caption{Detailed Procedure of Semantic Continuous-Sparse Attention (SCSA). In (a), S-AdaIN denotes that AdaIN~\cite{huang2017arbitrary} is applied individually for each semantic region. It initializes $F_c$, by matching the feature statistics of corresponding semantic regions in $F_c$ and $F_s$, to ensure that $F_c$ has semantically aligned color style information to some extent. Subsequently, $F_c$ can provide a more accurate query $f_q(\bar{F}_c)$ and more pure content structure features with less interference from the original color style. SCSA in (a) includes Semantic Continuous Attention (SCA) and Semantic Sparse Attention (SSA) two parts. “G1” in SCA sets the value of the specific product point generated by multiplying two points from different semantic categories to negative infinity. As shown in (b), only the product of points with the same semantics is retained. Therefore, SCA can fully account for the overall stylistic characteristics of regions with the same semantics. “G2” in SSA retains only the maximum value of the product of a specified query point with all key points in the same semantic region,  while all other values are set to negative infinity, as shown in (c). Hence, SSA can intently concentrate on the specific stylistic texture of the most similar structure in the regions with the same semantics. After SCSA, the generated $F_{cs} $from (a) not only reflects the overall stylistic characteristics of the semantic region corresponding to $F_s$ but also captures the specific textures of that semantic region. Meanwhile, $F_{cs}$  also includes the content structure of $F_c$. 
}
   \label{fig:2}
\end{figure}

\subsection{Semantic Continuous-Sparse Attention}

{\bf Revisit Universal Attention in Attn-AST.} Attn-AST approaches feed the encoded content feature and style feature into Universal Attention (UA), constructing intricate connections between various positions within the two feature sequences by directly utilizing the self-attention~\cite{bahdanau2014neural} computational paradigm, to capture the long-range local textures relationship. Typically, given the encoded content features $F_c$ and style features $F_s$, the query (Q), key (K), and value (V) of UA can be obtained as follows:
\begin{equation}
  Q=f_q(\bar{F}_c), \; K=f_k(\bar{F}_s), \; V=f_v(F_s),
\label{eq:1}
\end{equation}
where $\bar{F}_c$ and $\bar{F}_s$ denote the normalized $F_c$ and $F_s$, respectively. $f_q$, $f_k$, and $f_v$ are the project networks for getting query, key, and value in UA of the given Attn-AST model. 

After that, the attention map $S$, which represents a holistic relationship between each content query point of the content features and all style key points of the style features, can be obtained as follows: 
\begin{equation}
  S=Q^{\mathsf{T}} \otimes K,
\end{equation}
where $\otimes$ denotes matrix multiplication.

Finally, the stylized features $F_{cs}$ can be calculated as:
\begin{equation}
  F_{cs} = f_o(softmax(S) \otimes V) + F_c,
\end{equation}
where $f_o$ is the output project networks in UA of the specified Attn-AST model. 

Although Attn-AST methods utilizing UA can achieve high-quality stylization results, they still underperform on content and style images with identical semantics due to inherent limitations, as detailed analysis in Sec.~\ref{sec:intro}.

{\bf Semantic Continuous Attention.} To resolve the problem of the style discontinuity in adjacent regions within the identical semantic regions of the stylized image, as well as style inconsistency between corresponding semantic regions of stylized and style images, as analyzed in Sec.~\ref{sec:intro}, we propose semantic continuous attention, termed SCA, as shown in Fig.~\ref{fig:1}~(a) and detailed in Fig.~\ref{fig:2}. This method disregards the image structures while considering the relationship between local regions and semantic regions.

In particular, SCA first takes the encoded content semantic map features $F_{csem}$ and style semantic map features $F_{ssem}$ as inputs for the query ($Q_1$) and key ($K_1$), respectively, while the encoded style image features $F_{s}$ are used as inputs for the value ($V_1$):
\begin{equation}
  Q_1=f_q(\bar{F}_{csem}), \; K_1=f_k(\bar{F}_{ssem}), \; V_1=f_v(F_s),
\end{equation}
where $\bar{F}_{csem}$ and $\bar{F}_{ssem}$ represent the normalized form of $F_{csem}$ and $F_{ssem}$, respectively. $f_q$, $f_k$, and $f_v$ are the same as those in Eq.~\ref{eq:1}. 

Then, the preliminary attention map $\mathcal{A}$ can be obtained:
\begin{equation}
  \mathcal{A}=Q_1^{\mathsf{T}} \otimes K_1,
\end{equation}
where $\otimes$ denotes matrix multiplication. Since $F_{csem}$ and $F_{ssem}$ do not encompass the content structure of the content and style images, after attention computation, the correlations between content points within the same semantic region and all style points are identical. That is if the content query points $i$ and $j$ belong to the same semantic category, then $\mathcal{A}_{im} = \mathcal{A}_{jm} $, where $m$ represents any style key point. Consequently, the issue of stylization inconsistency in adjacent areas of the same semantic region, stemming from structural differences, is effectively addressed.

To make SCA focus on regions with the same semantics to transfer a coherent overall style, we modulate the attention map $\mathcal{A}$ with operation $G_1$, setting attention weights of two points from different semantic categories to negative infinity and those within the same category unchanged:
\begin{equation}
\begin{aligned}
& \qquad  \mathcal{\bar{A}}  = G_1(\mathcal{A}),\\
G_1(\mathcal{A}_{qk}) &=
\begin{cases} 
 \mathcal{A}_{qk}, \, \text{if } q \in c  \text{ and } k \in c\\
-\infty,   \, \text{ otherwise }
 \end{cases},
\end{aligned}
\end{equation}
where $q$ and $k$ represent the query and key point of $Q_1$ and $K_1$, respectively. $c$ is the semantic category. $\mathcal{\bar{A}}$ is the new attention map that captures the relationship exclusively between content query points and style key points of the same semantic category. This is visually shown in Fig.~\ref{fig:2} (b).

Ultimately, we can obtain the stylized features $F_{sca}$:
\begin{equation}
  F_{sca} = f_o(softmax( \mathcal{\bar{A}}) \otimes V_1),
\end{equation}
where $f_o$ is the same as that in Eq.~\ref{eq:1}. Following the above operation, the $softmax$ function could ensure that the weights of content and style points in different semantic regions become $0$, while the weights of different content points and all style points in the same semantic region are equal. Thus, each query point can match all continuous key points in the same semantic region. SCA can fully account for the overall stylistic characteristics (e.g., color and texture) of the same semantic regions in the style image while maintaining style consistency in adjacent areas of the identical semantic region in the stylized image.

\begin{figure*}[t]
  \centering
   \includegraphics[width=0.98\linewidth]{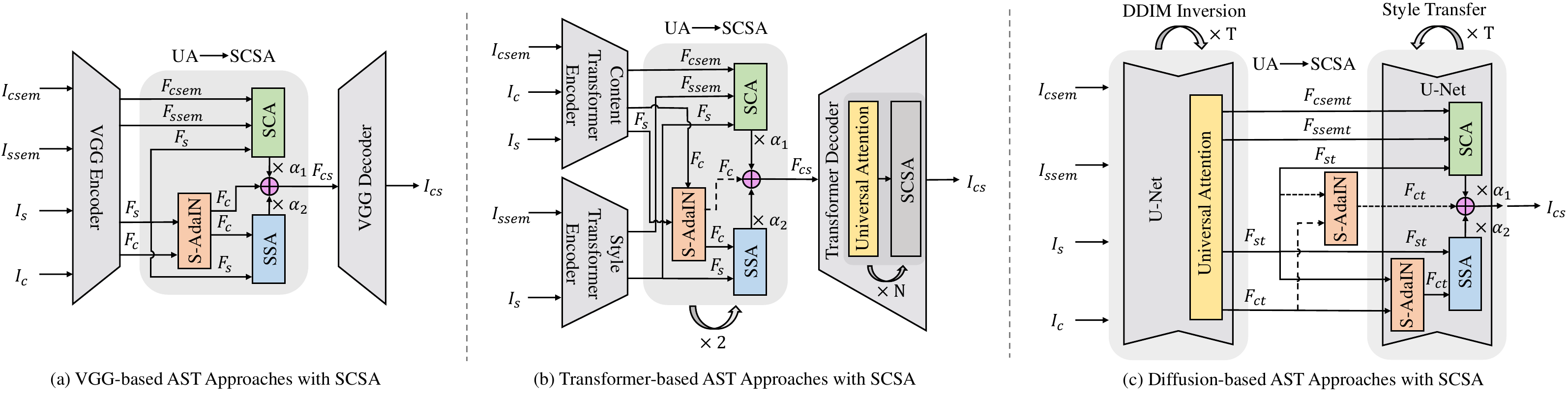}

   \caption{The overall frameworks of Attn-AST approaches--CNN-based, Transformer-based, and Diffusion-based methods--with semantic continuous-sparse attention (SCSA). SCA depicts semantic continuous attention. SSA denotes semantic sparse attention. S-AdaIN is semantic adaptive instance normalization. UA expresses universal attention in the Attn-AST methods. The dashed line in (b) shows encoded content features $F_c$ are only used in the first use of the feature transformation module, and the output features $F_{cs}$ are used as new content features in subsequent transformation. The dashed line in (c) indicates S-AdaIN is used only when $t$ is the maximum time step $T$.
}
   \label{fig:3}
\end{figure*}

{\bf Semantic Adaptive Instance Normalization.} To provide more pure content structure features with less interference from the original color style and a more accurate query for computing the relationships of local structures in corresponding semantic regions of the content and style images, we apply AdaIN~\cite{huang2017arbitrary} to the encoded content image features $F_c$ and style image features $F_s$ for each semantic region:
\begin{equation}
\begin{aligned}
I_{csem} \xrightarrow{KM} M_c^{i}, \,\, I_{ssem} &\xrightarrow{KM} M_s^{i}, \,\, i=1,...,N,\\
F_{c}^{i} = F_{c}^{i} \odot M_c^{i}, &\quad F_s^{i} = F_s^{i} \odot M_s^{i},\\
F_{c}^i = \frac{F_{c}^i-\mu(F_{c}^i)}{\sigma(F_{c}^i)}& \times \sigma(F_s^i) + \mu(F_s^i),
\label{eq:8}
\end{aligned}
\end{equation}
where $KM$ refers to K-Means clustering~\cite{lloyd1982least}. $M_c^{i}$ and $M_s^{i}$ are the masks containing only $0$ and $1$. $F_c^{i}$ and $F_s^{i}$ represent the features belong to corresponding semantic category $i$. Then, we obtain the features $F_{c}=\sum_{i} F_{c}^i$ whose overall styles of relevant semantic regions align with those of $F_s$.

{\bf Semantic Sparse Attention.} To preserve the original vivid stylistic texture of the style image and address the issue of style inconsistency between corresponding semantic regions of stylized and style images, as analyzed in Sec.~\ref{sec:intro}, we introduce semantic sparse attention, known as SSA, as shown in Fig.~\ref{fig:1}~(b) and Fig.~\ref{fig:2}. This method focuses solely on preserving the style texture with the highest weights, rather than weighting all the style textures, while considering the relationship between local and semantic regions.

SSA first employs the transformed content features $F_c$ as inputs for the query ($Q_2$) and the encoded style image features $F_s$ as inputs for both the key ($K_2$) and value ($V_2$):
\begin{equation}
  Q_2=f_q(\bar{F}_{c}), \; K_2=f_k(\bar{F}_{s}), \; V_2=f_v(F_s),
\end{equation}
where $\bar{F}_{c}$ and $\bar{F}_{s}$ represent the normalized $F_{c}$ and $F_{s}$, respectively. $f_q$, $f_k$, and $f_v$ are the same as those in Eq.~\ref{eq:1}. 

Then, the initial attention map $\mathcal{B}$ of SSA can be obtained:
\begin{equation}
  \mathcal{B}=Q_2^{\mathsf{T}} \otimes K_2,
\end{equation}
where $\otimes$ denotes matrix multiplication. Since $F_c$ and $F_s$ capture the structures of the content and style images, $\mathcal{B}$ can consider the feature correlations based on local structures.

We claim that the most accurate representation of the original stylistic texture resides in the encoded discrete feature point and the weighted style feature points struggle to preserve the original vivid stylistic textures. Thus, to get accurate and vivid textures, we adjust attention map $\mathcal{B}$ with operation $G_2$ to target the discrete feature point within related regions, setting the only maximum attention weight of a content query point and all style key points in the same semantic category unchanged and others to negative infinity:
\begin{equation}
\begin{aligned}
 & \qquad\qquad \mathcal{\overline{B}} = G_2(\mathcal{B}),\\
 G_{2}(\mathcal{B}_{qk}) =&
\begin{cases} 
 \mathcal{B}_{qk}, \, 
 \begin{aligned}
 &\text{if }\mathcal{B}_{qk}=max(\mathcal{B}_{qn}), \\
 & n\in \{1,...,len(K_2)\},\, q,k,n \in c 
 \end{aligned}\\
-\infty,   \, \text{otherwise}
 \end{cases},
\end{aligned}
\end{equation}
where $q$ represent the query point of $Q_2$. $k$ and $n$ represent the key points of $K_2$. $c$ denotes the semantic category. $\mathcal{\bar{B}}$ is the new attention map that can grasp the relationship between the content and style feature points, which share the most similar structure. This is visually shown in Fig.~\ref{fig:2} (c).

Eventually, we can obtain the stylized features $F_{ssa}$:
\begin{equation}
  F_{ssa} = f_o(softmax( \mathcal{\bar{B}}) \otimes V_2),
\end{equation}
where $f_o$ is the same as that in Eq.~\ref{eq:1}. Following the above operation, the $softmax$ function could ensure that the weights of content and style points in different semantic regions become $0$, while the weights of each content point and the most similar style point in the same semantic region are $1$ and weights of this content point and other style points in the same semantic region are $0$. Thus, each query point can match sparse key points in the same semantic region. SSA can intently concentrate on the specific stylistic texture of regions with the same semantics to transfer them to the matching semantic region in the stylized image.

{\bf Feature Fusion.} With the $F_{sca}$ and $F_{ssa}$ of SCA and SSA, the feature fusion can be executed:
\begin{equation}
  F_{cs} = \alpha_1 \times F_{sca} + \alpha_2 \times F_{ssa} + F_{c},
\end{equation}
where $\alpha_1$ and $\alpha_2$ denote the stylization degree of the overall style and vivid textures separately. $F_c$ are the outputs of Eq.~\ref{eq:8} for superior styling. $F_{cs}$ are the final stylized features.

\subsection{Overall Framework }

{\bf Revisit Attn-AST Approaches.} Given a content image $I_c$ and a style image $I_s$, the Attn-AST model $M$ could produce a stylized image $I_{cs}$, which preserves the content of $I_c$ and exhibit the style of $I_s$.

Firstly, they achieve the encoded content features $F_c$ and style features $F_s$ through the encoder $E$ of $M$:
\begin{equation}
F_c = E(I_c), \; F_s = E(I_s).
\end{equation}

Afterward, they utilize the feature transformation module $T$ with universal attention (UA) to generate the stylized features $F_{cs}$:
\begin{equation}
F_{cs} = T_{UA}(F_c, F_s).
\label{eq:15}
\end{equation}

Eventually, they could acquire the stylized image with the decoder $D$ of $M$: 
\begin{equation}
I_{cs} = D(F_{cs}),
\end{equation}
where $I_{cs}$ possesses the content of $I_c$ and the style of $I_s$, but the style of its corresponding semantic region is inconsistent with that of the style image.

{\bf Discussion.} Despite the three distinct Attn-AST frameworks—based on CNN, Transformer, and Diffusion—their pipelines for style transfer remain fundamentally consistent with the description above, differing only in minor model structures. The CNN-based method uses a simple encoder-decoder setup, while the Transformer-based approach employs two distinct encoders to process content and style separately. In contrast, the diffusion-based method also utilizes a single encoder-decoder structure but incorporates a time-step iteration mechanism for further refinement.

{\bf Attn-AST Approaches with our SCSA.} The proposed SCSA is a plug-and-play method that can be seamlessly incorporated into any existing Attn-AST method to facilitate semantic style transfer. Thus, given a content image $I_c$ along with its semantic map $I_{csem}$, a style image $I_s$ along with its semantic map $I_{ssem}$, and a specified Attn-AST model $M$, our objective is that $M_{SCSA}$, the approach obtained after incorporating SCSA into $M$, can generate a stylized image $I_{cs}$ using the quadruple \{$I_c, I_{csem}, I_s, I_{ssem}$\}, exhibiting a consistent style with the style image within the same semantic regions. The process of $M_{SCSA}$ for semantic style transfer is outlined below.

At first, we obtain the four encoded features of the quadruple through the encoder $E$:
\begin{equation}
\begin{aligned}
F_c = E(I_c), &\; F_s = E(I_s), \\
F_{csem} = E(I_{csem}), &\; F_{sem} = E(I_{sem}).
\end{aligned}
\end{equation}

Then, we replace UA in the feature transformation module $T$ of Eq.~\ref{eq:15} with SCSA, which could perform semantic feature conversion by enabling each query point to consider certain key points in the corresponding semantic region:
\begin{equation}
F_{cs} = T_{SCSA}(F_c, F_s, F_{csem}, F_{ssem}),
\end{equation}
where $F_{cs}$ are stylized features that meet semantic criteria.

Lastly, we get the stylized image with the decoder $D$:
\begin{equation}
I_{cs} = D(F_{cs}),
\end{equation}
where the style of the corresponding semantic region of the generated stylized image $I_{cs}$ is consistent with that of the style image $I_s$, while the content of the generated stylized image $I_{cs}$ is the same as that of the content image $I_c$.

\begin{figure*}
\centering
\resizebox{1\textwidth}{!}{
\setlength{\tabcolsep}{0.00cm} 
\renewcommand{\arraystretch}{0.001}  
\begin{tabular}{ccccccccccccc}
 Content & Style & SANet & SANet + SCSA & StyTR$^2$ & StyTR$^2$ + SCSA & StyleID & StyleID + SCSA & STROTSS & MAST & TR & DIA
& GLStyleNet\\
\includegraphics[width=0.14\linewidth]{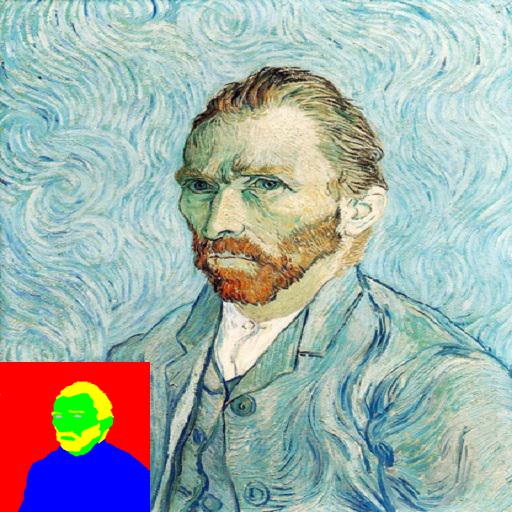} & \includegraphics[width=0.14\linewidth]{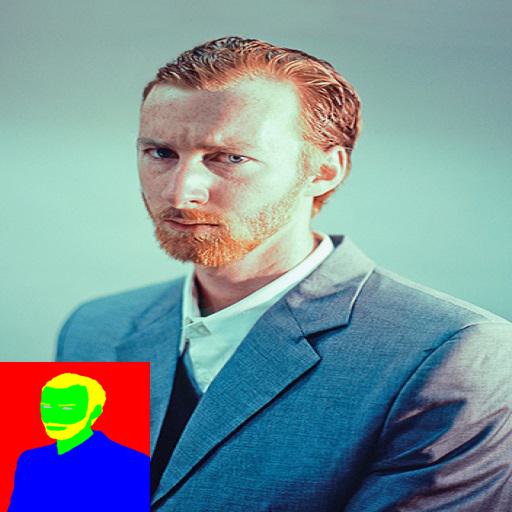}  & \includegraphics[width=0.14\linewidth]{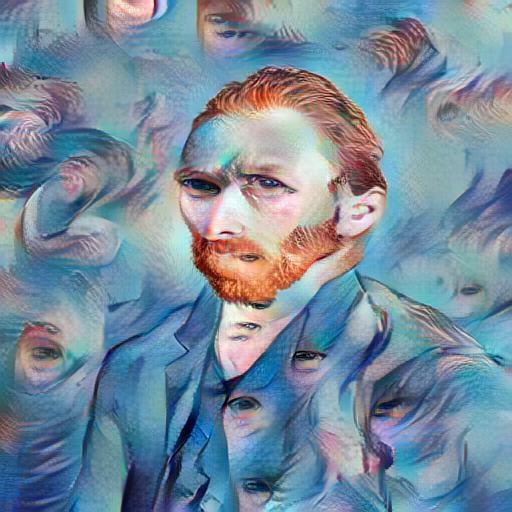} &
\includegraphics[width=0.14\linewidth]{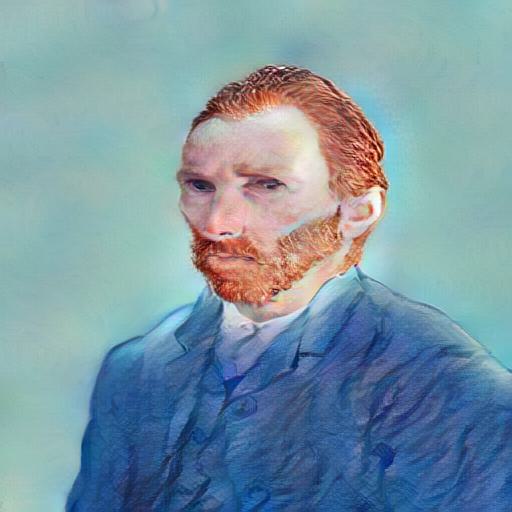} & \includegraphics[width=0.14\linewidth]{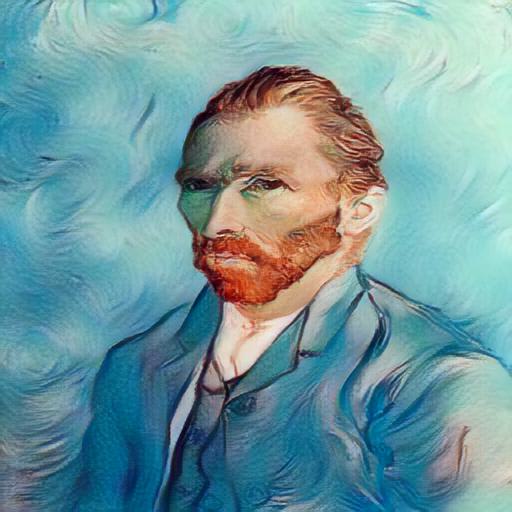} &
\includegraphics[width=0.14\linewidth]{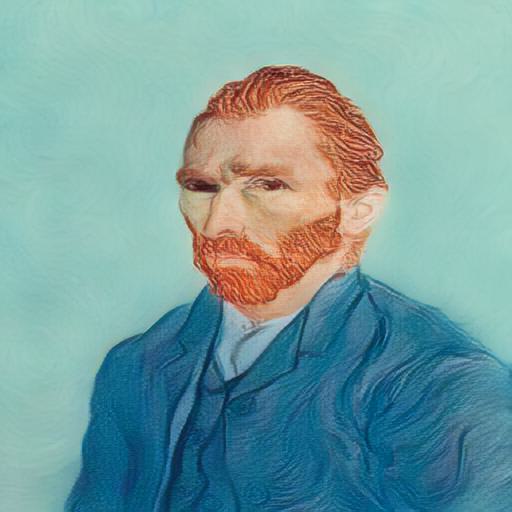} &  \includegraphics[width=0.14\linewidth]{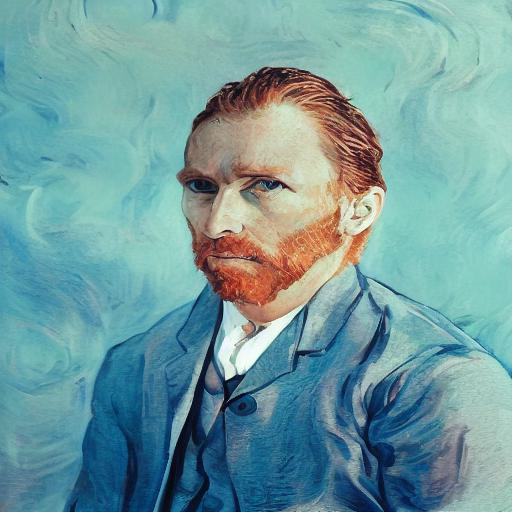} &  \includegraphics[width=0.14\linewidth]{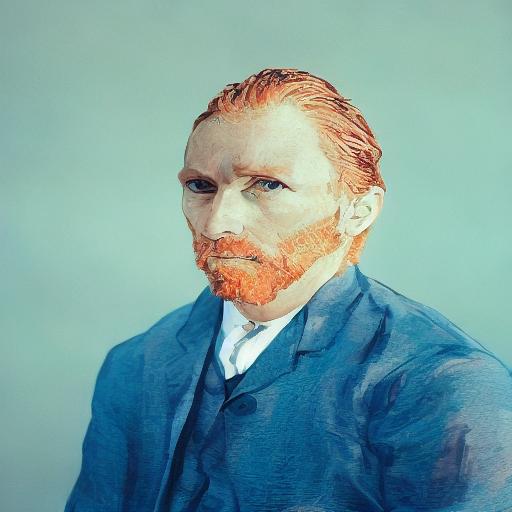} &  \includegraphics[width=0.14\linewidth]{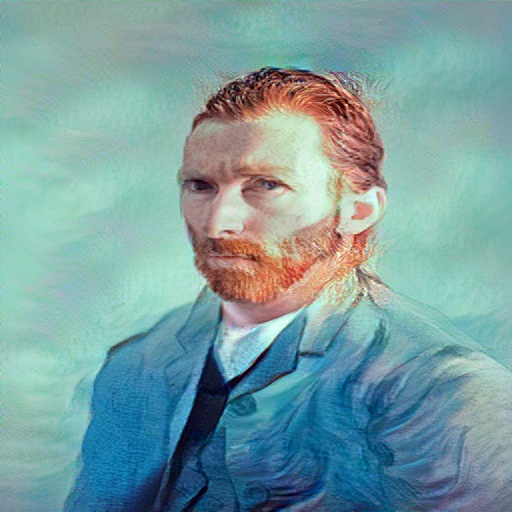}& \includegraphics[width=0.14\linewidth]{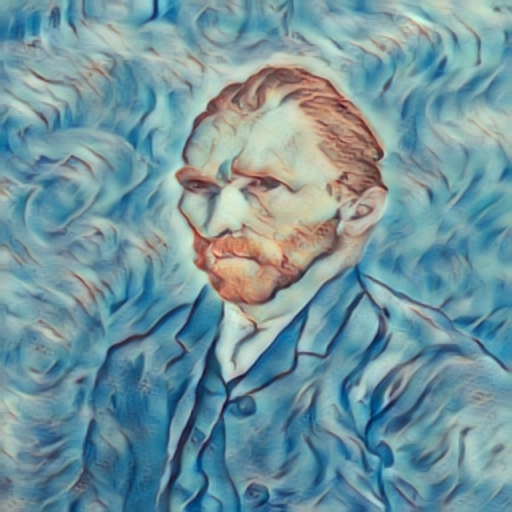} &  \includegraphics[width=0.14\linewidth]{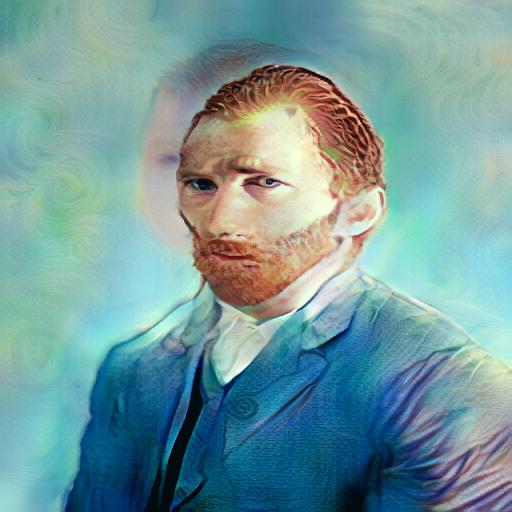} &  \includegraphics[width=0.14\linewidth]{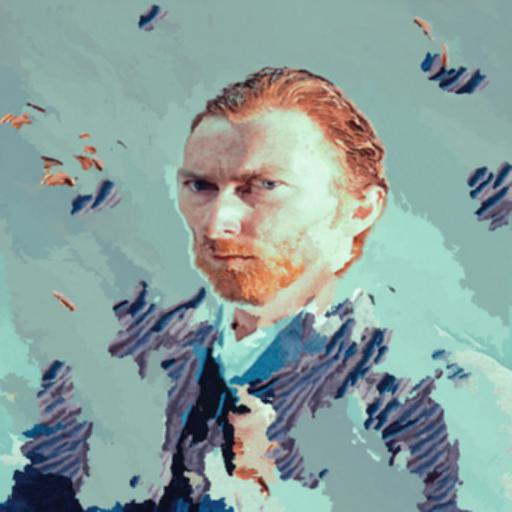} & \includegraphics[width=0.14\linewidth]{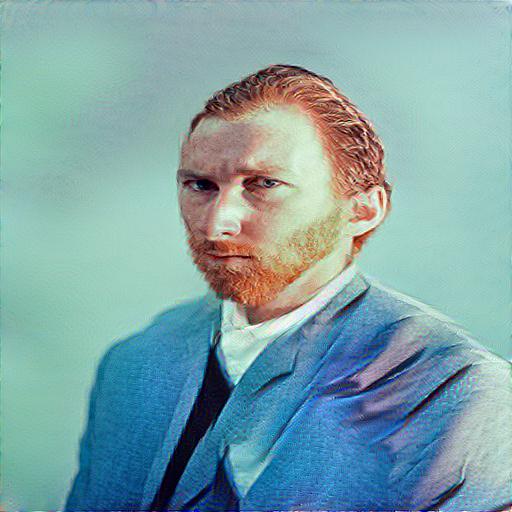}
\\
\includegraphics[width=0.14\linewidth]{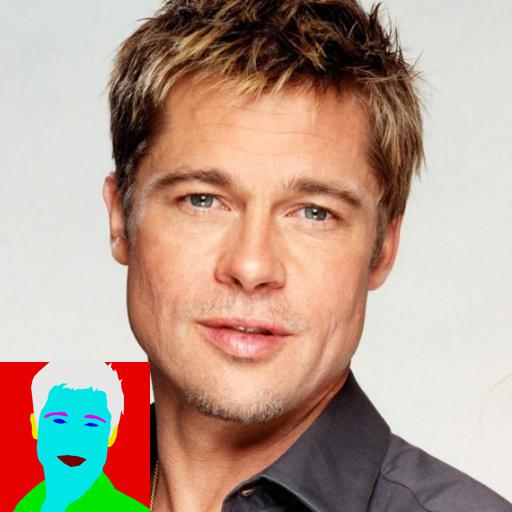} & \includegraphics[width=0.14\linewidth]{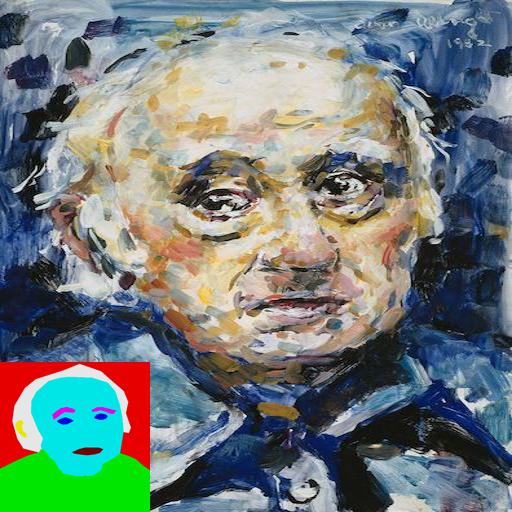}  & \includegraphics[width=0.14\linewidth]{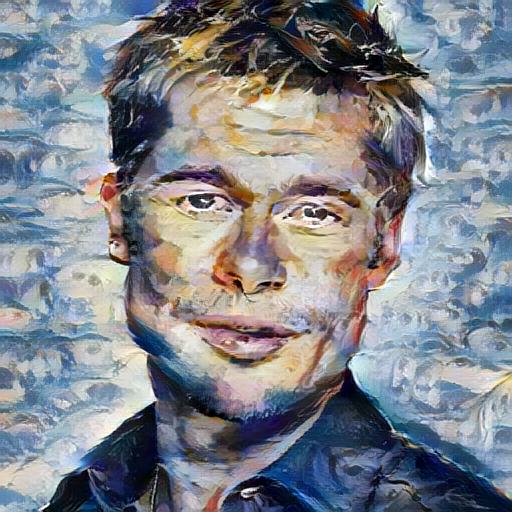} &
\includegraphics[width=0.14\linewidth]{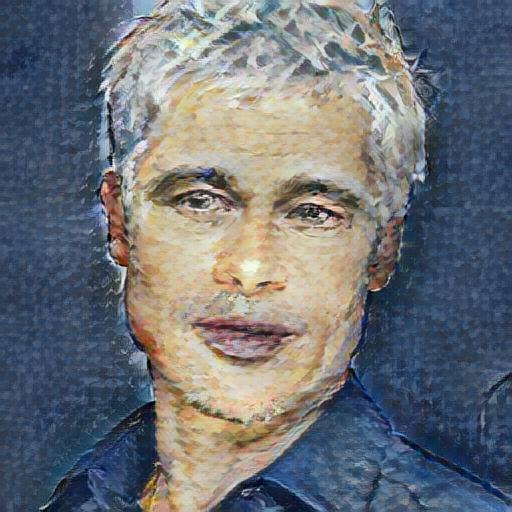} & \includegraphics[width=0.14\linewidth]{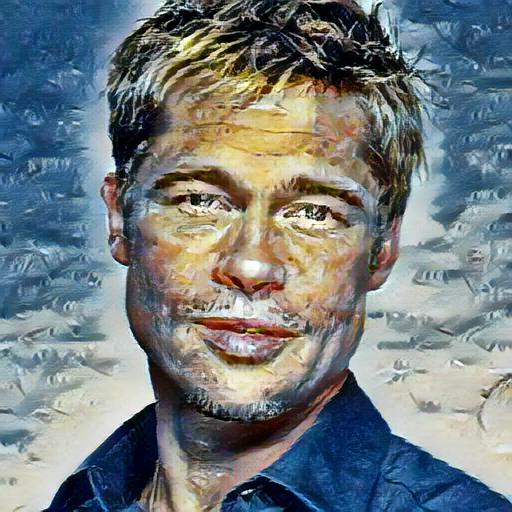} &
\includegraphics[width=0.14\linewidth]{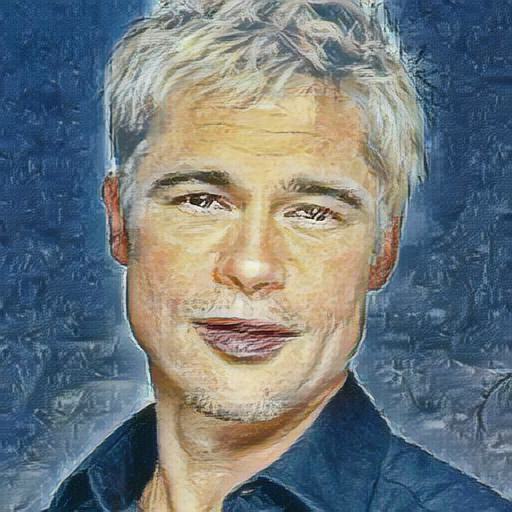} &  \includegraphics[width=0.14\linewidth]{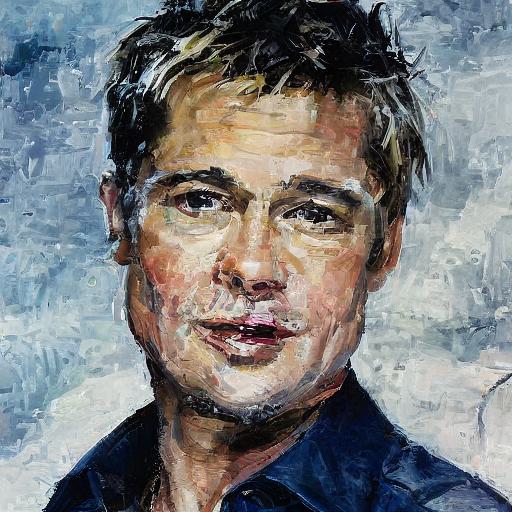} &  \includegraphics[width=0.14\linewidth]{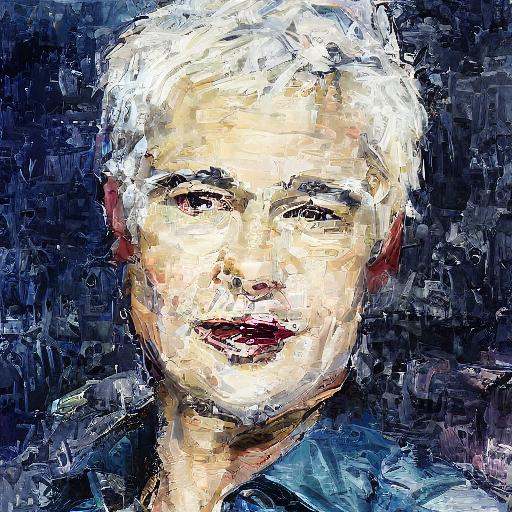} &  \includegraphics[width=0.14\linewidth]{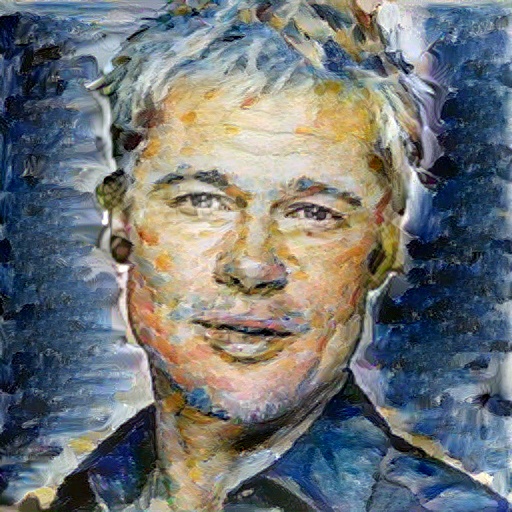}& \includegraphics[width=0.14\linewidth]{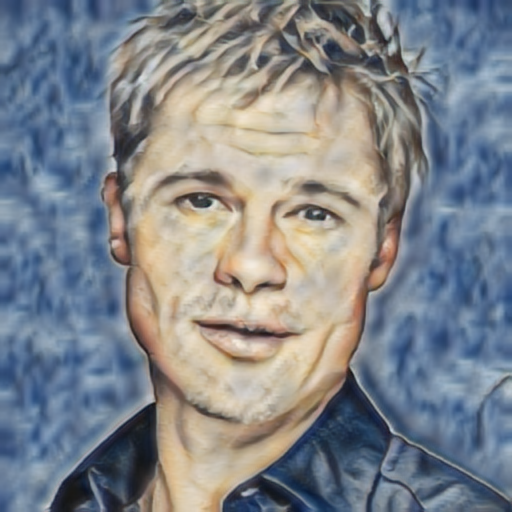} &  \includegraphics[width=0.14\linewidth]{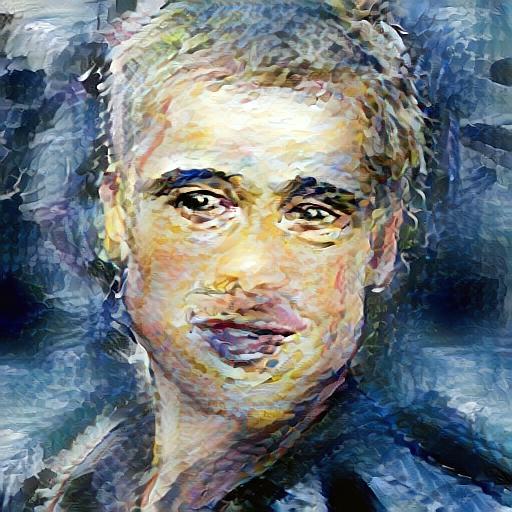} &  \includegraphics[width=0.14\linewidth]{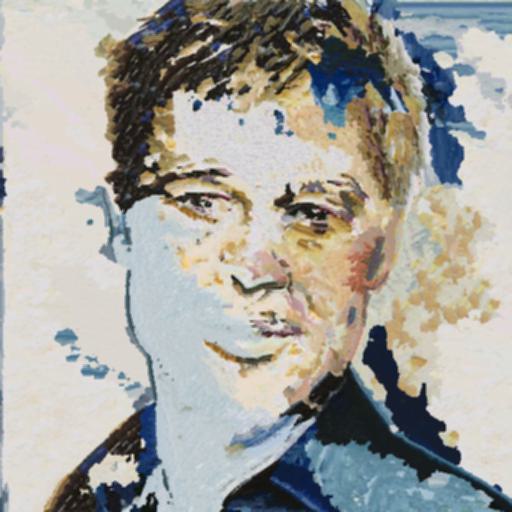} &
\includegraphics[width=0.14\linewidth]{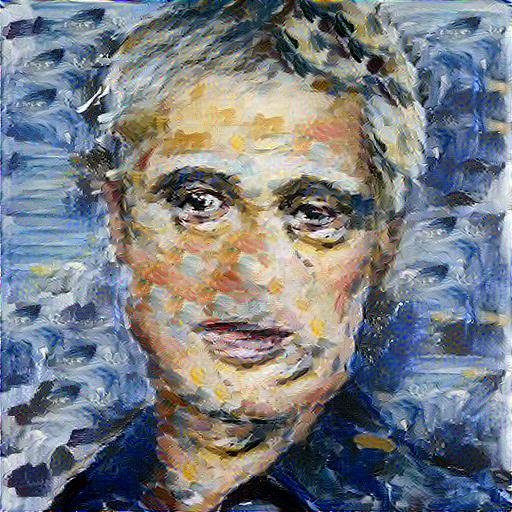}
\\
\includegraphics[width=0.14\linewidth]{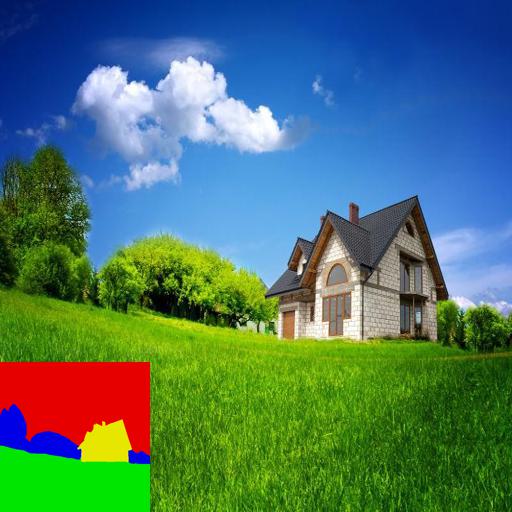} & \includegraphics[width=0.14\linewidth]{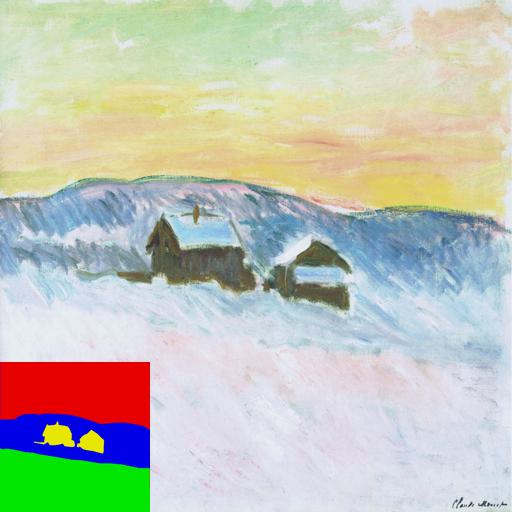}  & \includegraphics[width=0.14\linewidth]{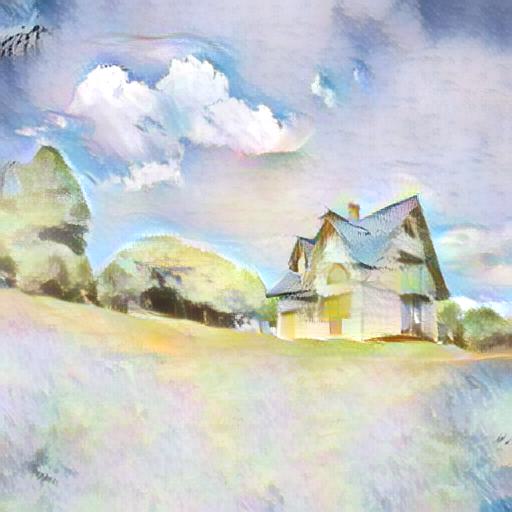} &
\includegraphics[width=0.14\linewidth]{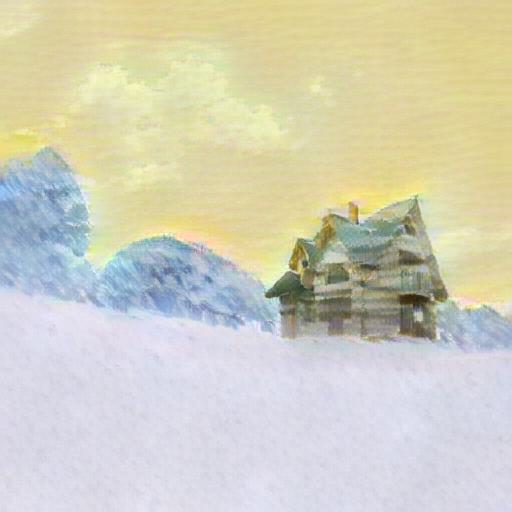} & \includegraphics[width=0.14\linewidth]{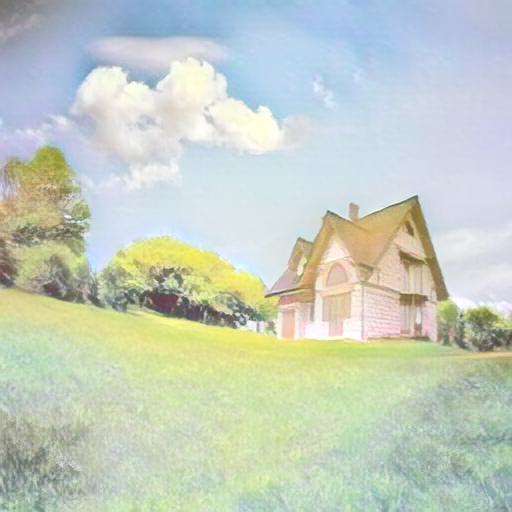} &
\includegraphics[width=0.14\linewidth]{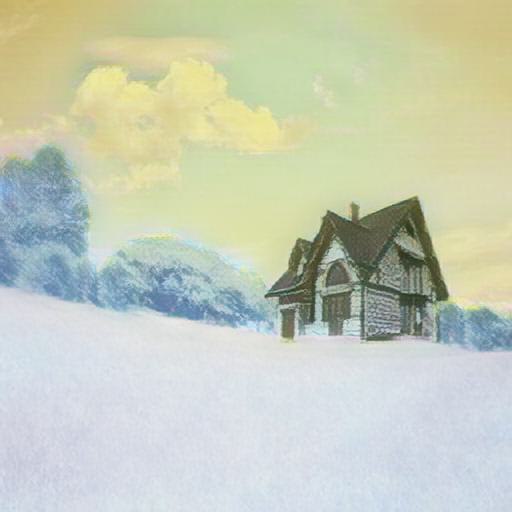} &  \includegraphics[width=0.14\linewidth]{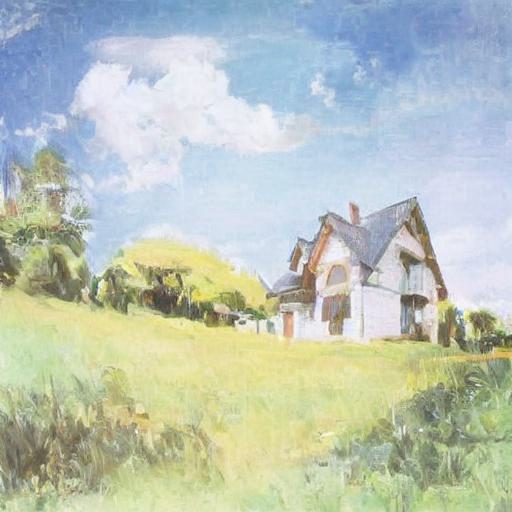} &  \includegraphics[width=0.14\linewidth]{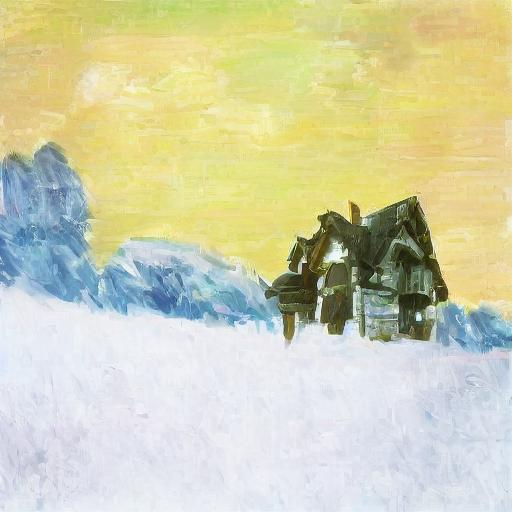} &  \includegraphics[width=0.14\linewidth]{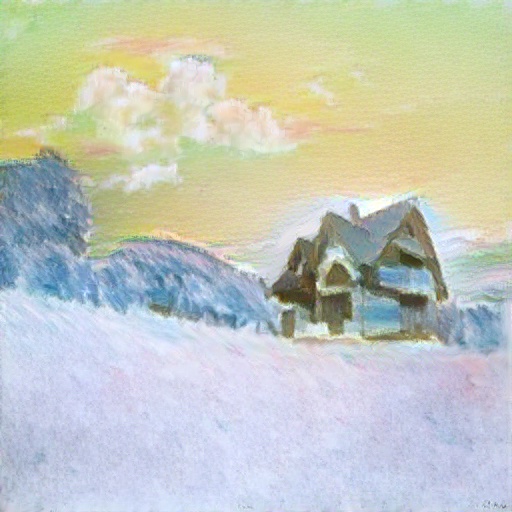}& \includegraphics[width=0.14\linewidth]{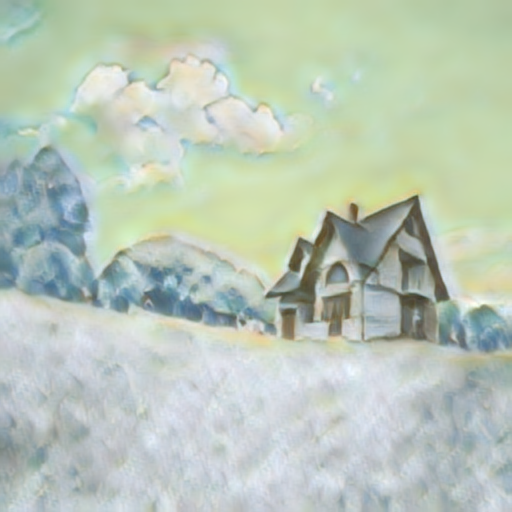} &  \includegraphics[width=0.14\linewidth]{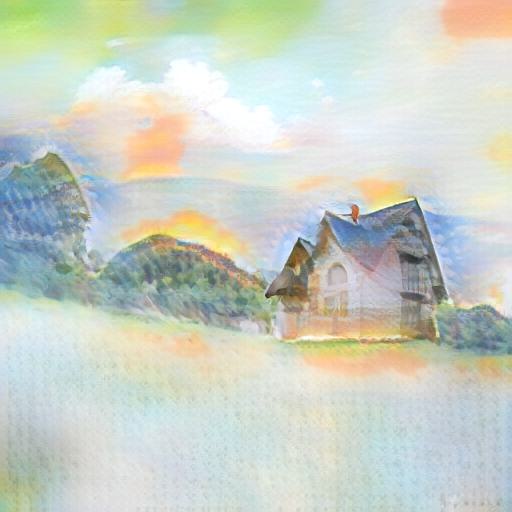} &  \includegraphics[width=0.14\linewidth]{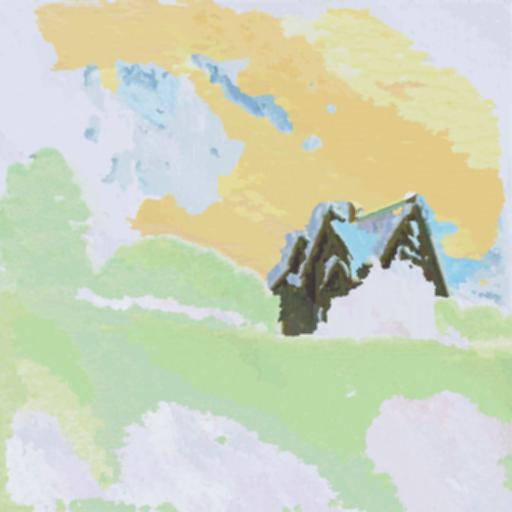} &
\includegraphics[width=0.14\linewidth]{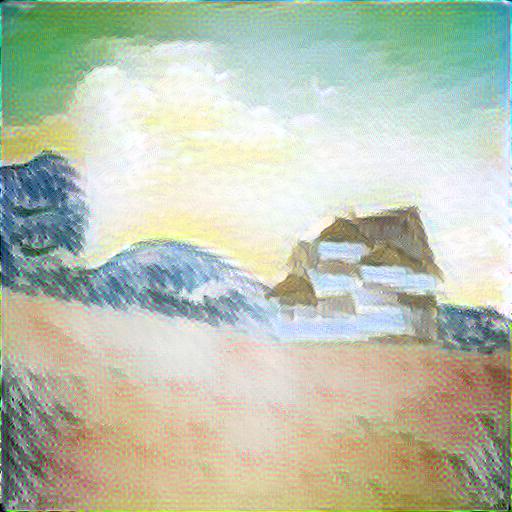}
\\
\includegraphics[width=0.14\linewidth]{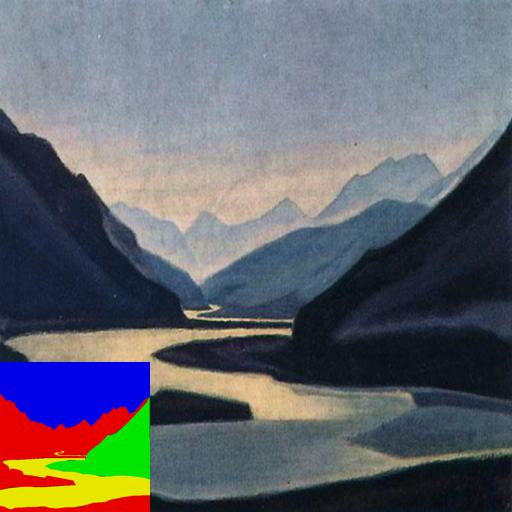} & \includegraphics[width=0.14\linewidth]{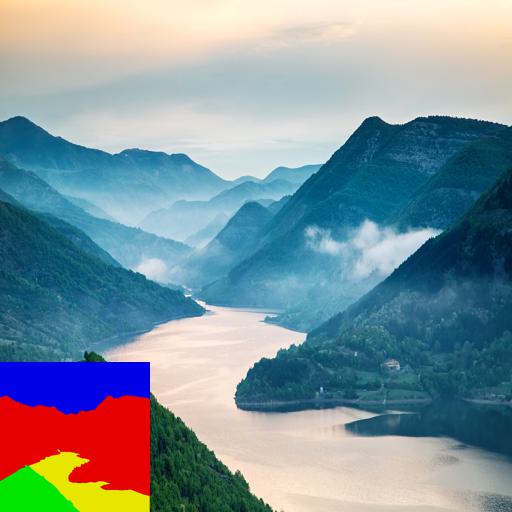}  & \includegraphics[width=0.14\linewidth]{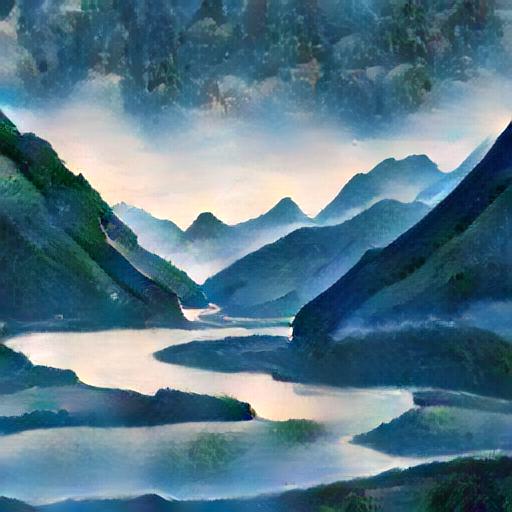} &
\includegraphics[width=0.14\linewidth]{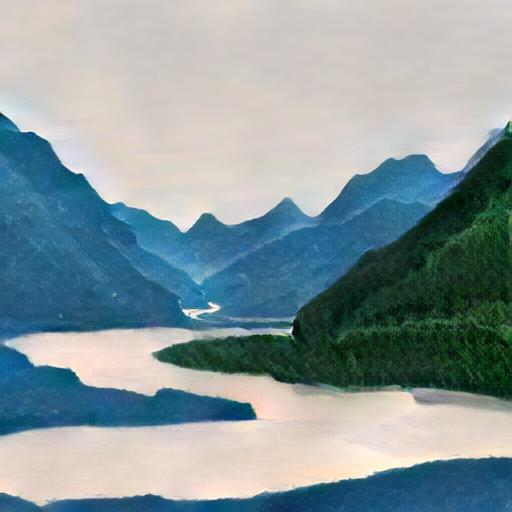} & \includegraphics[width=0.14\linewidth]{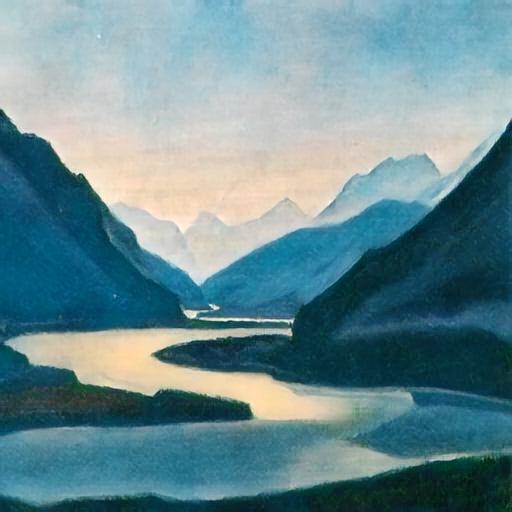} &
\includegraphics[width=0.14\linewidth]{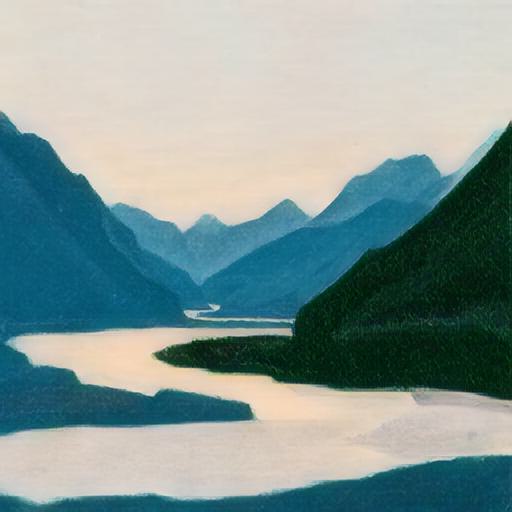} &  \includegraphics[width=0.14\linewidth]{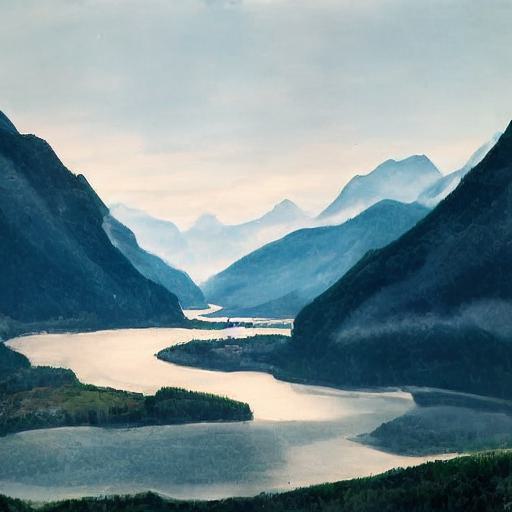} &  \includegraphics[width=0.14\linewidth]{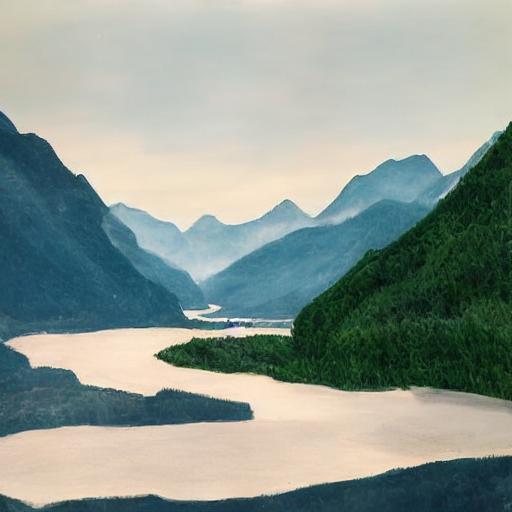} &  \includegraphics[width=0.14\linewidth]{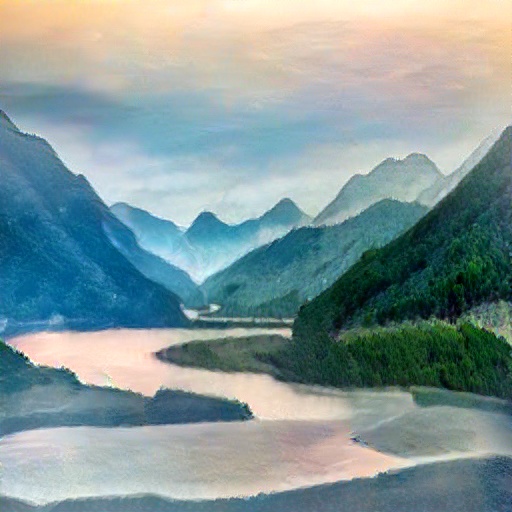}& \includegraphics[width=0.14\linewidth]{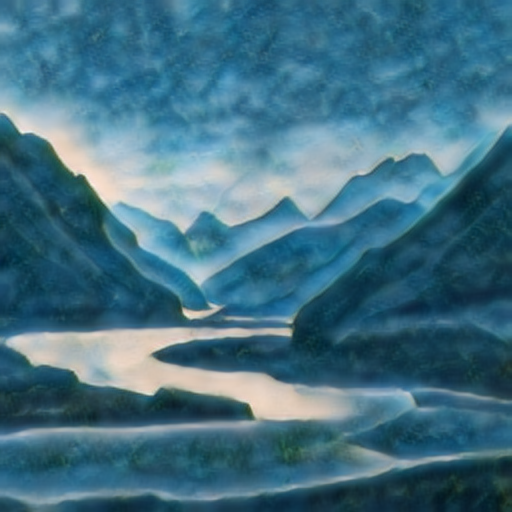} &  \includegraphics[width=0.14\linewidth]{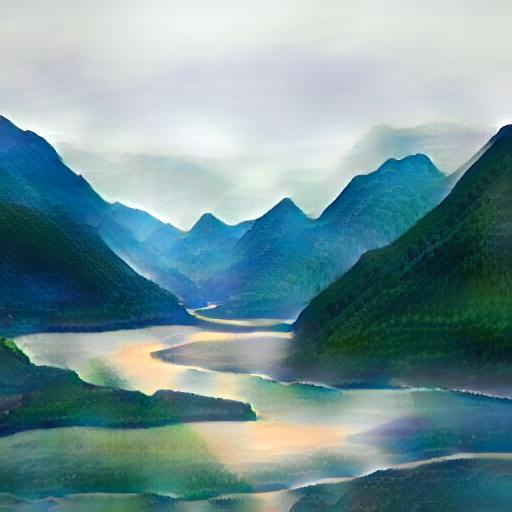} &  \includegraphics[width=0.14\linewidth]{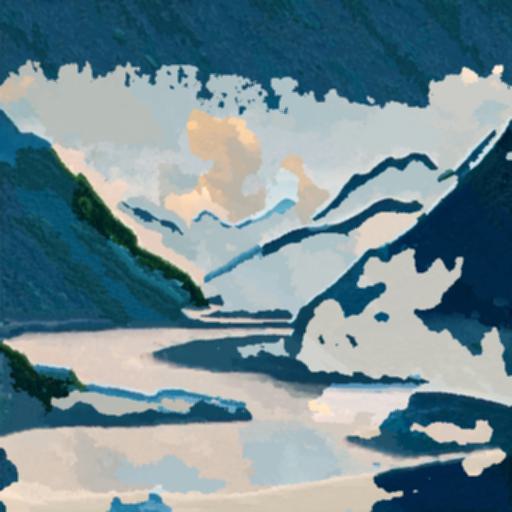} &
\includegraphics[width=0.14\linewidth]{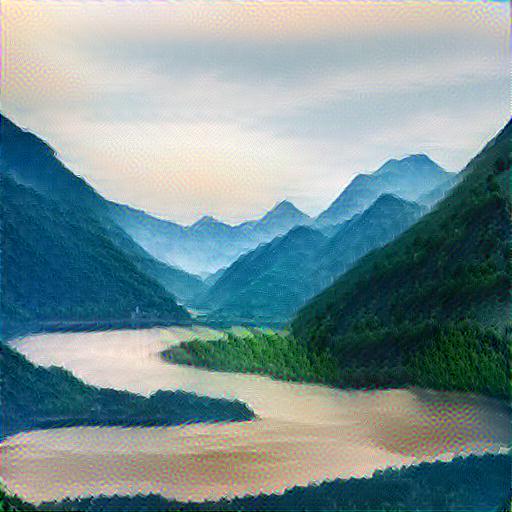}
\\
\includegraphics[width=0.14\linewidth]{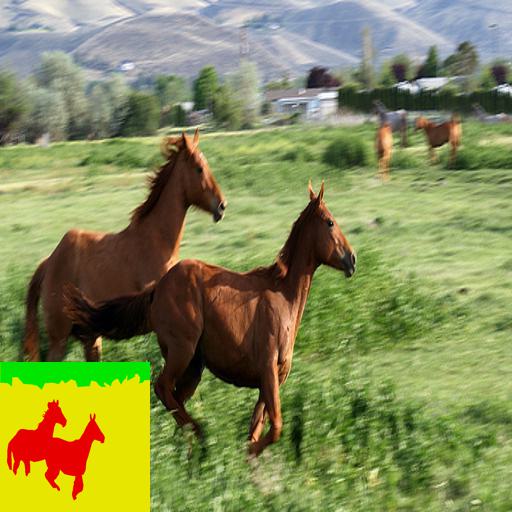} & \includegraphics[width=0.14\linewidth]{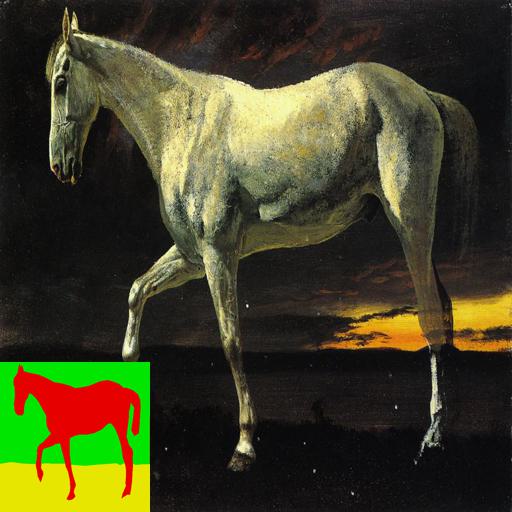}  & \includegraphics[width=0.14\linewidth]{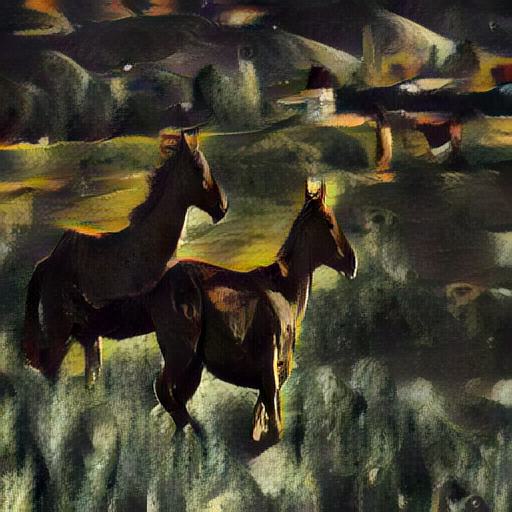} &
\includegraphics[width=0.14\linewidth]{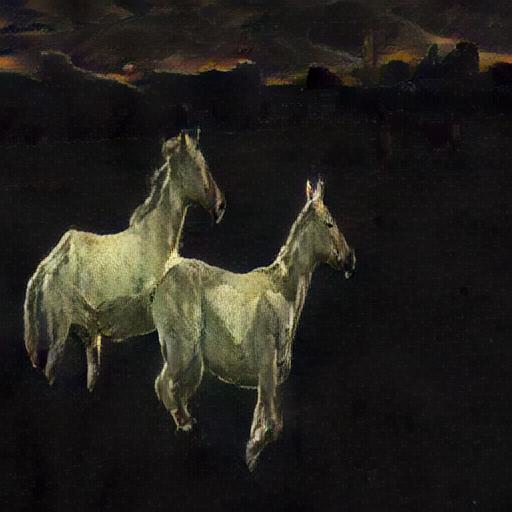} & \includegraphics[width=0.14\linewidth]{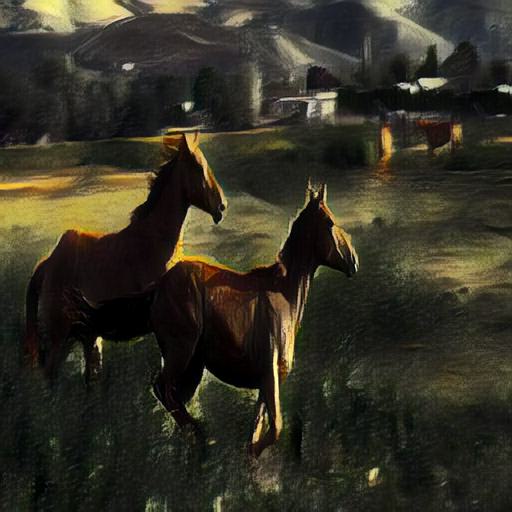} &
\includegraphics[width=0.14\linewidth]{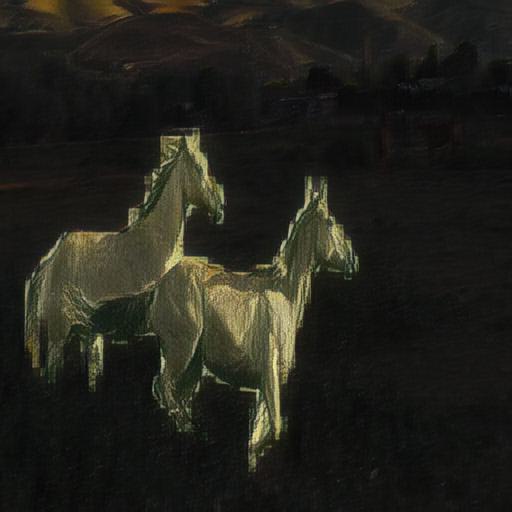} &  \includegraphics[width=0.14\linewidth]{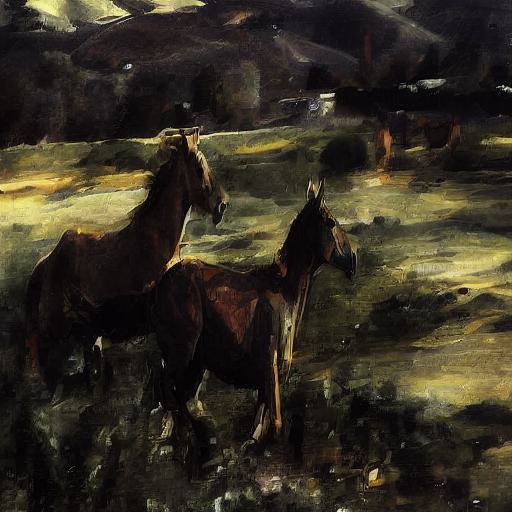} &  \includegraphics[width=0.14\linewidth]{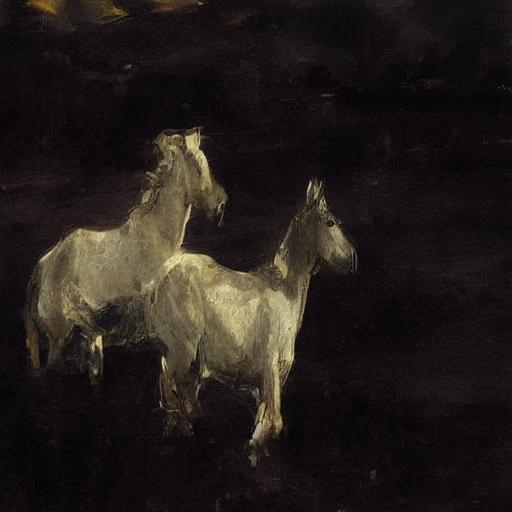} &  \includegraphics[width=0.14\linewidth]{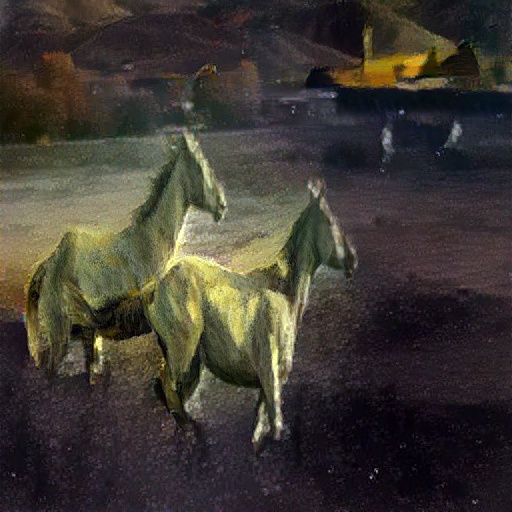}& \includegraphics[width=0.14\linewidth]{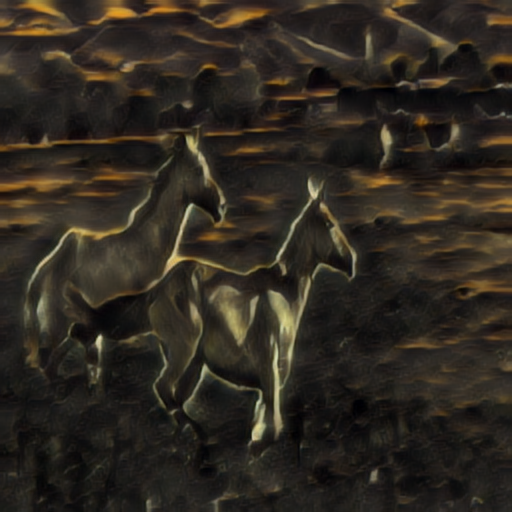} &  \includegraphics[width=0.14\linewidth]{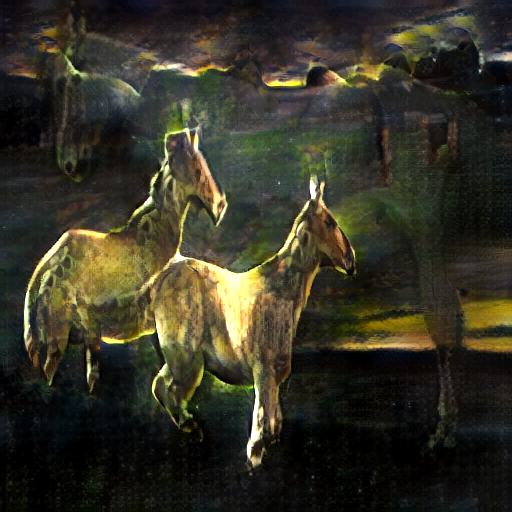} &  \includegraphics[width=0.14\linewidth]{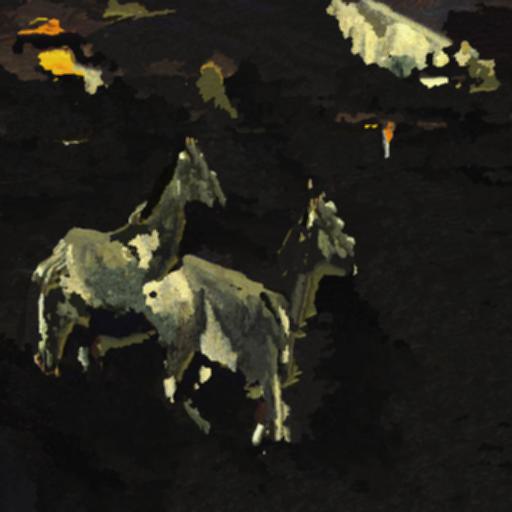} &
\includegraphics[width=0.14\linewidth]{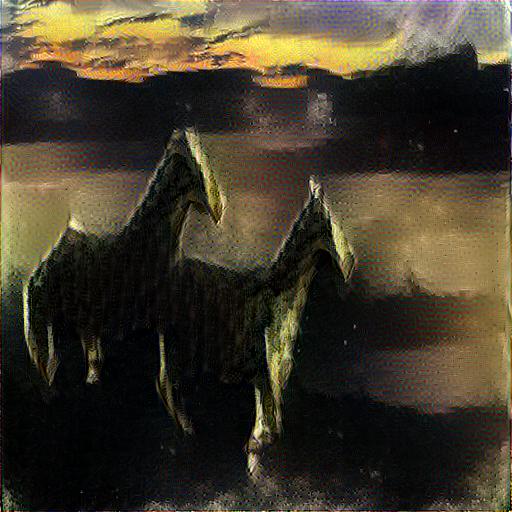}
\\
\includegraphics[width=0.14\linewidth]{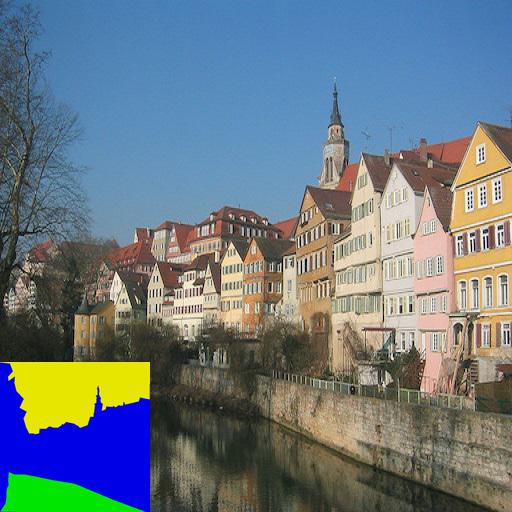} & \includegraphics[width=0.14\linewidth]{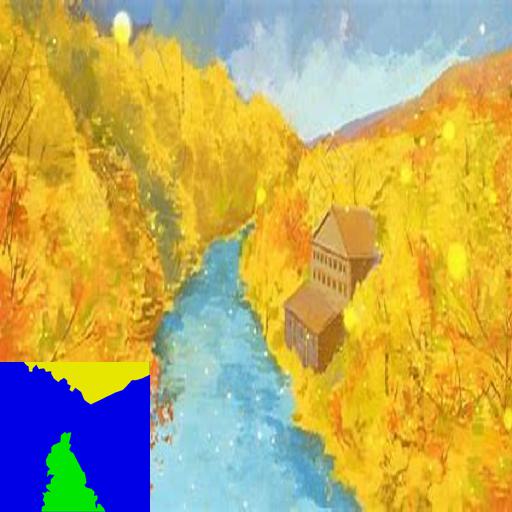}  & \includegraphics[width=0.14\linewidth]{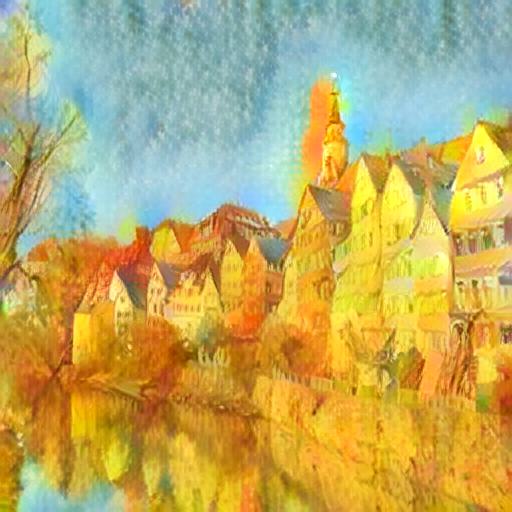} &
\includegraphics[width=0.14\linewidth]{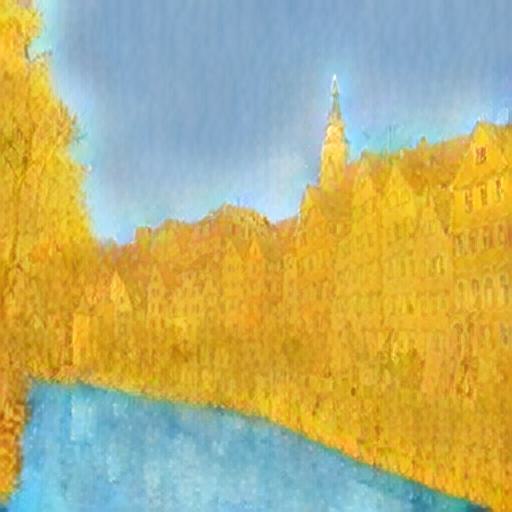} & \includegraphics[width=0.14\linewidth]{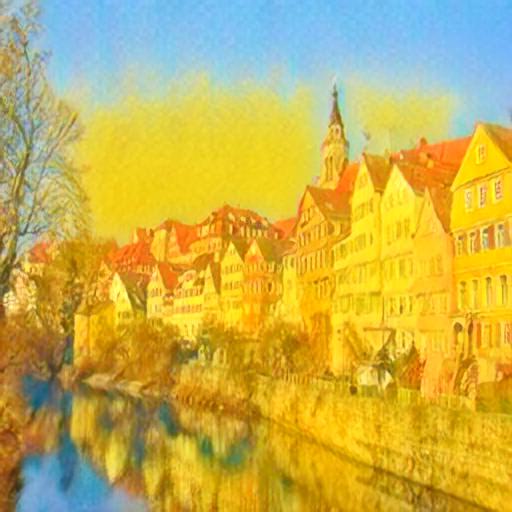} &
\includegraphics[width=0.14\linewidth]{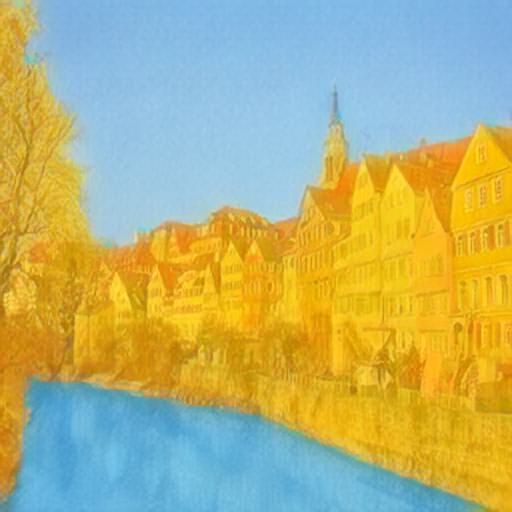} &  \includegraphics[width=0.14\linewidth]{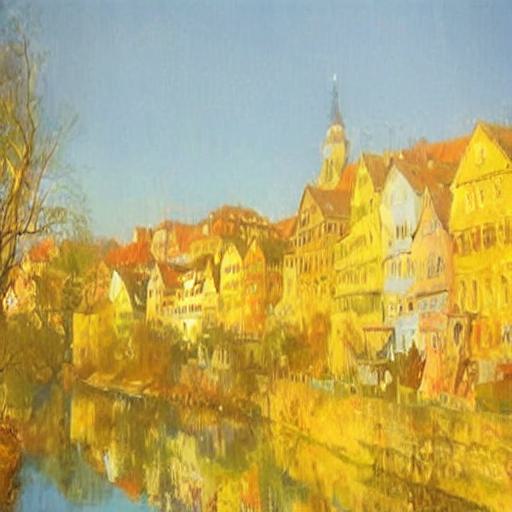} &  \includegraphics[width=0.14\linewidth]{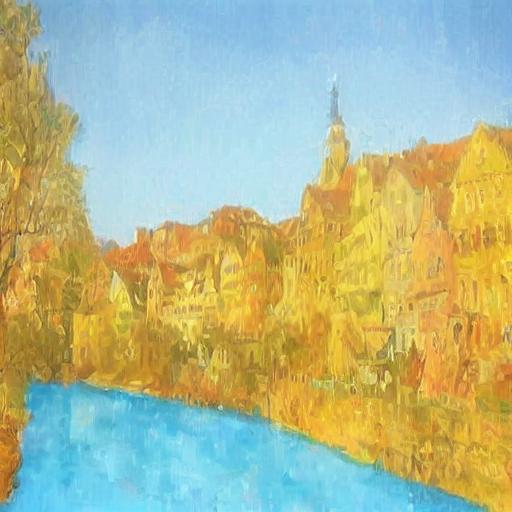} &  \includegraphics[width=0.14\linewidth]{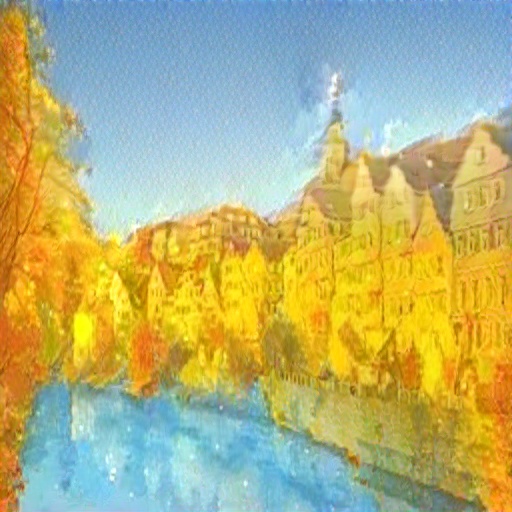}& \includegraphics[width=0.14\linewidth]{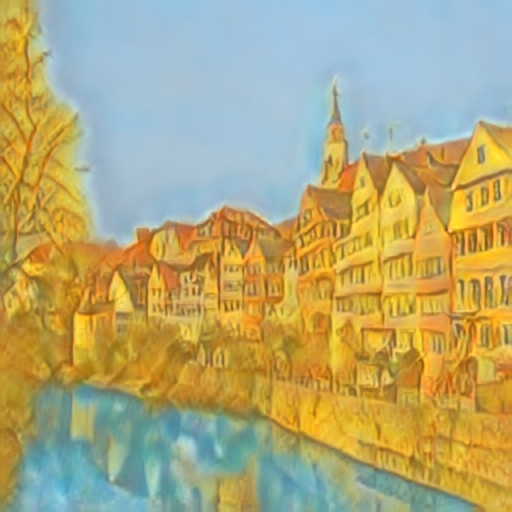} &  \includegraphics[width=0.14\linewidth]{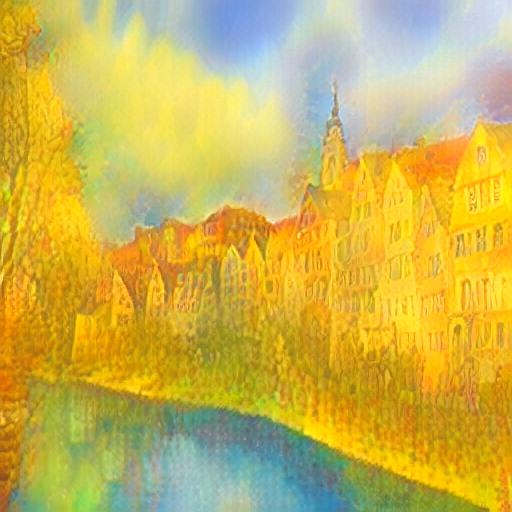} &  \includegraphics[width=0.14\linewidth]{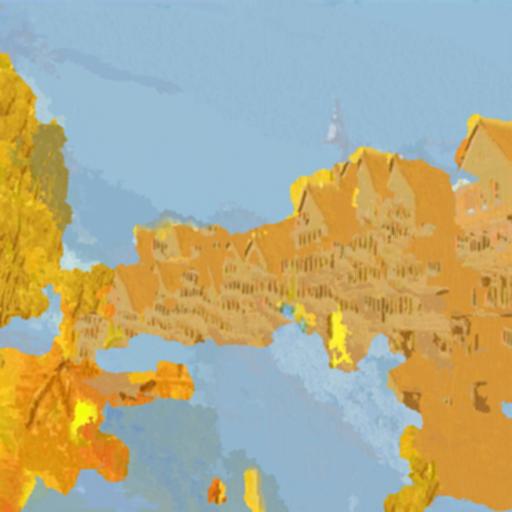} &
\includegraphics[width=0.14\linewidth]{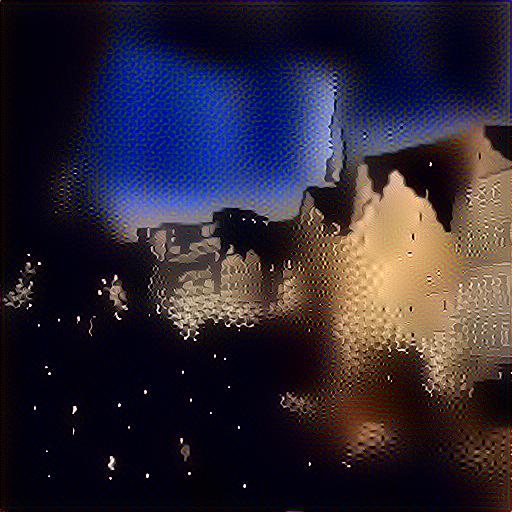}
\\
\includegraphics[width=0.14\linewidth]{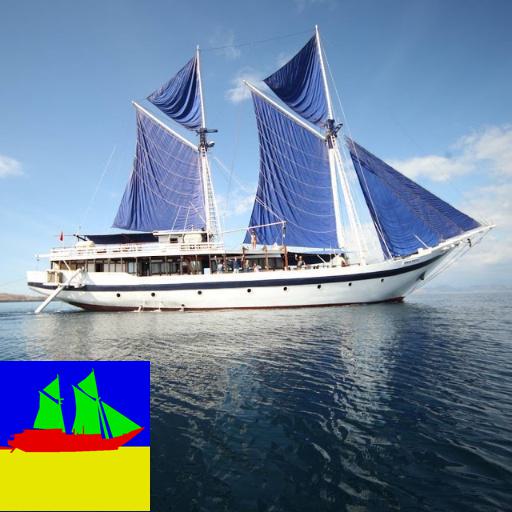} & \includegraphics[width=0.14\linewidth]{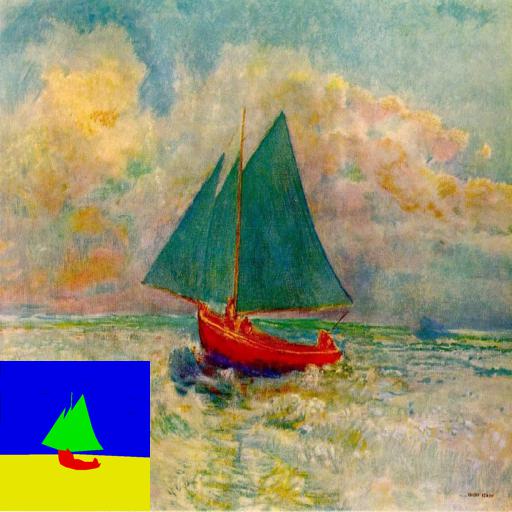}  & \includegraphics[width=0.14\linewidth]{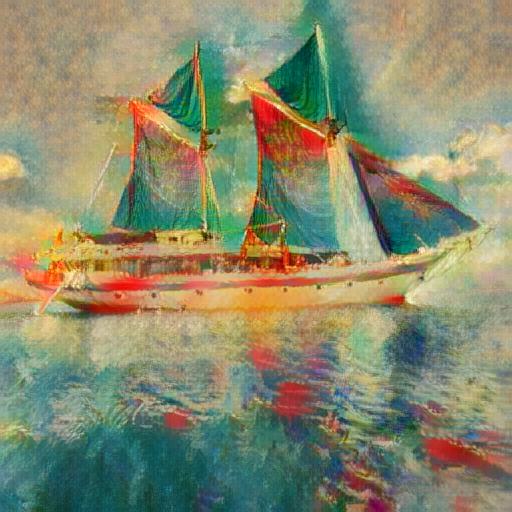} &
\includegraphics[width=0.14\linewidth]{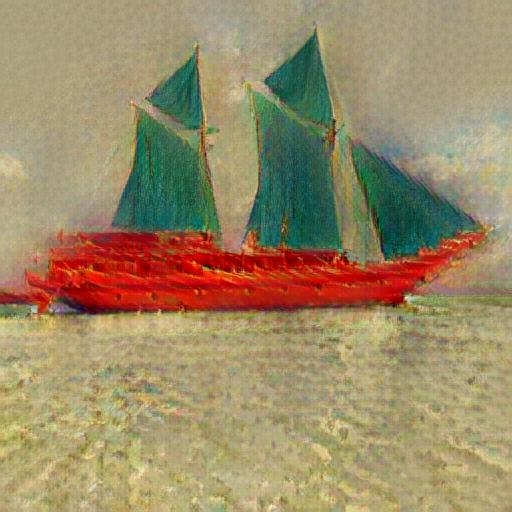} & \includegraphics[width=0.14\linewidth]{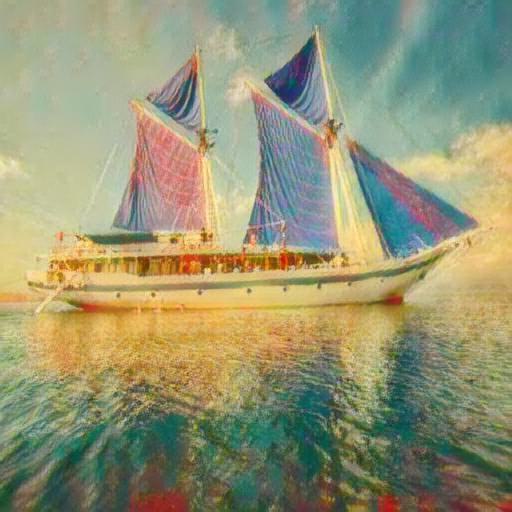} &
\includegraphics[width=0.14\linewidth]{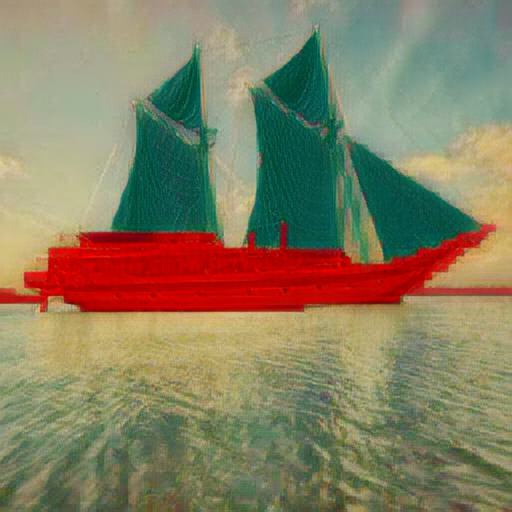} &  \includegraphics[width=0.14\linewidth]{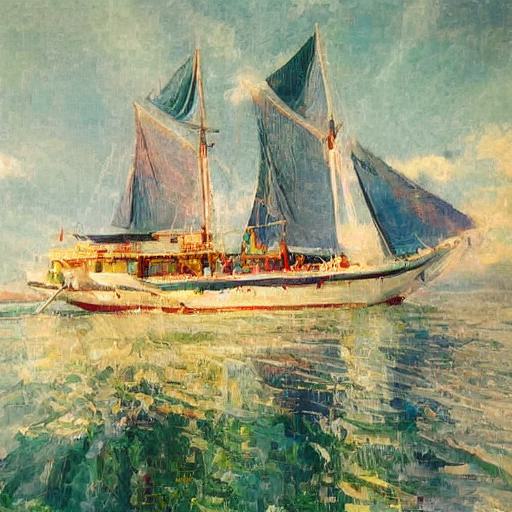} &  \includegraphics[width=0.14\linewidth]{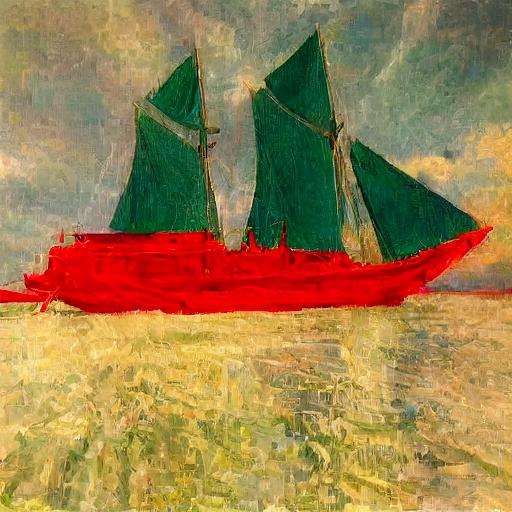} &  \includegraphics[width=0.14\linewidth]{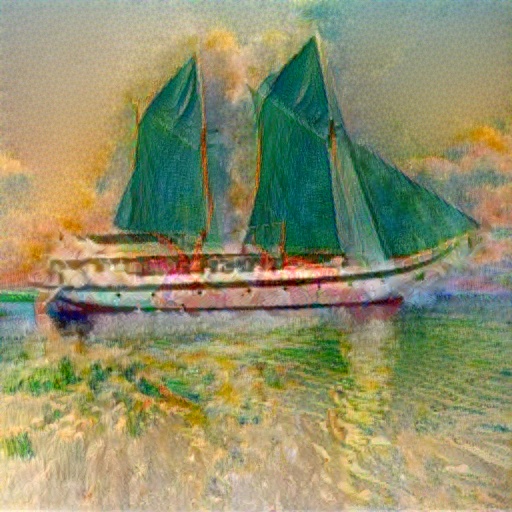}& \includegraphics[width=0.14\linewidth]{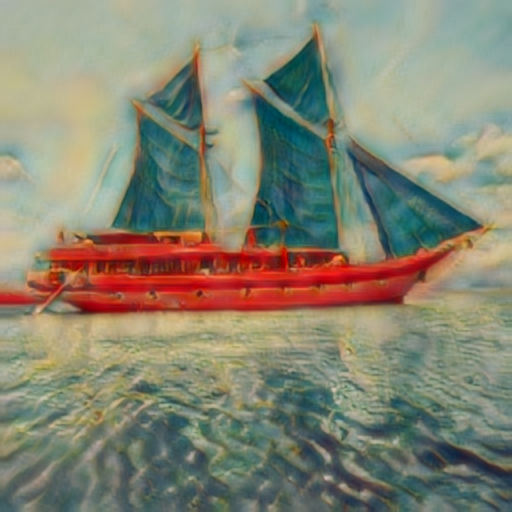} &  \includegraphics[width=0.14\linewidth]{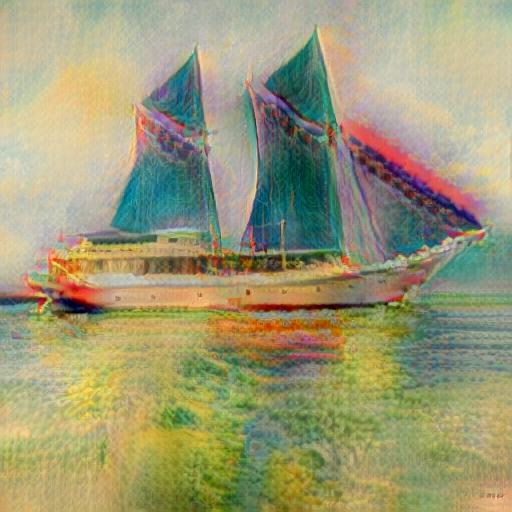} &  \includegraphics[width=0.14\linewidth]{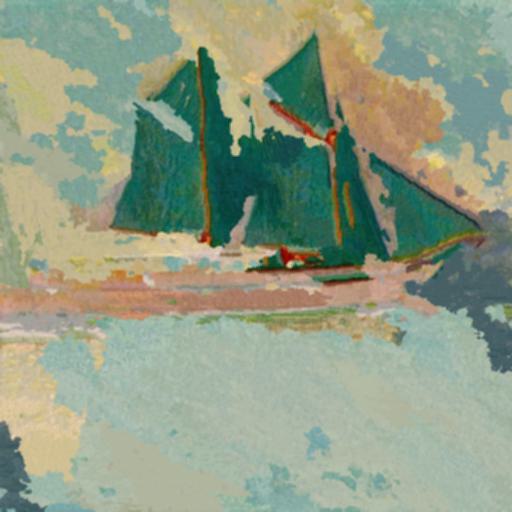} &
\includegraphics[width=0.14\linewidth]{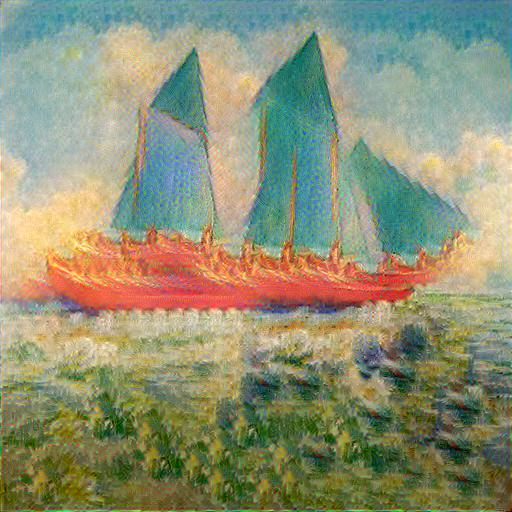}
\end{tabular}
}
\caption{Qualitative comparisons among Attn-AST approaches, those with SCSA, and SOTA methods.}
\label{fig:4}
\end{figure*}

\begin{table*}[t!]
\centering
\scalebox{0.75}{
\begin{tabular}{c|cc|cc|cc|ccccc}
\hline
& SANet & SANet + SCSA  & StyTR$^2$ & StyTR$^2$ + SCSA & StyleID & StyleID + SCSA & STROTSS & MAST & TR & DIA & GLStyleNet\\
\hline
SSL $\downarrow$ & 1.6583 & \underline{{\bf 0.8762}} & 1.9826 & {\bf 1.2228} & 1.7538 & {\bf 1.2447} & 1.0981 & 1.7320 & 1.2631 & 1.9398 & 1.0305 \\
FID $\downarrow$ & 14.3385 & {\bf 13.0788} & 12.5273 & \underline{{\bf 12.3963}} & 12.5944 & {\bf 12.4497} & 12.8400 & 16.5163 & 13.5846 & 20.7942 & 14.4880\\
CFSD $\downarrow$ & 0.1103 & {\bf 0.0874} & 0.0752 & \underline{{\bf 0.0705}} & {\bf 0.0916} &  0.1178 & 0.1008 & 0.0746 & 0.1149 & 0.1147 & 0.3991\\
Pref. $\uparrow$ & 0.1685 & {\bf 0.8315} & 0.1576  & {\bf 0.8424} & 0.2192  & {\bf 0.7808} & 0.1867 & 0.1100 & 0.0450 & 0.0167 & 0.1183\\

\hline
\end{tabular}
}
\caption{Quantitative comparisons of Attn-AST approaches, those with SCSA, and several SOTA semantic style transfer methods. The best results for Attn-AST methods and Attn-AST methods with SCSA are in \textbf{bold}, while the best results among all methods are \underline{underlined}.}
\label{tab:1}
\end{table*}

The frameworks of the CNN-based, Transformer-based, and Diffusion-based Attn-AST methods with our SCSA are shown in Fig.~\ref{fig:3}. The three methods differ only in the specific layers where UA is replaced with SCSA and the distinct structures of the encoder and decoder, while the other embedding operations remain largely consistent. 

\section{Experiments}

\subsection{Experiment Settings}

We carefully select three proxy Attn-AST approaches---CNN-based SANet~\cite{park2019arbitrary}, Transformer-based StyTR$^2$~\cite{deng2022stytr2}, and Diffusion-based StyleID~\cite{chung2024style}---and five state-of-the-art (SOTA) semantic style transfer methods--STROTSS~\cite{kolkin2019style}, MAST~\cite{huo2021manifold}, TR~\cite{wang2022texture}, DIA~\cite{liao2017visual}, GLStyleNet~\cite{wang2020glstylenet}--as baselines. We obtain content and style images from prior research, datasets, and the Internet. Then, we construct their semantic maps and generate validated quadruple data.

\subsection{Qualitative Comparison}
We conduct extensive comparison experiments to determine the effectiveness and generalization of our SCSA.

Attn-AST approaches combined with our SCSA can generate high-quality and vivid stylized images that better meet semantic needs compared to traditional Attn-AST approaches, showcasing rich textures and vibrant colors in corresponding semantic regions, e.g., the background in the $2nd$ row, along with the sky and ground in the $3rd$ row of Fig.~\ref{fig:4}. Likewise, compared to SOTA semantic style transfer methods, the stylized images generated by SCSA demonstrate superior semantic accuracy and content preservation, e.g. the river in the $4th$ row and the horses in the $5th$ row.

\begin{figure*}
\centering
\resizebox{0.99\textwidth}{!}{
\setlength{\tabcolsep}{0.04cm} 

\begin{tabular}{cc|cccc|cccc|cccc} 
  \Large Content & \Large Style &  \large SANet + SCSA  & \large - SSA  &  \large - SCA & \large - S-AdaIN & \large StyTR$^2$ + SCSA & \large - SSA  & \large - SCA & \large - S-AdaIN& \large StyleID + SCSA &  \large - SSA  & \large - SCA & \large - S-AdaIN\\
\includegraphics[width=0.16\linewidth]{img/25+sem.jpg}  &\includegraphics[width=0.16\linewidth]{img/25_paint+sem.jpg}&  
 \includegraphics[width=0.16\linewidth]{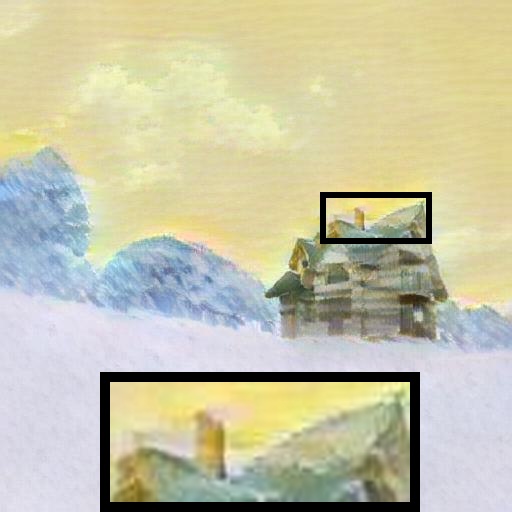} & 
 \includegraphics[width=0.16\linewidth]{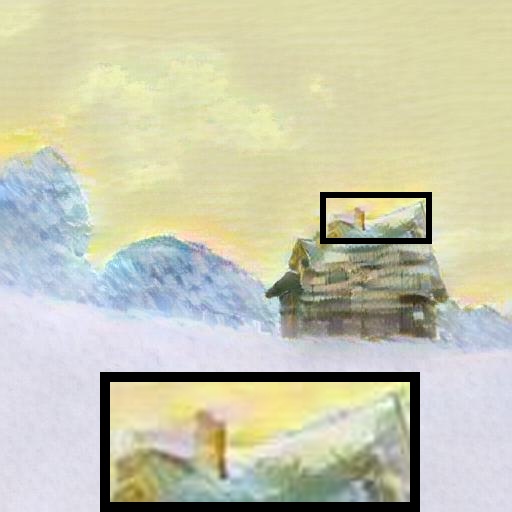} &
 \includegraphics[width=0.16\linewidth]{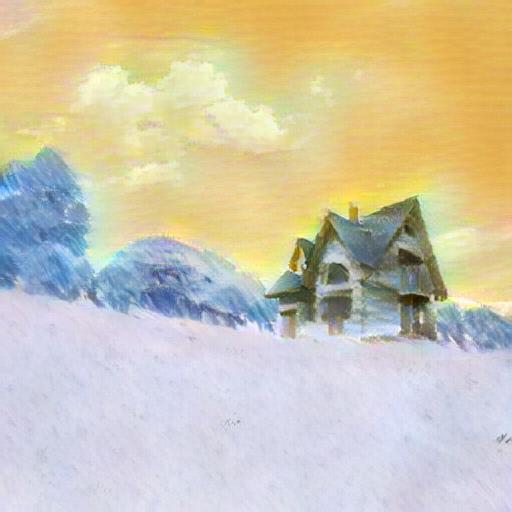} &
 \includegraphics[width=0.16\linewidth]{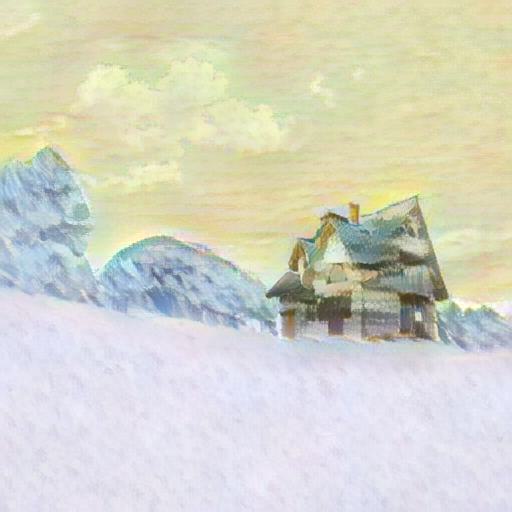} &
   \includegraphics[width=0.16\linewidth]{img/25_25_paint_StyTR2_sem.jpg}&
  \includegraphics[width=0.16\linewidth]{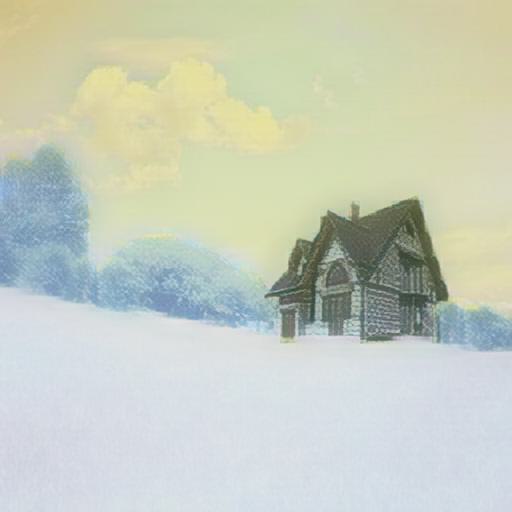}&
  \includegraphics[width=0.16\linewidth]{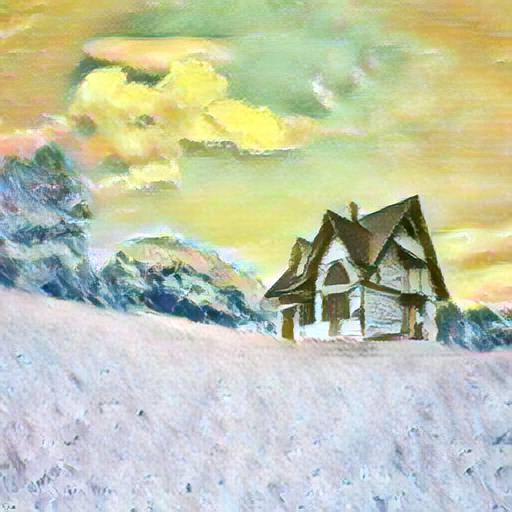}&
  \includegraphics[width=0.16\linewidth]{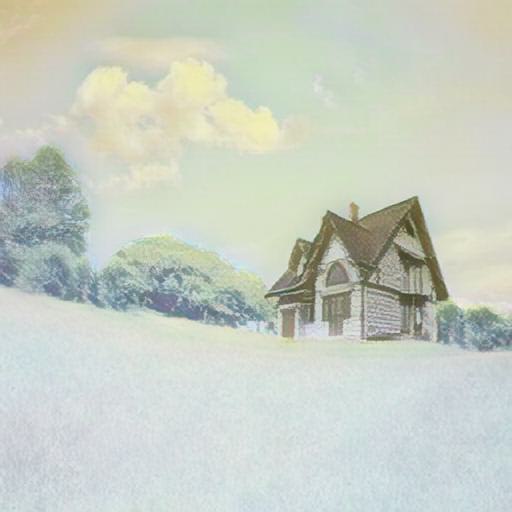}&
  \includegraphics[width=0.16\linewidth]{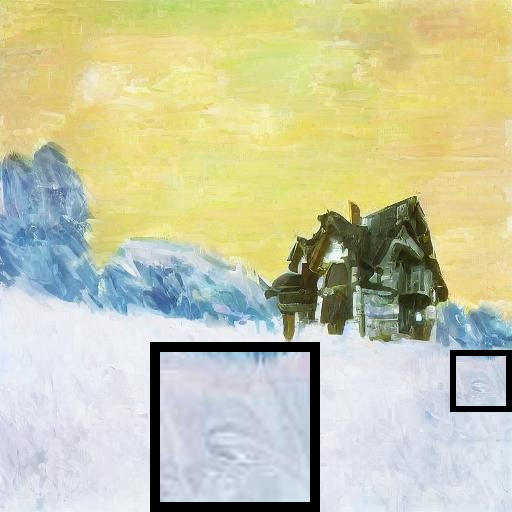}&
  \includegraphics[width=0.16\linewidth]{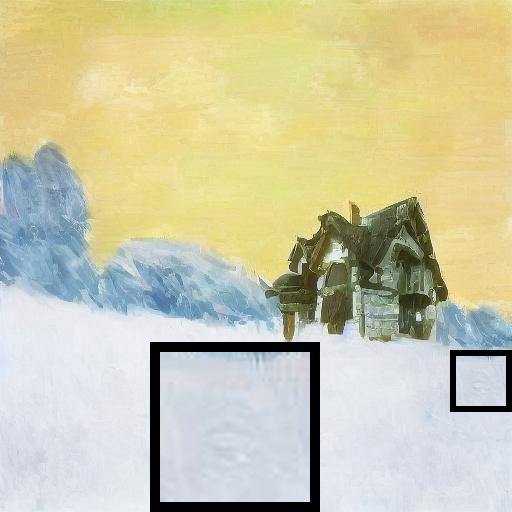}&
  \includegraphics[width=0.16\linewidth]{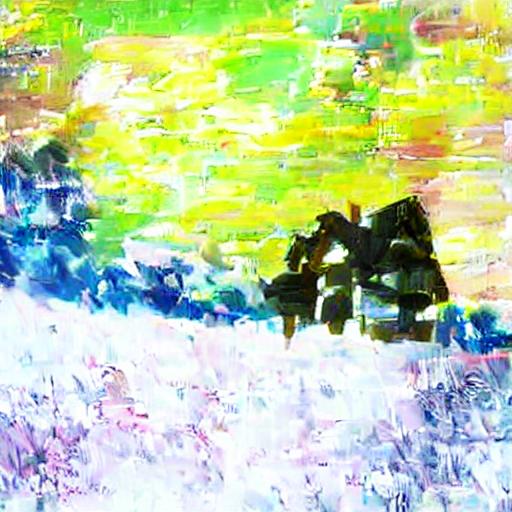}&
    \includegraphics[width=0.16\linewidth]{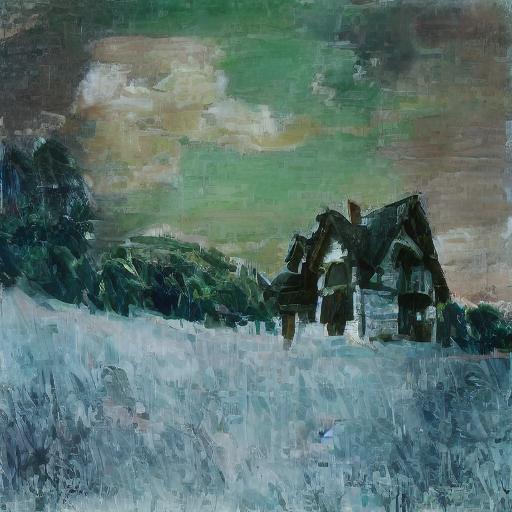}
\\
\includegraphics[width=0.16\linewidth]{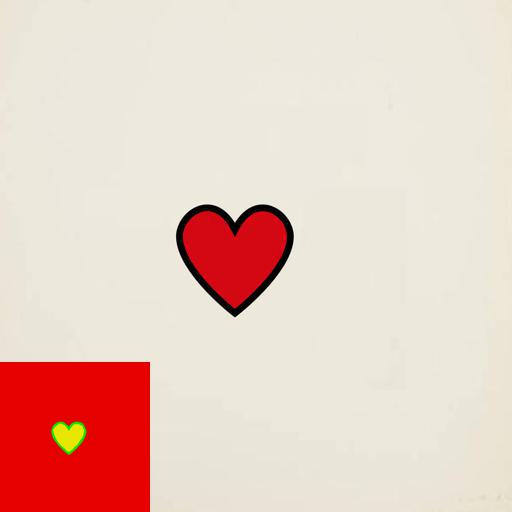} & \includegraphics[width=0.16\linewidth]{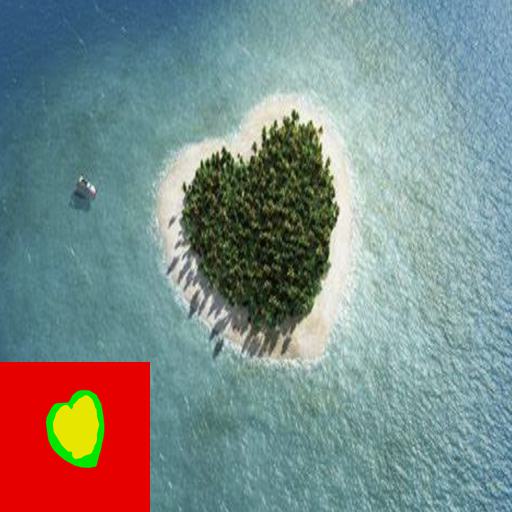}&  
  \includegraphics[width=0.16\linewidth]{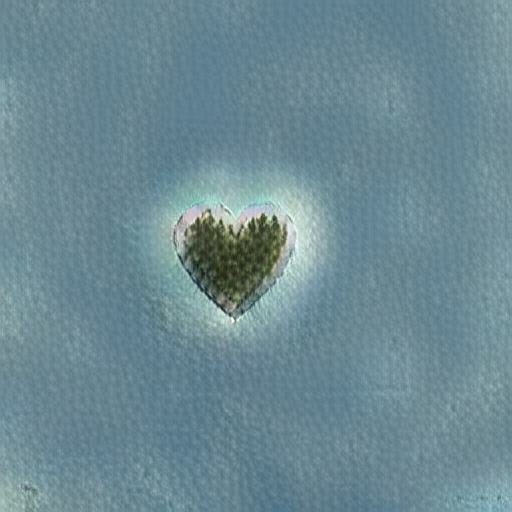} & 
\includegraphics[width=0.16\linewidth]{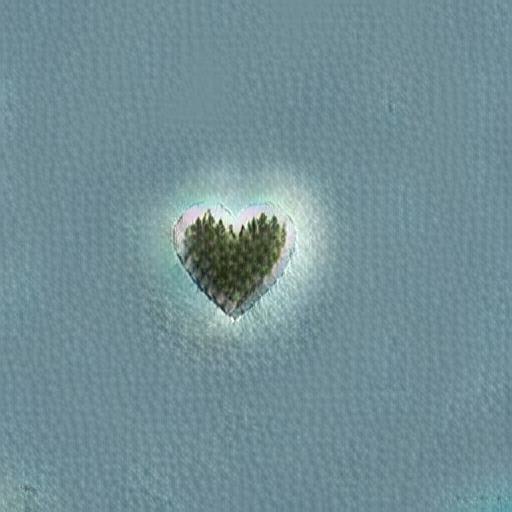} &
 \includegraphics[width=0.16\linewidth]{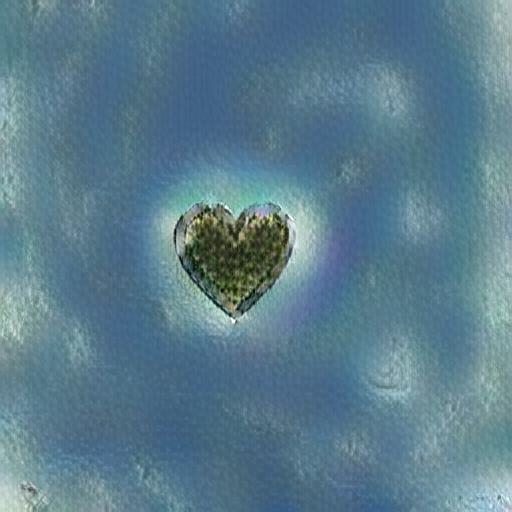} &
 \includegraphics[width=0.16\linewidth]{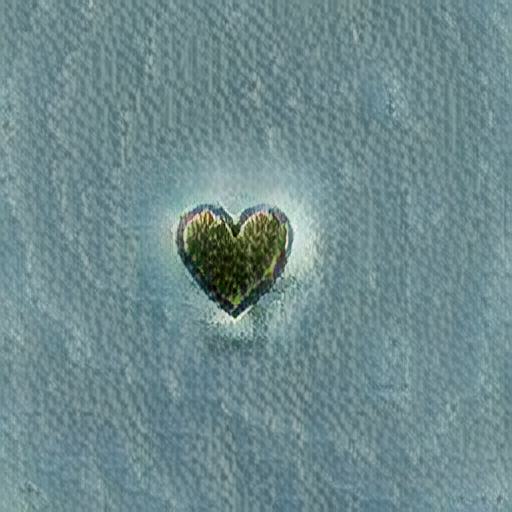} &
  \includegraphics[width=0.16\linewidth]{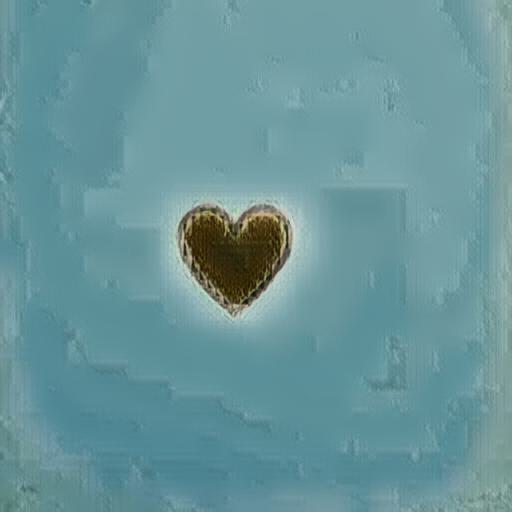}&
  \includegraphics[width=0.16\linewidth]{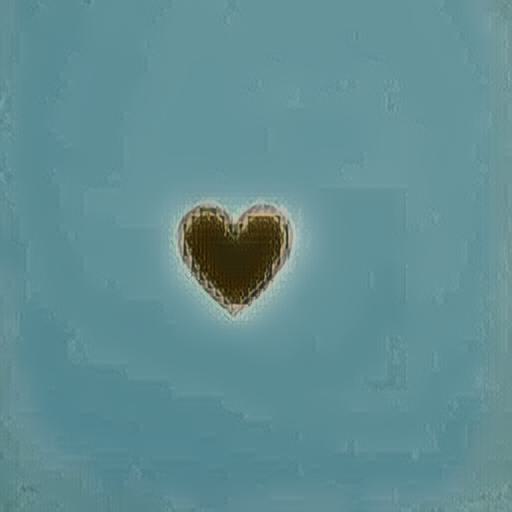}&
  \includegraphics[width=0.16\linewidth]{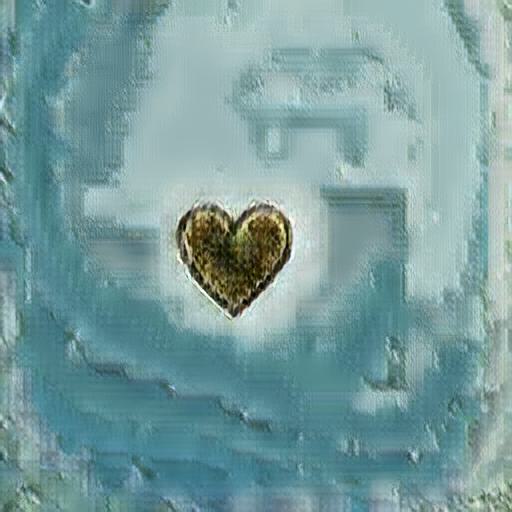}&
    \includegraphics[width=0.16\linewidth]{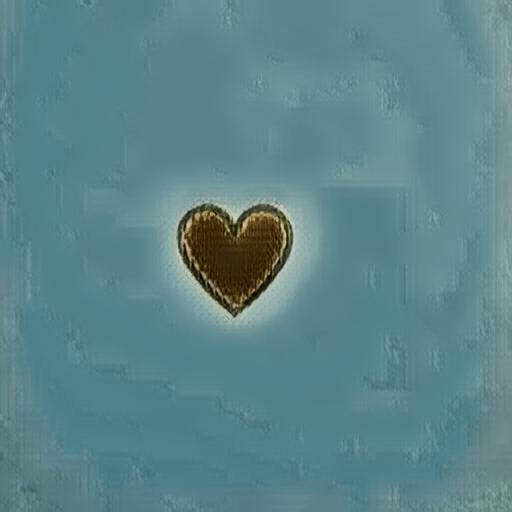}&
   \includegraphics[width=0.16\linewidth]{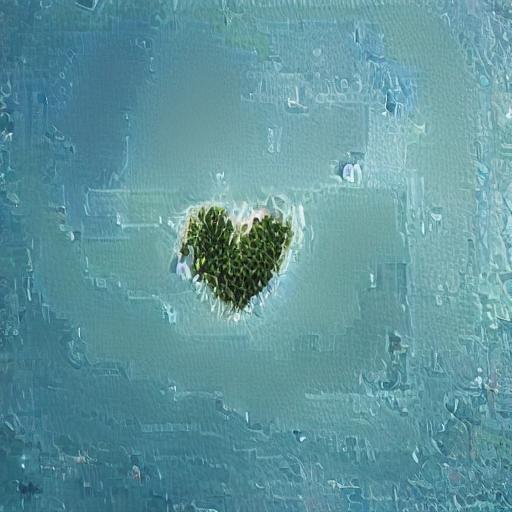}&
  \includegraphics[width=0.16\linewidth]{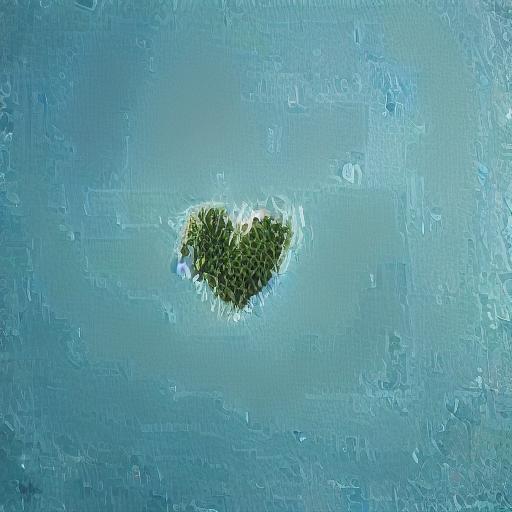}&
 \includegraphics[width=0.16\linewidth]{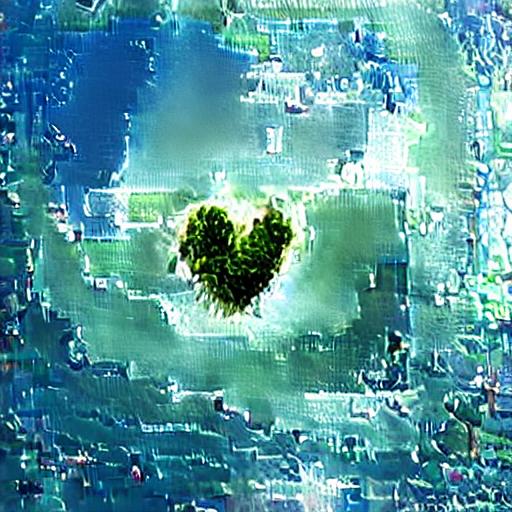}&
  \includegraphics[width=0.16\linewidth]{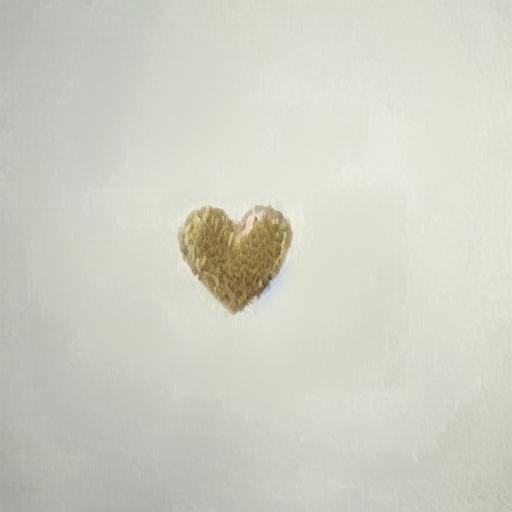}
\\
&&&&&&&&&&
\\
\multicolumn{2}{c|}{ \LARGE SSL$\downarrow$} &  \LARGE {\bf 0.8762} &  \LARGE 0.8840 &  \LARGE 0.9096 & \LARGE 0.8769 & \LARGE {\bf 1.2228} &  \LARGE 1.4157 & \LARGE 1.5714 & \LARGE 1.2832 & \LARGE {\bf 1.2447} &  \LARGE 1.3302 &  \LARGE 2.2981 & \LARGE 1.7558
\\
\multicolumn{2}{c|}{\LARGE FID$\downarrow$ \vphantom{100}} & \LARGE 13.0788 &  \LARGE 13.0814 &  \LARGE {\bf 12.0170} & \LARGE 12.0275 &  \LARGE  {\bf 12.3963} &  \LARGE 13.2059 &  \LARGE 14.1252 & \LARGE  12.4782 &  \LARGE {\bf 12.4497} &  \LARGE 13.0826 &  \LARGE 15.4705 & \LARGE 13.0359
\\
\multicolumn{2}{c|}{\LARGE CFSD$\downarrow$} & \LARGE {\bf 0.0874} &  \LARGE 0.0937 &  \LARGE 0.0994 & \LARGE 0.0922 &  \LARGE 0.0705 &  \LARGE 0.0694 & \LARGE 0.1668 & \LARGE {\bf 0.0657}&  \LARGE 0.1178 &  \LARGE  0.1066 &  \LARGE 0.3402 & \LARGE {\bf 0.0958}

\end{tabular}
}
\caption{Qualitative and quantitative ablation results on SCA, SSA, and S-AdaIN. Zoom in for better comparison.}
\label{fig:7}
\end{figure*}

In summary, SCSA is a plug-and-play solution that can be integrated into Attn-AST methods, addressing their semantic issues and offering a level of capability in semantic style transfer that surpasses that of existing SOTA methods.

\subsection{Quantitative Comparison}

We also conduct quantitative analyses for full validation.

{\it Semantic Style Loss (SSL).} Traditional style loss~\cite{huang2017arbitrary} is unsuitable for semantic style transfer. There are two main reasons: First, it focuses on the overall region, making it difficult to capture the semantic region style; second, the inconsistency in the sizes of the semantic regions between the style and stylized images affects the calculation of global style loss. Hence, we propose a semantic style loss to calculate each region style loss similar to Eq.~\ref{eq:8} using the style loss of ~\cite{huang2017arbitrary}. From the $1st$ row of Tab.~\ref{tab:1}, it is clear that the Attn-AST methods utilizing SCSA not only achieve the lowest SSL compared to traditional Attn-AST methods, but when SANet incorporates SCSA, it also reaches the best performance among all semantic style transfer methods. This indicates that SCSA can effectively facilitate semantic style transfer with its effectiveness and generalization.

{\it Fréchet Inception Distance (FID).} Similar to~\cite{chung2024style}, we use FID~\cite{heusel2017gans} to evaluate the overall style fidelity between the stylized and style images. As indicated by the $2nd$ row of Tab.~\ref{tab:1}, the Attn-AST methods integrated with SCSA achieve the lowest FID value as opposed to typical Attn-AST methods. Particularly, StyTR$^2$ with SCSA secures the optimal FID value among all the semantic style transfer methods. The above indicates that SCSA not only conducts the semantic style transfer but also markedly improves the overall style quality of the stylized images.

{\it Content Feature Structural Distance (CFSD).} As discussed in~\cite{chung2024style}, content loss~\cite{huang2017arbitrary} and LPIPS~\cite{zhang2018unreasonable} are affected by the stylization. Thus, we use the content feature structural distance~\cite{chung2024style} to assess content retention. As shown in the $3rd$ row of Tab.~\ref{tab:1}, the SCSA achieves the lowest CFSD value on CNN-based and Transformer-based methods. Although the CFSD value is slightly higher on the Diffusion-based approach with SCSA, both quantitative and qualitative results indicate that SCSA greatly improves semantic stylization, making this trade-off worthwhile. Besides, StyTR$^2$ with SCSA has the lowest CFSD value in all SOTA methods, showcasing its content preservation advantage.

{\it User Study.} Our user study is divided into two parts: the first compares traditional Attn-AST methods with SCSA-enhanced ones, and the second evaluates Attn-AST methods with SCSA against existing semantic style transfer methods. As shown in the $4th$ row of Tab.~\ref{tab:1}, the stylized images of the Attn-AST methods with SCSA are more favored by the public, and their popularity surpasses that of the SOTA methods. These results prove the superiority of SCSA.

\subsection{Ablation Study}

We conduct ablation experiments to verify the crucial roles of SCA and SSA and to affirm the potency of S-AdaIN. 

{\bf Effect of SCA and SSA.} We use SCA or SSA individually as the feature transformation module to assess their capabilities. As shown in Fig.~\ref{fig:7}, SCA alone facilitates overall color transfer in corresponding semantic regions but results in texture loss and reduced stylization, as reflected by the higher SSL and FID values. In contrast, SSA can effectively transfer specific textures but introduces color discrepancies between the semantic regions of the stylized and style images. This indicates that while SCA captures the overall style characteristics, SSA is crucial for transferring specific textures, highlighting the complementary roles of both.

{\bf Effect of S-AdaIN.} To validate the effectiveness of S-AdaIN, we conducted ablation experiments by removing S-AdaIN. As shown in Fig.~\ref{fig:7}, without S-AdaIN, the stylized images exhibit less accurate global style transfer in semantic regions and reduced texture details, particularly evident for the StyleID method. Although S-AdaIN was not employed, CFSD values of StyTR$^2$ and StyleID with SCSA are lower, which comes at the expense of stylization quality. Therefore, after considering the pros and cons, S-AdaIN is essential and effective for semantic style transfer.

\section{Conclusion}

In this paper, we highlight the challenge faced by existing arbitrary style transfer methods, with a particular emphasis on attention-based approaches, and provide an in-depth analysis of this issue, which is the failure to account for the relationship between local regions and semantic regions. To address this issue and generate high-quality stylized images that meet semantic needs, we propose a plug-and-play semantic continuous-sparse attention, dubbed SCSA. It achieves precise semantic style transfer by allowing each query point to consider certain key points in the corresponding semantic region. Inclusive experimental results demonstrate that our SCSA is effective and generalized.

{
    \small
    \bibliographystyle{ieeenat_fullname}
    \bibliography{main}
}

\clearpage
\setcounter{page}{1}
\maketitlesupplementary
\appendix

\Urlmuskip=-5mu plus 4mu

\section{Societal Impact}
To achieve a thorough understanding of the broader implications of our study, we assess the proposed method in terms of both positive and negative impacts, followed by suggestions to mitigate any adverse effects.

{\bf Positive Impact.} Our proposed method offers several societal benefits across different sectors. (1) This method provides researchers with innovative tools, encouraging them to explore challenges in semantic style transfer with fresh perspectives. By promoting the development of simple and effective solutions, our approach plays a role in advancing scientific discovery and accelerating innovation. (2) Artists can leverage this method to enhance their creative workflow and optimize productivity. The tool not only allows them to explore new artistic possibilities but also supports them in creating high-quality works more efficiently. Additionally, for general users, this method offers an accessible means to produce visually appealing outputs that align with semantic needs, enriching their creative experience. (3) By democratizing artistic tools and lowering barriers to creating stylistically unique digital art, this method encourages wider participation in creative expression. Such accessibility can lead to increased cultural exchange, as individuals from diverse backgrounds share and interpret art through new stylistic forms. This ultimately strengthens social cohesion and broadens avenues for cultural representation.

{\bf Negative Impact.} While beneficial, the method also presents potential risks that warrant attention. (1) As more people adopt AI-assisted tools, there may be a gradual erosion of unique artistic expression, with creators becoming reliant on automated processes instead of traditional, personal techniques. This could lead to a homogenization of digital art, where distinct styles and individual creativity are diminished. (2)  In the wrong hands, this technology could be utilized to produce manipulated images or videos that appear authentic, fostering misinformation. Misuse of stylization techniques in this manner poses ethical concerns and risks misleading audiences if safeguards are not established.

{\bf Mitigation Strategies.} To counteract these negative impacts, we propose the following measures: (1) By supporting customization features and encouraging users to incorporate personal touches in AI-assisted creations, we aim to preserve artistic diversity and prevent homogenization. Enabling flexibility in the method’s application allows artists to maintain their unique creative identities while benefiting from the technology. (2) Educating users on the ethical implications and potential misuse of stylization techniques is vital. By fostering an understanding of responsible AI use, we can mitigate the risk of harmful applications and promote ethical practices.

\section{Used Assets}
We utilize the following assets for our experiments:

\begin{itemize}[leftmargin=2em]
    \item SANet~\cite{park2019arbitrary}: \url{https://github.com/GlebSBrykin/SANET}, MIT license.
    \item StyTR$^2$~\cite{deng2022stytr2}: \url{https://github.com/diyiiyiii/StyTR-2}, No License.
    \item StyleID~\cite{chung2024style}: \url{https://github.com/jiwoogit/StyleID}, MIT license.
    \item STROTSS~\cite{kolkin2019style}:\url{https://github.com/nkolkin13/STROTSS}, No License.
    \item MAST~\cite{huo2021manifold}:\url{https://github.com/NJUHuoJing/MAST}, No License.
    \item TR~\cite{wang2022texture}:\url{https://github.com/EndyWon/Texture-Reformer}, MIT license.
    \item DIA~\cite{liao2017visual}:\url{https://github.com/harveyslash/Deep-Image-Analogy-PyTorch}, MIT license.
    \item GLStyleNet~\cite{wang2020glstylenet}:\url{https://github.com/EndyWon/GLStyleNet}, MIT license.
\end{itemize}

To the best of our knowledge, there are no moral or ethical concerns associated with these assets. We have thoroughly reviewed their use to ensure compliance with ethical standards, confirming that their implementation aligns with responsible research practices. This careful consideration reinforces our commitment to conducting research that adheres to both ethical guidelines and scientific integrity.

\section{Application Details}
We select three representative Attn-AST methods for SCSA embedding experiments: SANet~\cite{park2019arbitrary}, built on the CNN framework, StyTR2~\cite{deng2022stytr2}, based on the Transformer architecture, and StyleID~\cite{chung2024style}, utilizing the Diffusion model. By integrating SCSA into these Attn-AST methods, we aim to investigate its performance in semantic style transfer tasks.

\subsection{SANet with SCSA}
Given a quadruple data $\{I_c, I_{csem}, I_s, I_{ssem}\}$, consisting of a content image $I_c$ and its semantic map $I_{csem}$, a style image $I_s$ and its semantic map $I_{ssem}$, and the SANet model $M=\{E, T_{UA}, D\}$, which is composed of an encoder $E$, a feature transformation module $T_{UA}$, and a decoder $D$, we aim to generate a stylized image $I_{cs}$ that meets the semantic needs by replacing $T_{UA}$ with our $T_{SCSA}$.

In the initial stage, we get the encoded quadruple features $\{F_c, F_{csem}, F_s, F_{ssem}\}$:
\begin{equation}
\begin{aligned}
F_c = E(I_c), &\; F_s = E(I_s), \\
F_{csem} = E(I_{csem}), &\; F_{sem} = E(I_{sem}),
\end{aligned}
\end{equation}
where $E$ is a VGG-19~\cite{simonyan2014very} encoder.

Subsequently, we feed the quadruple features into our feature transformation module $T_{SCSA}$. Thus, the following operations will be performed:
\begin{equation}
\begin{aligned}
  Q_1=f_q(\bar{F}_{csem}), \; &K_1=f_k(\bar{F}_{ssem}), \; V_1=f_v(F_s), \\
  & \mathcal{\bar{A}}  = G_1(Q_1^{\mathsf{T}} \otimes K_1), \\
   F_{sca} = &f_o(softmax( \mathcal{\bar{A}}) \otimes V_1),
\end{aligned}
\label{eq:21}
\end{equation}
where $\bar{F}_{x}$ represents the normalized form of the features $F_{x}$ using its mean and standard deviation. $f_q$, $f_k$, $f_v$, and $f_o$ are the projection networks in $T_{UA}$. $G_1$ is the modulation function in Eq.6 of the main paper. $F_{sca}$ are the features that contain the overall style characteristics of the corresponding semantic regions. Then, we obtain the new encoded content features using semantic adaptive instance normalization in Sec.3.1 of the main paper: 
\begin{equation}
 F_c = \;S\text{-}AdaIN(F_c, F_s),
\end{equation}
where the new content features $F_c$ partially eliminate the influence of the inherent style of the original encoded features of the content image, thereby offering a more precise query for subsequent SSA and content features with distinct stylistic characteristics of the style features. Hence, the following formulas exist for SSA:
\begin{equation}
\begin{aligned}
  Q_2=f_q(\bar{F}_{c}), \; &K_2=f_k(\bar{F}_{s}), \; V_2=f_v(F_s),\\
   &\mathcal{\bar{B}}=G_2(Q_2^{\mathsf{T}} \otimes K_2),\\
   F_{ssa} = &f_o(softmax( \mathcal{\bar{B}}) \otimes V_2),
\end{aligned}
\end{equation}
where $\bar{F}_{x}$, $f_q$, $f_k$, $f_v$, and $f_o$ are the same as those in Eq.~\ref{eq:21}. $G_2$ is the modulation function in Eq.11 of the main paper. $F_{ssa}$ are the features characterized by fine and specific style textures in the corresponding semantic regions. Then, the stylized features through the feature transformation module $T_{SCSA}$ can be obtained:
\begin{equation}
  F_{cs} = \alpha_1 \times F_{sca} + \alpha_2 \times F_{ssa} + F_{c},
\end{equation}
where $\alpha_1$ and $\alpha_2$ indicate separately the stylization degree for the overall style and vivid textures in semantic regions. We set $\alpha_1 = 0.7$ and $\alpha_2 = 0.3$ in the main paper.

Ultimately, the stylized image can be produced:
\begin{equation}
I_{cs} = D(F_{cs}),
\end{equation}
where the decoder $D$ structure mirrors that of VGG-19.

\subsection{StyTR$^2$ with SCSA}
\label{sec:c2}
Given a quadruple data $\{I_c, I_{csem}, I_s, I_{ssem}\}$, consisting of a content image $I_c$ and its semantic map $I_{csem}$, a style image $I_s$ and its semantic map $I_{ssem}$, and the StyTR$^2$ model $M=\{E_c, E_s, T_{UA}, D\}$, which is composed of a content encoder $E_c$, a style encoder $E_s$, some feature transformation transformer modules $T_{UA}$, and a decoder $D$, we aim to generate a stylized image $I_{cs}$ that meets the semantic needs by replacing $T_{UA}$ with our $T_{SCSA}$.

As a first step, we get the encoded quintuple features $\{F_c, F_{csem}, F_s^{c}, F_s^{s}, F_{ssem}\}$:
\begin{equation}
\begin{aligned}
F_c = E_c(I_c), \; F_{s}^{c} = E&_c(I_s), \; F_{s}^{s} = E_s(I_s),\\
F_{csem} = E_c(I_{csem}), &\; F_{sem} = E_s(I_{sem}),
\end{aligned}
\end{equation}
where $E_c$ and $E_s$ are the content and style transformer~\cite{vaswani2017attention} encoders, respectively.

Consequently, we feed the quintuple features into our feature transformation module $T_{SCSA}$. Thus, the following operations will be performed:
\begin{equation}
\begin{aligned}
  Q_1=f_q(F_{csem}), \; &K_1=f_k(F_{ssem}), \; V_1=f_v(F_s^{s}), \\
  & \mathcal{\bar{A}}  = G_1(Q_1^{\mathsf{T}} \otimes K_1), \\
   F_{sca} = &f_o(softmax( \mathcal{\bar{A}}) \otimes V_1),
\end{aligned}
\label{eq:27}
\end{equation}
where $f_q$, $f_k$, $f_v$, and $f_o$ are the projection networks in $T_{UA}$. $G_1$ is the modulation function in Eq.6 of the main paper. $F_{sca}$ are the features that contain the overall style characteristics of the corresponding semantic regions. Then, we obtain the new encoded content features using semantic adaptive instance normalization in Sec.3.1 of the main paper:
\begin{equation}
 F_c^{S\text{-}AdaIN} = \;S\text{-}AdaIN(F_c, F_s^c),
\end{equation}
where the new content features $F_c^{S\text{-}AdaIN}$ partially eliminate the influence of the inherent style of the original encoded features of the content image, thereby offering a more precise query for subsequent SSA and content features with distinct stylistic characteristics of the style features. Hence, the following formulas exist for SSA:
\begin{equation}
\begin{aligned}
  Q_2=f_q(F_c^{S\text{-}AdaIN}&), \; K_2=f_k(F_{s}^s), \; V_2=f_v(F_s^s),\\
   \mathcal{\bar{B}}&=G_2(Q_2^{\mathsf{T}} \otimes K_2),\\
   F_{ssa} = f_o&(softmax( \mathcal{\bar{B}}) \otimes V_2),
\end{aligned}
\end{equation}
where $f_q$, $f_k$, $f_v$, and $f_o$ are the same as those in Eq.~\ref{eq:27}. $G_2$ is the modulation function in Eq.11 of the main paper. $F_{ssa}$ are the features characterized by fine and specific style textures in the corresponding semantic regions. Then, the stylized features through the feature transformation module $T_{SCSA}$ can be obtained:
\begin{equation}
  F_{cs} = \alpha_1 \times F_{sca} + \alpha_2 \times F_{ssa} + b \times F_{c}^{S\text{-}AdaIN} + (1-b)\times F_{c},
\end{equation}
where $\alpha_1$ and $\alpha_2$ indicate separately the stylization degree for the overall style and vivid textures in semantic regions. We set $\alpha_1 = 1.2$ and $\alpha_2 = 0.5$. $b$ is used to trade off the degree of the semantic style initialization of the content features and the degree of content preservation. We set $b=0.7$. $F_{cs}$ are utilized as the new content features for the subsequent feature transformation modules. It is important to note that we focus on aligning the semantic style of content features only in the first feature transformation module, while $b=0$ is set for the remaining feature transformation modules.

Ultimately, the stylized image can be produced:
\begin{equation}
I_{cs} = D(F_{cs}),
\end{equation}
where the even-numbered transformers of the decoder $D$ are replaced with our $T_{SCSA}$.

\subsection{StyleID with SCSA}
\label{sec:c3}
Given a quadruple data $\{I_c, I_{csem}, I_s, I_{ssem}\}$, consisting of a content image $I_c$ and its semantic map $I_{csem}$, a style image $I_s$ and its semantic map $I_{ssem}$, and the StyleID model $M=\{E, U\text{-}Net, D\}$, which is composed of a encoder $E$, a denoising model $U\text{-}Net$~\cite{ronneberger2015u}, and a decoder $D$, we aim to generate a stylized image $I_{cs}$ that meets the semantic needs by replacing certain $T_{UA}$ of $U\text{-}Net$ module with $T_{SCSA}$.

First of all, we obtain the encoded quadruple features $\{X_{c0}, X_{csem0}, X_{s0}, X_{ssem0}\}$ at the time step $t=0$:
\begin{equation}
\begin{aligned}
X_{c0} = E(I_c), &\; X_{s0} = E(I_s), \\
X_{csem0} = E(I_{csem}), &\; X_{sem0} = E(I_{sem}),
\end{aligned}
\end{equation}
where $E$ is a VAE~\cite{kingma2013auto} encoder.

Following that, we acquire the noisy quadruple features $\{F_{cT}, F_{csemT}, F_{sT}, F_{ssemT}\}$ from certain $T_{UA}$ layers of $U\text{-}Net$ at the time step $t=T$ via DDIM inversion~\cite{song2020denoising}:
\begin{equation}
\begin{aligned}
F_{cT} = DD\text{-}SA(X_{c0}), &\; F_{sT} = DD\text{-}SA(X_{s0}), \\
F_{csemT} = DD\text{-}SA(X_{csem0}), &\; F_{semT} = DD\text{-}SA(X_{sem0}),
\end{aligned}
\label{eq:34}
\end{equation}
where $DD\text{-}SA$ represents the DDIM inversion and extracting features from the $T_{UA}$ layers.

As a next step, we obtain the new noisy content features using semantic adaptive instance normalization in Sec.3.1 of the main paper: 
\begin{equation}
 F_{cT} = \;S\text{-}AdaIN(F_{cT}, F_{sT}),
\end{equation}
where the new features $F_{cT}$ partially eliminate the influence of the inherent style of the original noisy features. This provides a more precise query for the subsequent SCA and SSA, along with content features that exhibit distinct stylistic characteristics from the style features.

Following this, we feed the noisy quadruple features into our feature transformation module $T_{SCSA}$. Thus, the following operations will be performed:
\begin{equation}
\begin{aligned}
  Q_1=f_q(t_1 & \times \bar{F}_{csemT} + (1-t_1) \times \bar{F}_{cT}), \\
  K_1=f_k(t_1 & \times \bar{F}_{ssemT} + (1-t_1) \times \bar{F}_{sT}),\\
  & \quad V_1=f_v(F_{sT}), \\
  & \mathcal{\bar{A}}  = G_1(Q_1^{\mathsf{T}} \otimes K_1), \\
   F_{sca} = &f_o(softmax( \mathcal{\bar{A}}) \otimes V_1),
\end{aligned}
\end{equation}
where $\bar{F}_{x}$ represents the normalized form of the features $F_{x}$ using its normalized networks in $T_{UA}$. $t_1$ represents the trade-off between semantic stylization and content preservation. We set $t_1 = 0.3$. $f_q$, $f_k$, $f_v$, and $f_o$ are the projection networks in $T_{UA}$. $G_1$ is the modulation function in Eq.6 of the main paper. $F_{sca}$ are the features that contain the overall style characteristics of the corresponding semantic regions. Similarly, the following formulas exist for SSA:
\begin{equation}
\begin{aligned}
  Q_2&=f_q(t_2 \times \bar{F}_{cT} + (1-t_2) \times \tilde{F}_T),\\ \; &K_2=f_k(\bar{F}_{sT}), \; V_2=f_v(F_{sT}),\\
  & \quad\; \mathcal{\bar{B}}=G_2(Q_2^{\mathsf{T}} \otimes K_2),\\
 & F_{ssa} = f_o(softmax( \mathcal{\bar{B}}) \otimes V_2),
\end{aligned}
\end{equation}
where $\bar{F}_{x}$, $f_q$, $f_k$, $f_v$, and $f_o$ are the same as those in Eq.~\ref{eq:34}. $F_T$ represent the input features of $T_{UA}$. $\tilde{F}_{T}$ represents the normalized form of the features $F_T$ using its mean and standard deviation. $t_2$ represents the trade-off between content preservation and semantic stylization, similar to that in~\cite{chung2024style}. We set $t_2=0.5$. $G_2$ is the modulation function in Eq.11 of the main paper. $F_{ssa}$ are the features characterized by fine and specific style textures in the corresponding semantic regions. Then, the stylized features through the feature transformation module $T_{SCSA}$ can be obtained:
\begin{equation}
  F_{cs} = \alpha_1 \times F_{sca} + \alpha_2 \times F_{ssa} + F_{cT},
\end{equation}
where $\alpha_1$ and $\alpha_2$ indicate separately the stylization degree for the overall style and vivid textures in semantic regions. We set $\alpha_1 = 0.8$ and $\alpha_2 = 0.2$ in the main paper. It is worth noting that we use the features $F_{cT}$ processed by semantic adaptive instance normalization only at the time step $t=T$, while replacing it with the input features of $T_{UA}$ at other time steps.

Ultimately, the stylized image can be produced:
\begin{equation}
I_{cs} = D(X_{cs0}),
\end{equation}
where $X_{cs0}$ represents the output of the $U\text{-}Net$ at the time step $t=0$ through the above DDIM sample. $D$ is a VAE~\cite{kingma2013auto} decoder.

\section{Parameter Analysis}
To achieve a more comprehensive understanding of the effects of experimental parameters, we also perform additional experiments that delve into their specific roles and contributions, providing deeper insights into the effectiveness of our method across various conditions.

{\bf Overall Style-Local Texture Intensity.} Our SCSA enables dynamic adjustment of the intensity of overall style and local textures in corresponding semantic regions. To demonstrate this, we conduct relevant experiments for comprehensive analyses. 

\begin{figure}
\centering
\resizebox{0.435\textwidth}{!}{
\setlength{\tabcolsep}{0.005cm} 
\renewcommand{\arraystretch}{0.25}  
\begin{tabular}{>{\centering\arraybackslash}m{0.35cm} 
 >{\centering\arraybackslash}m{1.6cm} >{\centering\arraybackslash}m{1.6cm} >{\centering\arraybackslash}m{1.6cm} >{\centering\arraybackslash}m{1.6cm} >{\centering\arraybackslash}m{1.6cm}}
  &&  Content & & Style &
 \\
 &&  \includegraphics[width=1.0\linewidth]{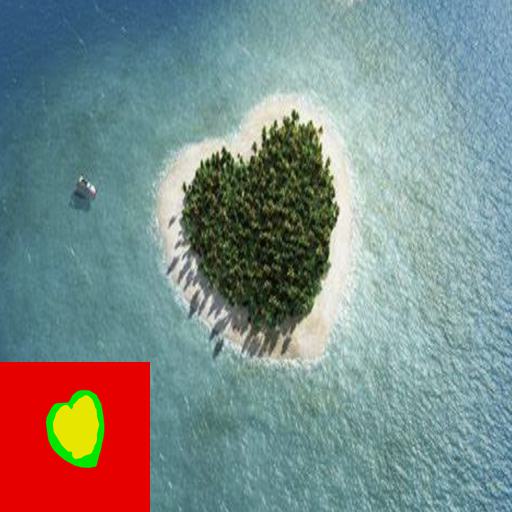} & & \includegraphics[width=1.0\linewidth]{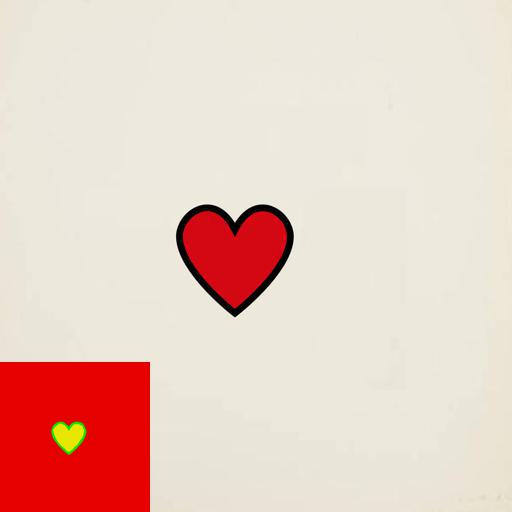} &
 \\
  & $ \alpha_{1} = 0.4$ &  $0.8$ &  $1.2$ &  $1.6$ &  $2.0$
 \\
\rotatebox[origin=rb]{270}{$\alpha_{2} = 0.1$} & 
 \includegraphics[width=1.0\linewidth]{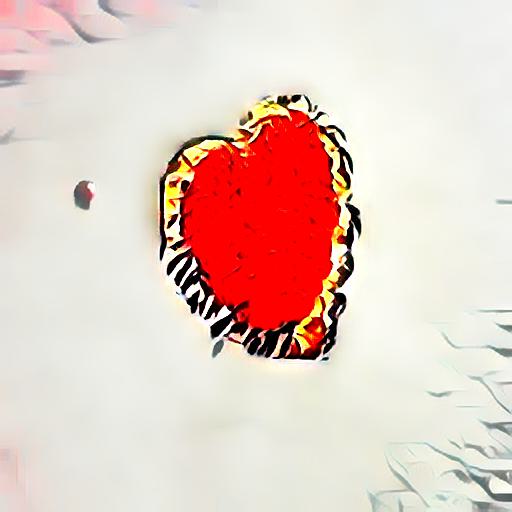} & \includegraphics[width=1\linewidth]{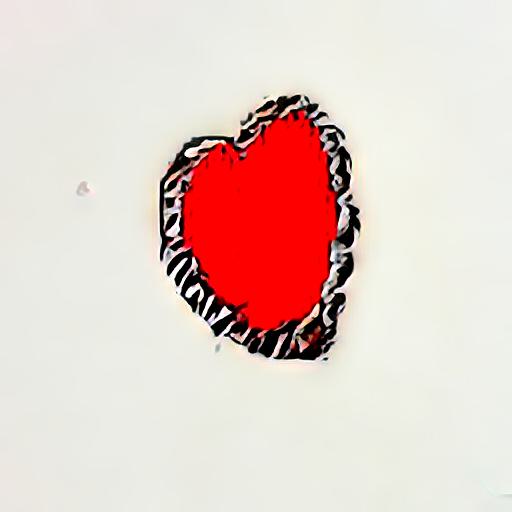} & \includegraphics[width=1\linewidth]{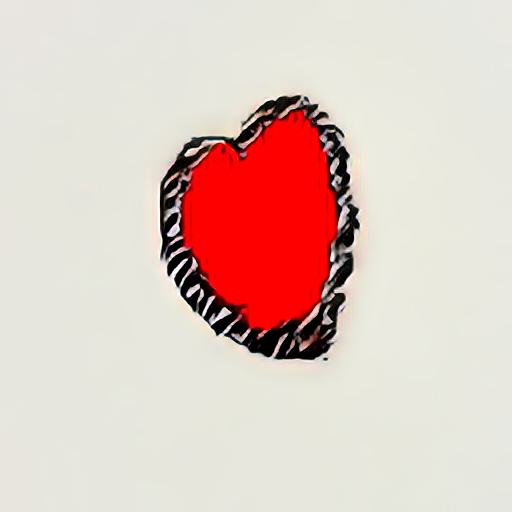} & \includegraphics[width=1\linewidth]{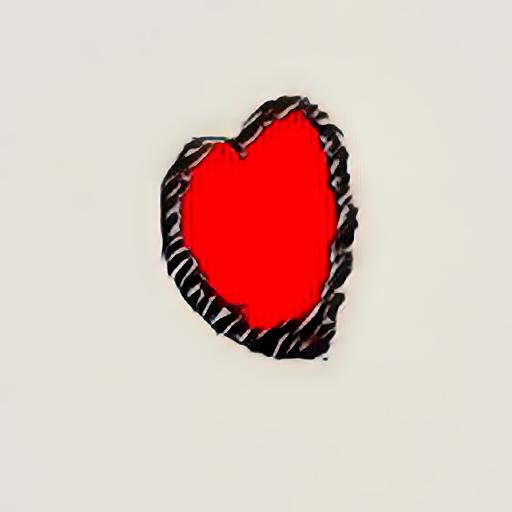} & \includegraphics[width=1\linewidth]{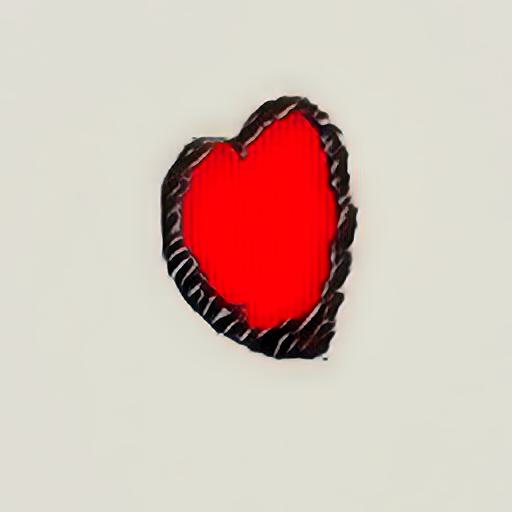} \\
 \rotatebox[origin=rb]{270}{$0.3$} &
 \includegraphics[width=1\linewidth]{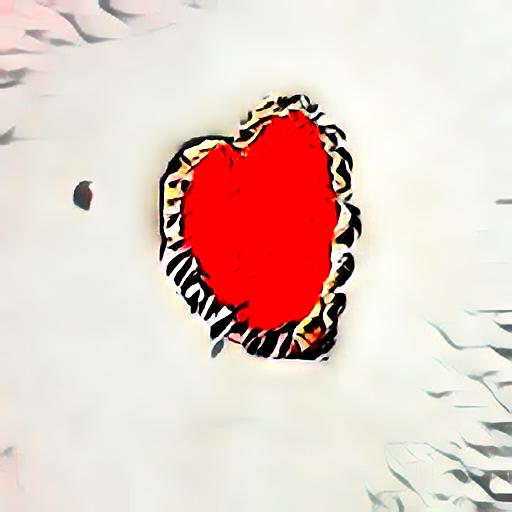} & \includegraphics[width=1\linewidth]{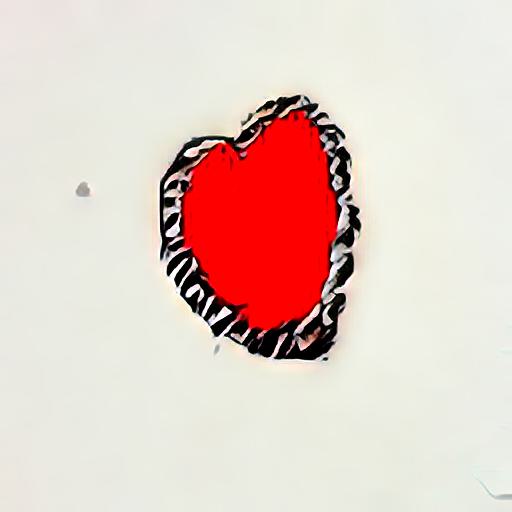} & \includegraphics[width=1\linewidth]{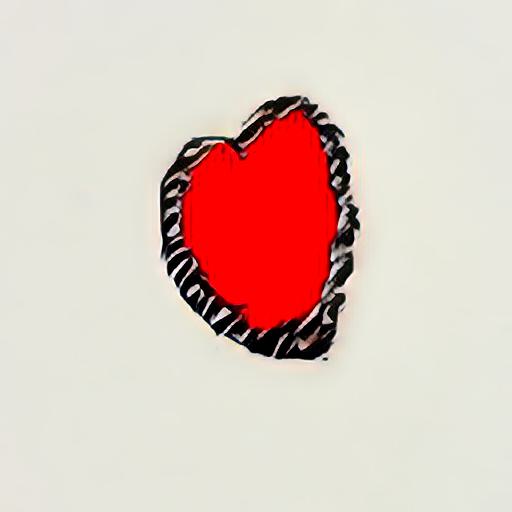} & \includegraphics[width=1\linewidth]{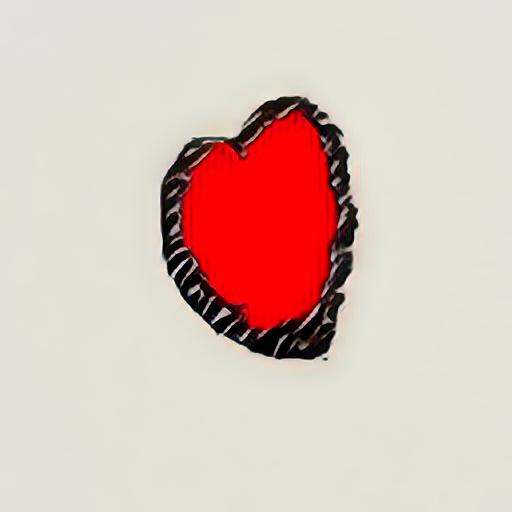} & \includegraphics[width=1\linewidth]{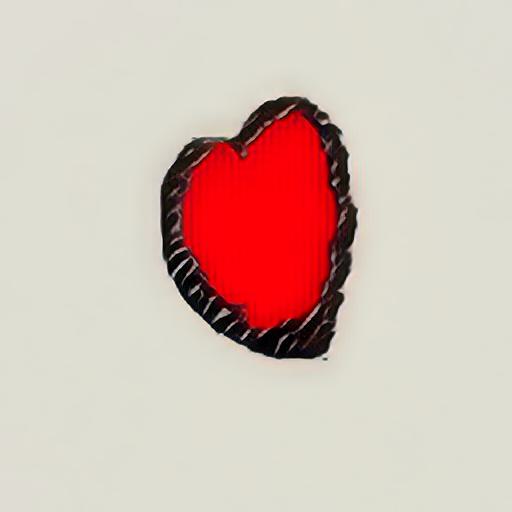} \\
  \rotatebox[origin=rb]{270}{$0.5$} & 
 \includegraphics[width=1\linewidth]{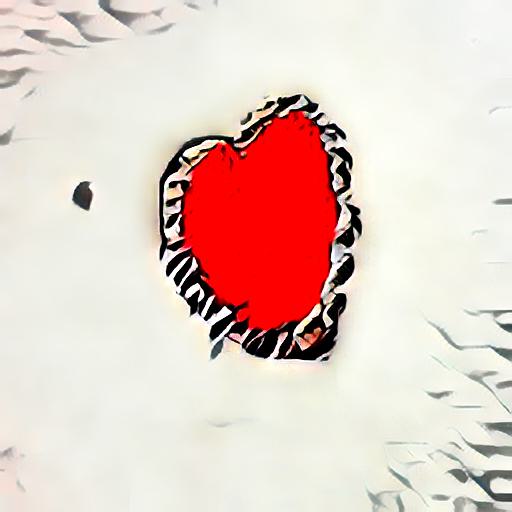} & \includegraphics[width=1\linewidth]{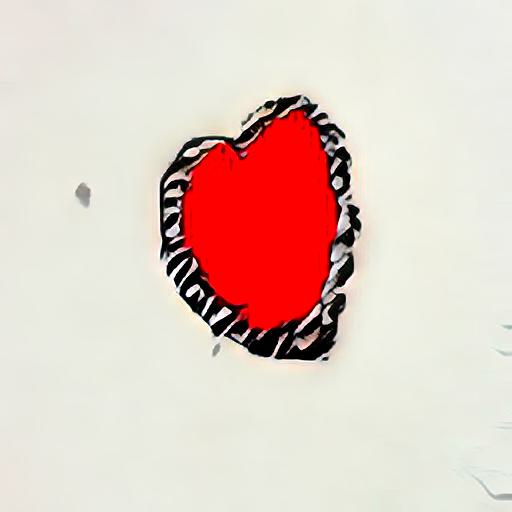} & \includegraphics[width=1\linewidth]{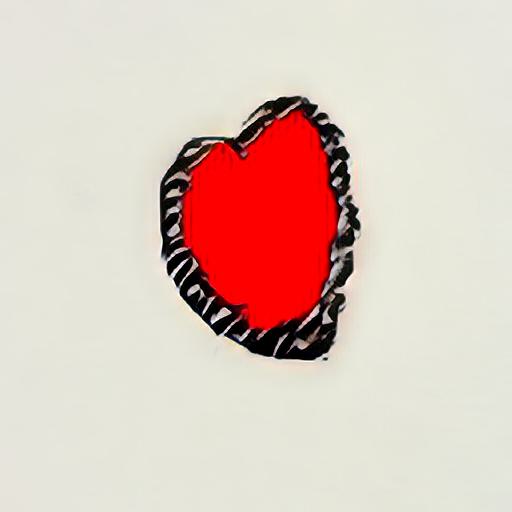} & \includegraphics[width=1\linewidth]{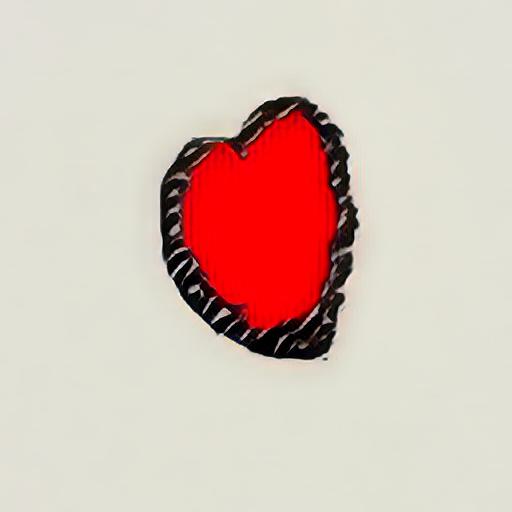} & \includegraphics[width=1\linewidth]{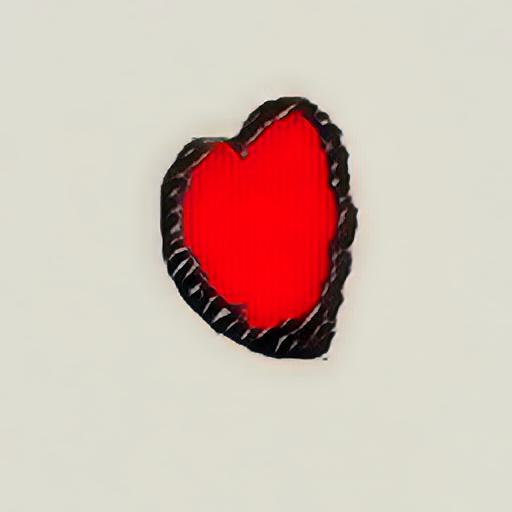} \\
    \rotatebox[origin=rb]{270}{$0.7$} &
 \includegraphics[width=1\linewidth]{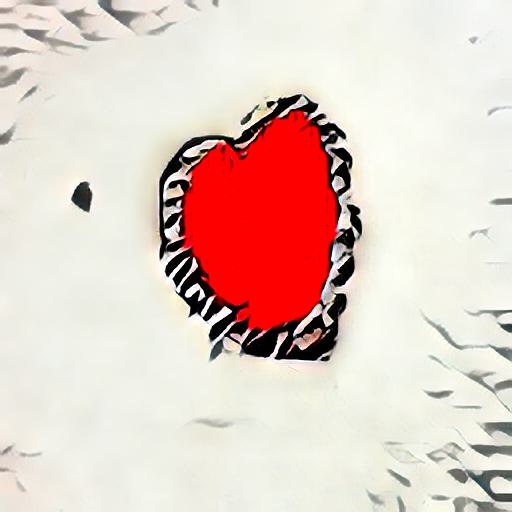} & \includegraphics[width=1\linewidth]{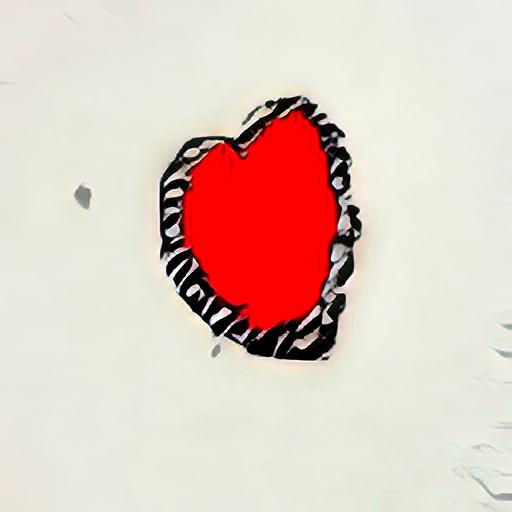} & \includegraphics[width=1\linewidth]{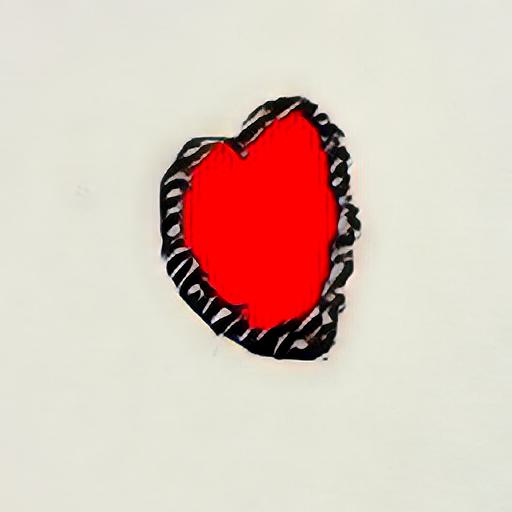} & \includegraphics[width=1\linewidth]{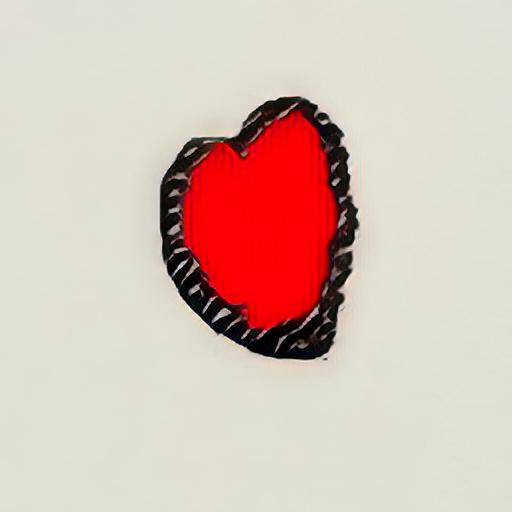} & \includegraphics[width=1\linewidth]{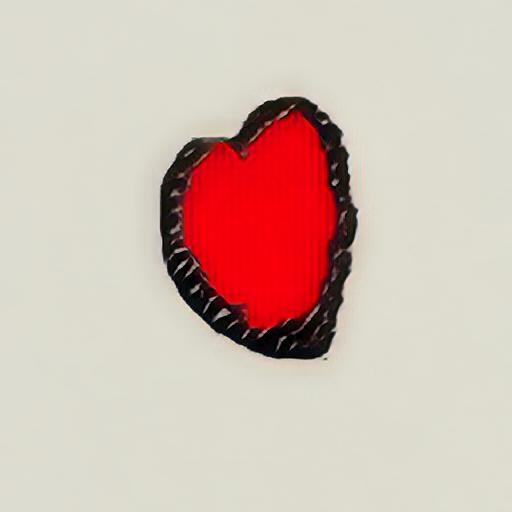} \\
      \rotatebox[origin=rb]{270}{$0.9$} & \includegraphics[width=1\linewidth]{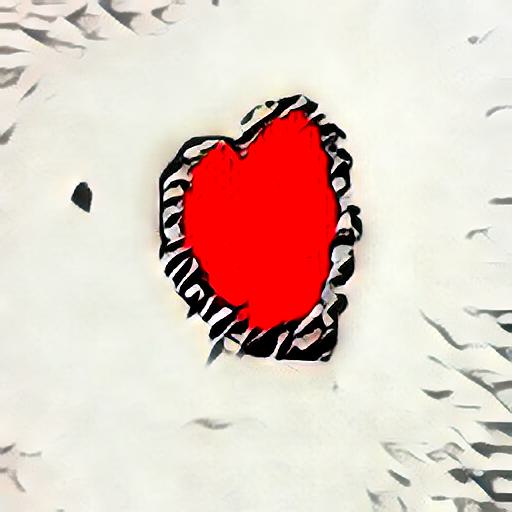} & \includegraphics[width=1\linewidth]{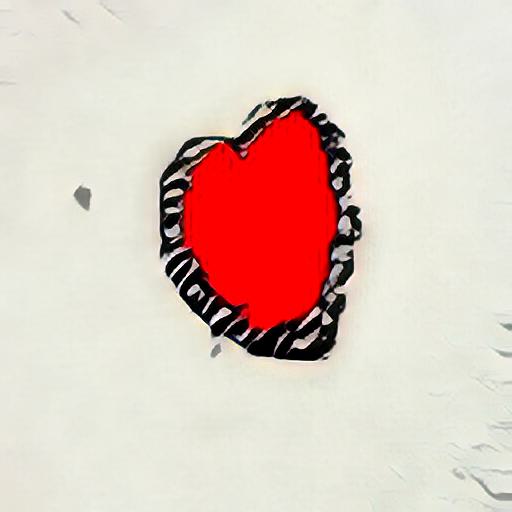} & \includegraphics[width=1\linewidth]{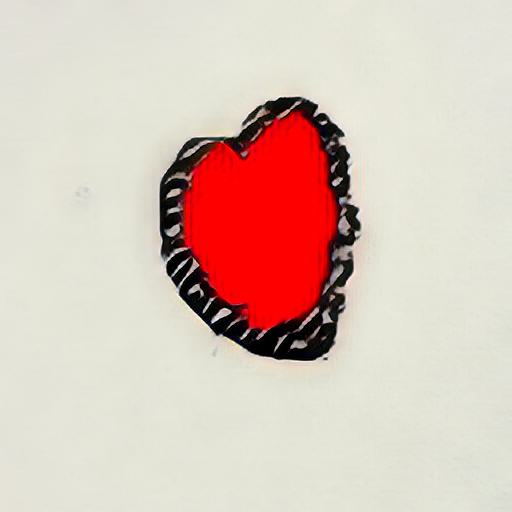} & \includegraphics[width=1\linewidth]{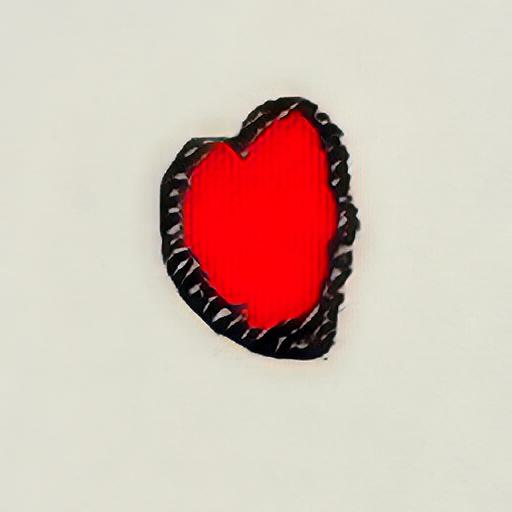} & \includegraphics[width=1\linewidth]{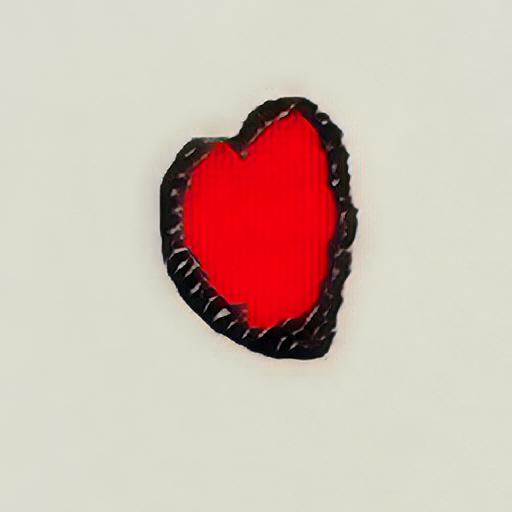}\\
\end{tabular}}
\caption{Comparisons of the overall style and local texture intensity in semantic regions of StyTR$^2$ embedded with our SCSA.}
\label{fig:9}
\end{figure}

\begin{figure}
\centering
\resizebox{0.435\textwidth}{!}{
\setlength{\tabcolsep}{0.005cm} 
\renewcommand{\arraystretch}{0.25}  
\begin{tabular}{>{\centering\arraybackslash}m{0.35cm} 
 >{\centering\arraybackslash}m{1.6cm} >{\centering\arraybackslash}m{1.6cm} >{\centering\arraybackslash}m{1.6cm} >{\centering\arraybackslash}m{1.6cm} >{\centering\arraybackslash}m{1.6cm}}
  &&  Content & & Style &
 \\
 &&  \includegraphics[width=1.0\linewidth]{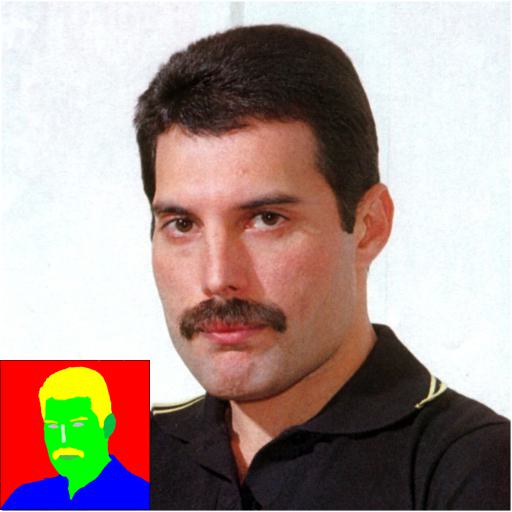} & & \includegraphics[width=1.0\linewidth]{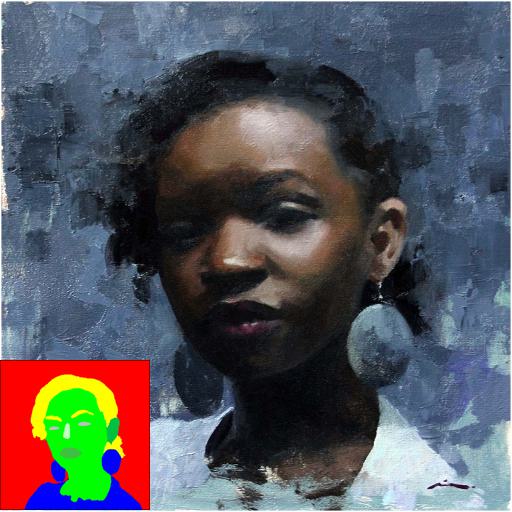} &
 \\
  & $ \alpha_{1} = 0.2$ &  $0.4$ &  $0.6$ &  $0.8$ &  $1.0$
 \\
\rotatebox[origin=rb]{270}{$\alpha_{2} = 0.1$} & 
 \includegraphics[width=1.0\linewidth]{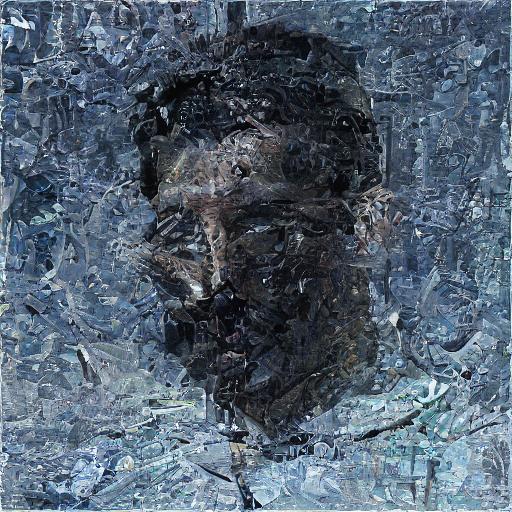} & \includegraphics[width=1\linewidth]{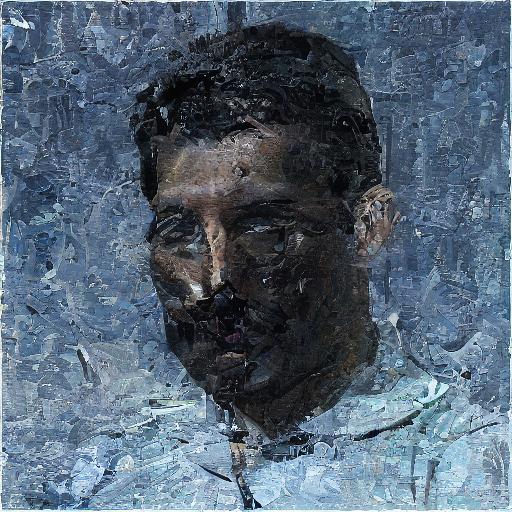} & \includegraphics[width=1\linewidth]{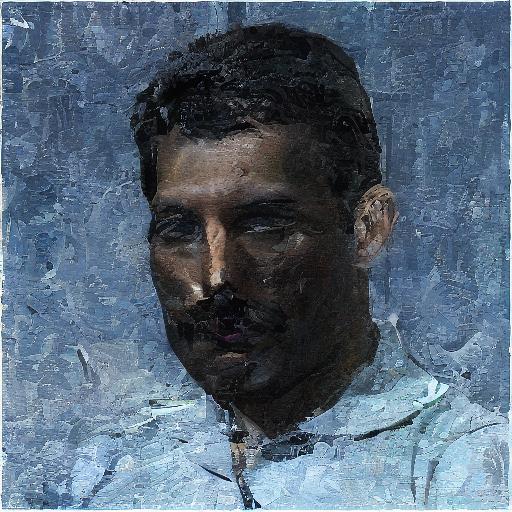} & \includegraphics[width=1\linewidth]{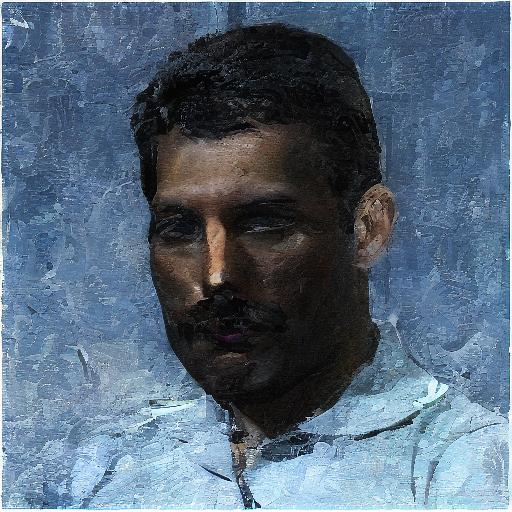} & \includegraphics[width=1\linewidth]{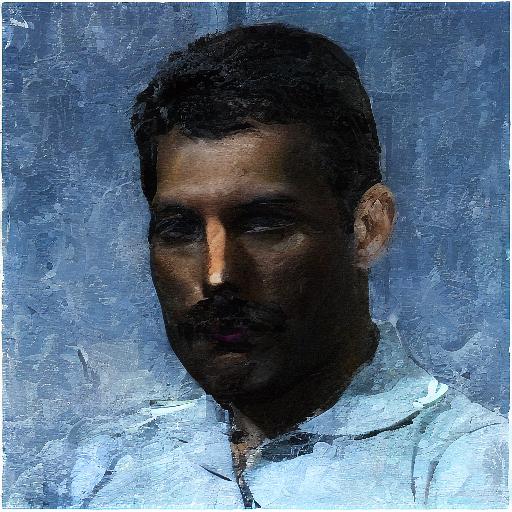} \\
 \rotatebox[origin=rb]{270}{$0.2$} &
 \includegraphics[width=1.0\linewidth]{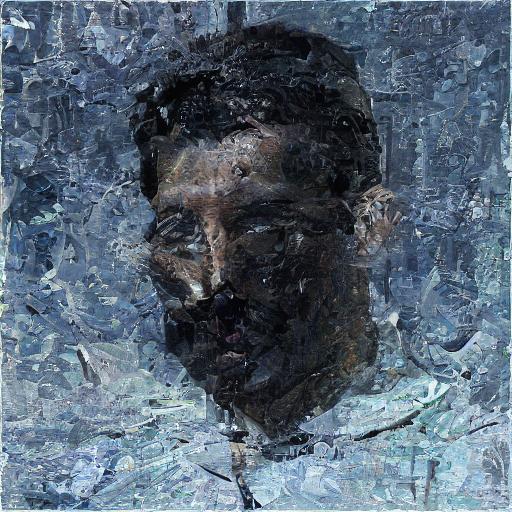} & \includegraphics[width=1\linewidth]{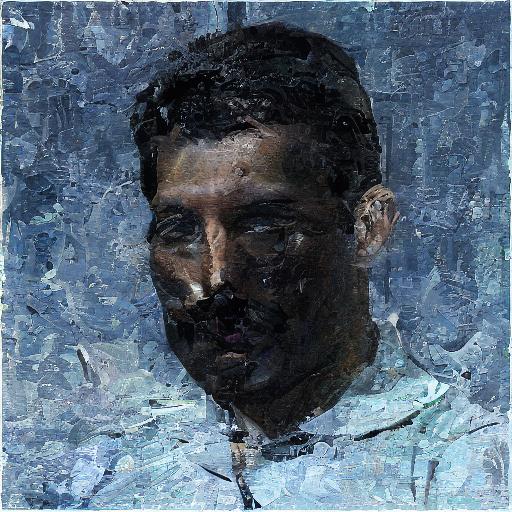} & \includegraphics[width=1\linewidth]{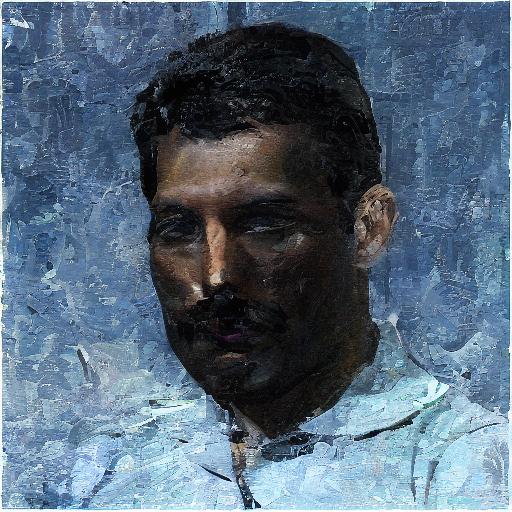} & \includegraphics[width=1\linewidth]{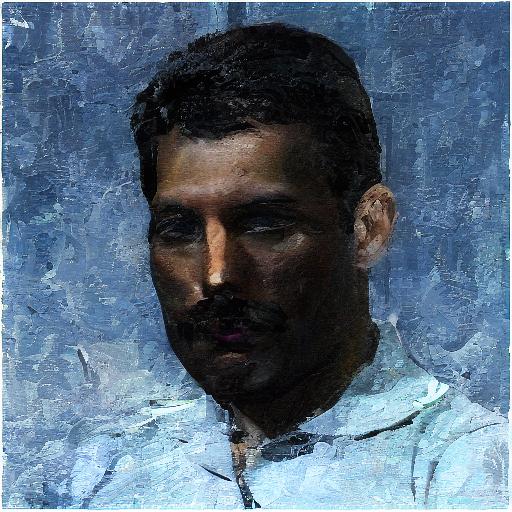} & \includegraphics[width=1\linewidth]{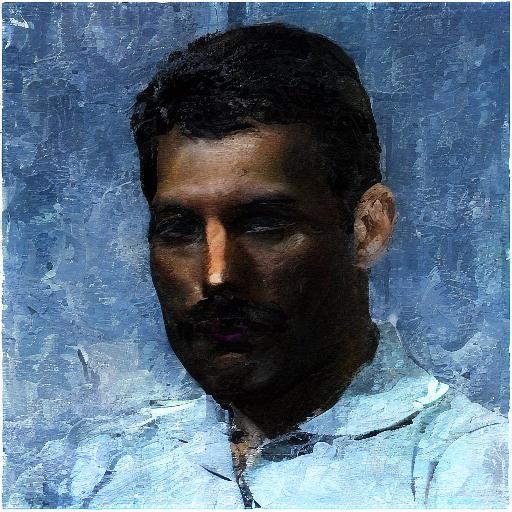} \\
  \rotatebox[origin=rb]{270}{$0.3$} & 
 \includegraphics[width=1.0\linewidth]{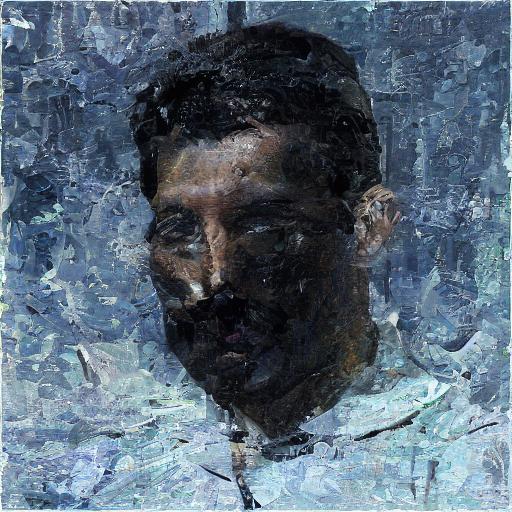} & \includegraphics[width=1\linewidth]{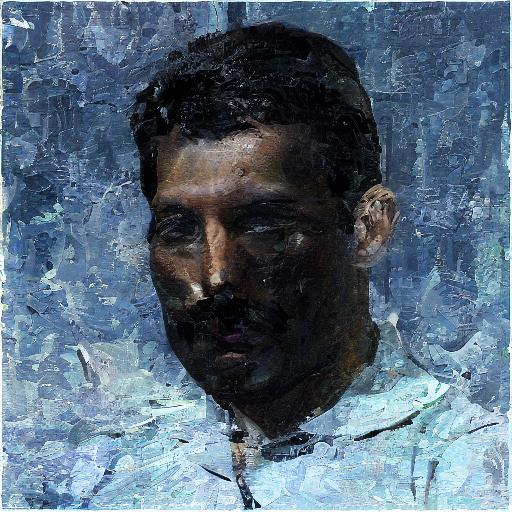} & \includegraphics[width=1\linewidth]{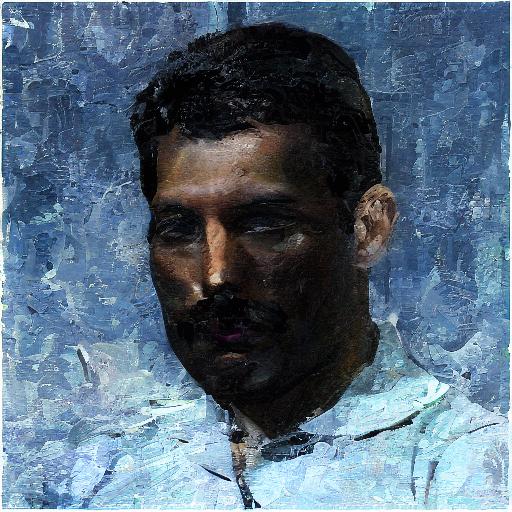} & \includegraphics[width=1\linewidth]{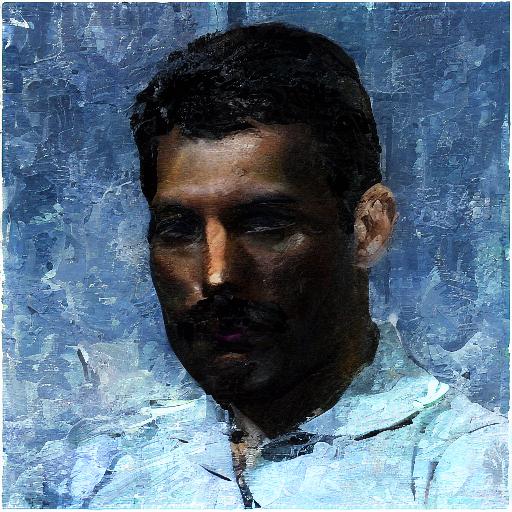} & \includegraphics[width=1\linewidth]{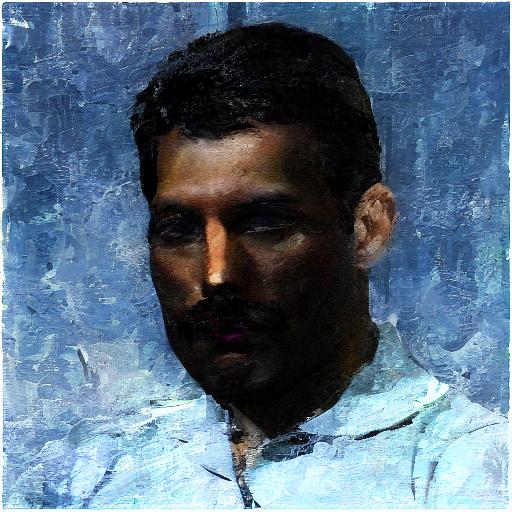} \\
    \rotatebox[origin=rb]{270}{$0.4$} &
 \includegraphics[width=1.0\linewidth]{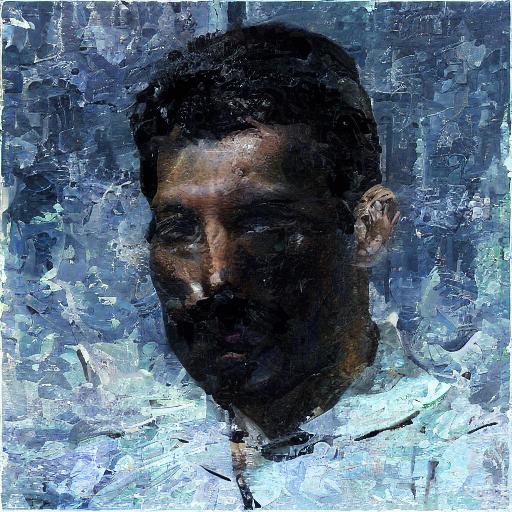} & \includegraphics[width=1\linewidth]{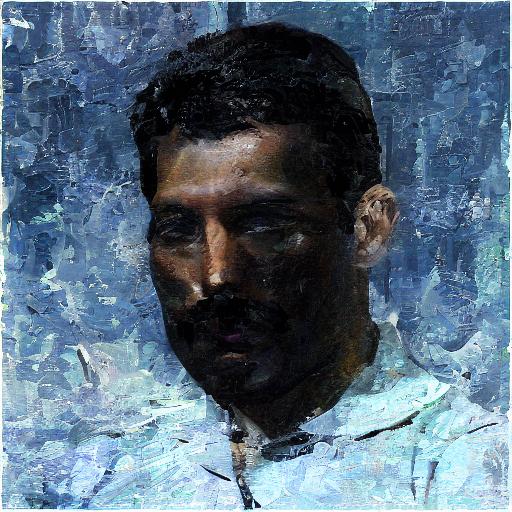} & \includegraphics[width=1\linewidth]{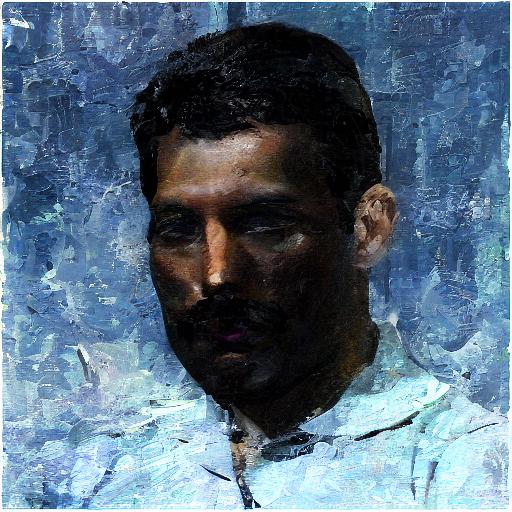} & \includegraphics[width=1\linewidth]{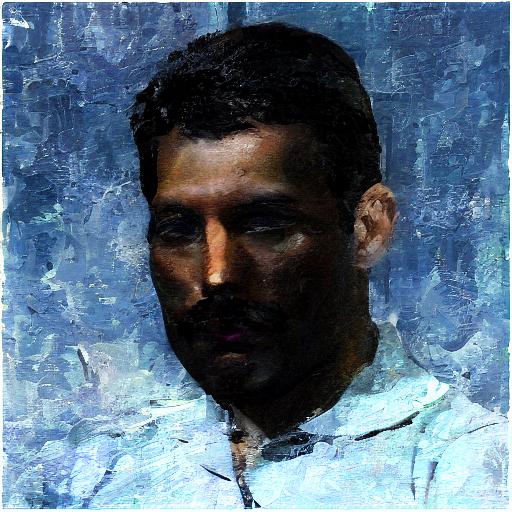} & \includegraphics[width=1\linewidth]{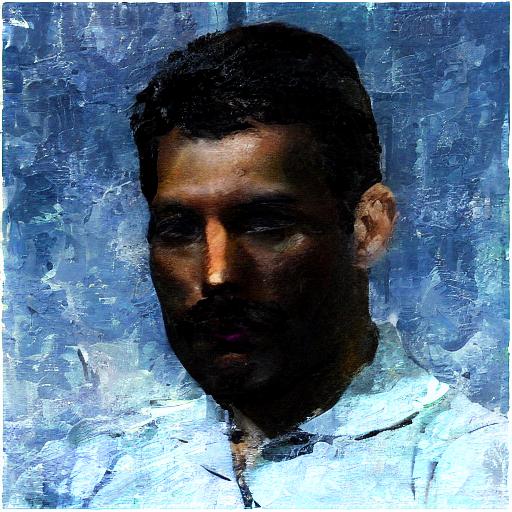} \\
      \rotatebox[origin=rb]{270}{$0.5$} &  \includegraphics[width=1.0\linewidth]{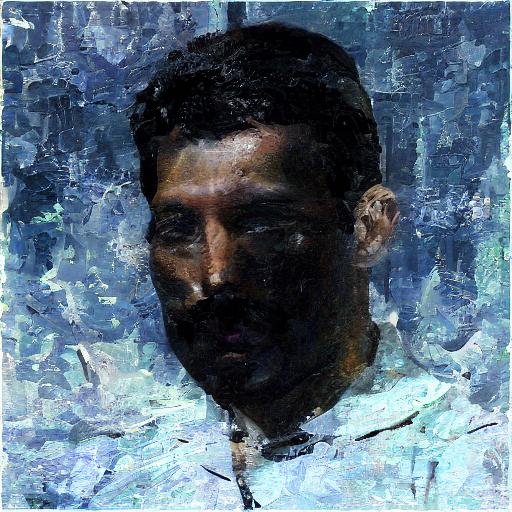} & \includegraphics[width=1\linewidth]{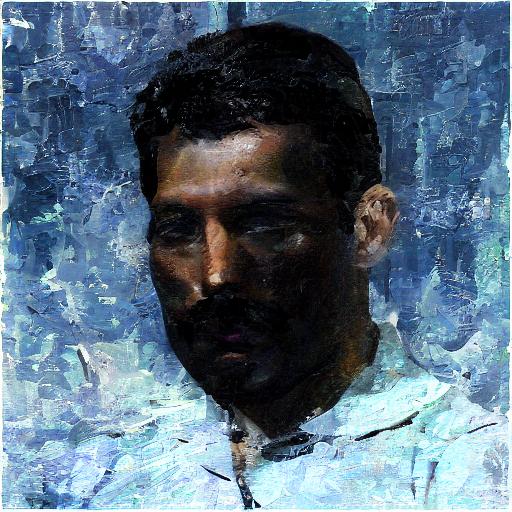} & \includegraphics[width=1\linewidth]{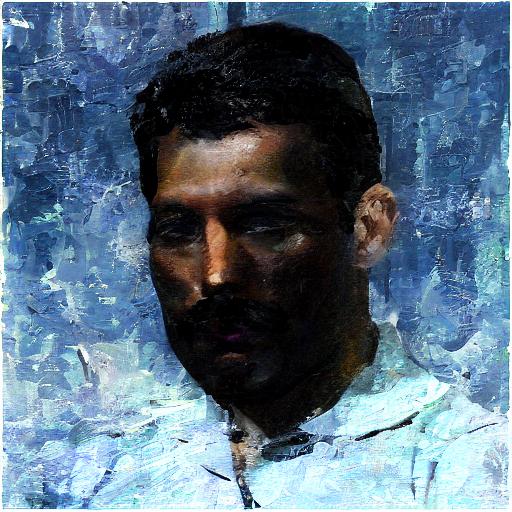} & \includegraphics[width=1\linewidth]{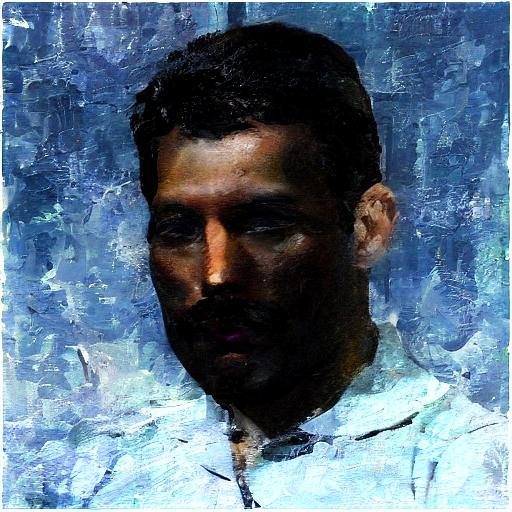} & \includegraphics[width=1\linewidth]{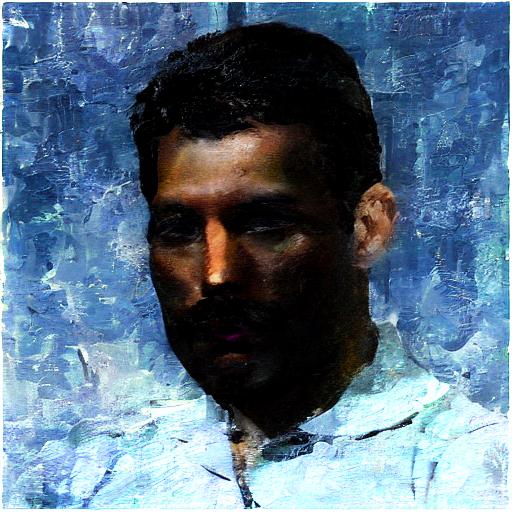} \\
\end{tabular}}
\caption{Comparisons of the overall style and local texture intensity in semantic regions of StyleID embedded with our SCSA.}
\label{fig:10}
\end{figure}

As shown in Fig.~\ref{fig:8}, Fig~\ref{fig:9}, and Fig~\ref{fig:10} an increase in the parameter $\alpha_1$ enhances the overall style expression of the semantic regions. In contrast, a rise in parameter $\alpha_2$ brings greater clarity to the textures within these regions.

\begin{figure}
\centering
\resizebox{0.435\textwidth}{!}{
\setlength{\tabcolsep}{0.005cm} 
\renewcommand{\arraystretch}{0.25}  
\begin{tabular}{>{\centering\arraybackslash}m{0.35cm} 
 >{\centering\arraybackslash}m{1.6cm} >{\centering\arraybackslash}m{1.6cm} >{\centering\arraybackslash}m{1.6cm} >{\centering\arraybackslash}m{1.6cm} >{\centering\arraybackslash}m{1.6cm}}
  &&  Content & & Style &
 \\
 &&  \includegraphics[width=1.0\linewidth]{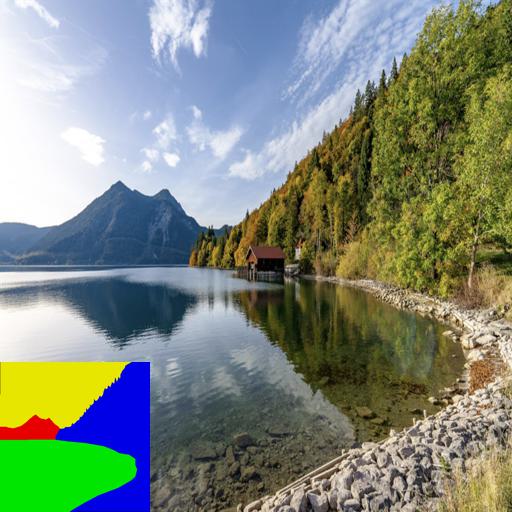} & & \includegraphics[width=1.0\linewidth]{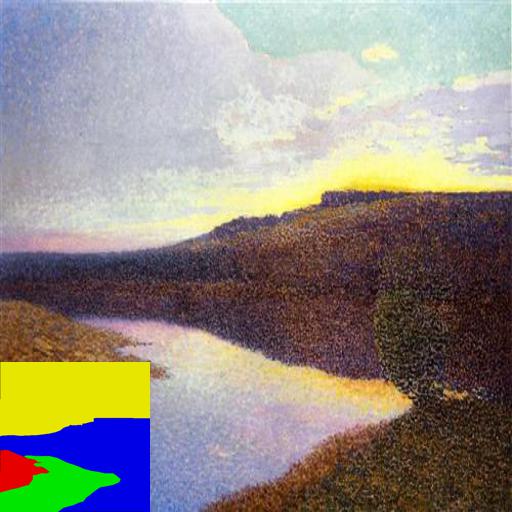} &
 \\
  & $ \alpha_{1} = 0.1$ &  $0.3$ &  $0.5$ &  $0.7$ &  $0.9$
 \\
\rotatebox[origin=rb]{270}{$\alpha_{2} = 0.1$} & 
 \includegraphics[width=1.0\linewidth]{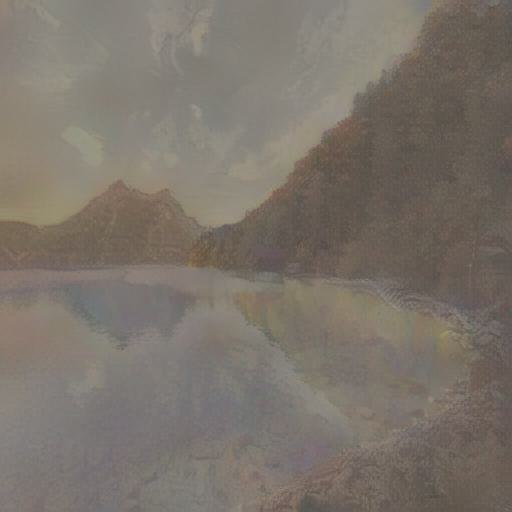} & \includegraphics[width=1\linewidth]{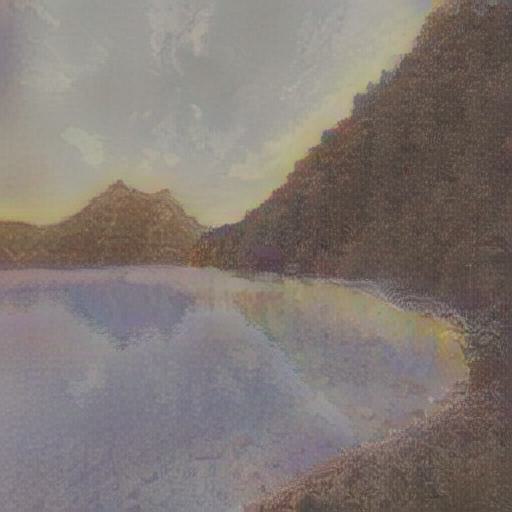} & \includegraphics[width=1\linewidth]{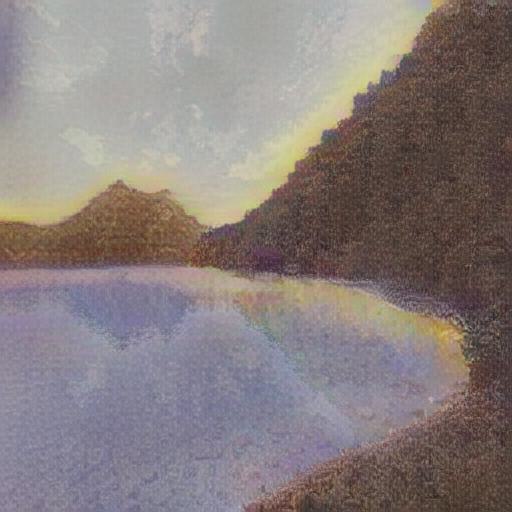} & \includegraphics[width=1\linewidth]{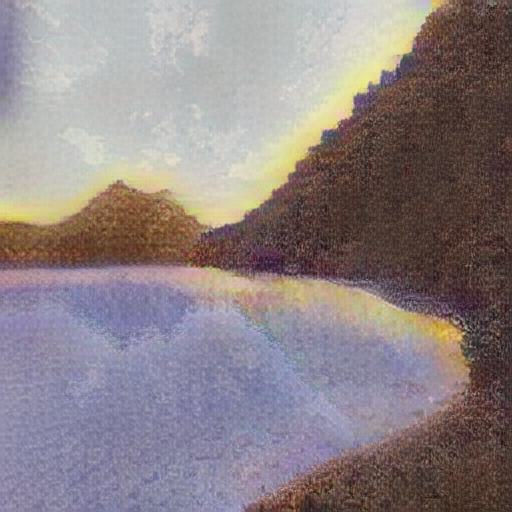} & \includegraphics[width=1\linewidth]{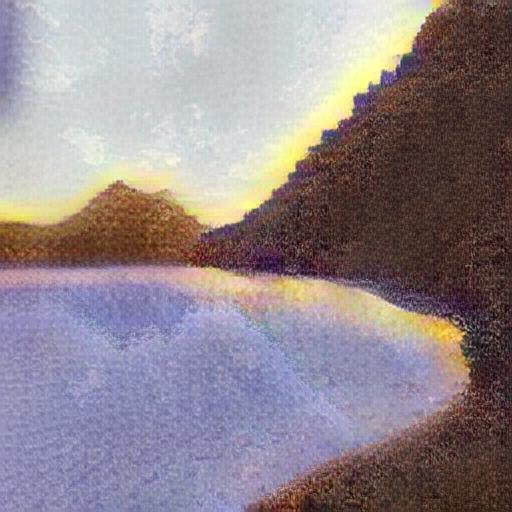} \\
 \rotatebox[origin=rb]{270}{$0.3$} &
 \includegraphics[width=1\linewidth]{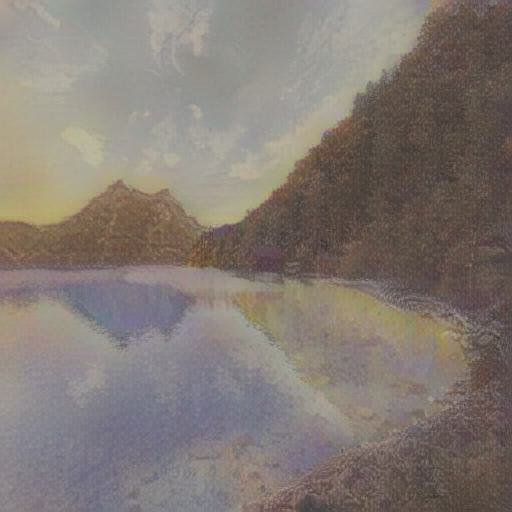} & \includegraphics[width=1\linewidth]{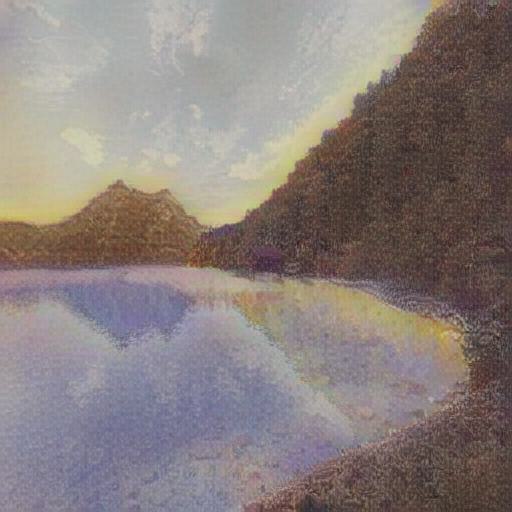} & \includegraphics[width=1\linewidth]{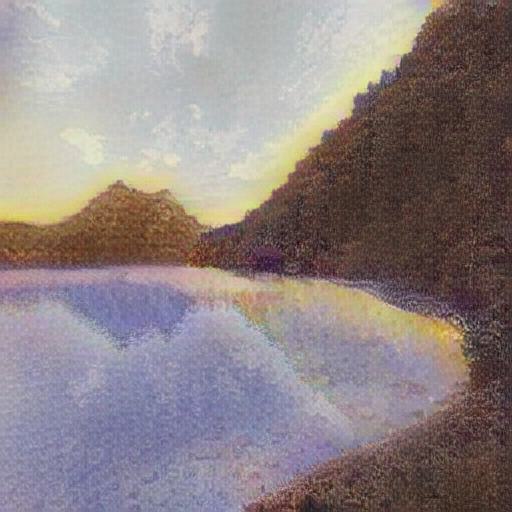} & \includegraphics[width=1\linewidth]{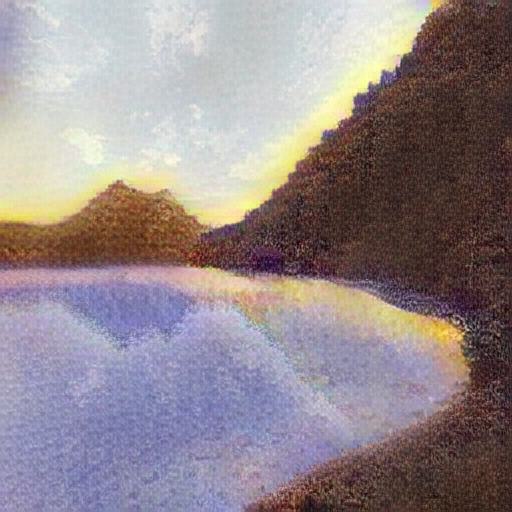} & \includegraphics[width=1\linewidth]{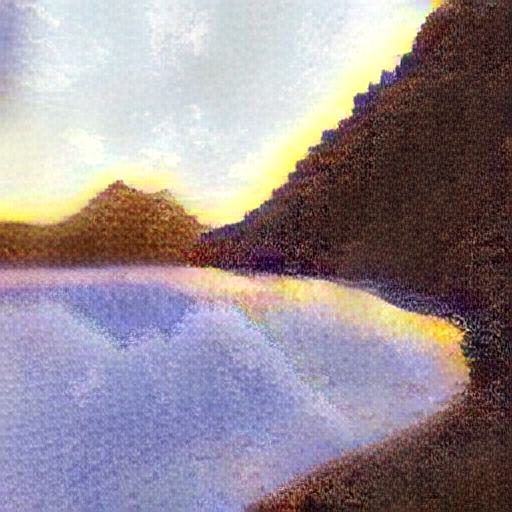} \\
  \rotatebox[origin=rb]{270}{$0.5$} & 
 \includegraphics[width=1\linewidth]{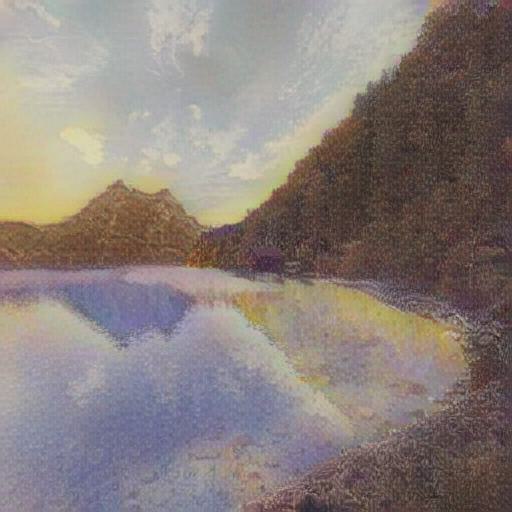} & \includegraphics[width=1\linewidth]{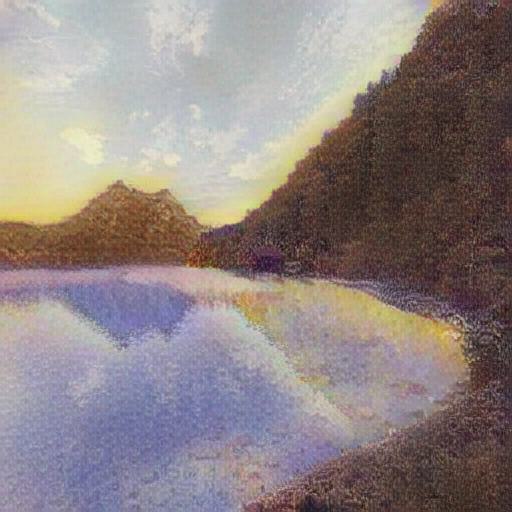} & \includegraphics[width=1\linewidth]{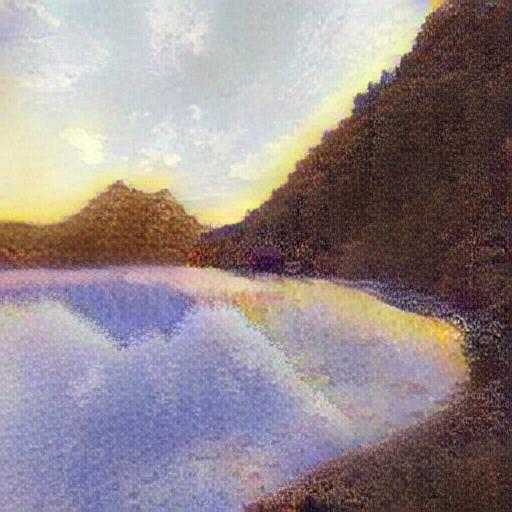} & \includegraphics[width=1\linewidth]{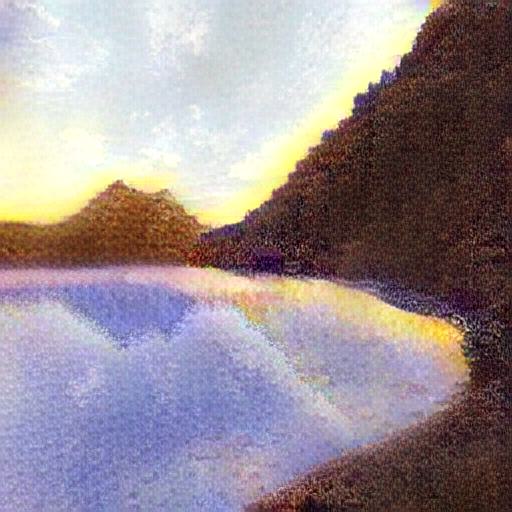} & \includegraphics[width=1\linewidth]{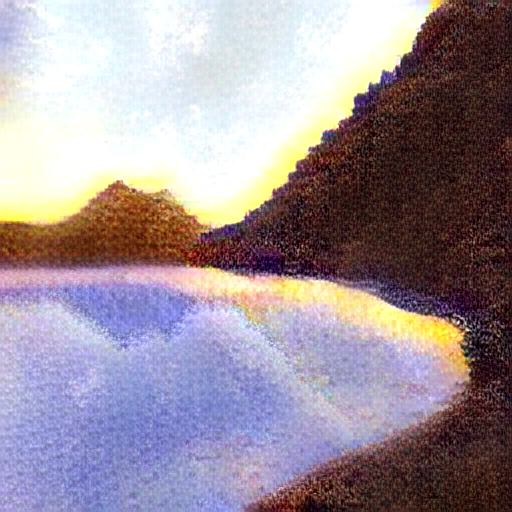} \\
    \rotatebox[origin=rb]{270}{$0.7$} &
 \includegraphics[width=1\linewidth]{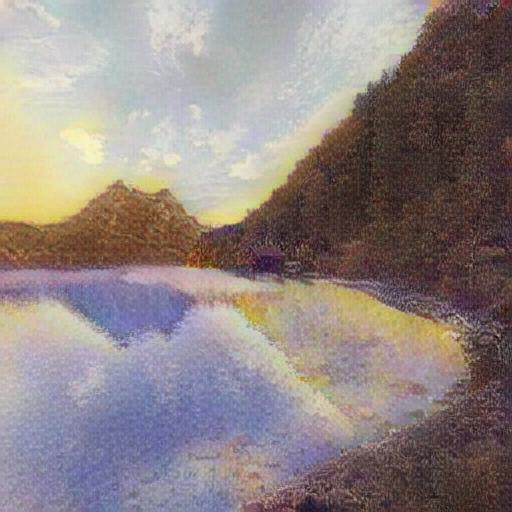} & \includegraphics[width=1\linewidth]{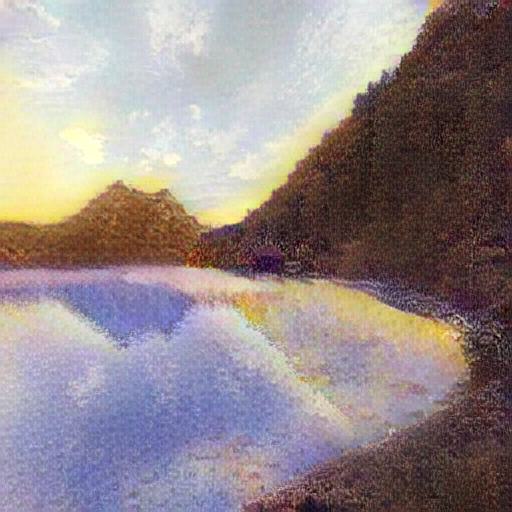} & \includegraphics[width=1\linewidth]{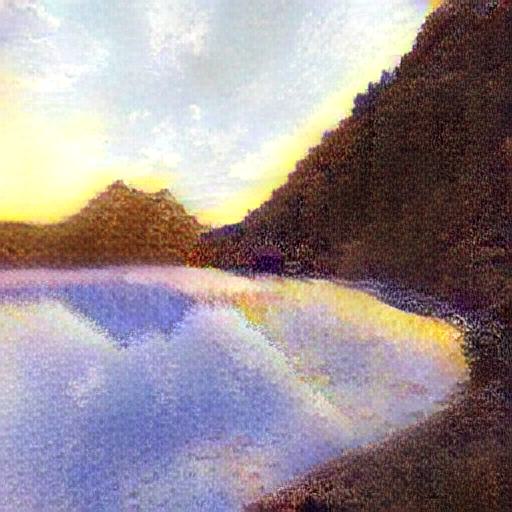} & \includegraphics[width=1\linewidth]{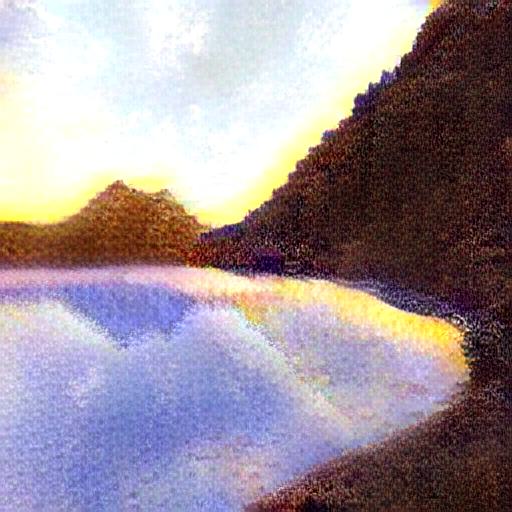} & \includegraphics[width=1\linewidth]{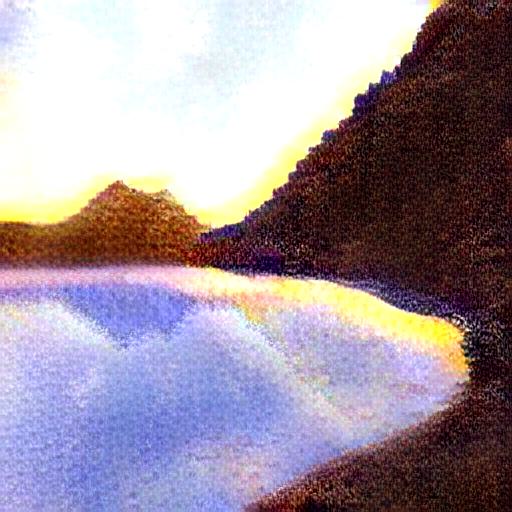} \\
      \rotatebox[origin=rb]{270}{$0.9$} & \includegraphics[width=1\linewidth]{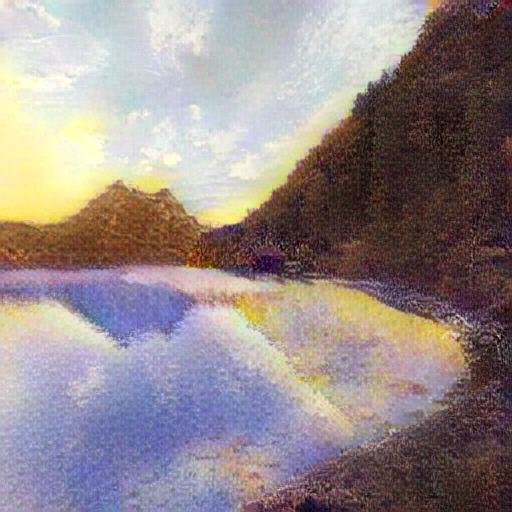} & \includegraphics[width=1\linewidth]{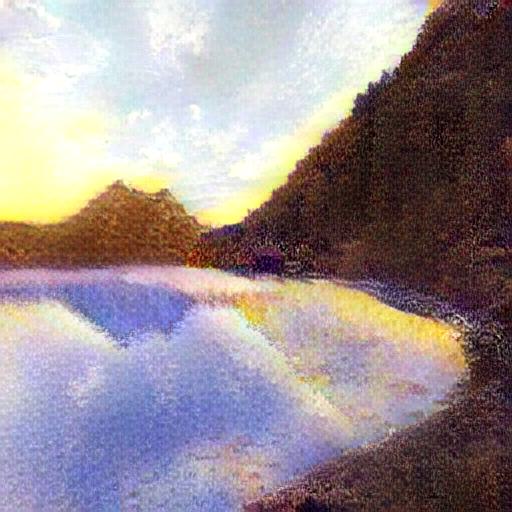} & \includegraphics[width=1\linewidth]{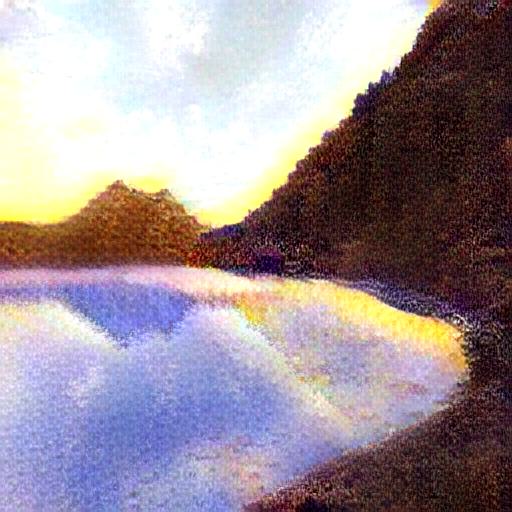} & \includegraphics[width=1\linewidth]{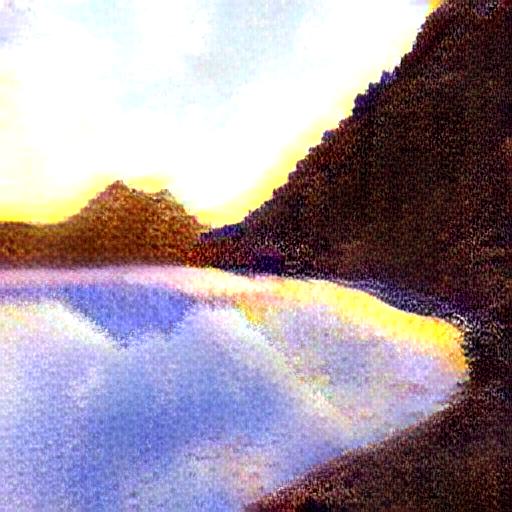} & \includegraphics[width=1\linewidth]{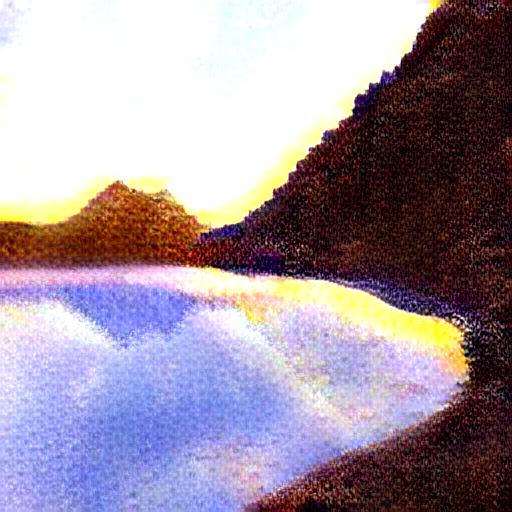}\\
\end{tabular}}
\caption{Comparisons of the overall style and local texture intensity in semantic regions of SANet embedded with our SCSA.}
\label{fig:8}
\end{figure}

{\bf Content-Style Trade-off.} Through adjustments to parameter $b$ in Sec.~\ref{sec:c2} and parameters $t_1$ and $t_2$ in Sec.~\ref{sec:c3}, our method offers a flexible balance between stylization intensity and content preservation. To illustrate this, we perform a series of targeted experiments, allowing for an in-depth analysis and comprehensive evaluation. It is important to highlight that the effectiveness of $t_2$ has already been established in~\cite{chung2024style}. Thus, our focus here will be solely on demonstrating the content preservation capability of $t_1$, thoroughly examining its impact on maintaining the integrity of the original content during the stylization process.

\begin{figure}
\centering
\resizebox{0.46\textwidth}{!}{
\setlength{\tabcolsep}{0.006cm} 
\renewcommand{\arraystretch}{0.25}  
\begin{tabular}{>{\centering\arraybackslash}m{0.8cm} 
 >{\centering\arraybackslash}m{1.6cm} >{\centering\arraybackslash}m{1.6cm} >{\centering\arraybackslash}m{1.6cm} >{\centering\arraybackslash}m{1.6cm} >{\centering\arraybackslash}m{1.6cm}}
 \\
  Inputs & $ b = 0.1$ &  $0.3$ &  $0.5$ &  $0.7$ &  $0.9$
\\
\includegraphics[width=1.0\linewidth]{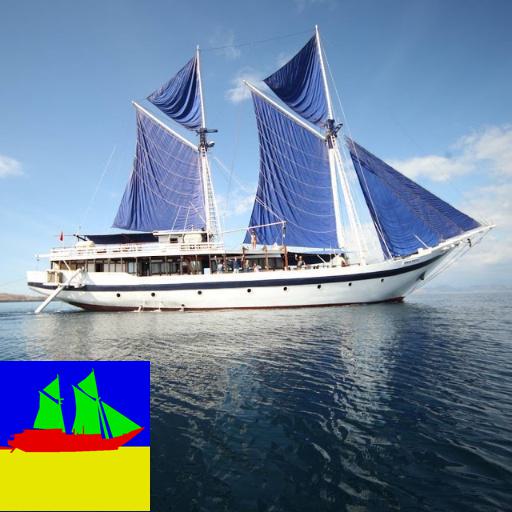} \includegraphics[width=1.0\linewidth]{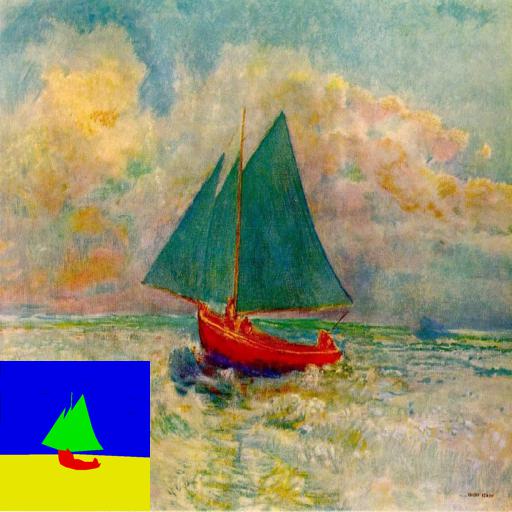} & 
 \includegraphics[width=1.0\linewidth]{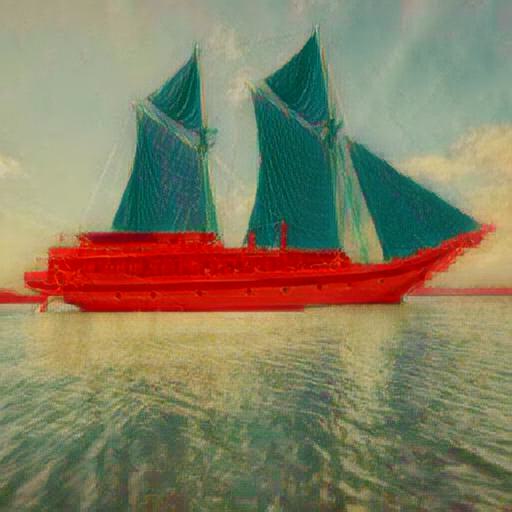} & \includegraphics[width=1.0\linewidth]{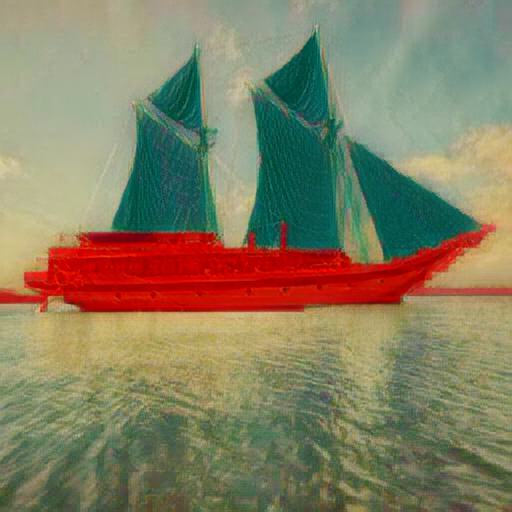} & \includegraphics[width=1.0\linewidth]{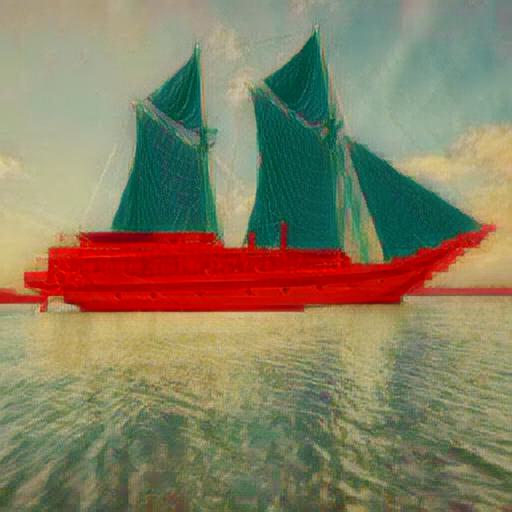} & \includegraphics[width=1.0\linewidth]{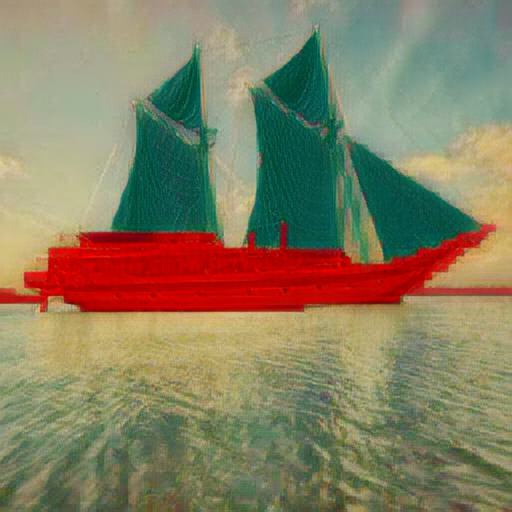} & \includegraphics[width=1.0\linewidth]{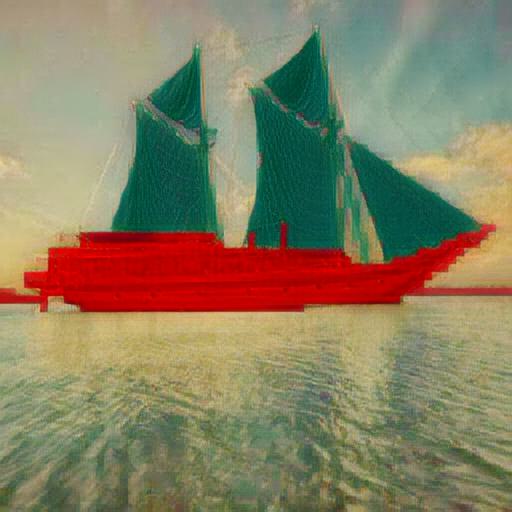} \\
 \includegraphics[width=1.0\linewidth]{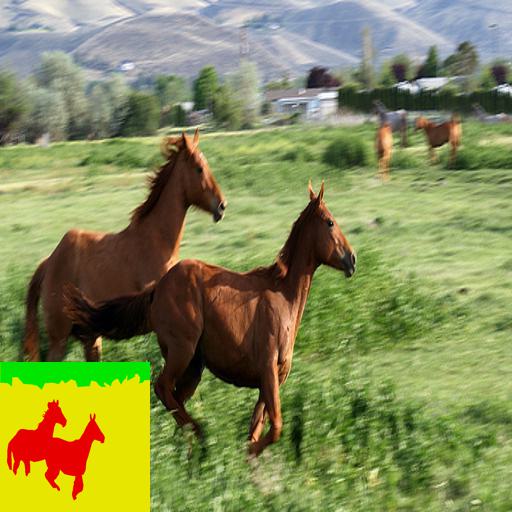} \includegraphics[width=1.0\linewidth]{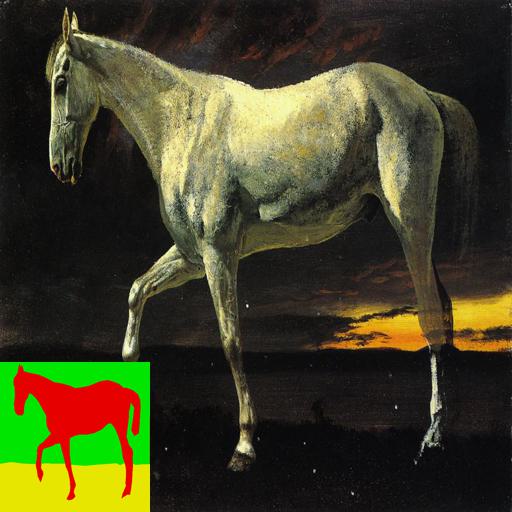} & 
 \includegraphics[width=1.0\linewidth]{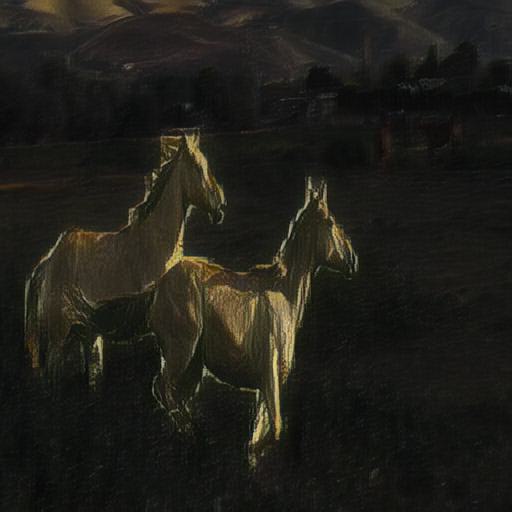} & \includegraphics[width=1.0\linewidth]{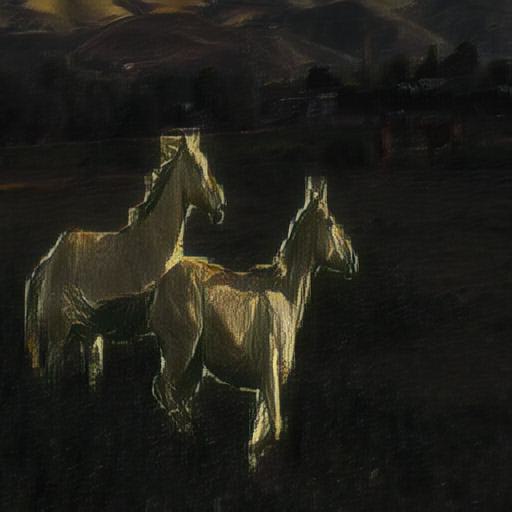} & \includegraphics[width=1.0\linewidth]{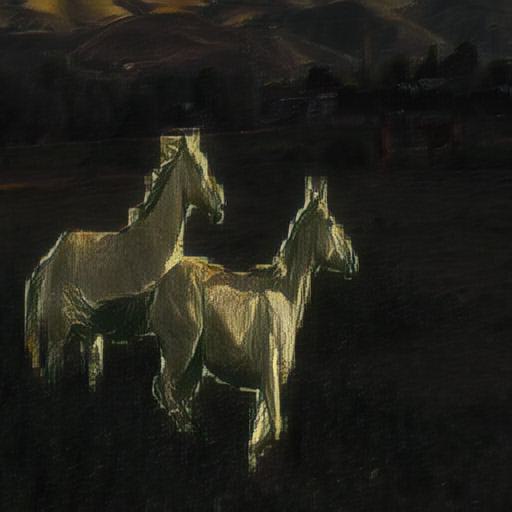} & \includegraphics[width=1.0\linewidth]{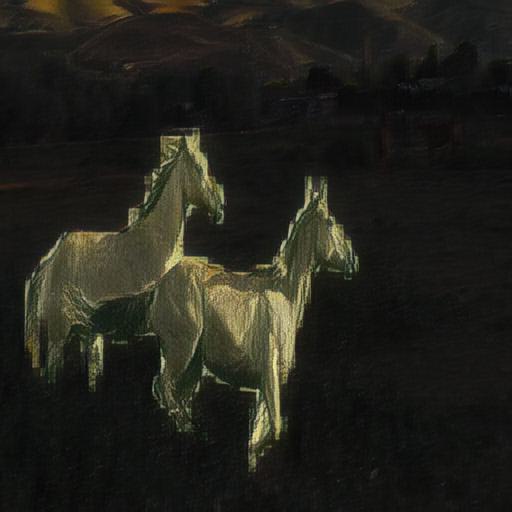} & \includegraphics[width=1.0\linewidth]{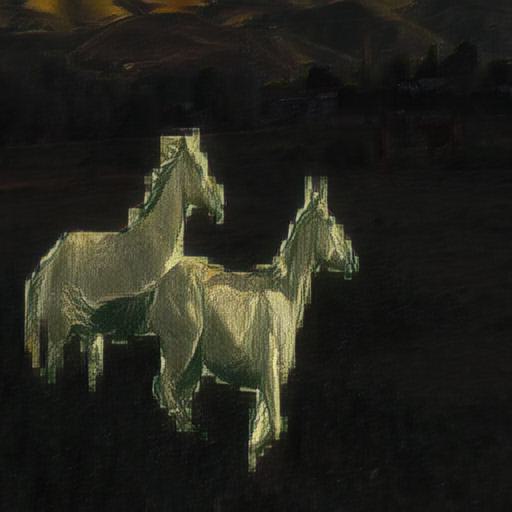} \\
  \includegraphics[width=1.0\linewidth]{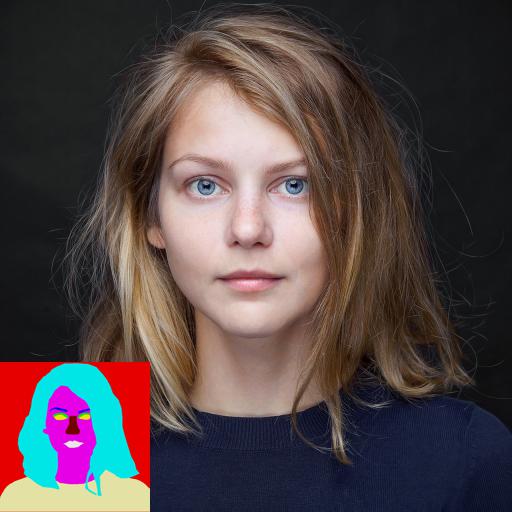} \includegraphics[width=1.0\linewidth]{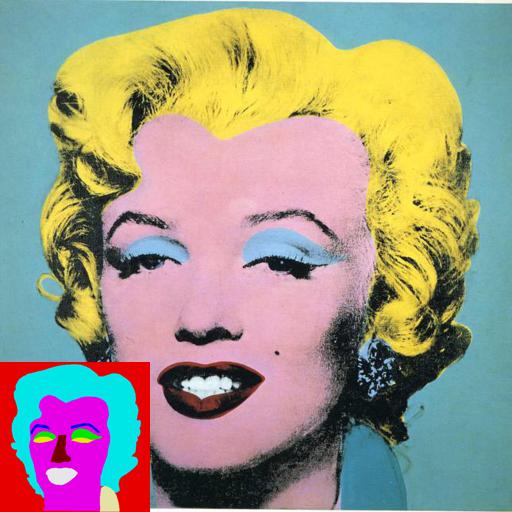} & 
 \includegraphics[width=1.0\linewidth]{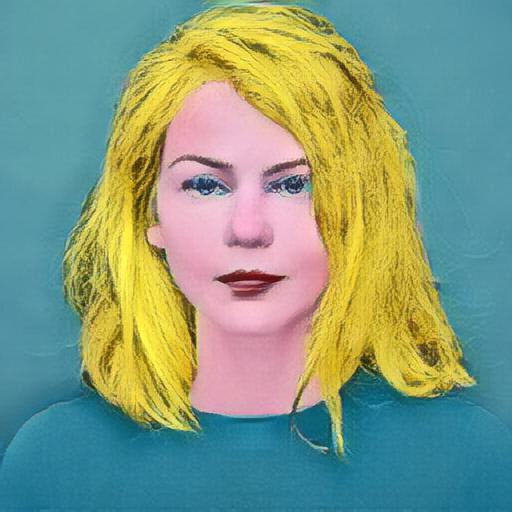} & \includegraphics[width=1.0\linewidth]{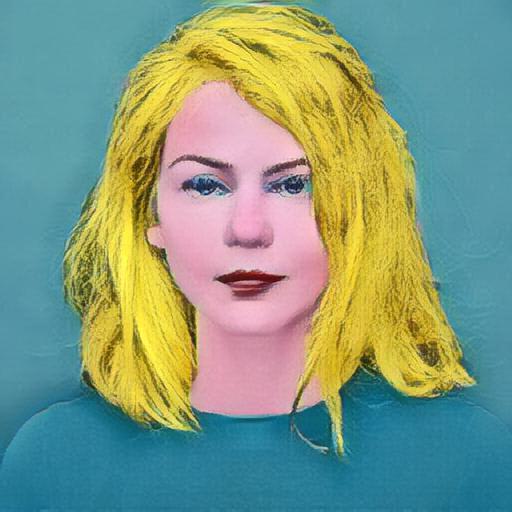} & \includegraphics[width=1.0\linewidth]{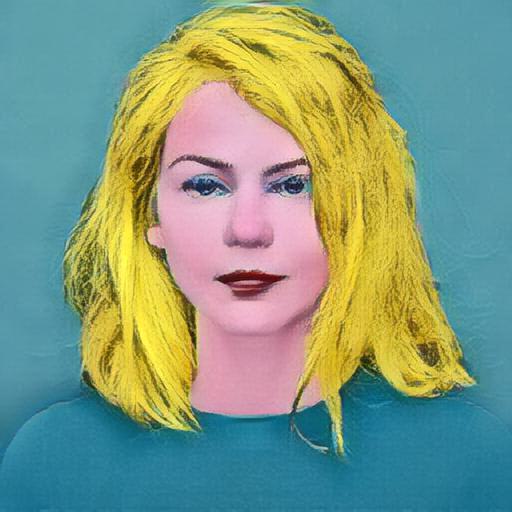} & \includegraphics[width=1.0\linewidth]{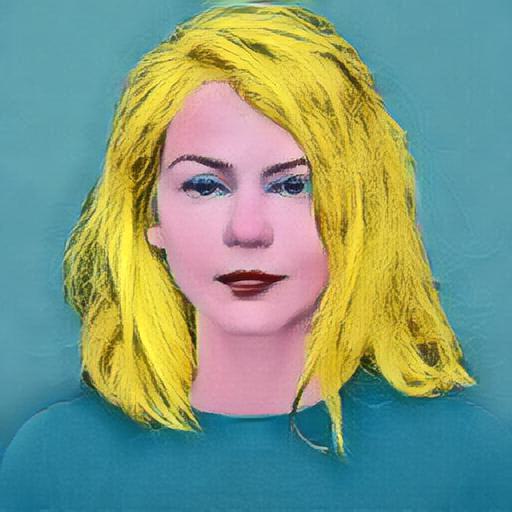} & \includegraphics[width=1.0\linewidth]{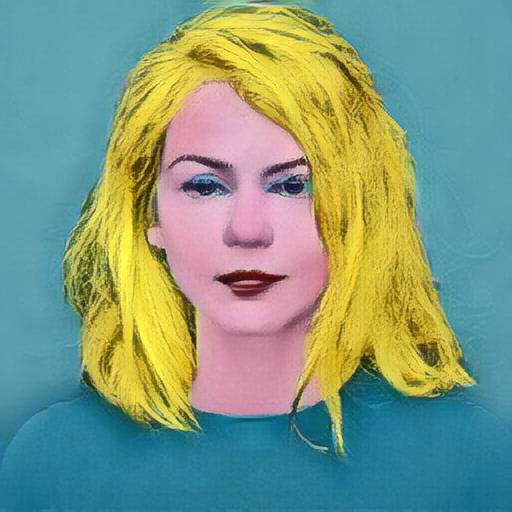} \\
\end{tabular}}
\caption{Comparisons of the trade-off between content preservation and stylization of StyTR$^2$ embedded with SCSA.}
\label{fig:11}
\end{figure}

As illustrated in Fig.~\ref{fig:11}, with the increase of the parameter $b$, the degree of stylization progressively intensifies, while the level of content preservation experiences a slight decline. This phenomenon is particularly evident across different subjects. For instance, in the $1st$ row, the boat's structure becomes more pronounced in the stylization process; while its basic form remains discernible, the accuracy of content preservation decreases slightly. In the $2nd$ row, the eyes of the horse exhibit more distinctive stylistic features, although their original characteristics are somewhat retained, the details appear less clear. In the $3rd$ row, the eyes of the person maintain their fundamental structure while integrating more stylistic elements, which also affects their level of content preservation. 

\begin{figure}
\centering
\resizebox{0.46\textwidth}{!}{
\setlength{\tabcolsep}{0.006cm} 
\renewcommand{\arraystretch}{0.25}  
\begin{tabular}{>{\centering\arraybackslash}m{0.8cm} 
 >{\centering\arraybackslash}m{1.6cm} >{\centering\arraybackslash}m{1.6cm} >{\centering\arraybackslash}m{1.6cm} >{\centering\arraybackslash}m{1.6cm} >{\centering\arraybackslash}m{1.6cm}}
 \\
  Inputs & $ t_1 = 0.1$ &  $0.3$ &  $0.5$ &  $0.7$ &  $0.9$
\\
\includegraphics[width=1.0\linewidth]{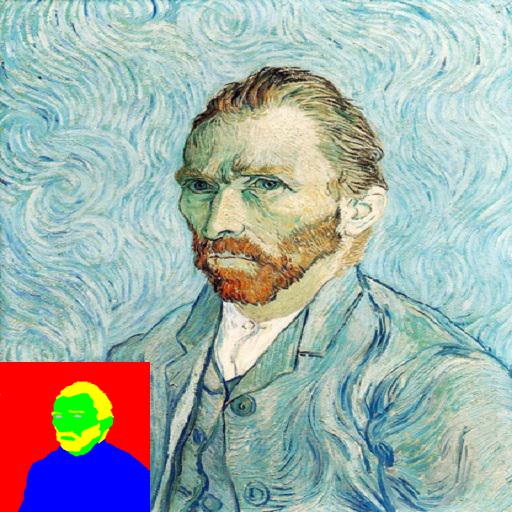} \includegraphics[width=1.0\linewidth]{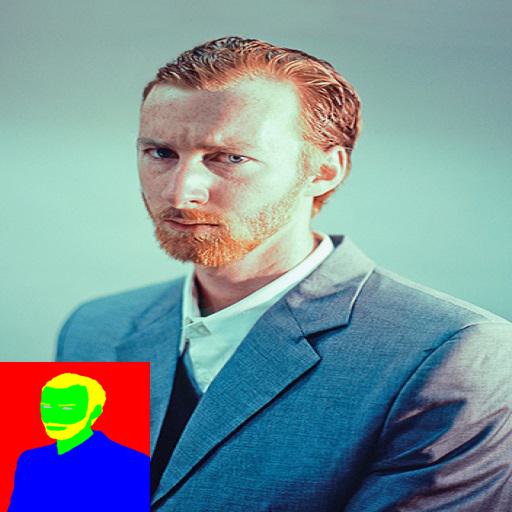} & 
 \includegraphics[width=1.0\linewidth]{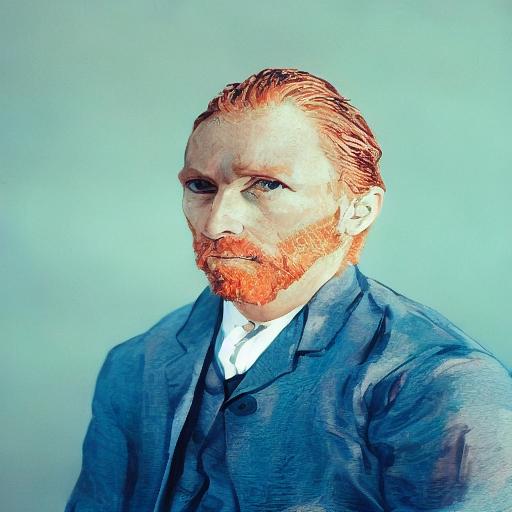} & \includegraphics[width=1.0\linewidth]{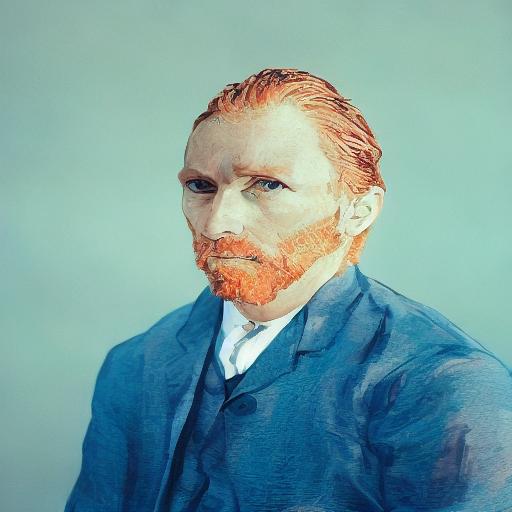} & \includegraphics[width=1.0\linewidth]{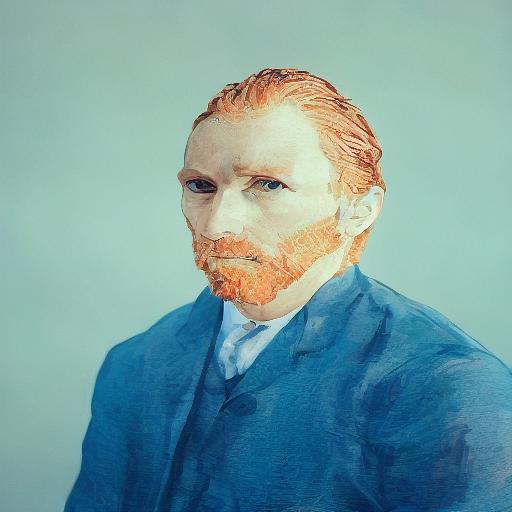} & \includegraphics[width=1.0\linewidth]{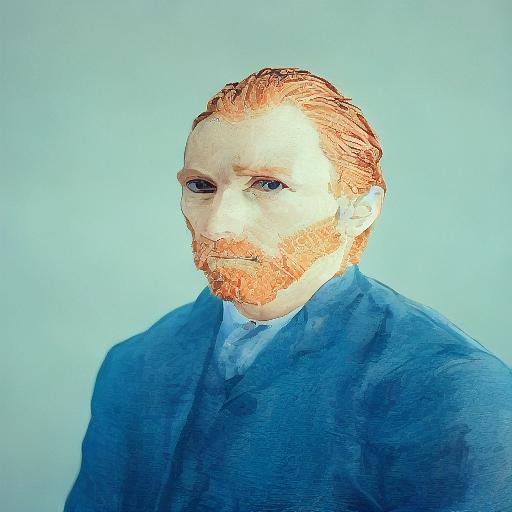} & \includegraphics[width=1.0\linewidth]{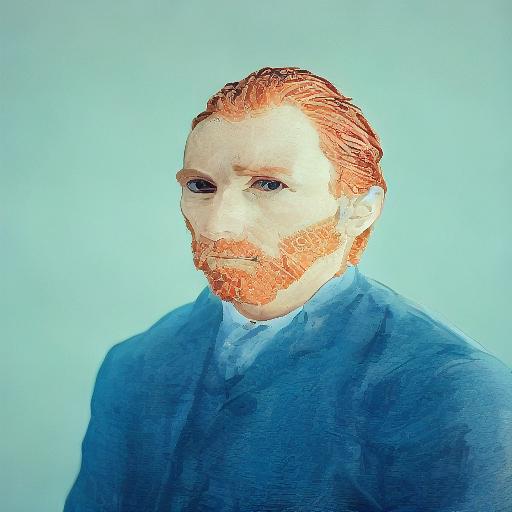} \\
 \includegraphics[width=1.0\linewidth]{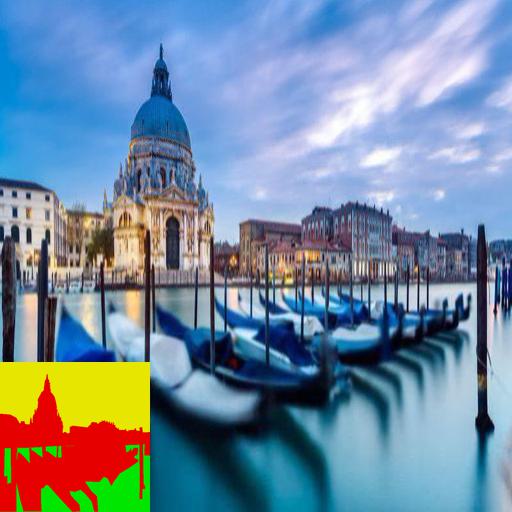} \includegraphics[width=1.0\linewidth]{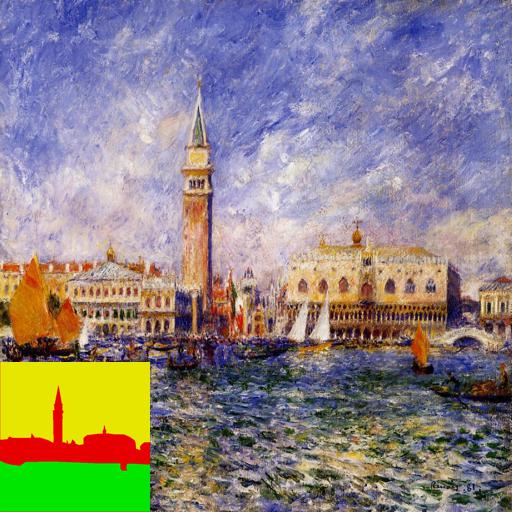} & 
 \includegraphics[width=1.0\linewidth]{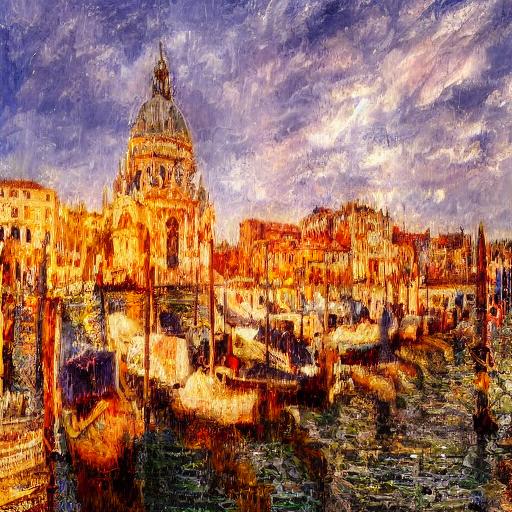} & \includegraphics[width=1.0\linewidth]{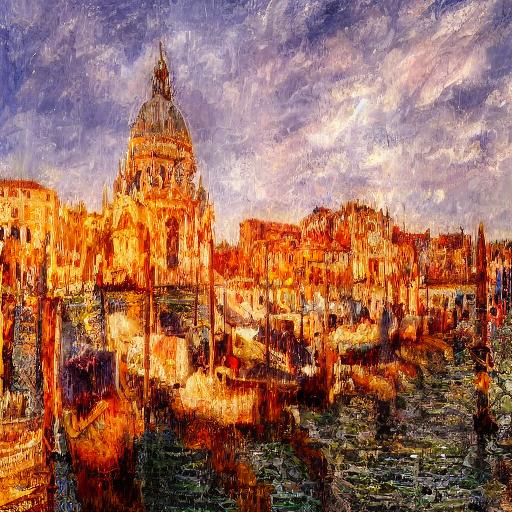} & \includegraphics[width=1.0\linewidth]{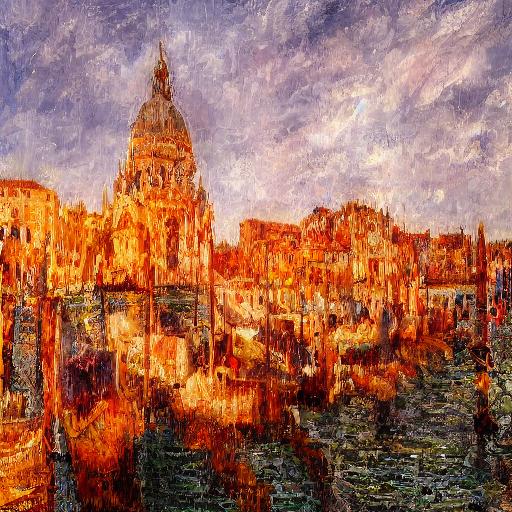} & \includegraphics[width=1.0\linewidth]{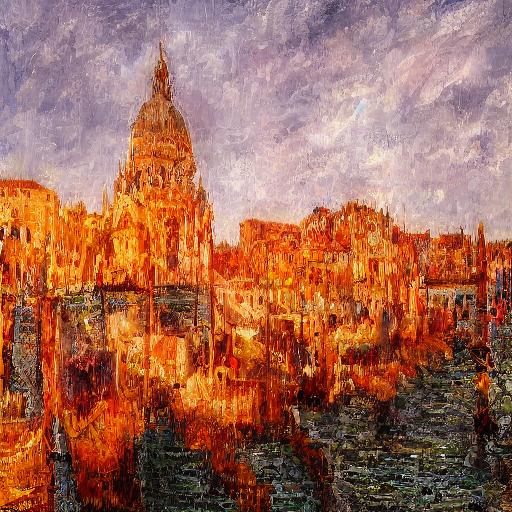} & \includegraphics[width=1.0\linewidth]{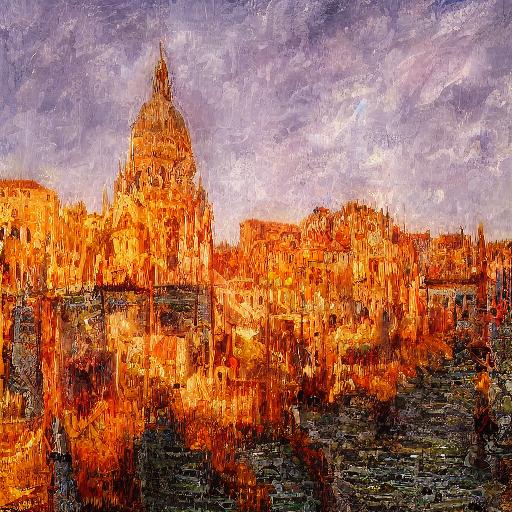} \\
  \includegraphics[width=1.0\linewidth]{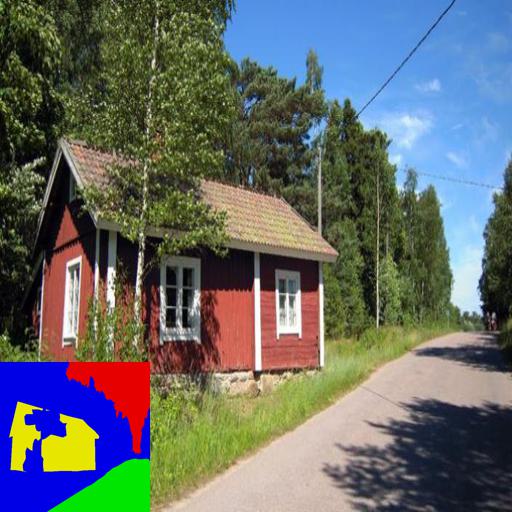} \includegraphics[width=1.0\linewidth]{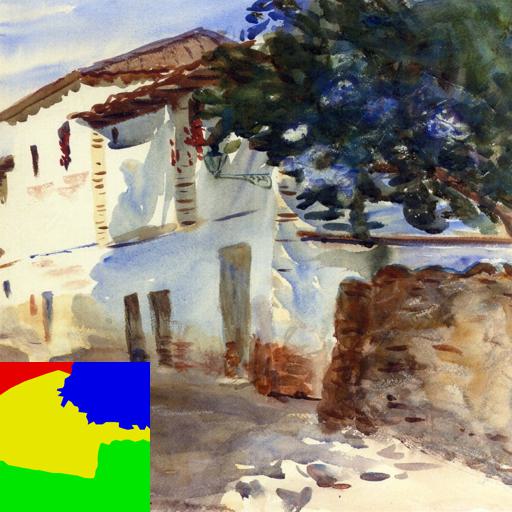} & 
 \includegraphics[width=1.0\linewidth]{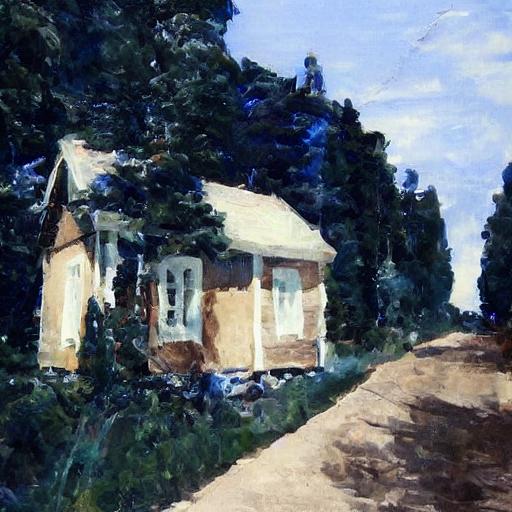} & \includegraphics[width=1.0\linewidth]{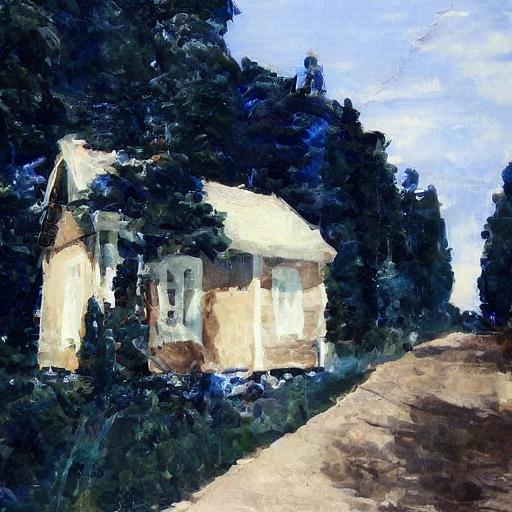} & \includegraphics[width=1.0\linewidth]{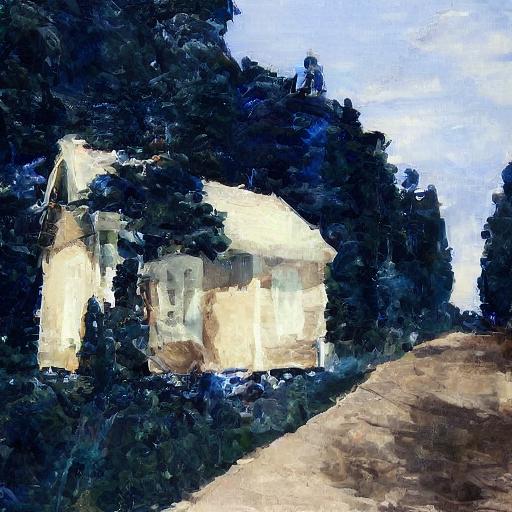} & \includegraphics[width=1.0\linewidth]{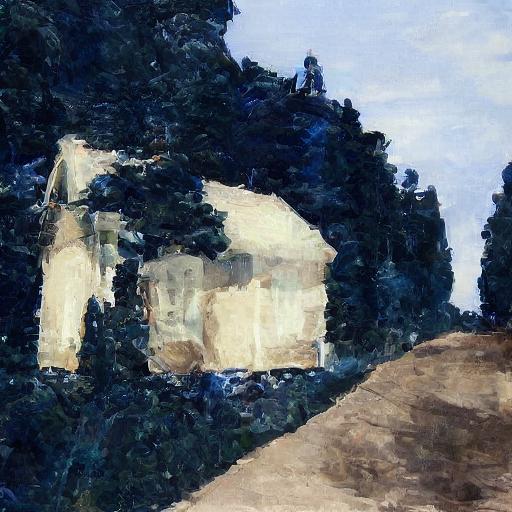} & \includegraphics[width=1.0\linewidth]{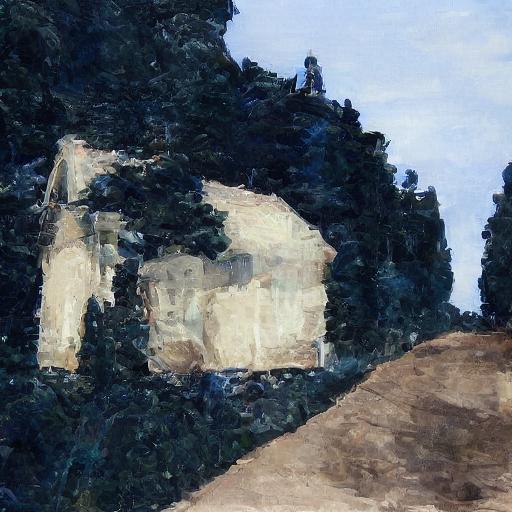} \\
\end{tabular}}
\caption{Comparisons of the trade-off between content preservation and stylization of StyleID embedded with SCSA.}
\label{fig:12}
\end{figure}

As shown in Fig.~\ref{fig:12}, as $t_1$ increases, the degree of stylization in semantic regions gradually intensifies, while the ability to maintain content declines, e.g., the building structures in the $1st$ row, the cloth in the $2nd$ row the house in the $3rd$ row. Therefore, $t_1$ allows for a trade-off between semantic stylization and content preservation.

\begin{figure}
\centering
\resizebox{0.45\textwidth}{!}{
\setlength{\tabcolsep}{0.006cm} 
\renewcommand{\arraystretch}{0.25}  
\begin{tabular}{>{\centering\arraybackslash}m{1.2cm} 
 >{\centering\arraybackslash}m{2.4cm} >{\centering\arraybackslash}m{2.4cm} >{\centering\arraybackslash}m{2.4cm} >{\centering\arraybackslash}m{2.4cm}}
 \\
  Inputs & SANet &   + SCSA &   + S-AdaIN &    + Style-Swap 
\\
\includegraphics[width=1.0\linewidth]{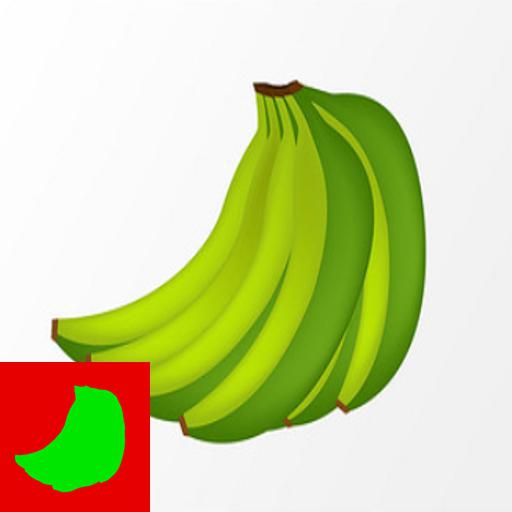} \includegraphics[width=1.0\linewidth]{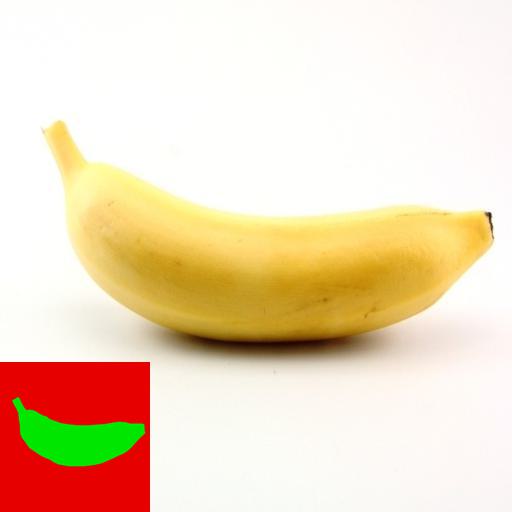} & 
 \includegraphics[width=1.0\linewidth]{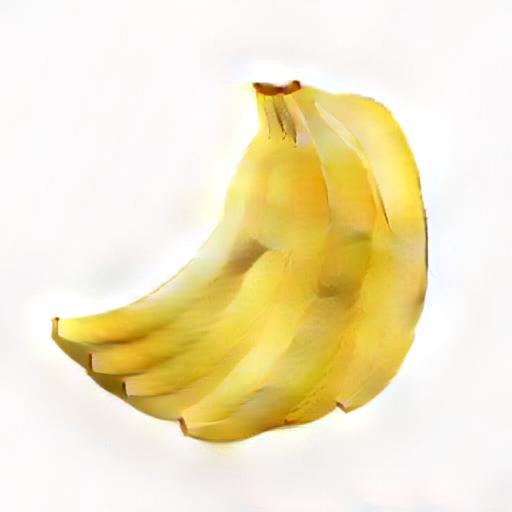} & \includegraphics[width=1.0\linewidth]{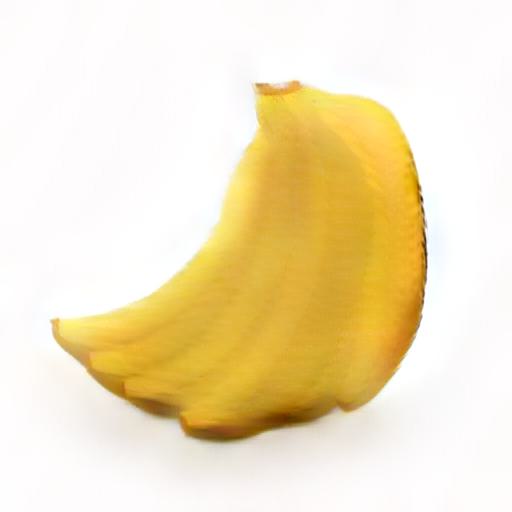} & \includegraphics[width=1.0\linewidth]{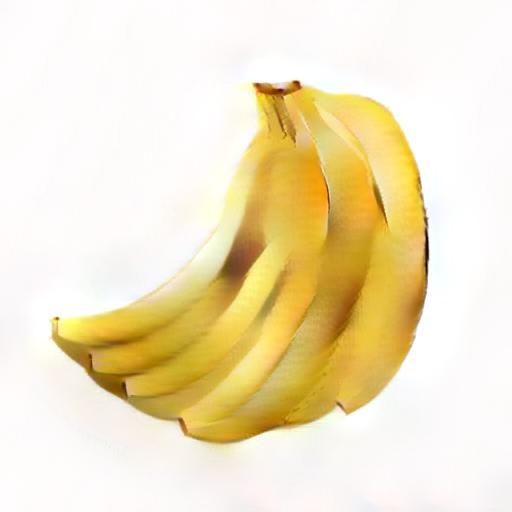} & \includegraphics[width=1.0\linewidth]{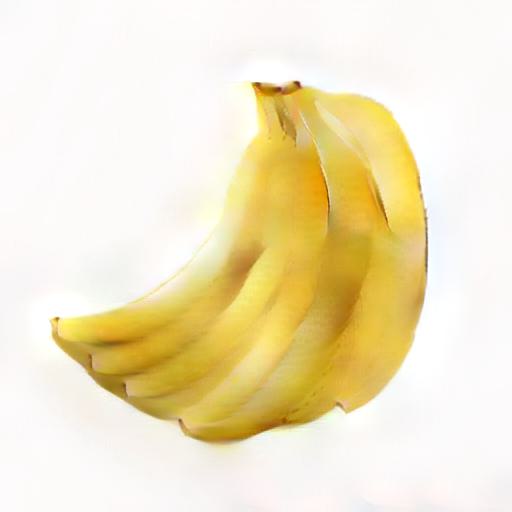} \\
 \includegraphics[width=1.0\linewidth]{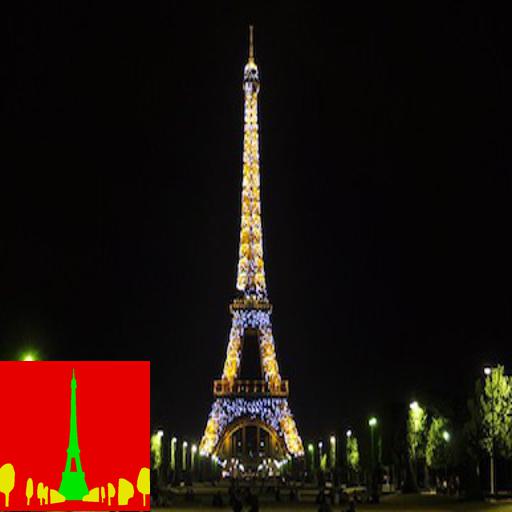} \includegraphics[width=1.0\linewidth]{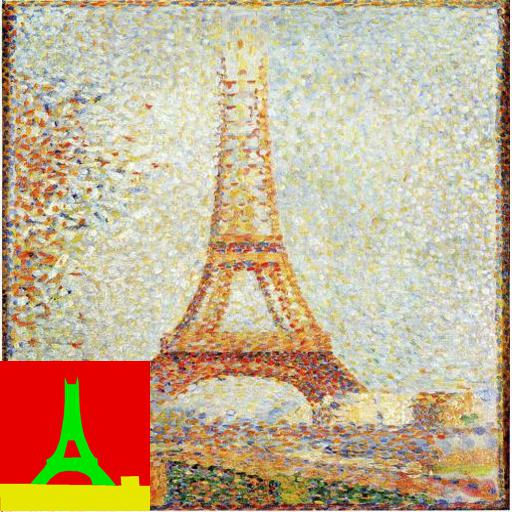} & 
 \includegraphics[width=1.0\linewidth]{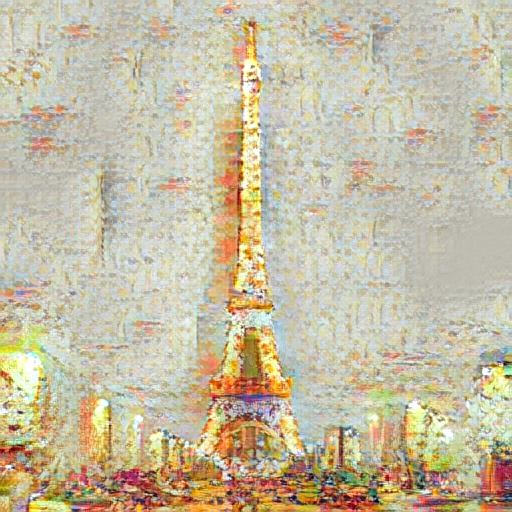} & \includegraphics[width=1.0\linewidth]{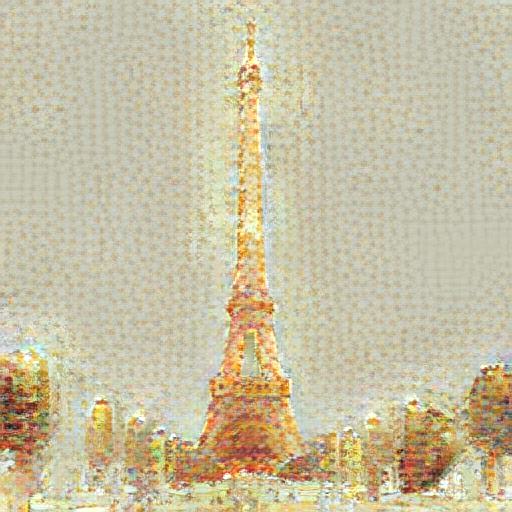} & \includegraphics[width=1.0\linewidth]{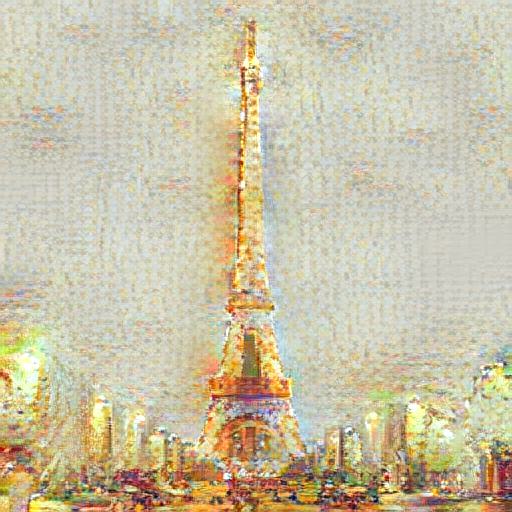} & \includegraphics[width=1.0\linewidth]{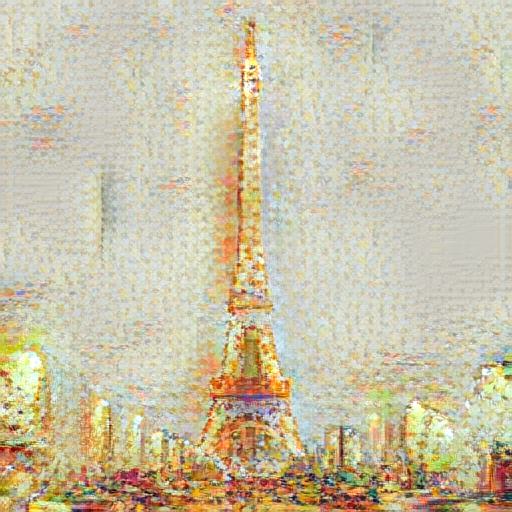}  \\
  \includegraphics[width=1.0\linewidth]{sm/img/27_paint+sem.jpg} \includegraphics[width=1.0\linewidth]{sm/img/27+sem.jpg} & 
 \includegraphics[width=1.0\linewidth]{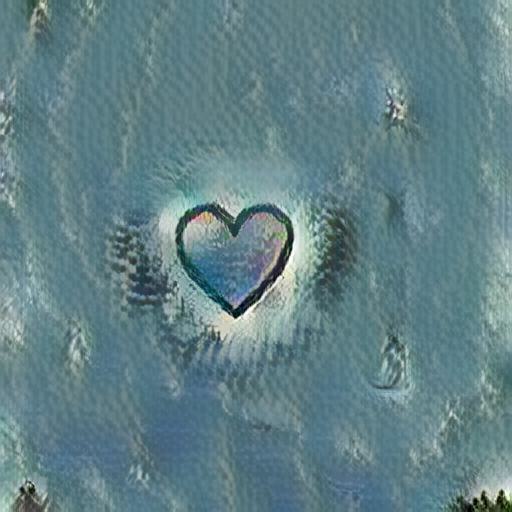} & \includegraphics[width=1.0\linewidth]{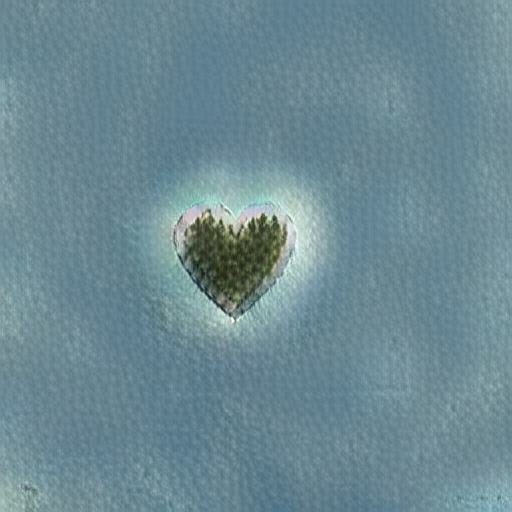} & \includegraphics[width=1.0\linewidth]{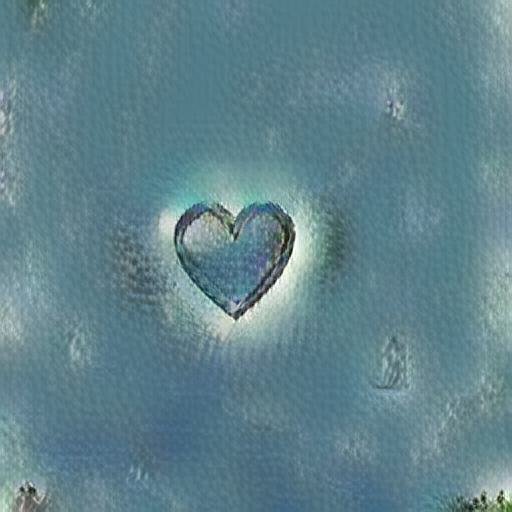} & \includegraphics[width=1.0\linewidth]{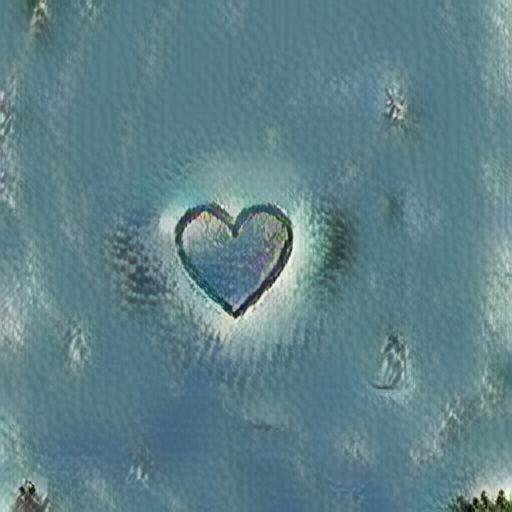}  \\
 &
 \\
 SSL $\downarrow$ & 1.6583 & {\bf 0.8762} & 1.7289 & 1.5751\\
 &
 \\
 FID $\downarrow$ & 14.3385 & {\bf 13.0788} & 14.2748 & 14.1523\\
  &
 \\
 CFSD $\downarrow$ & 0.1103 & {\bf 0.0874} & 0.1057 & 0.1077\\
\end{tabular}}
\caption{Qualitative and quantitative comparisons among CNN-based SANet, SANet with SCSA, SANet with S-AdaIN, and SANet with Style-Swap.}
\label{fig:19}
\end{figure}

\section{Dataset Details}

We select content and style images from prior research~\cite{champandard2016semantic,wang2022aesust,huang2023quantart,wu2022ccpl,zhang2022domain,ahn2024dreamstyler,zhang2024artbank,tarres2024parasol,Wen2023CAPVSTNetCA,li2019neural,xue2021end}, publicly available datasets~\cite{phillips2011wiki,zhu2017unpaired,5206848,everingham2010pascal}, and the Internet to ensure a diverse and comprehensive set of images. These sources provide a broad spectrum of content and style representations, encompassing various domains and visual styles. We then construct the semantic maps for each image, capturing their intrinsic features and structural elements. Based on these maps, we generate 91 validated quadruple data, which will be made publicly available upon acceptance of the paper to foster further research and development in semantic style transfer.

\section{Ablation Study}

To comprehensively validate the superiority of our proposed SCSA method, we incorporate the existing S-AdaIN~\cite{lu_closed-form_2019} and Style-Swap~\cite{chen2016fast} techniques into the universal attention module of the Attn-AST framework.

\begin{figure}
\centering
\resizebox{0.45\textwidth}{!}{
\setlength{\tabcolsep}{0.006cm} 
\renewcommand{\arraystretch}{0.25}  
\begin{tabular}{>{\centering\arraybackslash}m{1.2cm} 
 >{\centering\arraybackslash}m{2.4cm} >{\centering\arraybackslash}m{2.4cm} >{\centering\arraybackslash}m{2.4cm} >{\centering\arraybackslash}m{2.4cm}}
 \\
  Inputs & StyTR$^2$ &  + SCSA &   + S-AdaIN &   + Style-Swap 
\\
\includegraphics[width=1.0\linewidth]{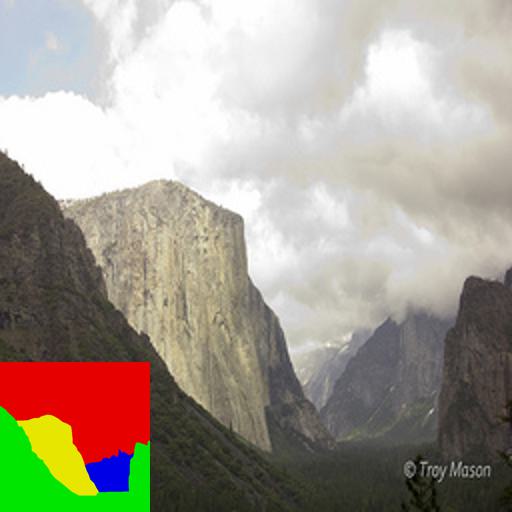} \includegraphics[width=1.0\linewidth]{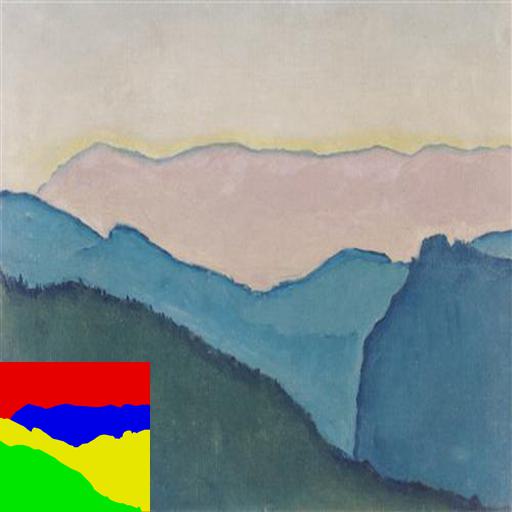} & 
 \includegraphics[width=1.0\linewidth]{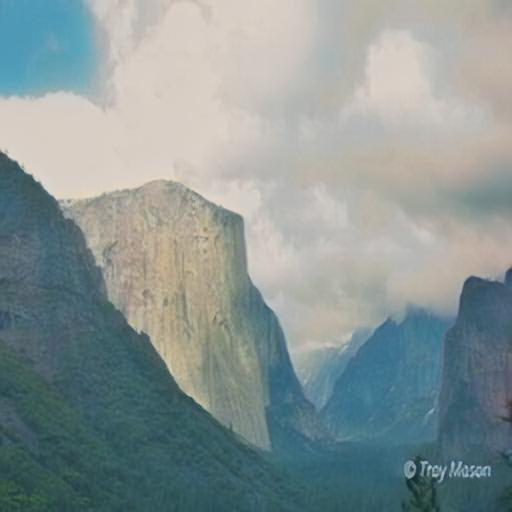} & \includegraphics[width=1.0\linewidth]{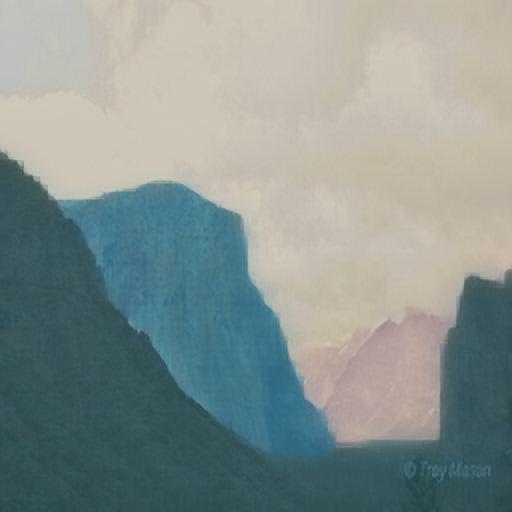} & \includegraphics[width=1.0\linewidth]{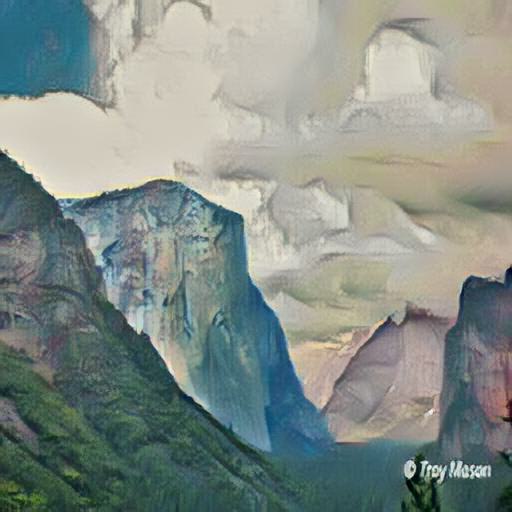} & \includegraphics[width=1.0\linewidth]{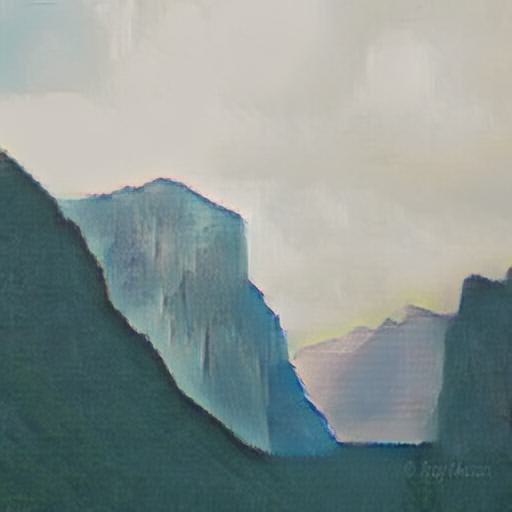} \\
 \includegraphics[width=1.0\linewidth]{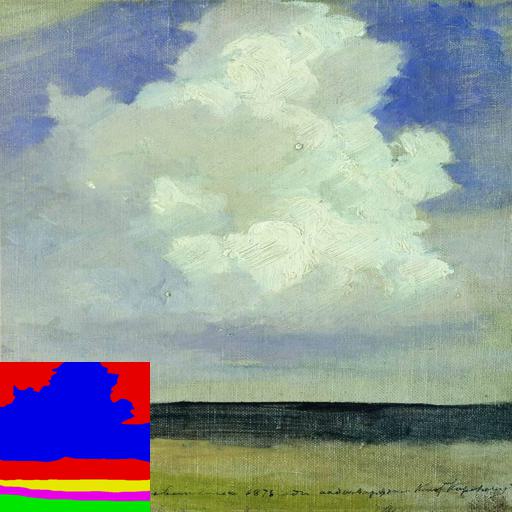} \includegraphics[width=1.0\linewidth]{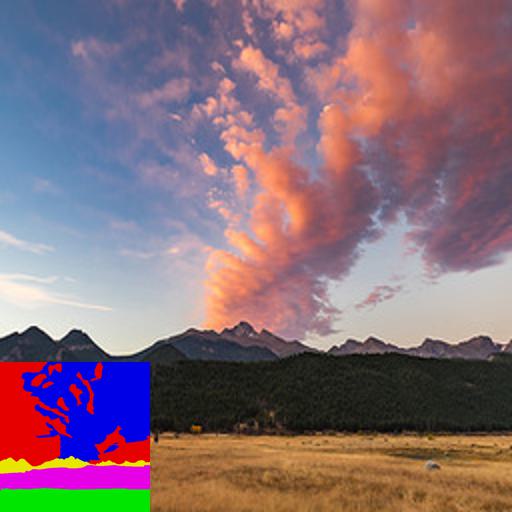} & 
 \includegraphics[width=1.0\linewidth]{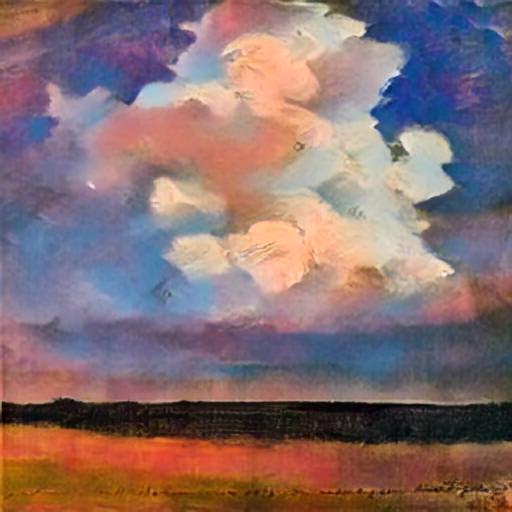} & \includegraphics[width=1.0\linewidth]{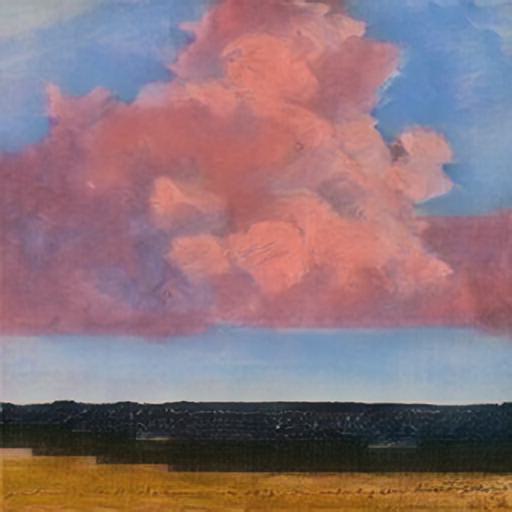} & \includegraphics[width=1.0\linewidth]{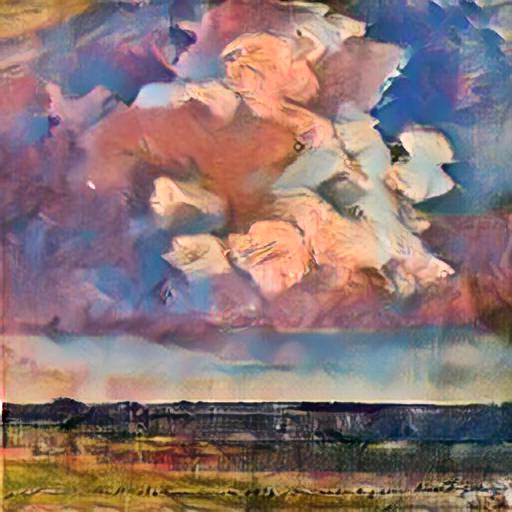} & \includegraphics[width=1.0\linewidth]{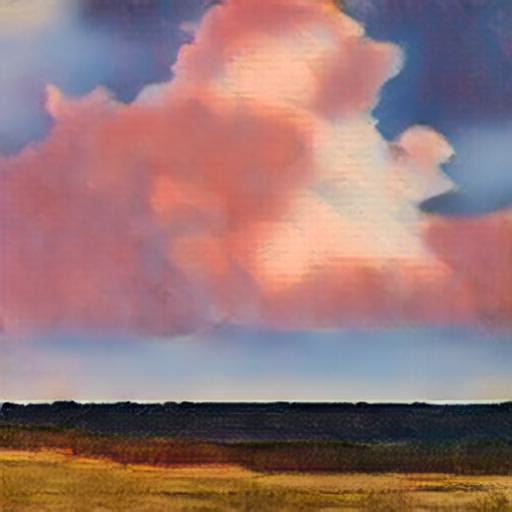}  \\
  \includegraphics[width=1.0\linewidth]{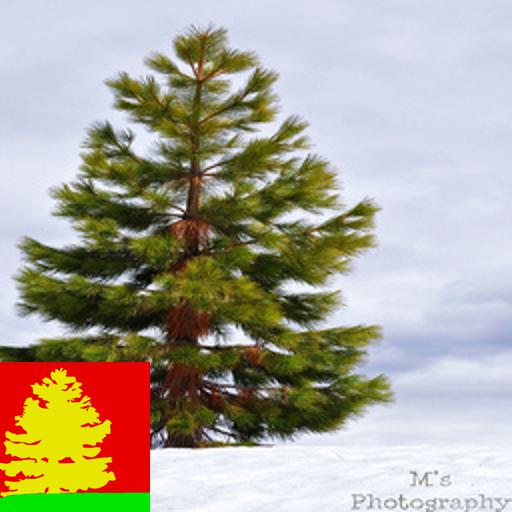} \includegraphics[width=1.0\linewidth]{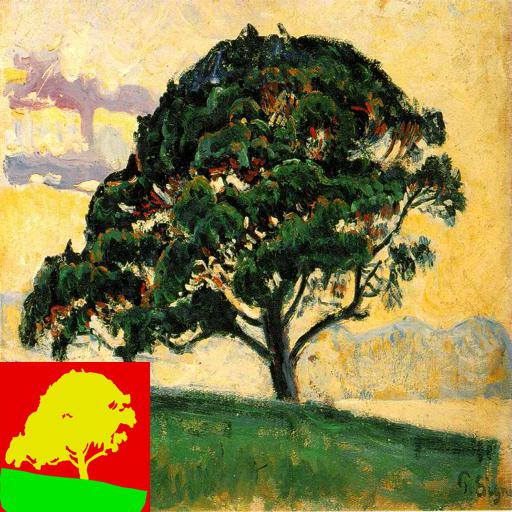} & 
 \includegraphics[width=1.0\linewidth]{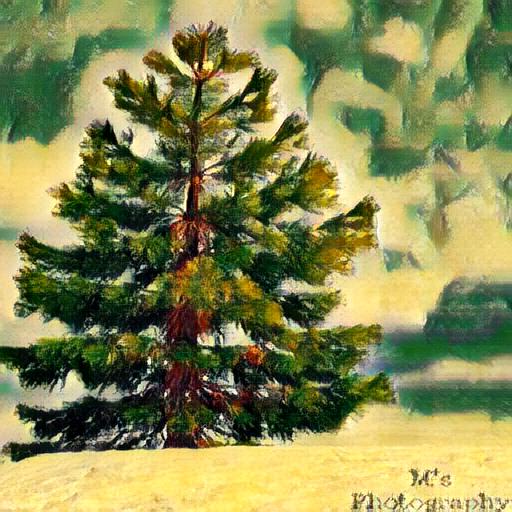} & \includegraphics[width=1.0\linewidth]{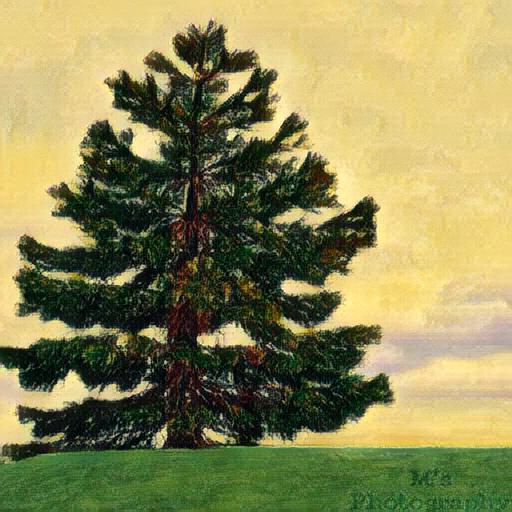} & \includegraphics[width=1.0\linewidth]{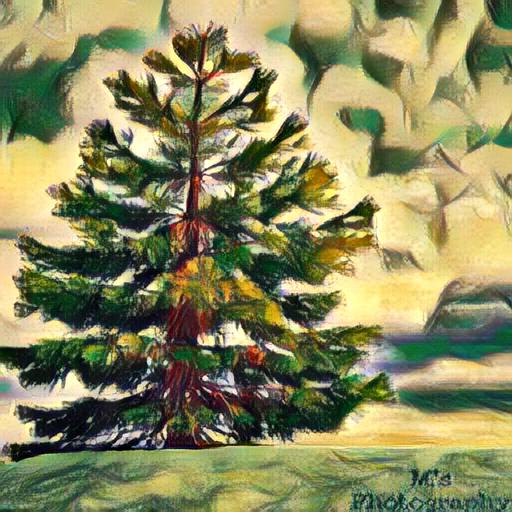} & \includegraphics[width=1.0\linewidth]{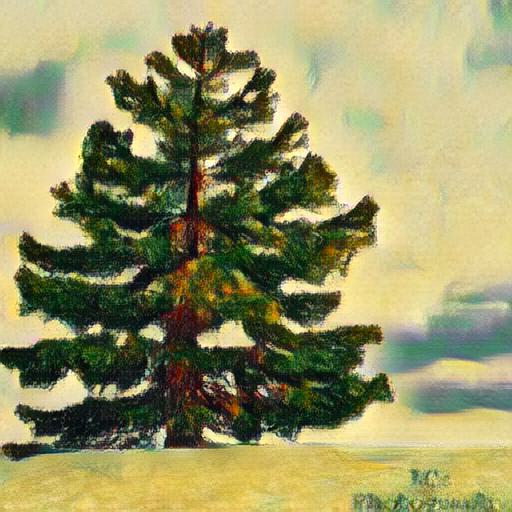}  \\
   &
 \\
 SSL $\downarrow$ & 1.9826 & {\bf 1.2228} & 1.7318 & 1.3824 \\
 &
 \\
 FID $\downarrow$ & 12.5273 & {\bf 12.3963} & 14.9473 & 13.9573 \\
  &
 \\
 CFSD $\downarrow$ & 0.0752 & {\bf 0.0705} & 0.1007 & 0.0818 \\
\end{tabular}}
\caption{Qualitative and quantitative comparisons among Transformer-based StyTR$^2$, StyTR$^2$ with SCSA, StyTR$^2$ with S-AdaIN, and StyTR$^2$ with Style-Swap.}
\label{fig:20}
\end{figure}

\begin{figure}
\centering
\resizebox{0.45\textwidth}{!}{
\setlength{\tabcolsep}{0.006cm} 
\renewcommand{\arraystretch}{0.25}  
\begin{tabular}{>{\centering\arraybackslash}m{1.2cm} 
 >{\centering\arraybackslash}m{2.4cm} >{\centering\arraybackslash}m{2.4cm} >{\centering\arraybackslash}m{2.4cm} >{\centering\arraybackslash}m{2.4cm}}
 \\
  Inputs & StyleID &   + SCSA &  + S-AdaIN  &   + Style-Swap
\\
\includegraphics[width=1.0\linewidth]{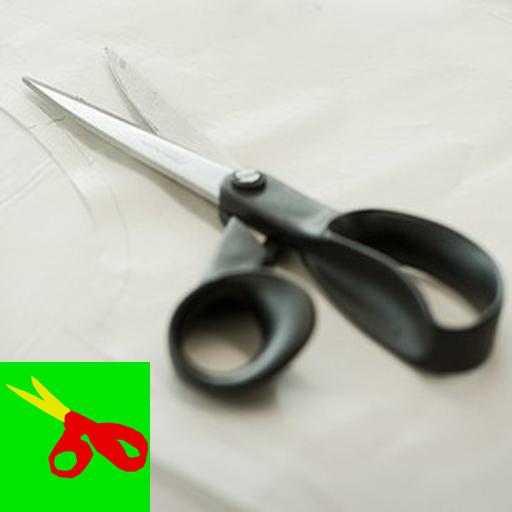} \includegraphics[width=1.0\linewidth]{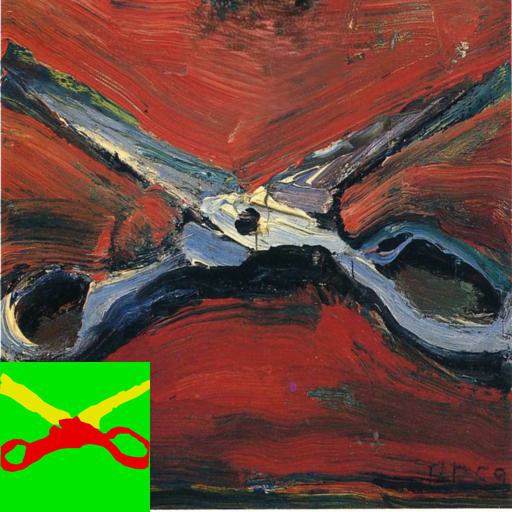} & 
 \includegraphics[width=1.0\linewidth]{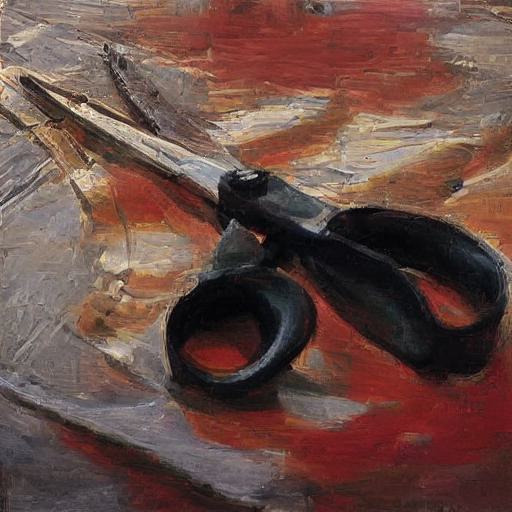} & \includegraphics[width=1.0\linewidth]{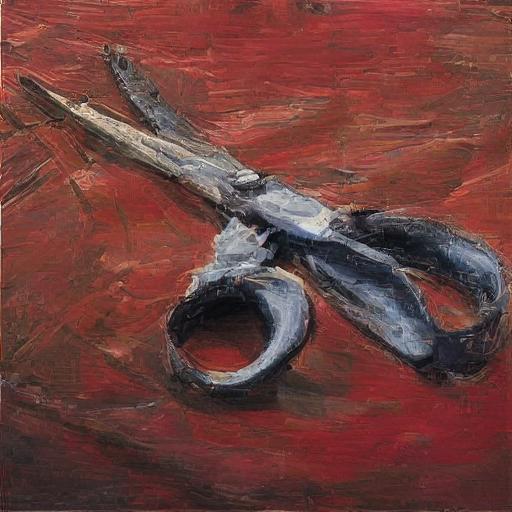} & \includegraphics[width=1.0\linewidth]{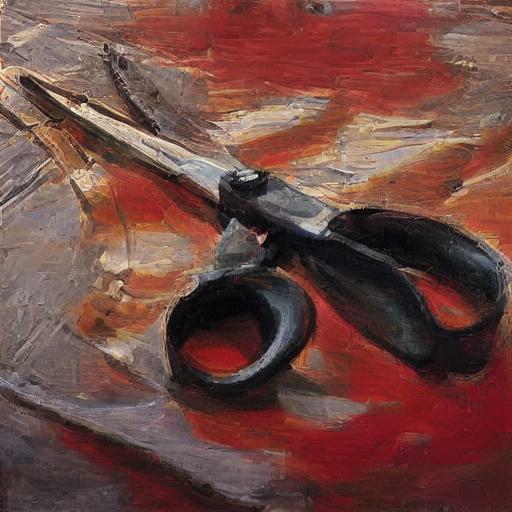} & \includegraphics[width=1.0\linewidth]{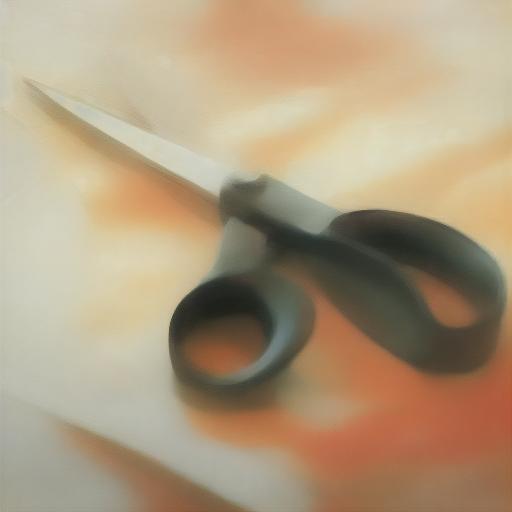} \\
 \includegraphics[width=1.0\linewidth]{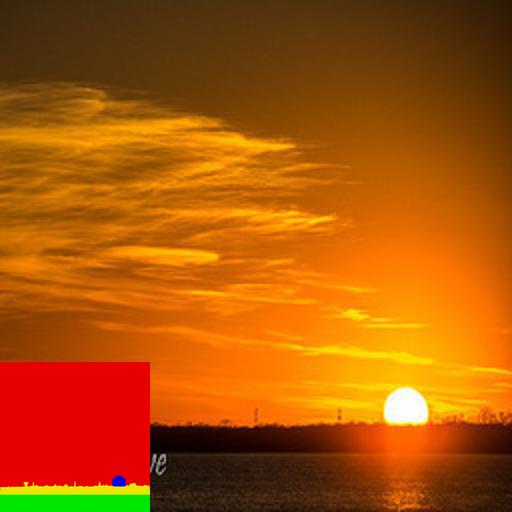} \includegraphics[width=1.0\linewidth]{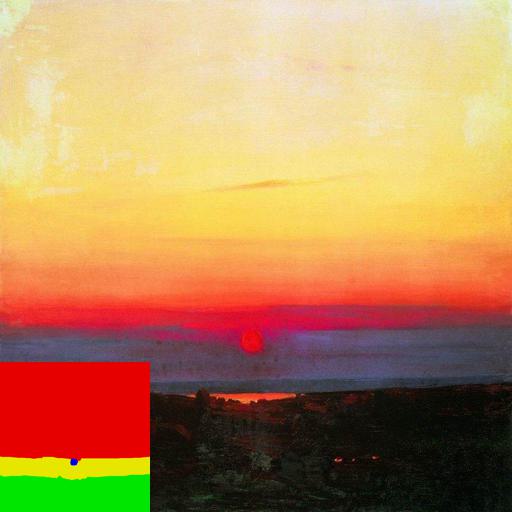} & 
 \includegraphics[width=1.0\linewidth]{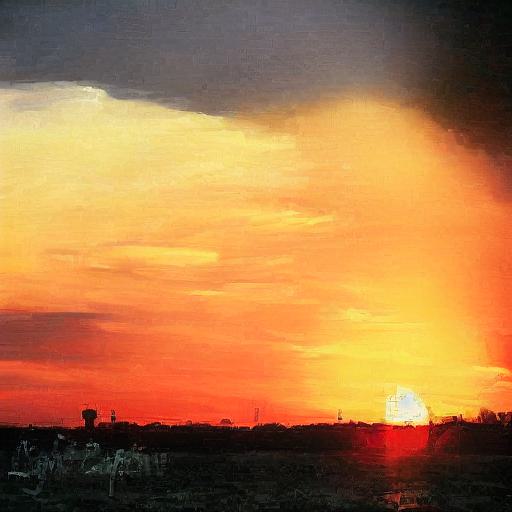} & \includegraphics[width=1.0\linewidth]{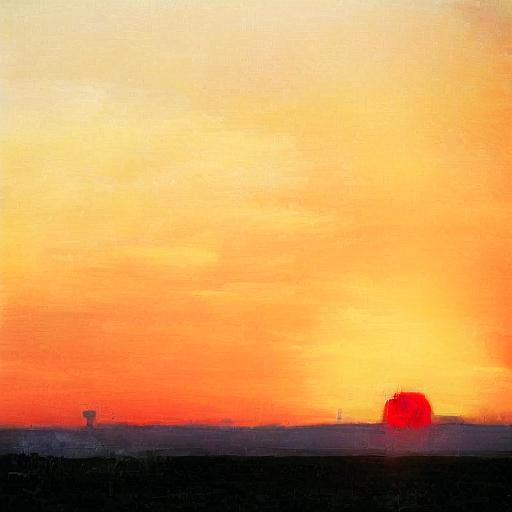} & \includegraphics[width=1.0\linewidth]{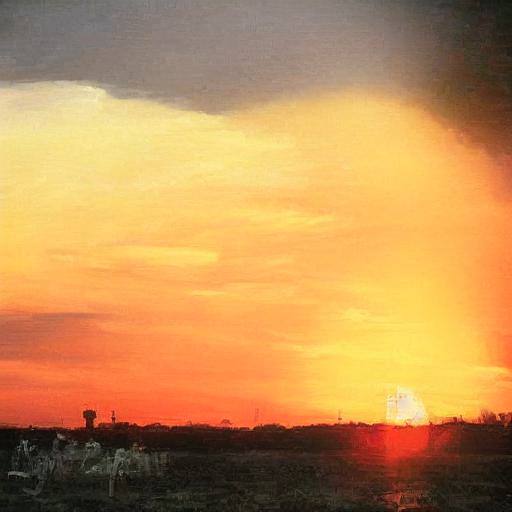} & \includegraphics[width=1.0\linewidth]{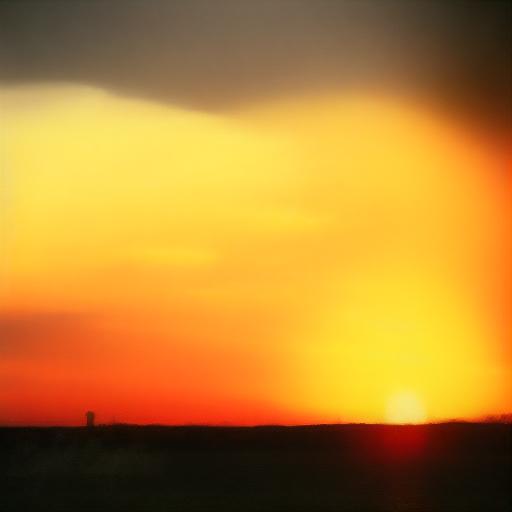}  \\
  \includegraphics[width=1.0\linewidth]{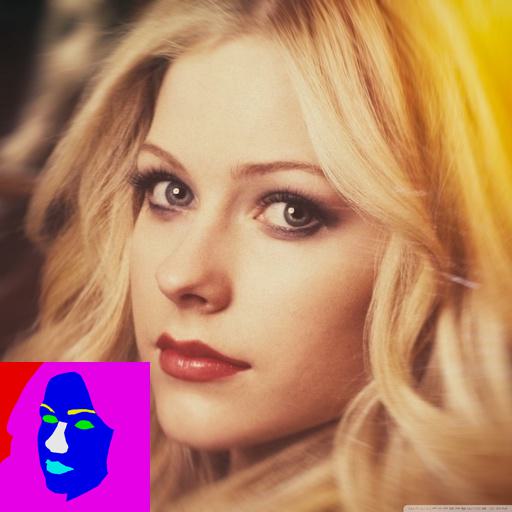} \includegraphics[width=1.0\linewidth]{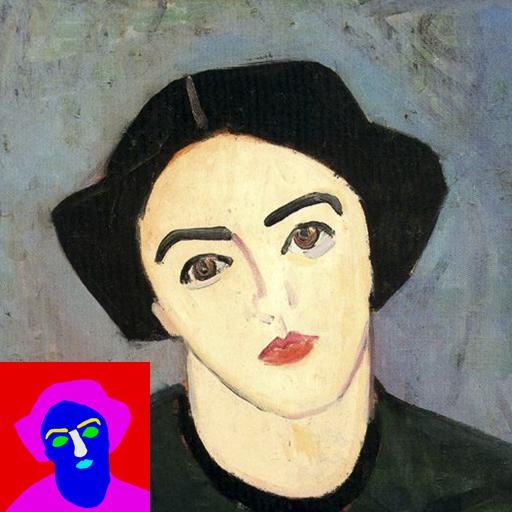} & 
 \includegraphics[width=1.0\linewidth]{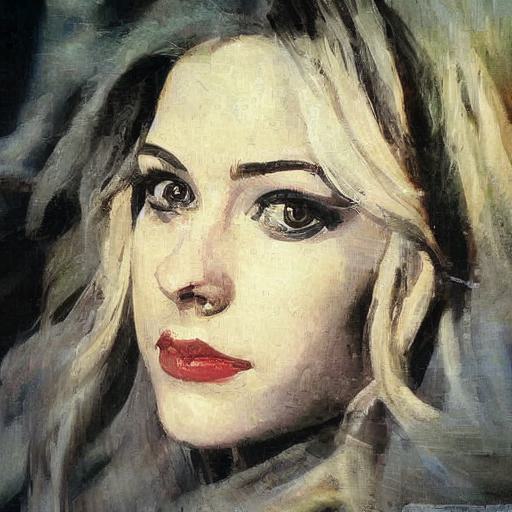} & \includegraphics[width=1.0\linewidth]{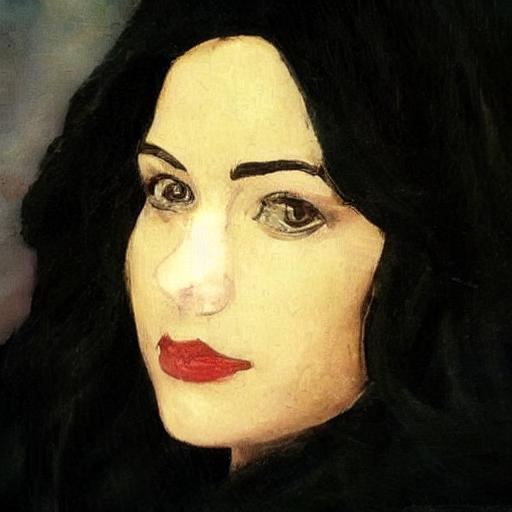} & \includegraphics[width=1.0\linewidth]{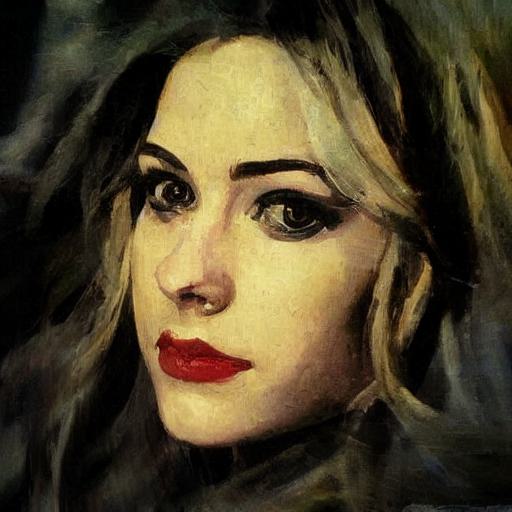} & \includegraphics[width=1.0\linewidth]{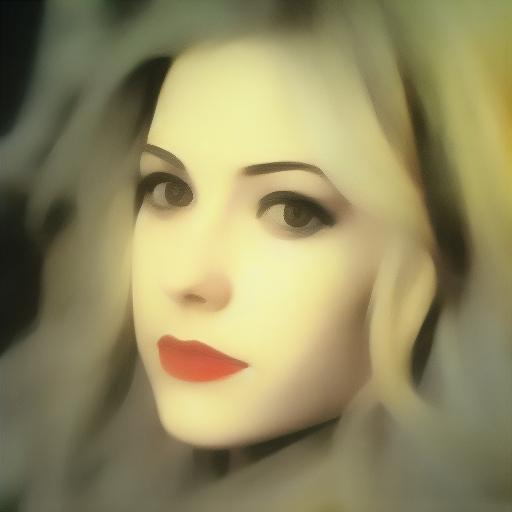}  \\
  &
 \\
 SSL $\downarrow$ & 1.7538 & {\bf 1.2447} & 1.6054 & 2.5032 \\
 &
 \\
 FID $\downarrow$ & 12.5944 & {\bf 12.4497} & 12.5771 & 19.3049 \\
  &
 \\
 CFSD $\downarrow$ & 0.0916 & 0.1178 & 0.0939 & {\bf 0.0732}\\
\end{tabular}}
\caption{Qualitative and quantitative comparisons of Diffusion-based StyleID, StyleID with SCSA, StyleID with S-AdaIN, and StyleID with Style-Swap.}
\label{fig:21}
\end{figure}

{\bf SCSA vs. S-AdaIN.} As shown in Fig.~\ref{fig:19}, Fig.~\ref{fig:20}, and Fig~\ref{fig:21}, S-AdaIN can achieve semantic style transfer to some extent, but the degree of semantic stylization is far inferior to ours. For example, in the $3rd$ row of Fig.~\ref{fig:19}, our heart features dense tree-like textures, while S-AdaIN lacks these, and its global style of semantic regions is not effectively transferred. In the $1st$ row of Fig.~\ref{fig:20}, the overall style of our mountains is continuous, whereas S-AdaIN lacks this. A similar difference is in the $2nd$ row in Fig.~\ref{fig:21}. In addition to the qualitative comparison, the quantitative results from the three figures further demonstrate that the degree of stylization in stylized images generated by S-AdaIN is notably lower than that of our method, as reflected in the higher SSL and FID values. Also, incorporating S-AdaIN into SANet and StyTR$^2$ leads to inferior content preservation compared to our method, as indicated by higher CFSD values.

{\bf SCSA vs. Style-Swap.} As shown in Fig.~\ref{fig:19}, Fig.~\ref{fig:20}, and Fig.~\ref{fig:21}, although Style-Swap can achieve some degree of semantic style transfer, it still lacks the accuracy of the transferred semantic style and the continuity within the same semantic region. A case in point is the $3rd$ row of Fig.~\ref{fig:19} (the heart), the $3rd$ row of Fig.\ref{fig:20} (the grass), and the $1st$ row of Fig.~\ref{fig:21} (the background), where the corresponding semantic regions fail to undergo accurate and continuous style transfer. Furthermore, the overall stylization effect of stylized images generated by Style-Swap is significantly inferior to that of our method, as evidenced by its higher SSL and FID values in the three figures compared to SCSA.

The above analysis demonstrates that our SCSA greatly outperforms both S-AdaIN and Style-Swap in qualitative and quantitative perspectives, proving its superiority.

\section{User Study}
Our user study is primarily divided into two main parts. 

In one part of the user study, we presented participants with corresponding data quadruples \{$I_c, I_{csem}, I_s, I_{ssem}$\} along with a stylized image generated by a traditional Attn-AST method, randomly selected from CNN-based, Transformer-based, and Diffusion-based approaches, and one produced by the corresponding traditional Attn-AST method with our SCSA embedded, both displayed in random order. We asked participants to select, “Which image do you believe represents the most satisfactory result of semantic style transfer?” This part aims to compare user preferences between the SCSA-embedded Attn-AST approaches and the traditional Attn-AST methods, thereby validating the effectiveness and generalization of our SCSA in semantic style transfer. 

In another part, we also presented participants with the corresponding data quadruples \{$I_c, I_{csem}, I_s, I_{ssem}$\}, along with three stylized images generated by the traditional Attn-AST method with our SCSA embedded, CNN-based, Transformer-based, and diffusion-based methods, as well as stylized images produced by five SOTA semantic style transfer methods. The display order of eight stylized images was randomized. Again, we asked participants to select the stylized image they found most satisfactory. This part aims to validate the effectiveness of our SCSA in comparison to SOTA semantic style transfer methods, providing a subjective basis for establishing our approach as a new benchmark for semantic style transfer.

We invited 40 participants to take part in our user study, with each participant responding to a total of 30 questions-15 in each of the two previously mentioned sections. This comprehensive survey design enabled us to gather detailed insights into user preferences, ultimately resulting in the collection of 1,200 votes, which will be essential for our analysis and validation of the findings.

\section{Additional Experiment}
In the main paper, we have verified the generalization of SCSA when integrated in the Attn-AST methods. To further validate the generalization of our proposed SCSA concerning data, we carry out an additional experiment. Specifically, we broaden the scope of semantics, extending them beyond individual instances to encompass regions that share the same semantic mask. In this context, the instance semantics within these regions targeted for style transfer may not be entirely uniform.

\begin{figure*}
\centering
\resizebox{0.975\textwidth}{!}{
\setlength{\tabcolsep}{0.02cm} 
\renewcommand{\arraystretch}{1}  
\begin{tabular}{cccccccc}
 Content & Style & SANet & SANet + SCSA & StyTR$^2$ & StyTR$^2$ + SCSA & StyleID & StyleID + SCSA \\
\includegraphics[width=0.14\linewidth]{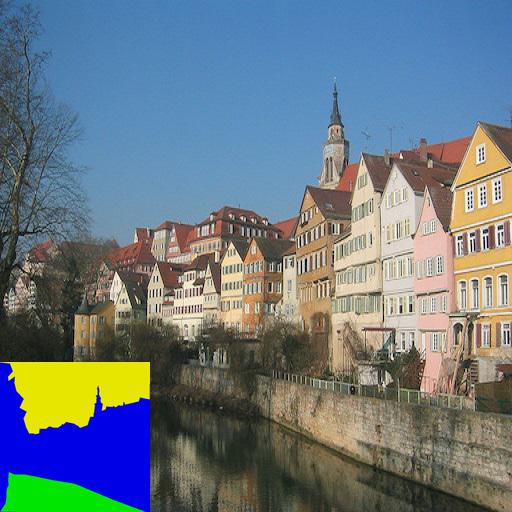} & \includegraphics[width=0.14\linewidth]{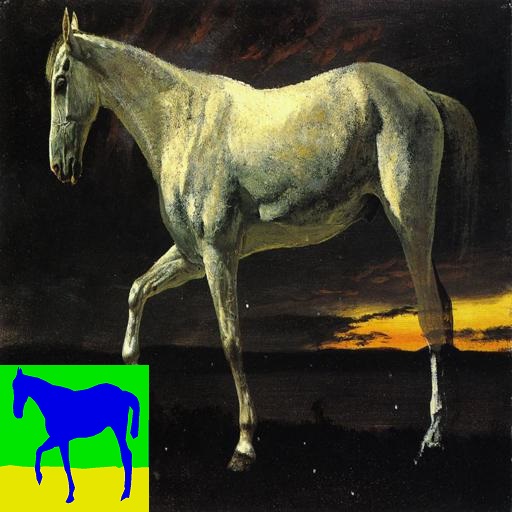}  & \includegraphics[width=0.14\linewidth]{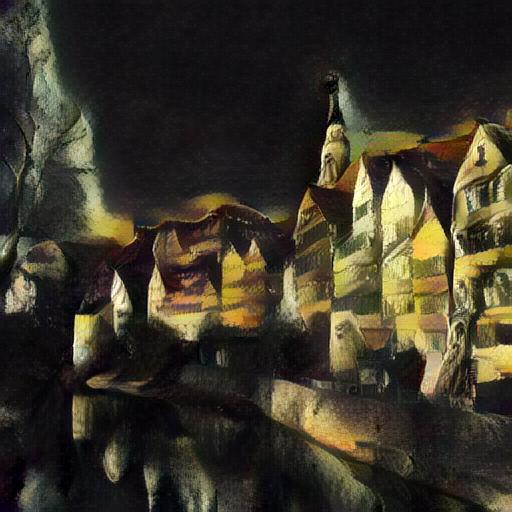}  &
\includegraphics[width=0.14\linewidth]{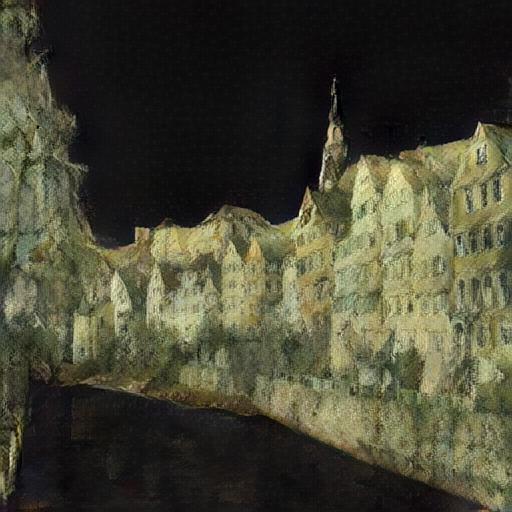}& \includegraphics[width=0.14\linewidth]{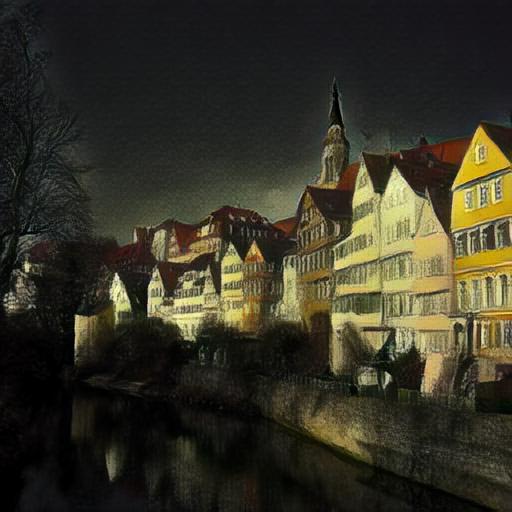} &
\includegraphics[width=0.14\linewidth]{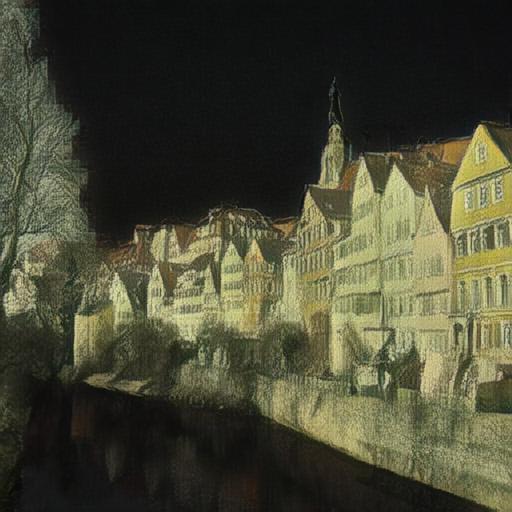} &  \includegraphics[width=0.14\linewidth]{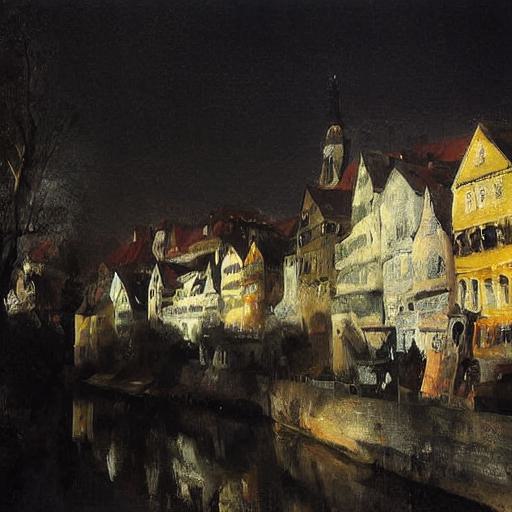} &  \includegraphics[width=0.14\linewidth]{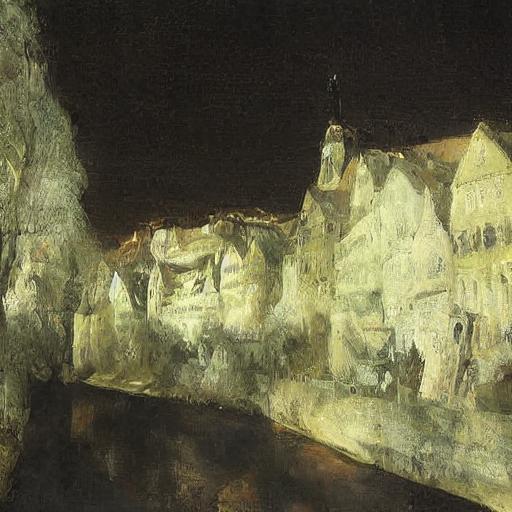} 
\\
\includegraphics[width=0.14\linewidth]{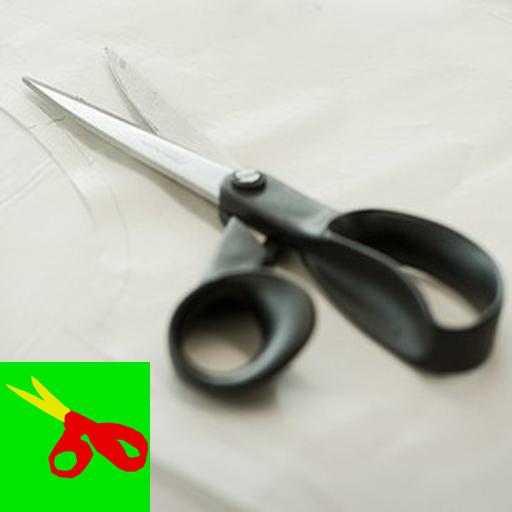} & \includegraphics[width=0.14\linewidth]{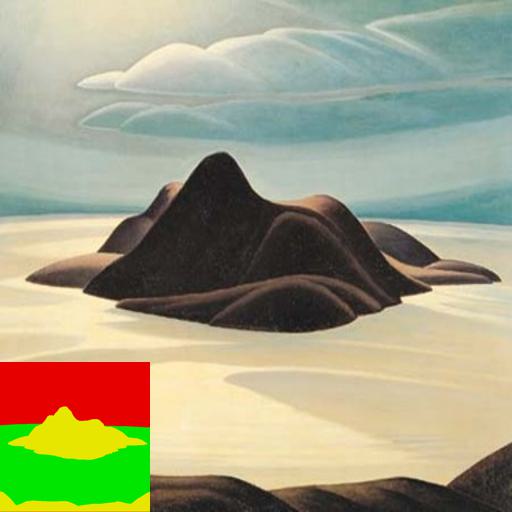}  & \includegraphics[width=0.14\linewidth]{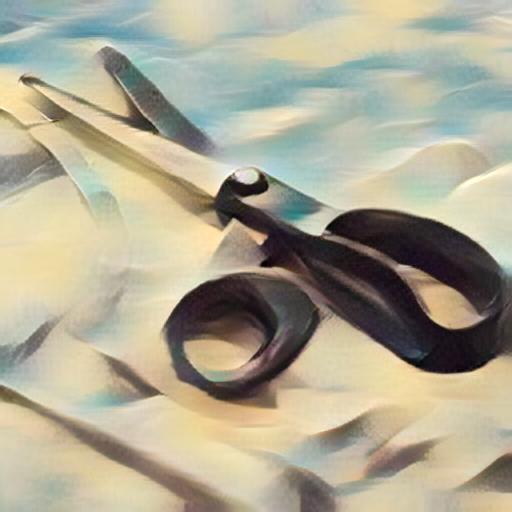}  &
\includegraphics[width=0.14\linewidth]{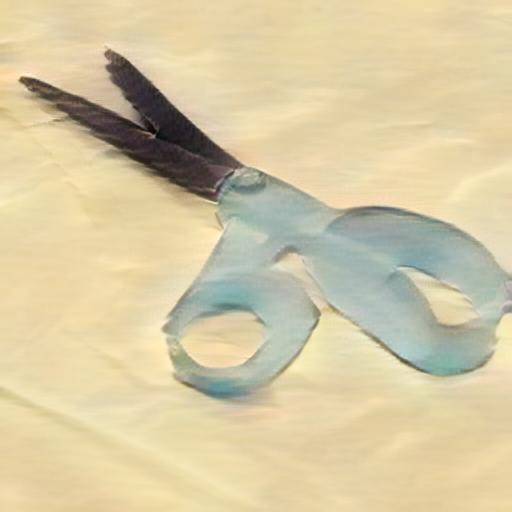}& \includegraphics[width=0.14\linewidth]{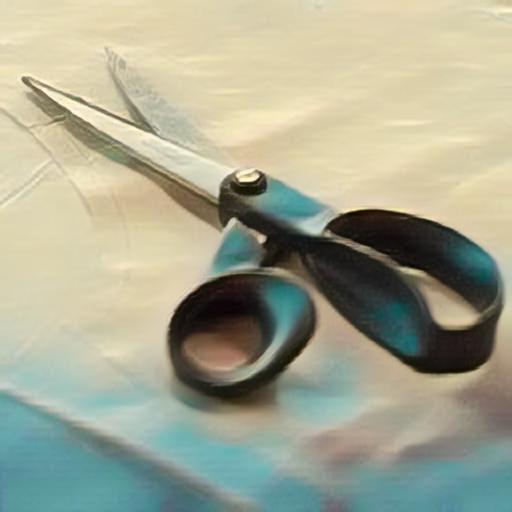} &
\includegraphics[width=0.14\linewidth]{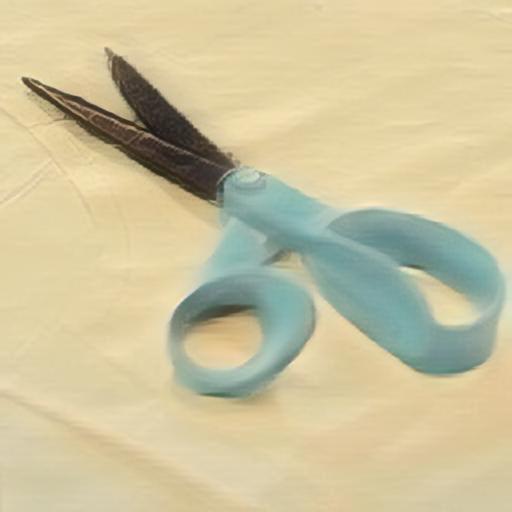} &  \includegraphics[width=0.14\linewidth]{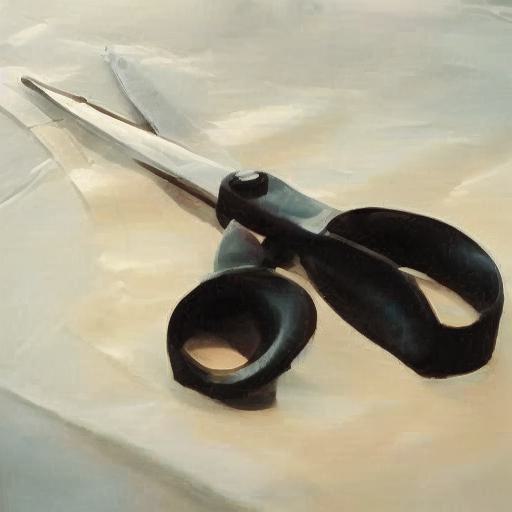} &  \includegraphics[width=0.14\linewidth]{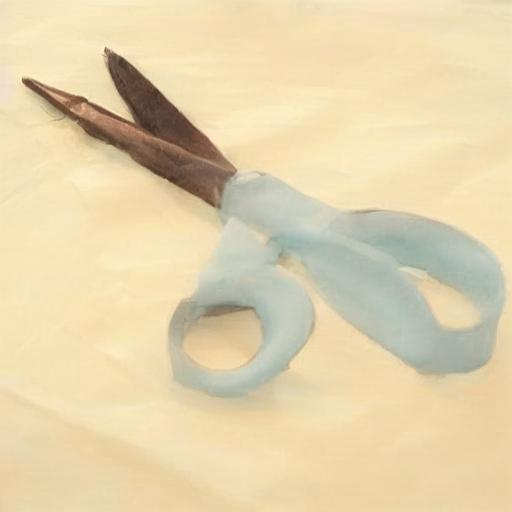} 
\\
\includegraphics[width=0.14\linewidth]{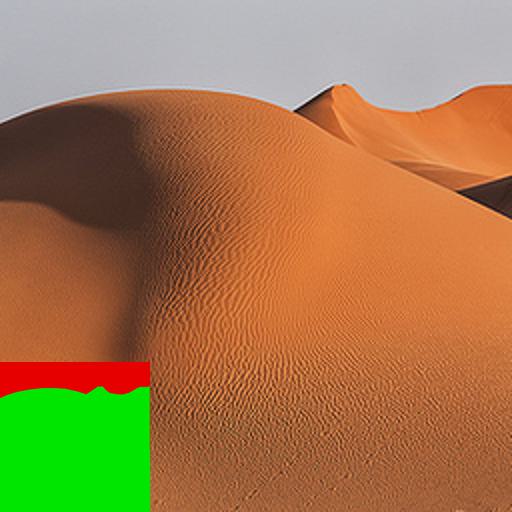}  & 
\includegraphics[width=0.14\linewidth]{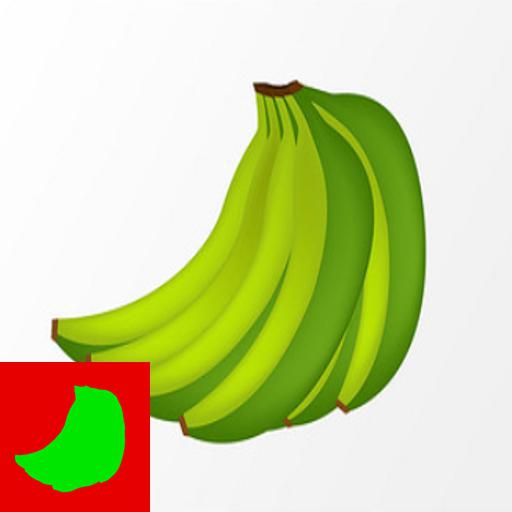} & \includegraphics[width=0.14\linewidth]{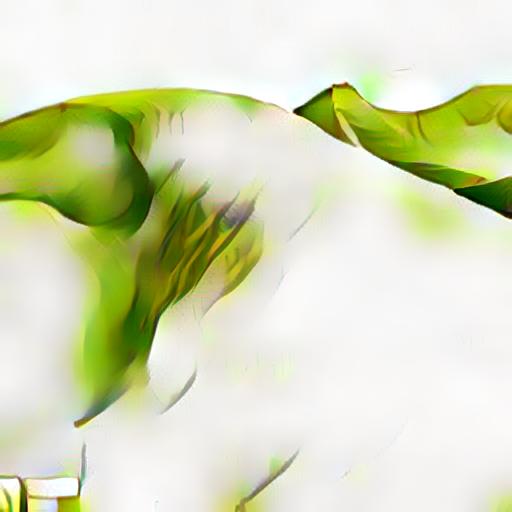}  &
\includegraphics[width=0.14\linewidth]{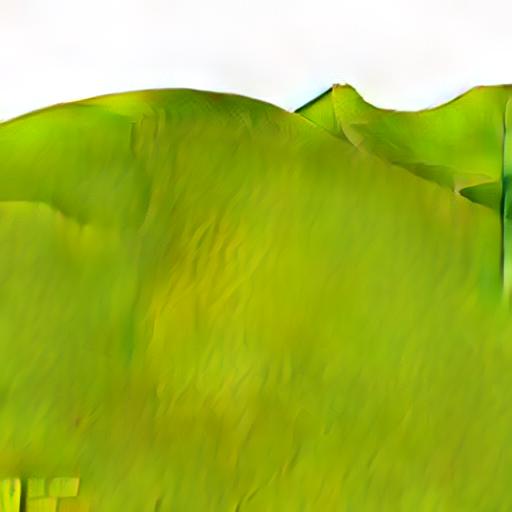}& \includegraphics[width=0.14\linewidth]{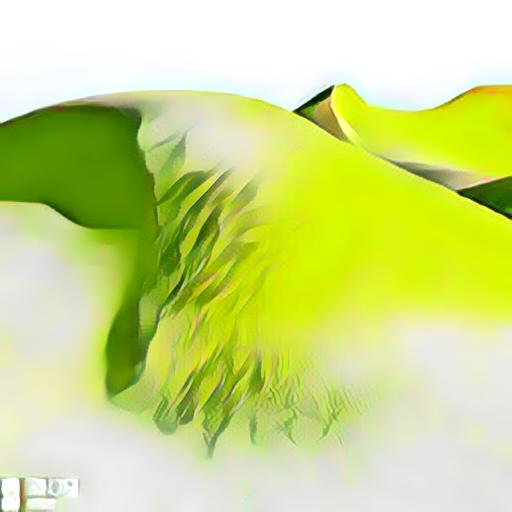} &
\includegraphics[width=0.14\linewidth]{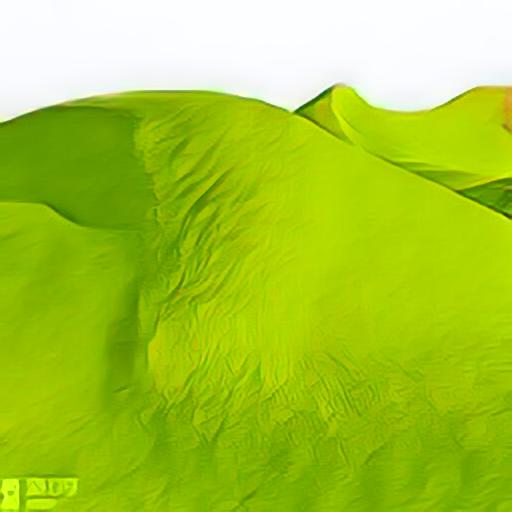} &  \includegraphics[width=0.14\linewidth]{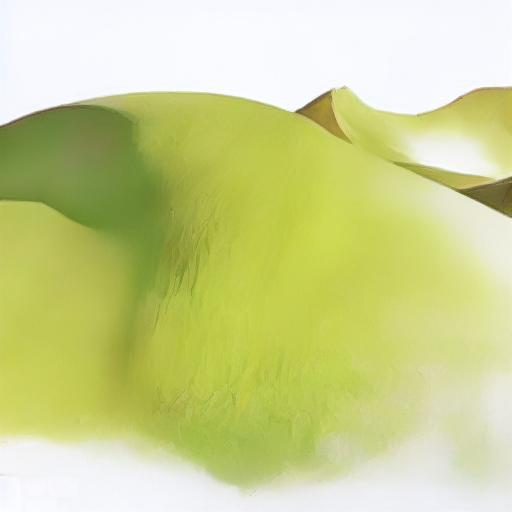} &  \includegraphics[width=0.14\linewidth]{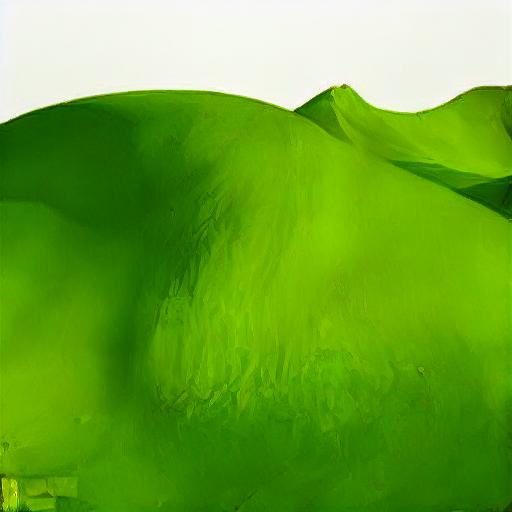} 
\\
\includegraphics[width=0.14\linewidth]{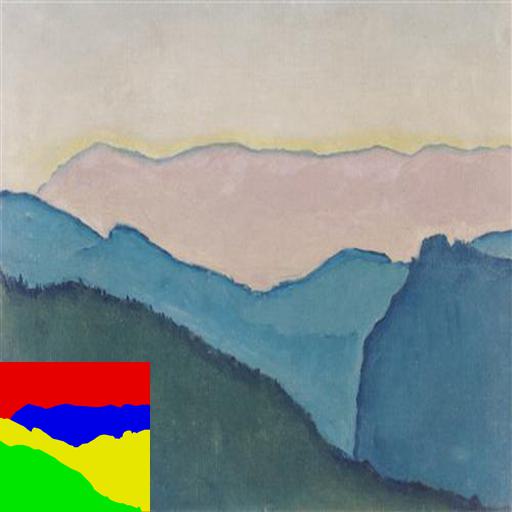} & \includegraphics[width=0.14\linewidth]{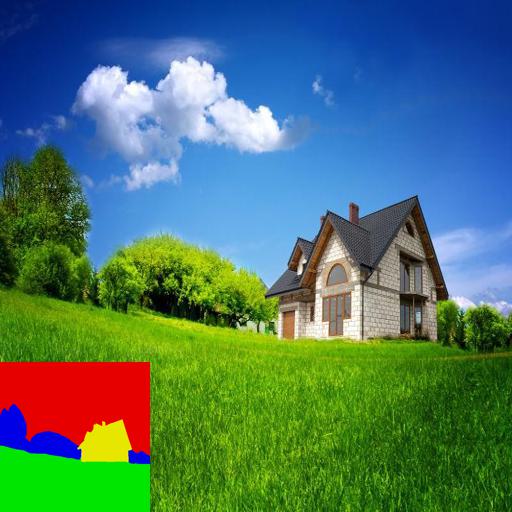}  & \includegraphics[width=0.14\linewidth]{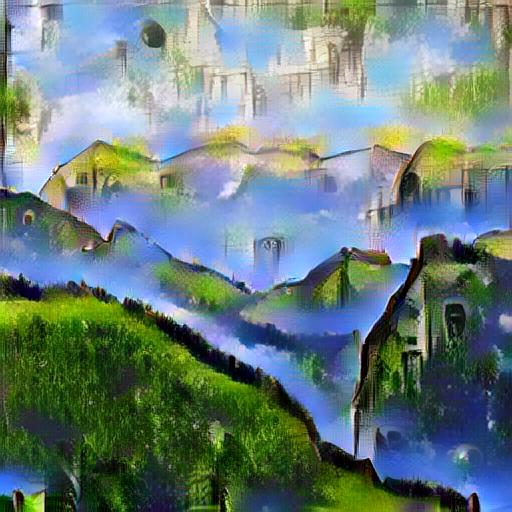}  &
\includegraphics[width=0.14\linewidth]{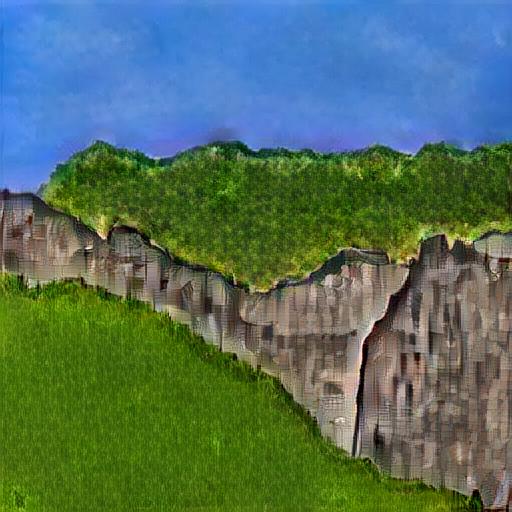}& \includegraphics[width=0.14\linewidth]{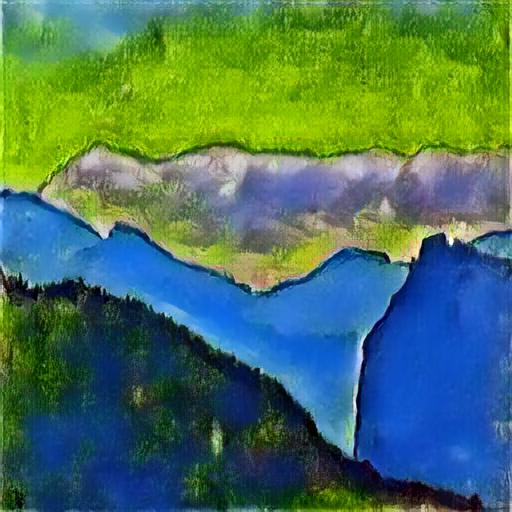} &
\includegraphics[width=0.14\linewidth]{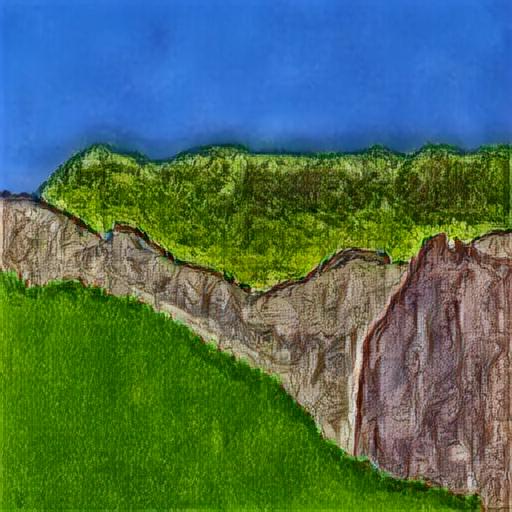} &  \includegraphics[width=0.14\linewidth]{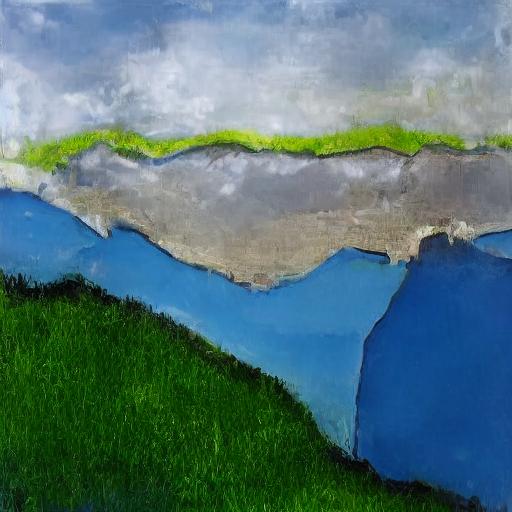} &  \includegraphics[width=0.14\linewidth]{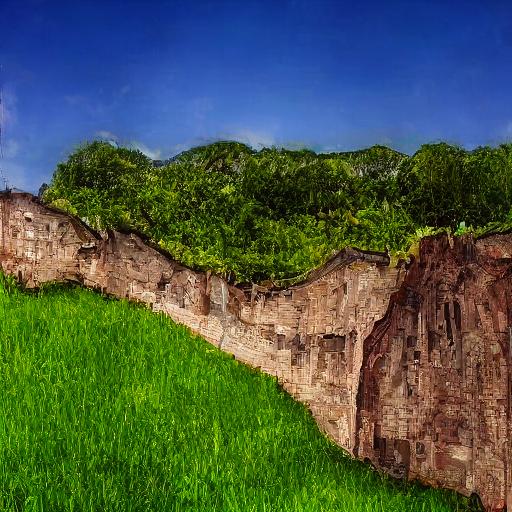} 
\\
\includegraphics[width=0.14\linewidth]{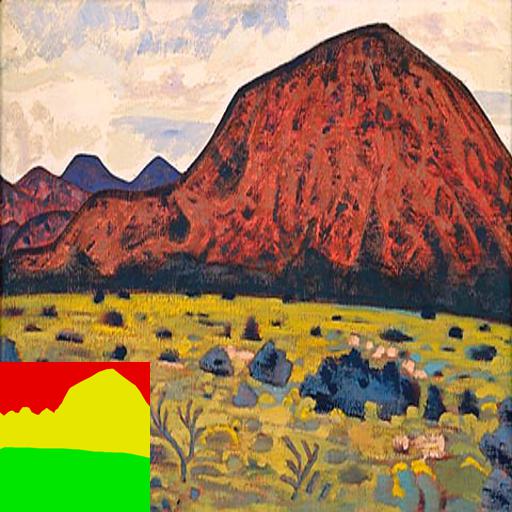} & \includegraphics[width=0.14\linewidth]{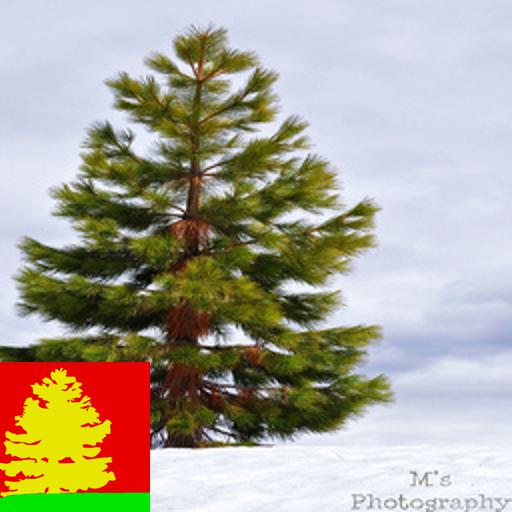}  & \includegraphics[width=0.14\linewidth]{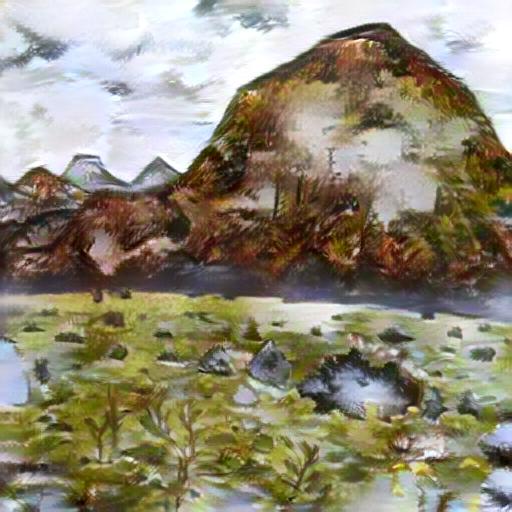}  &
\includegraphics[width=0.14\linewidth]{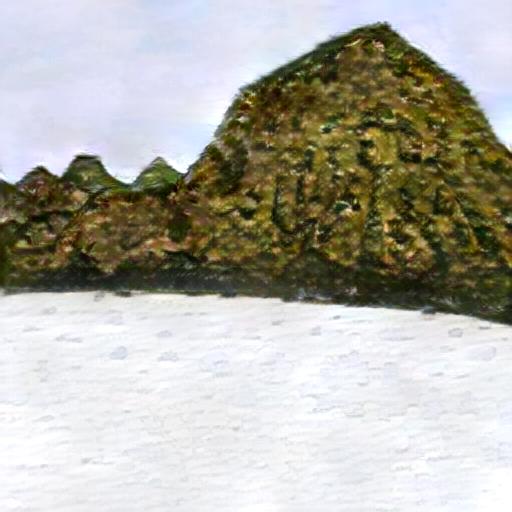}& \includegraphics[width=0.14\linewidth]{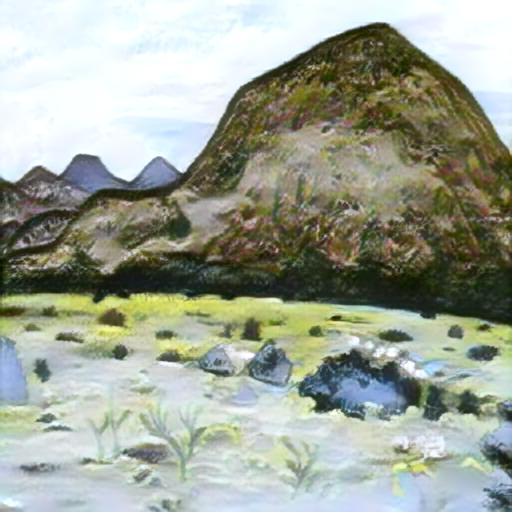} &
\includegraphics[width=0.14\linewidth]{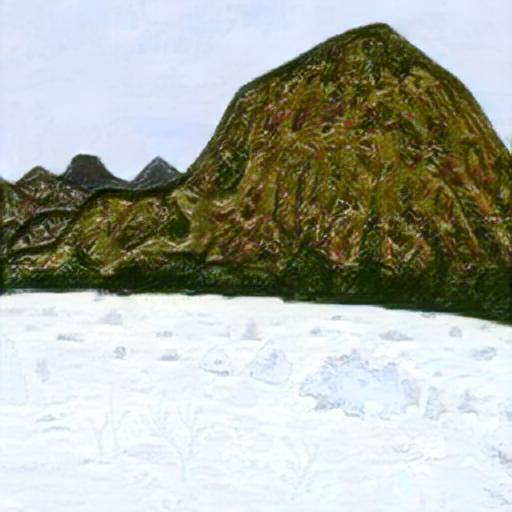} &  \includegraphics[width=0.14\linewidth]{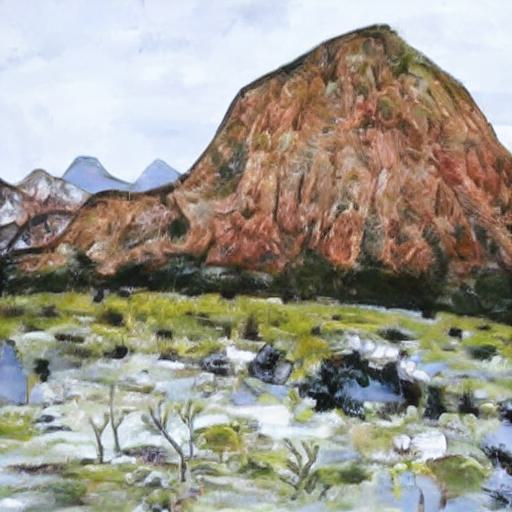} &  \includegraphics[width=0.14\linewidth]{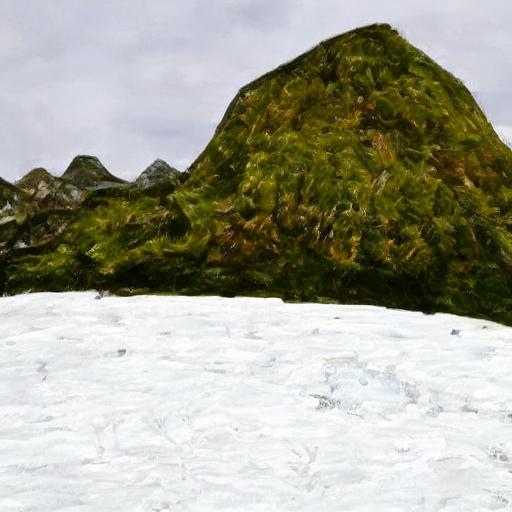} 
\\
\includegraphics[width=0.14\linewidth]{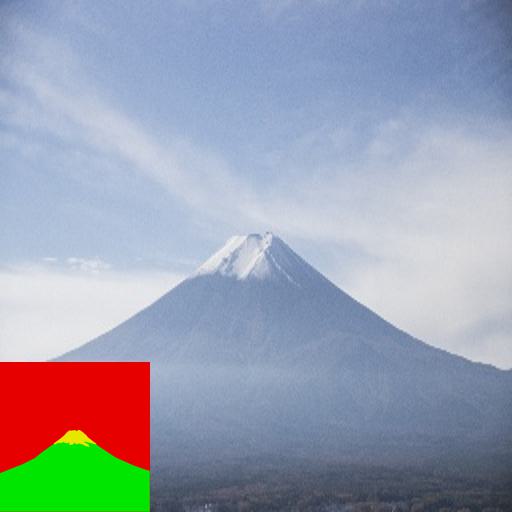} & \includegraphics[width=0.14\linewidth]{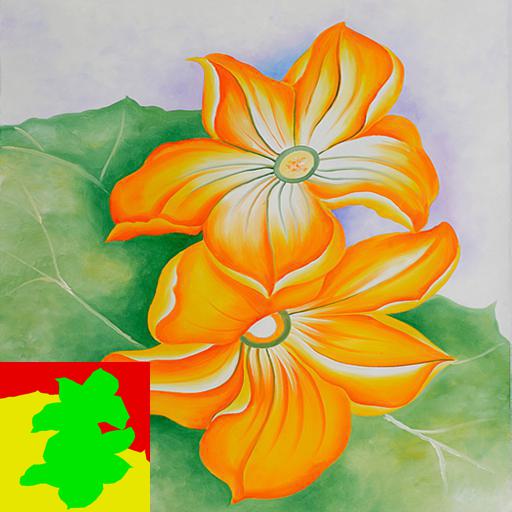}  & \includegraphics[width=0.14\linewidth]{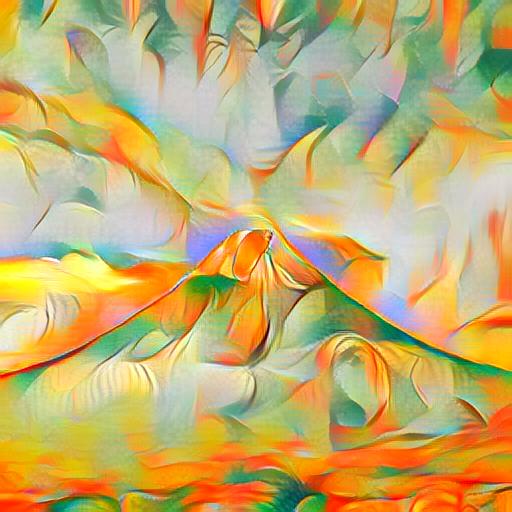}  &
\includegraphics[width=0.14\linewidth]{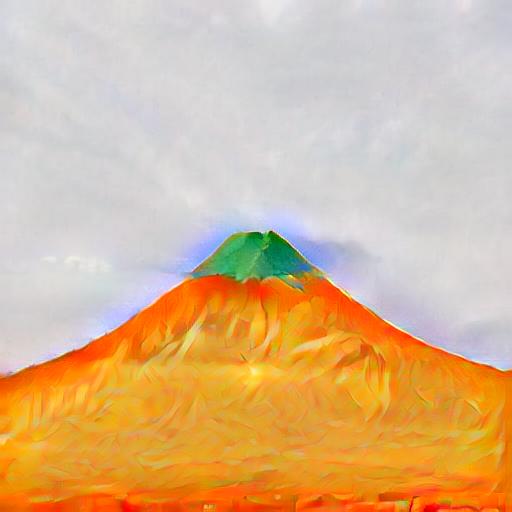}& \includegraphics[width=0.14\linewidth]{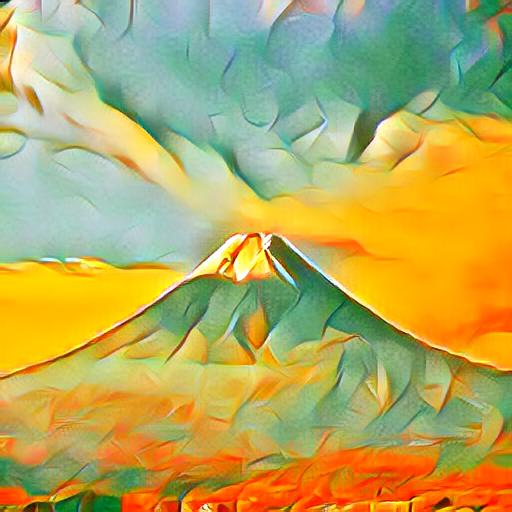} &
\includegraphics[width=0.14\linewidth]{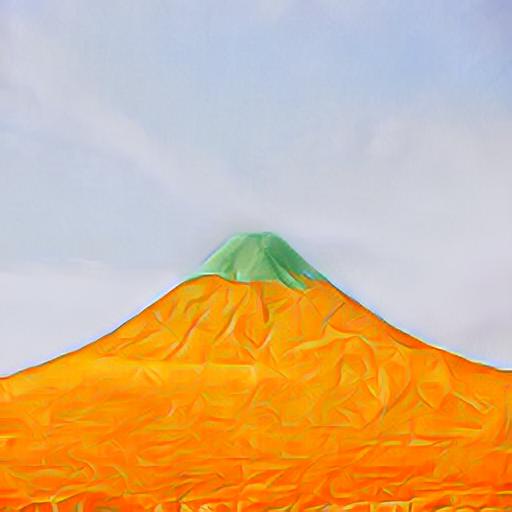} &  \includegraphics[width=0.14\linewidth]{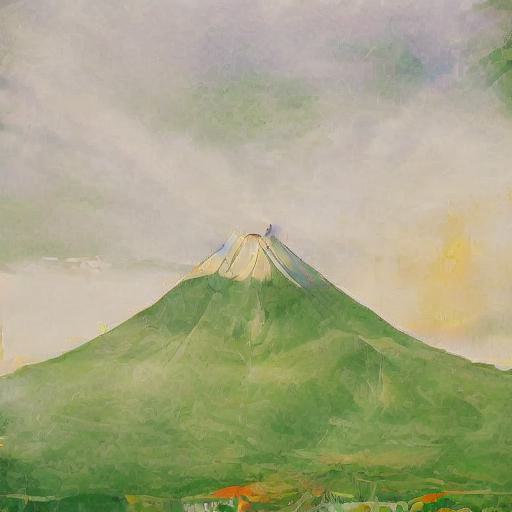} &  \includegraphics[width=0.14\linewidth]{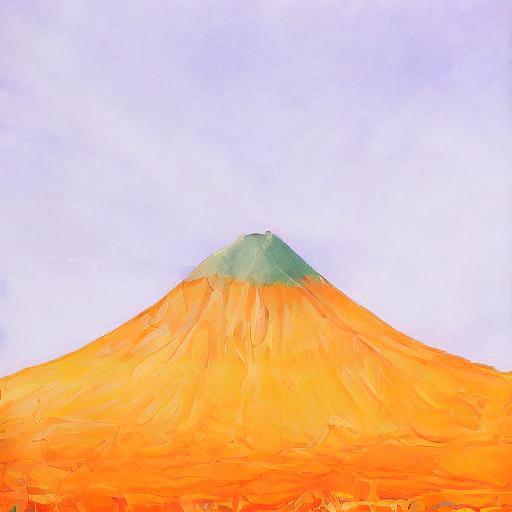}

\end{tabular}
}
\caption{Qualitative comparisons between Attn-AST approaches and they with our SCSA for different semantics.}
\label{fig:23}
\end{figure*}

As shown in Fig.~\ref{fig:23}, our SCSA exhibits remarkable data generalization. It is capable of processing not only data with consistent semantic instances but also data with distinct semantic instances, as long as users can provide corresponding identical mask labels.

\section{More Discussions}

{\bf Effects of Attn-AST methods with SCSA.} As stated in the main paper, SCSA is a plug-and-play semantic style transfer method. The quality of the stylized images it produces is directly influenced by the capability of the underlying Attn-AST model. Especially, the stronger the transfer style ability of the Attn-AST model, the more remarkable the semantic stylization effects achieved when integrated with SCSA.

{\bf Processing costs of Attn-AST methods with SCSA.} As SCSA incorporates the processing and guidance of semantics, Attn-AST methods with SCSA require more time and memory than the original Attn-AST. However, our primary focus is currently on the effectiveness of semantic stylization, with efficiency optimization planned as a future research direction.

\section{Qualitative Comparison}

To conduct a more in-depth evaluation of SCSA, we present a broader array of qualitative results.

\begin{figure*}
\centering
\resizebox{1.0\textwidth}{!}{
\setlength{\tabcolsep}{0.02cm} 
\renewcommand{\arraystretch}{1}  
\begin{tabular}{cccccccc}
 Content & Style & SANet & SANet + SCSA & StyTR$^2$ & StyTR$^2$ + SCSA & StyleID & StyleID + SCSA \\
\includegraphics[width=0.14\linewidth]{sm/img/7_paint+sem.jpg} & \includegraphics[width=0.14\linewidth]{sm/img/7+sem.jpg}  & \includegraphics[width=0.14\linewidth]{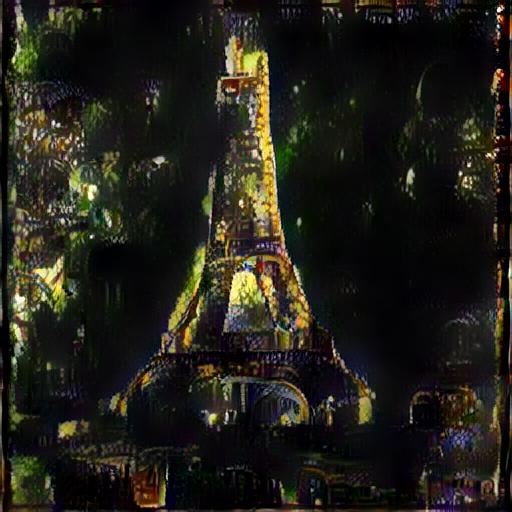}  &
\includegraphics[width=0.14\linewidth]{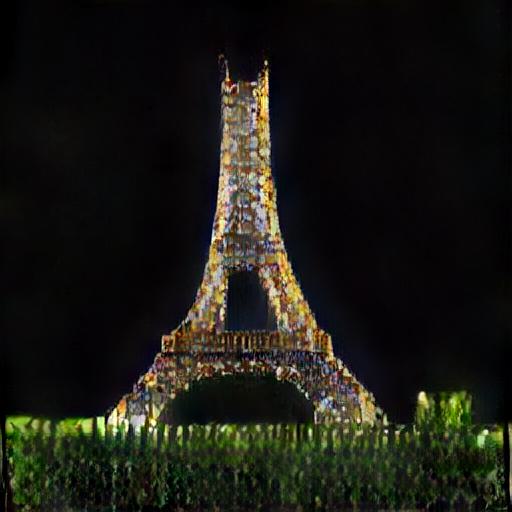}& \includegraphics[width=0.14\linewidth]{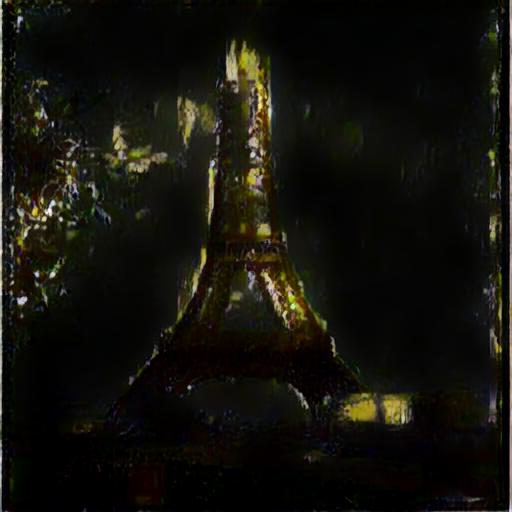} &
\includegraphics[width=0.14\linewidth]{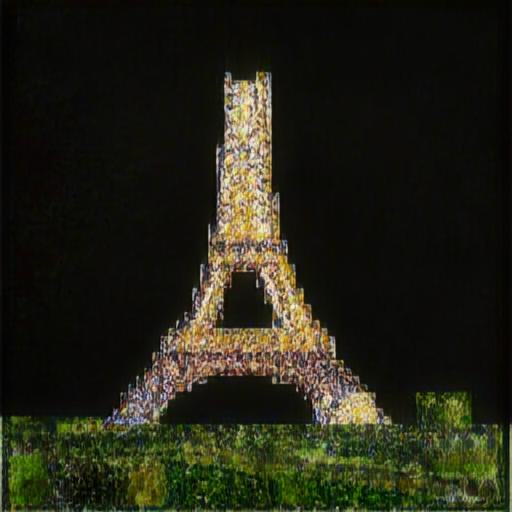} &  \includegraphics[width=0.14\linewidth]{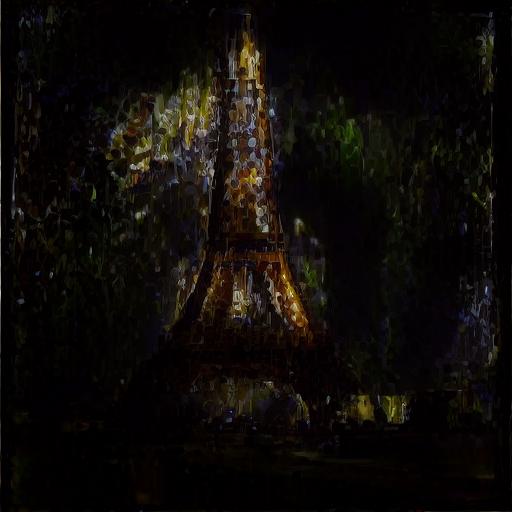} &  \includegraphics[width=0.14\linewidth]{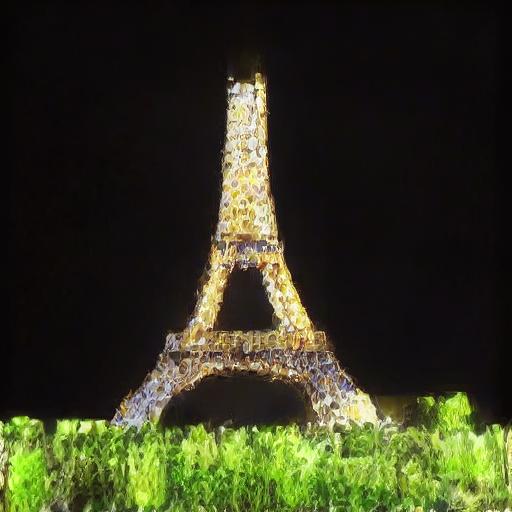} \\
& & & STROTSS & MAST & TR & DIA & GLStyleNet \\
& & & \includegraphics[width=0.14\linewidth]{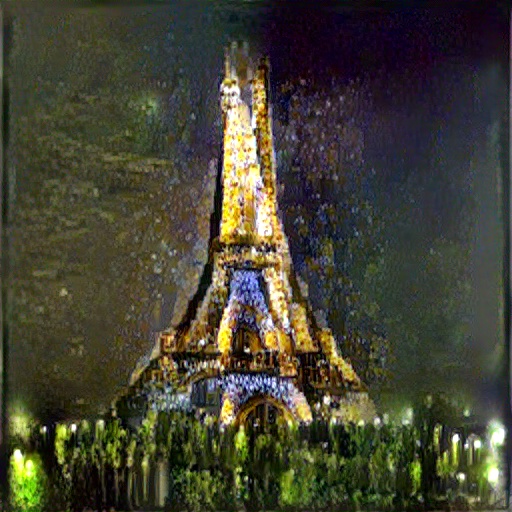} & \includegraphics[width=0.14\linewidth]{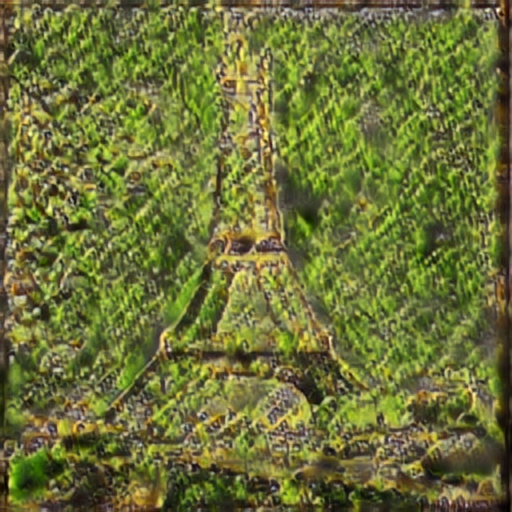} & 
\includegraphics[width=0.14\linewidth]{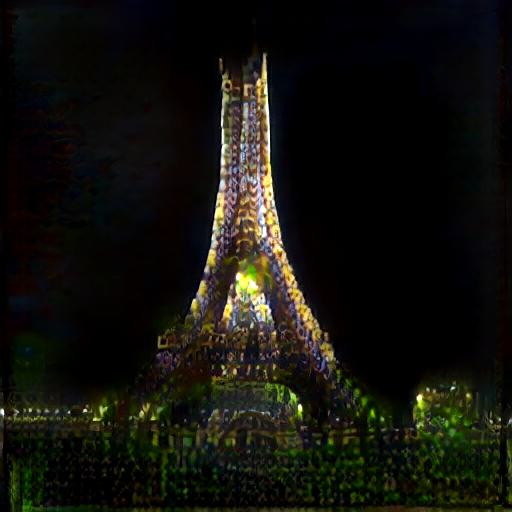} &
\includegraphics[width=0.14\linewidth]{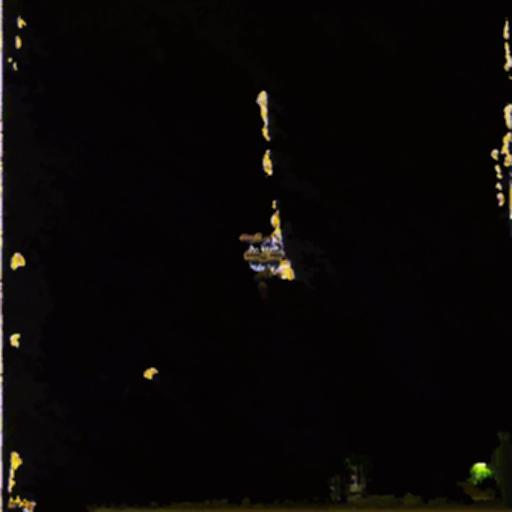} & 
\includegraphics[width=0.14\linewidth]{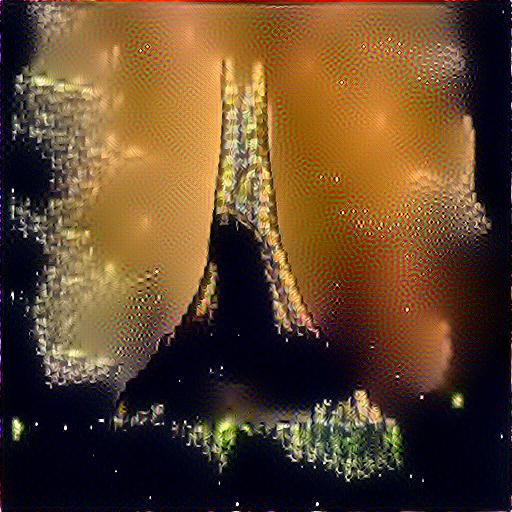} \\
& 
\\
 Content & Style & SANet & SANet + SCSA & StyTR$^2$ & StyTR$^2$ + SCSA & StyleID & StyleID + SCSA \\

\includegraphics[width=0.14\linewidth]{sm/img/1_paint+sem.jpg} & \includegraphics[width=0.14\linewidth]{sm/img/1+sem.jpg}  & \includegraphics[width=0.14\linewidth]{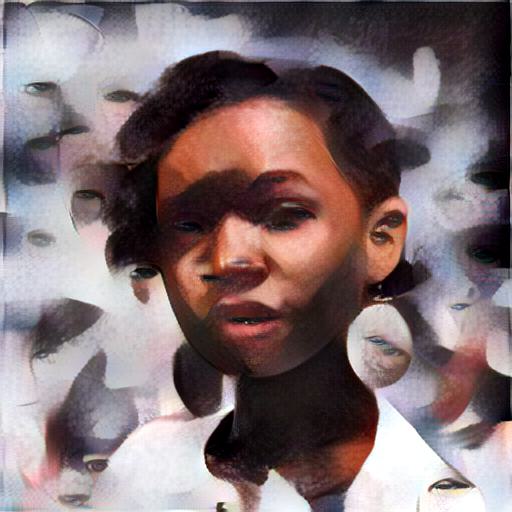}  &
\includegraphics[width=0.14\linewidth]{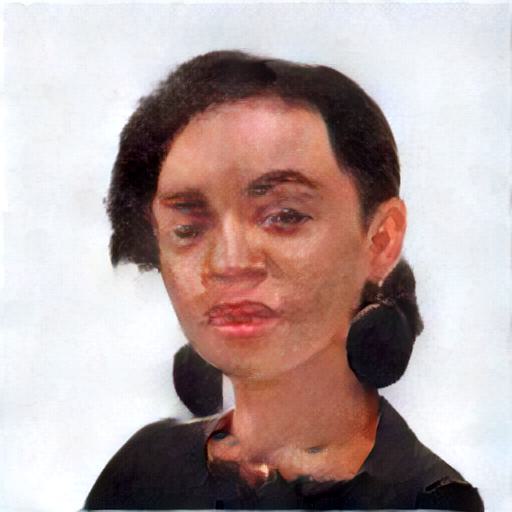}& \includegraphics[width=0.14\linewidth]{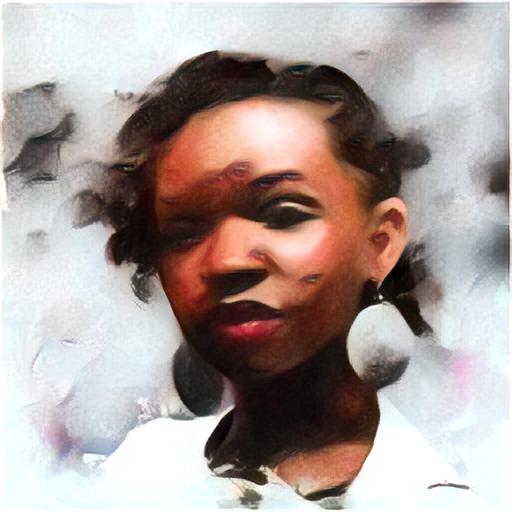} &
\includegraphics[width=0.14\linewidth]{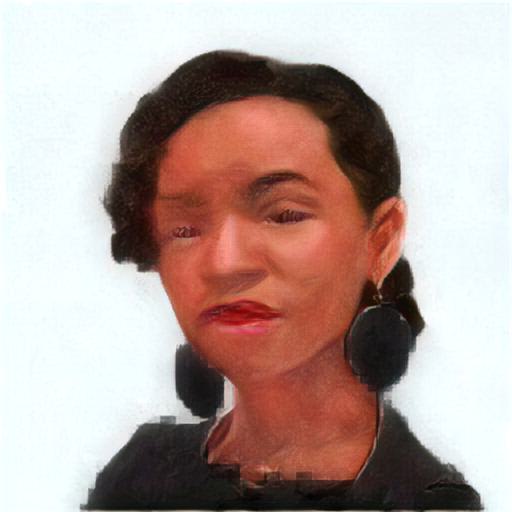} &  \includegraphics[width=0.14\linewidth]{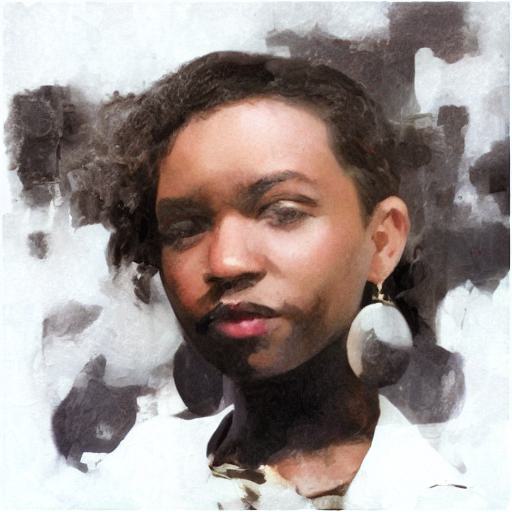} &  \includegraphics[width=0.14\linewidth]{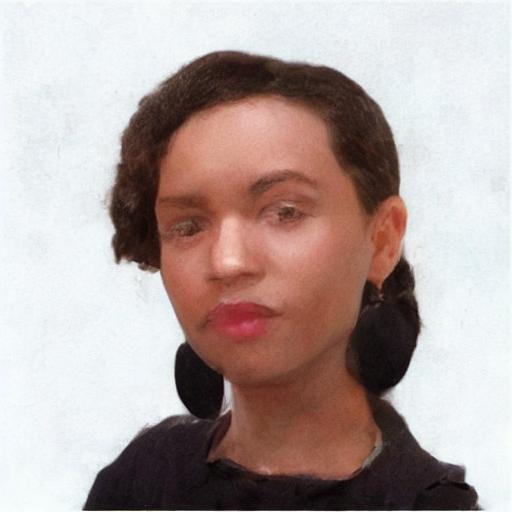} \\
& & & STROTSS & MAST & TR & DIA & GLStyleNet \\
& & & \includegraphics[width=0.14\linewidth]{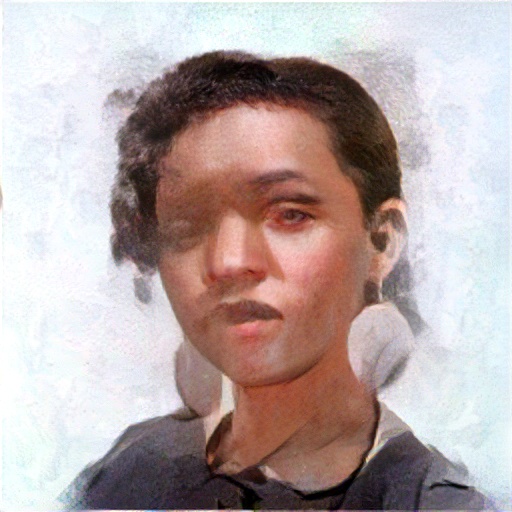} & \includegraphics[width=0.14\linewidth]{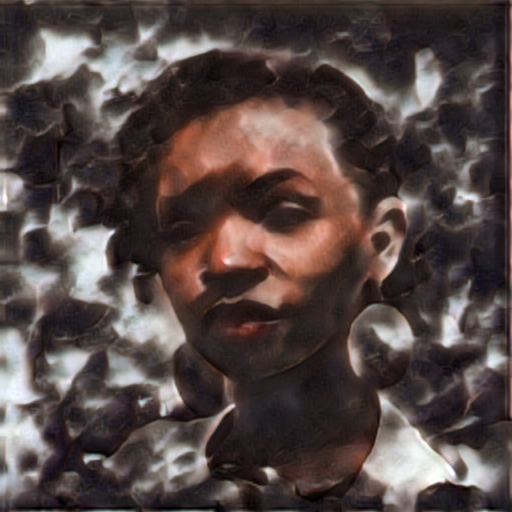} & 
\includegraphics[width=0.14\linewidth]{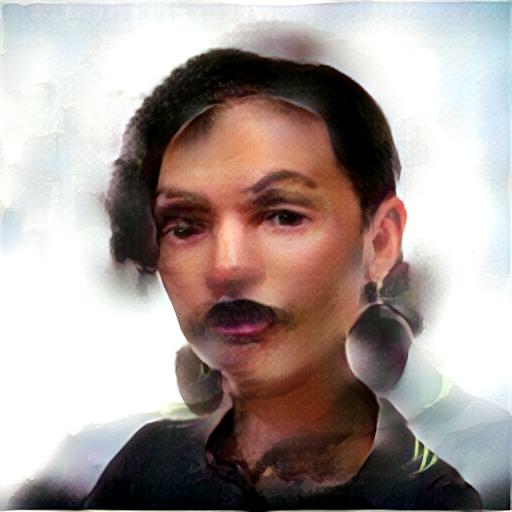} &
\includegraphics[width=0.14\linewidth]{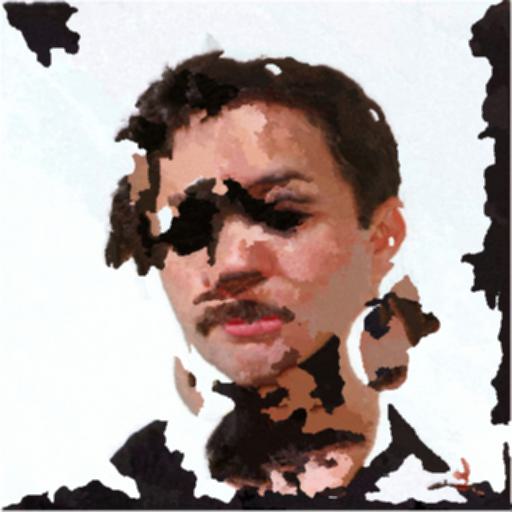} & 
\includegraphics[width=0.14\linewidth]{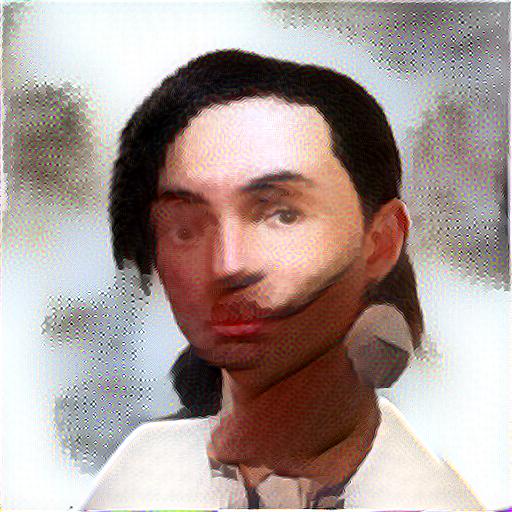} \\
&
\\
 Content & Style & SANet & SANet + SCSA & StyTR$^2$ & StyTR$^2$ + SCSA & StyleID & StyleID + SCSA \\

\includegraphics[width=0.14\linewidth]{sm/img/3_paint+sem.jpg} & \includegraphics[width=0.14\linewidth]{sm/img/3+sem.jpg}  & \includegraphics[width=0.14\linewidth]{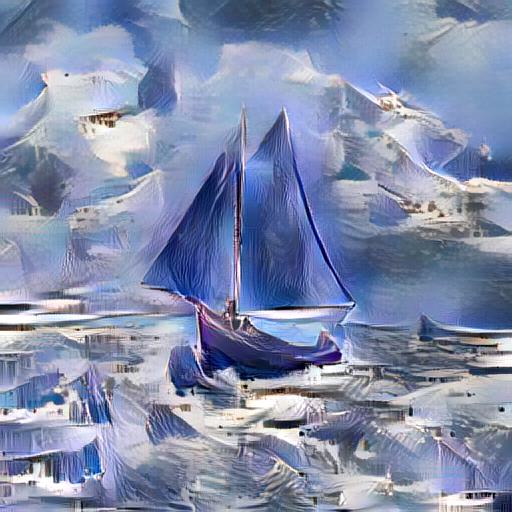}  &
\includegraphics[width=0.14\linewidth]{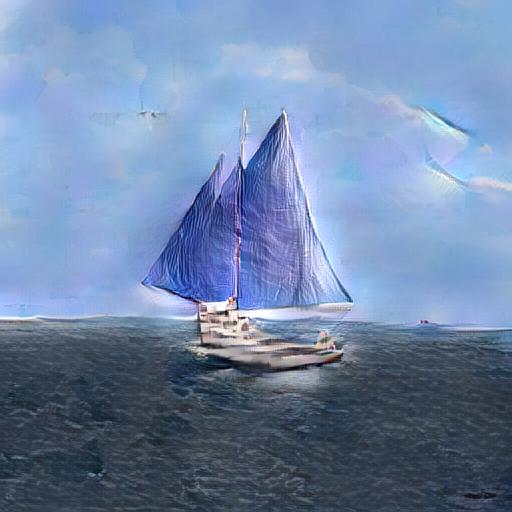}& \includegraphics[width=0.14\linewidth]{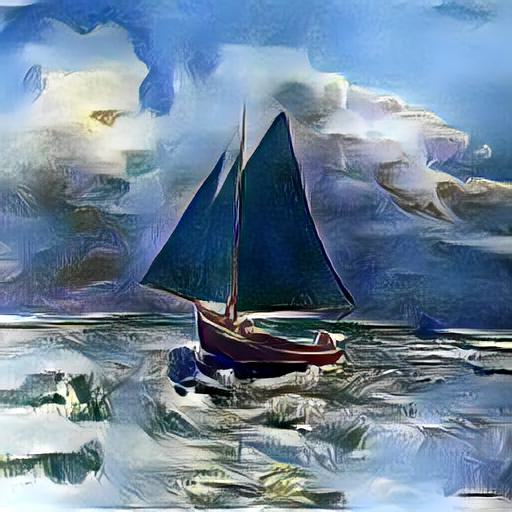} &
\includegraphics[width=0.14\linewidth]{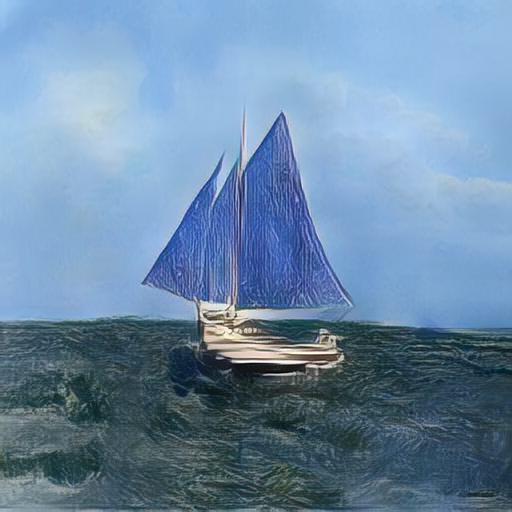} &  \includegraphics[width=0.14\linewidth]{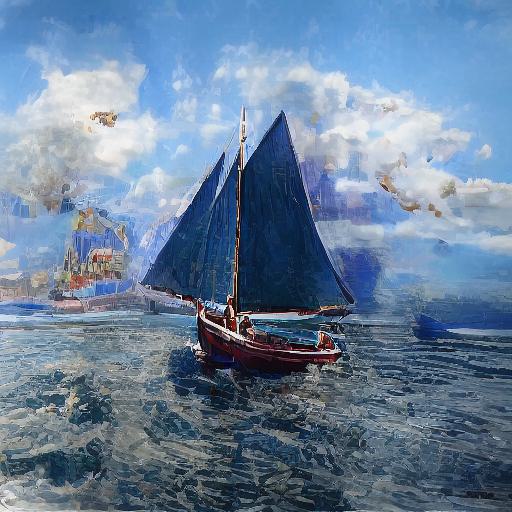} &  \includegraphics[width=0.14\linewidth]{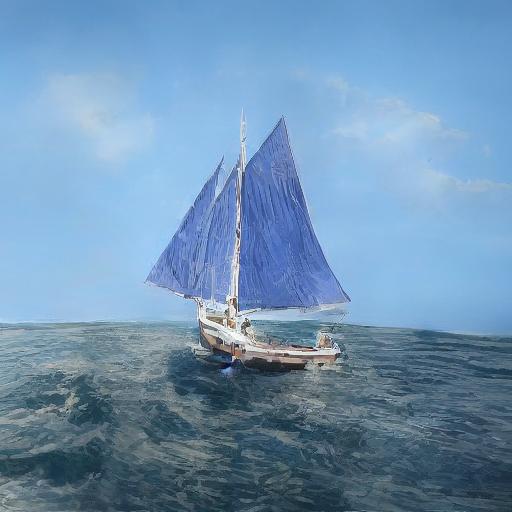} \\
& & & STROTSS & MAST & TR & DIA & GLStyleNet \\
& & & \includegraphics[width=0.14\linewidth]{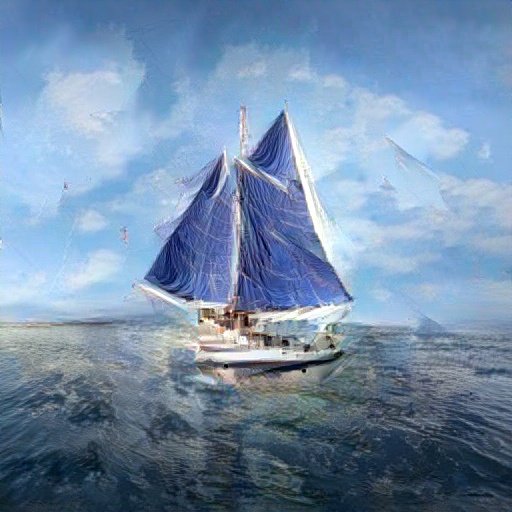} & \includegraphics[width=0.14\linewidth]{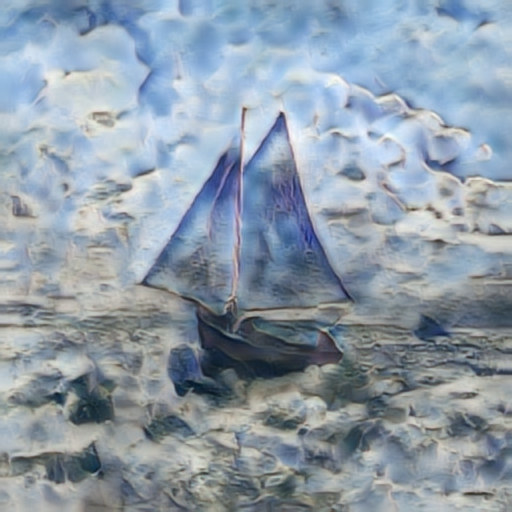} & 
\includegraphics[width=0.14\linewidth]{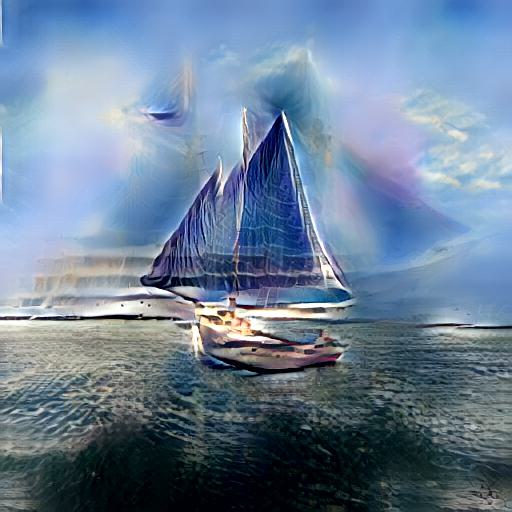} &
\includegraphics[width=0.14\linewidth]{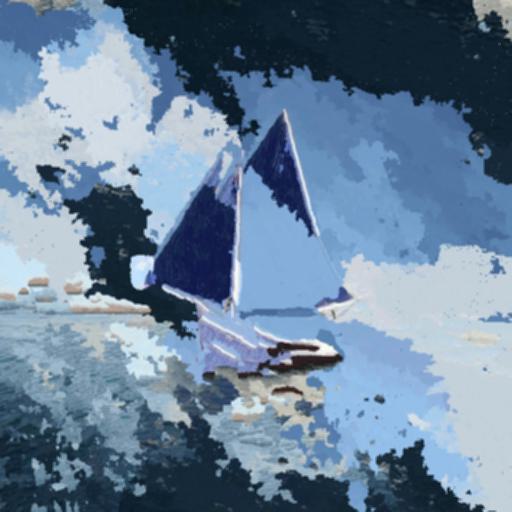} & 
\includegraphics[width=0.14\linewidth]{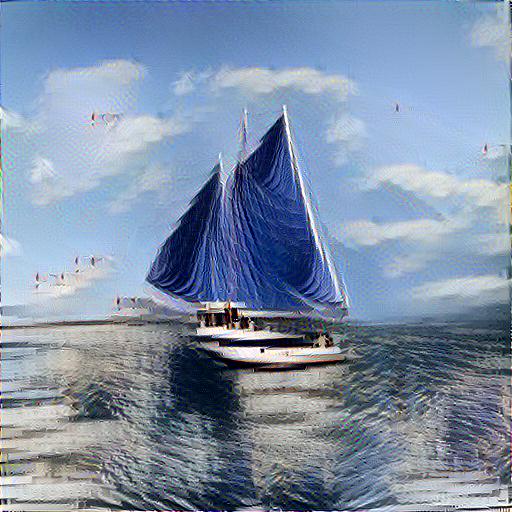} \\

&
\\
 Content & Style & SANet & SANet + SCSA & StyTR$^2$ & StyTR$^2$ + SCSA & StyleID & StyleID + SCSA \\

\includegraphics[width=0.14\linewidth]{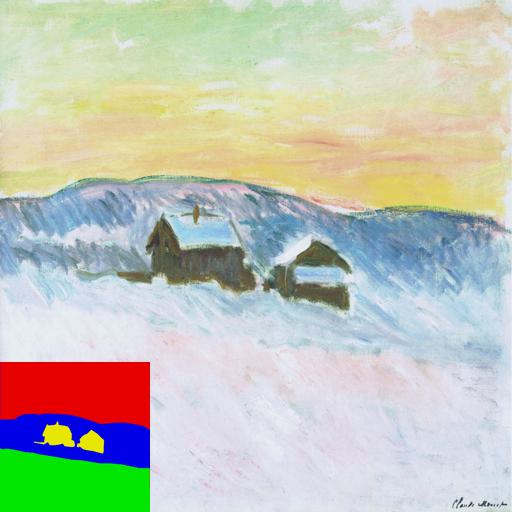} & \includegraphics[width=0.14\linewidth]{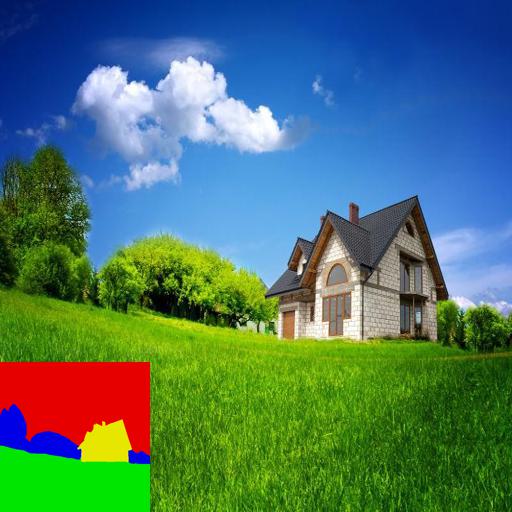}  & \includegraphics[width=0.14\linewidth]{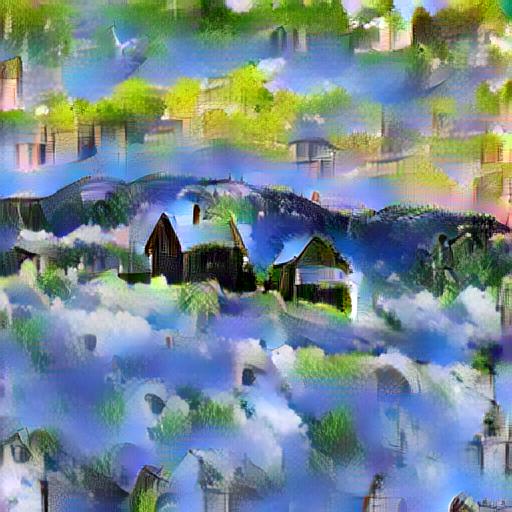}  &
\includegraphics[width=0.14\linewidth]{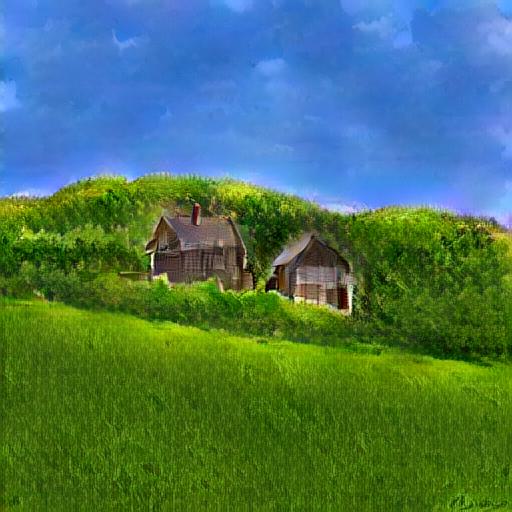}& \includegraphics[width=0.14\linewidth]{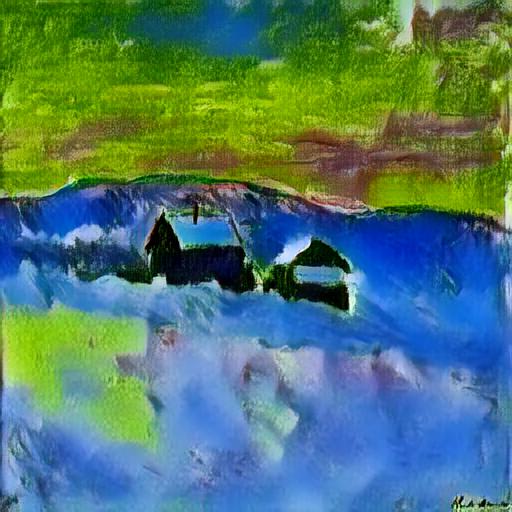} &
\includegraphics[width=0.14\linewidth]{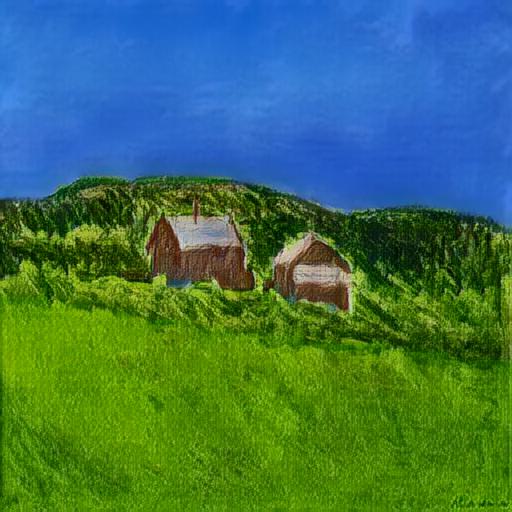} &  \includegraphics[width=0.14\linewidth]{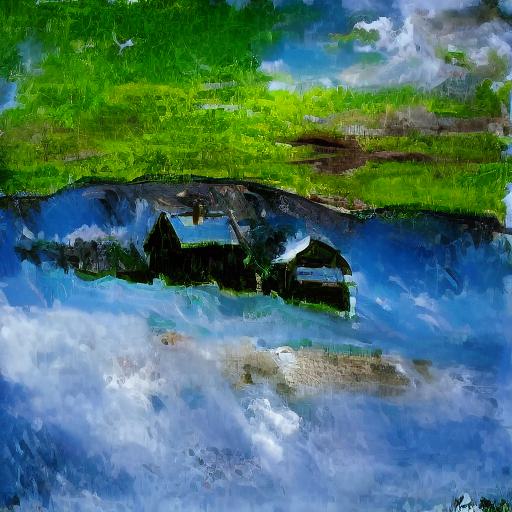} &  \includegraphics[width=0.14\linewidth]{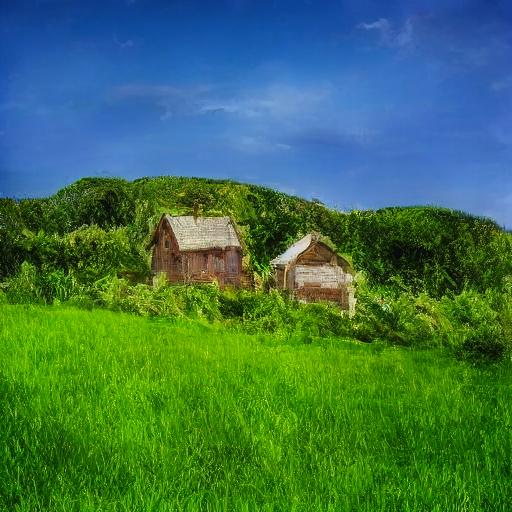} \\
& & & STROTSS & MAST & TR & DIA & GLStyleNet \\
& & & \includegraphics[width=0.14\linewidth]{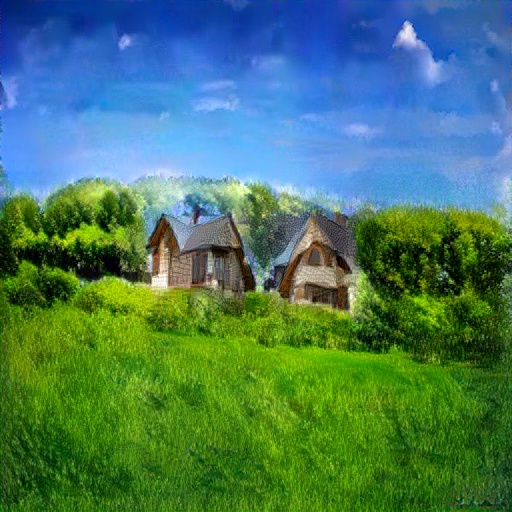} & \includegraphics[width=0.14\linewidth]{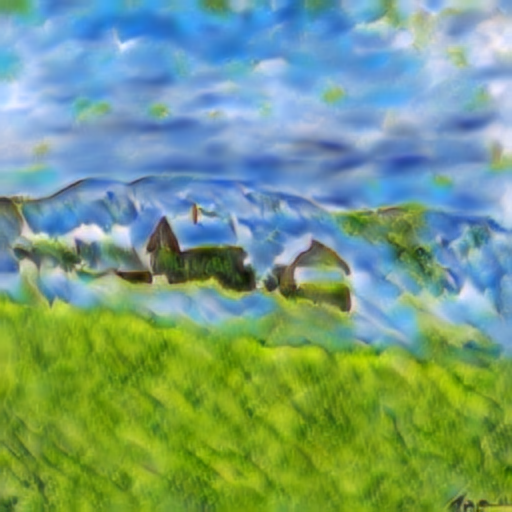} & 
\includegraphics[width=0.14\linewidth]{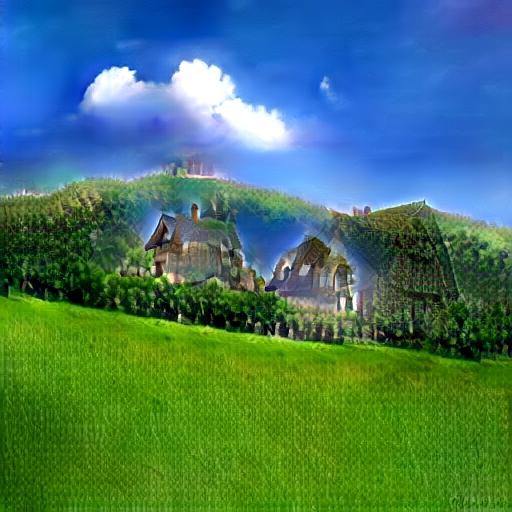} &
\includegraphics[width=0.14\linewidth]{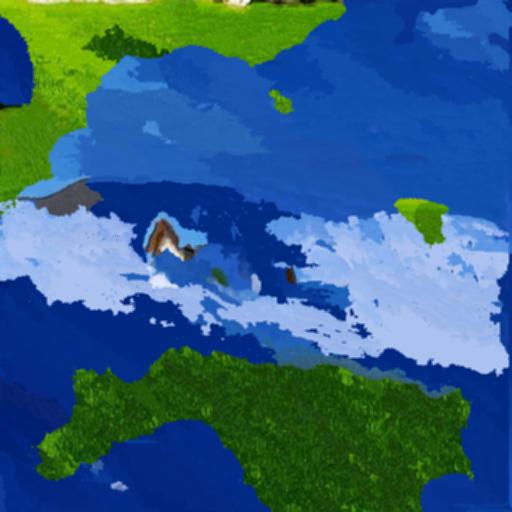} & 
\includegraphics[width=0.14\linewidth]{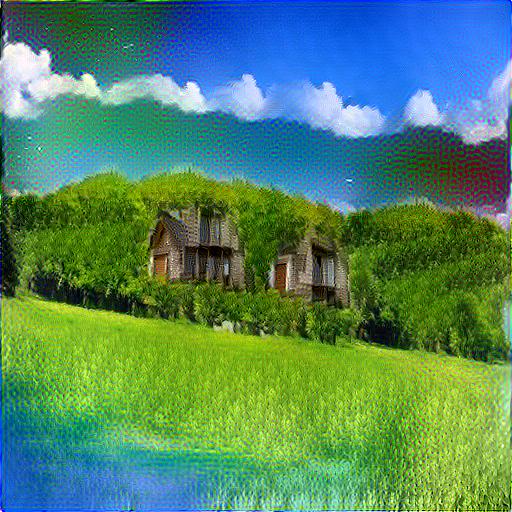} \\

\end{tabular}
}
\caption{Qualitative comparisons among Attn-AST approaches, those with SCSA, and SOTA methods.}
\label{fig:13}
\end{figure*}

\begin{figure*}
\centering
\resizebox{1.0\textwidth}{!}{
\setlength{\tabcolsep}{0.02cm} 
\renewcommand{\arraystretch}{1}  
\begin{tabular}{cccccccc}
 Content & Style & SANet & SANet + SCSA & StyTR$^2$ & StyTR$^2$ + SCSA & StyleID & StyleID + SCSA \\
\includegraphics[width=0.14\linewidth]{sm/img/15+sem.jpg} & \includegraphics[width=0.14\linewidth]{sm/img/15_paint+sem.jpg}  & \includegraphics[width=0.14\linewidth]{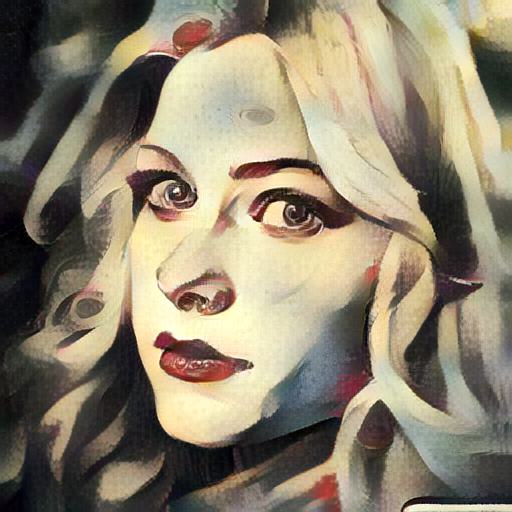}  &
\includegraphics[width=0.14\linewidth]{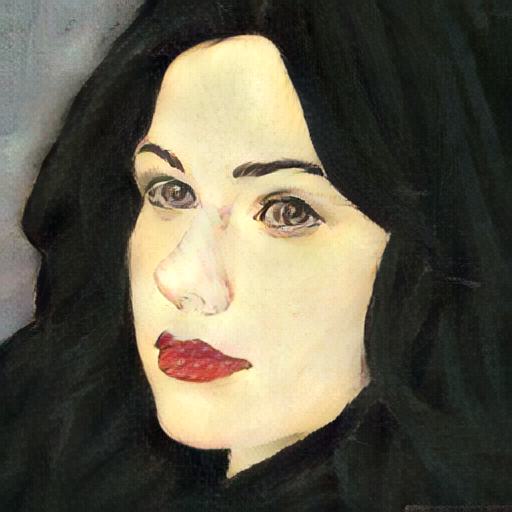}& \includegraphics[width=0.14\linewidth]{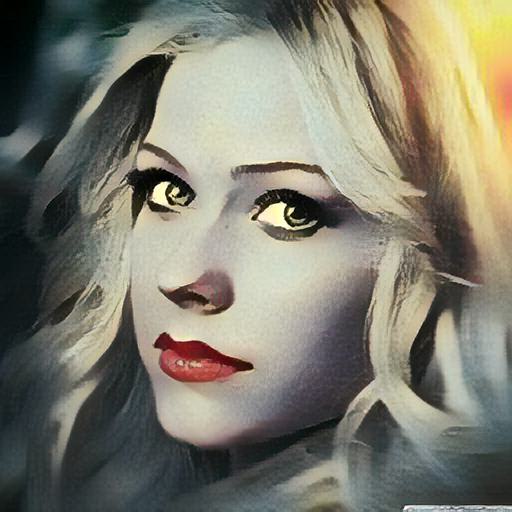} &
\includegraphics[width=0.14\linewidth]{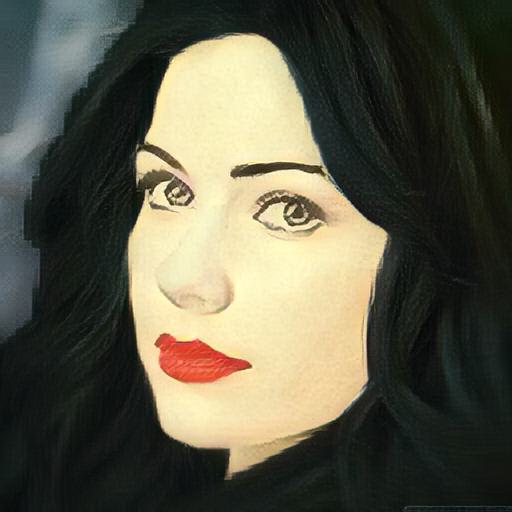} &  \includegraphics[width=0.14\linewidth]{sm/img/15_15_paint_StyleID.jpg} &  \includegraphics[width=0.14\linewidth]{sm/img/15_15_paint_StyleID_sem.jpg} \\
& & & STROTSS & MAST & TR & DIA & GLStyleNet \\
& & & \includegraphics[width=0.14\linewidth]{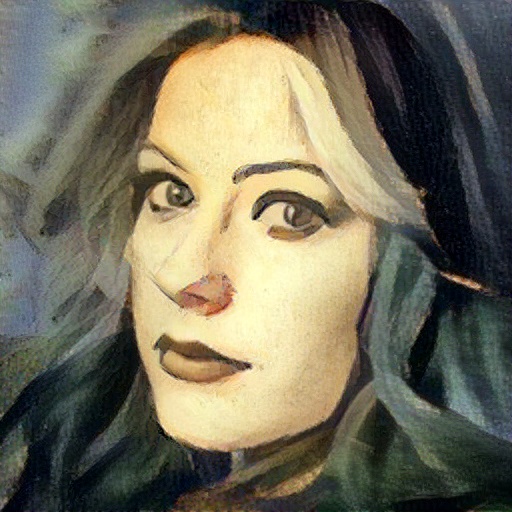} & \includegraphics[width=0.14\linewidth]{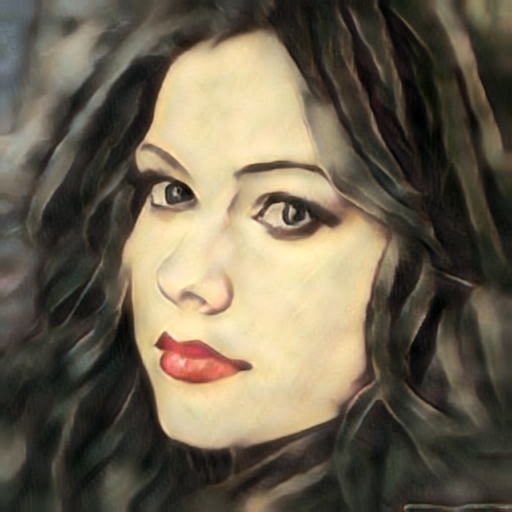} & 
\includegraphics[width=0.14\linewidth]{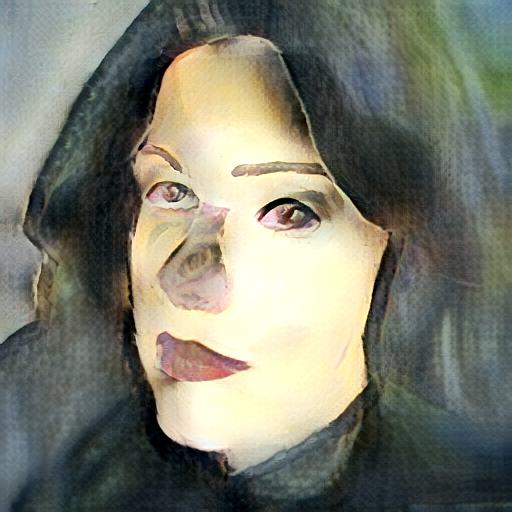} &
\includegraphics[width=0.14\linewidth]{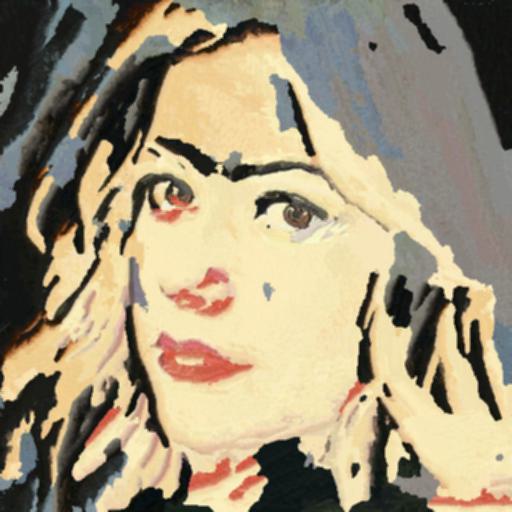} & 
\includegraphics[width=0.14\linewidth]{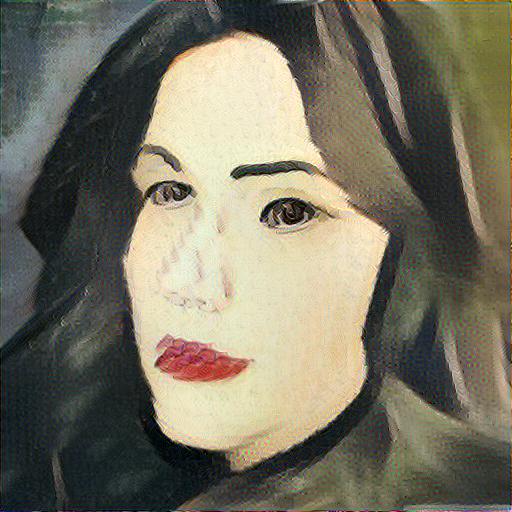} \\
& 
\\
 Content & Style & SANet & SANet + SCSA & StyTR$^2$ & StyTR$^2$ + SCSA & StyleID & StyleID + SCSA \\

\includegraphics[width=0.14\linewidth]{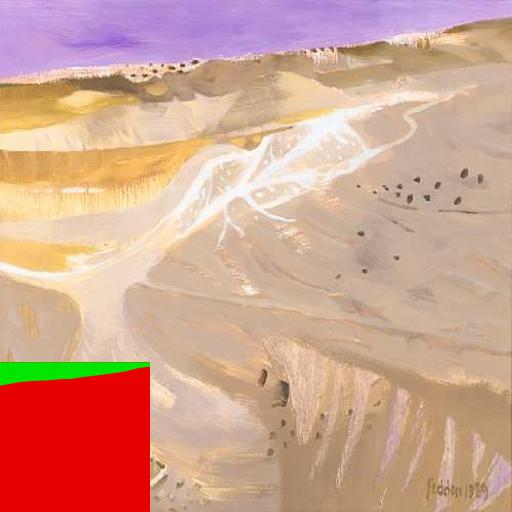} & \includegraphics[width=0.14\linewidth]{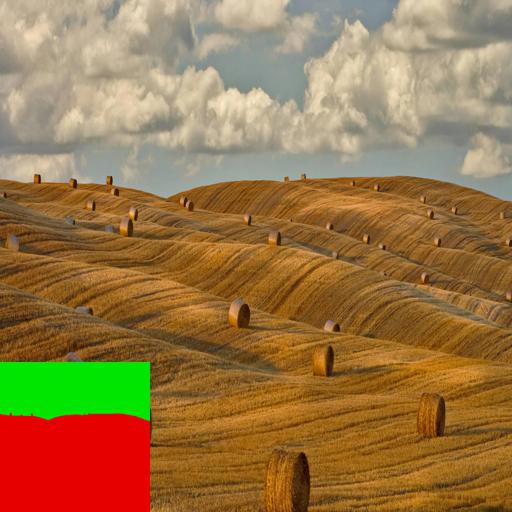}  & \includegraphics[width=0.14\linewidth]{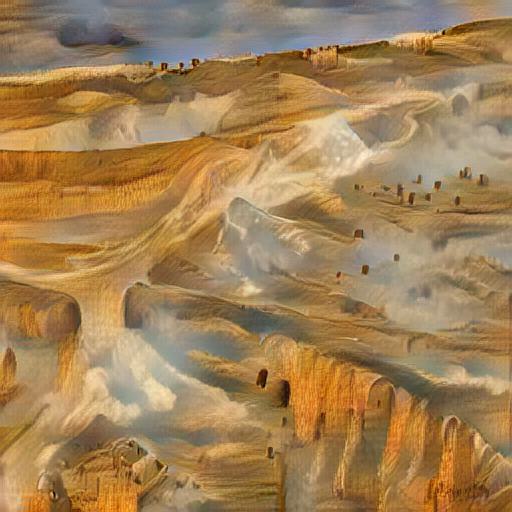}  &
\includegraphics[width=0.14\linewidth]{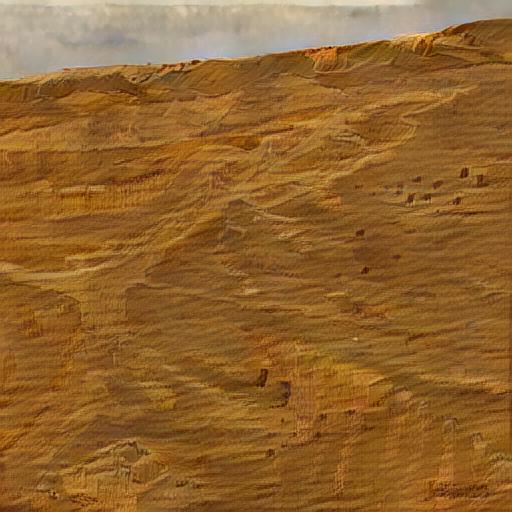}& \includegraphics[width=0.14\linewidth]{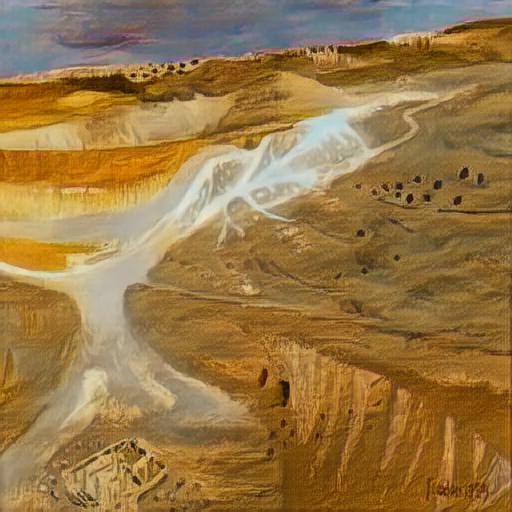} &
\includegraphics[width=0.14\linewidth]{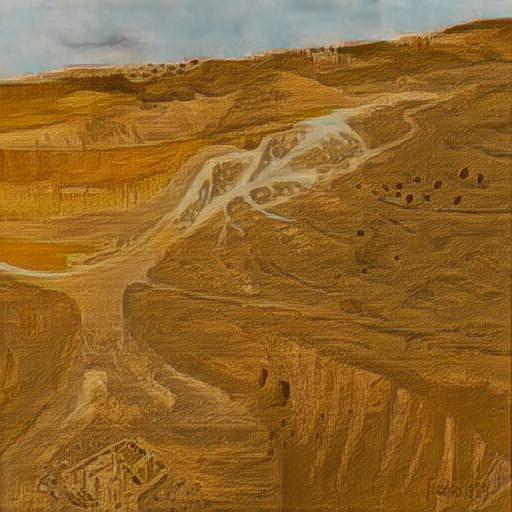} &  \includegraphics[width=0.14\linewidth]{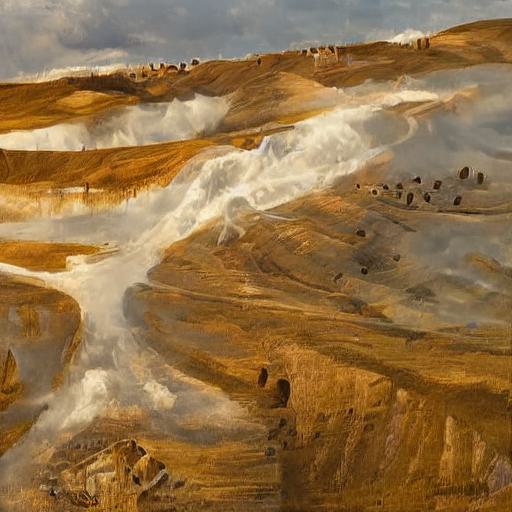} &  \includegraphics[width=0.14\linewidth]{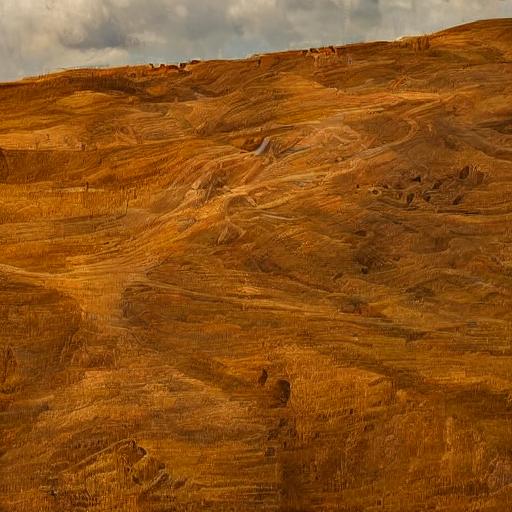} \\
& & & STROTSS & MAST & TR & DIA & GLStyleNet \\
& & & \includegraphics[width=0.14\linewidth]{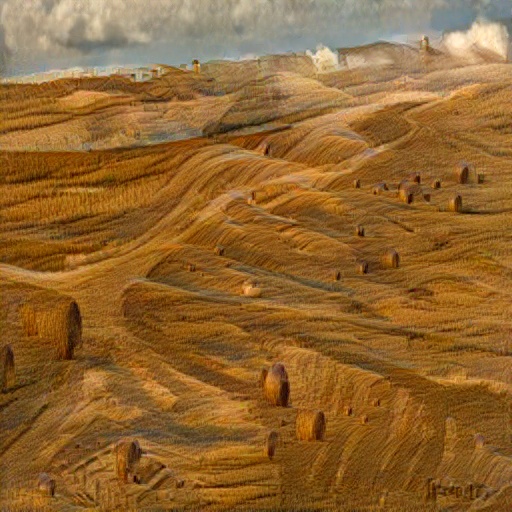} & \includegraphics[width=0.14\linewidth]{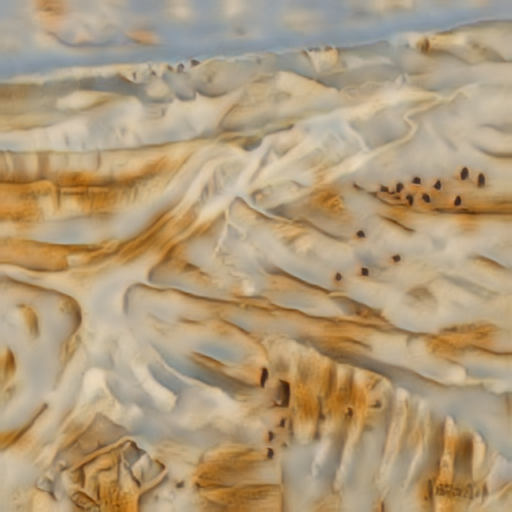} & 
\includegraphics[width=0.14\linewidth]{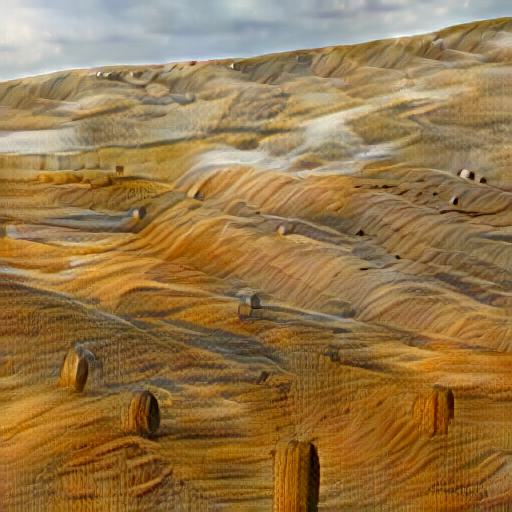} &
\includegraphics[width=0.14\linewidth]{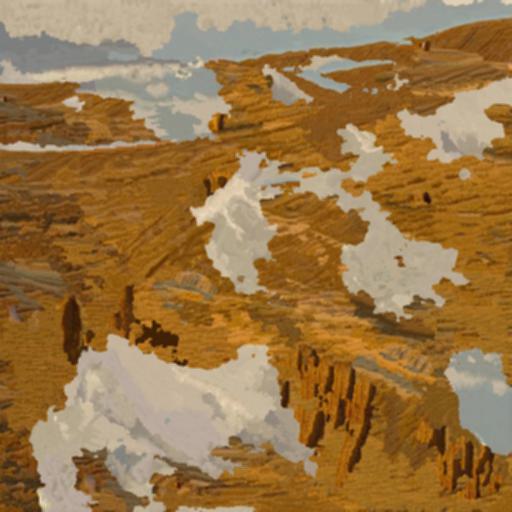} & 
\includegraphics[width=0.14\linewidth]{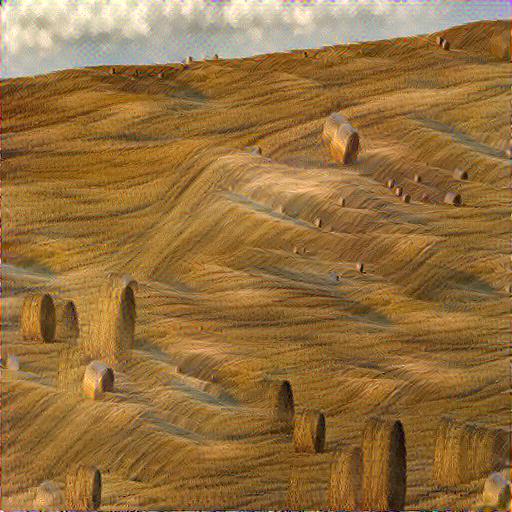} \\
&
\\
 Content & Style & SANet & SANet + SCSA & StyTR$^2$ & StyTR$^2$ + SCSA & StyleID & StyleID + SCSA \\
\includegraphics[width=0.14\linewidth]{sm/img/12+sem.jpg} & \includegraphics[width=0.14\linewidth]{sm/img/12_paint+sem.jpg}  & \includegraphics[width=0.14\linewidth]{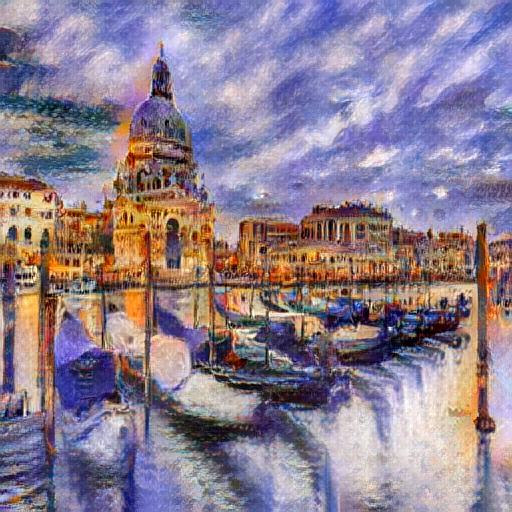}  &
\includegraphics[width=0.14\linewidth]{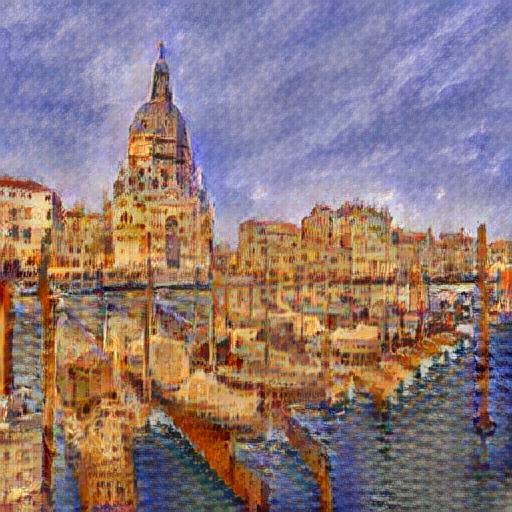}& \includegraphics[width=0.14\linewidth]{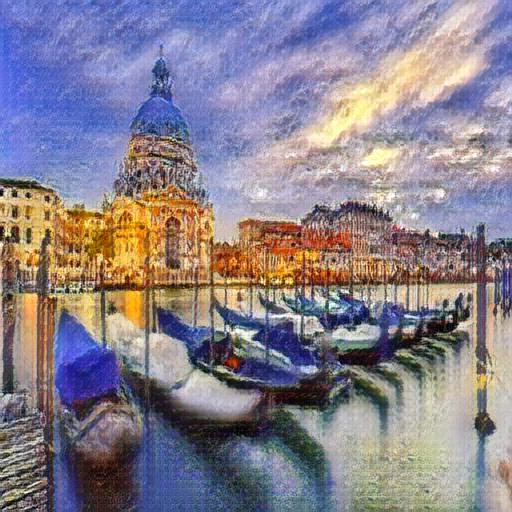} &
\includegraphics[width=0.14\linewidth]{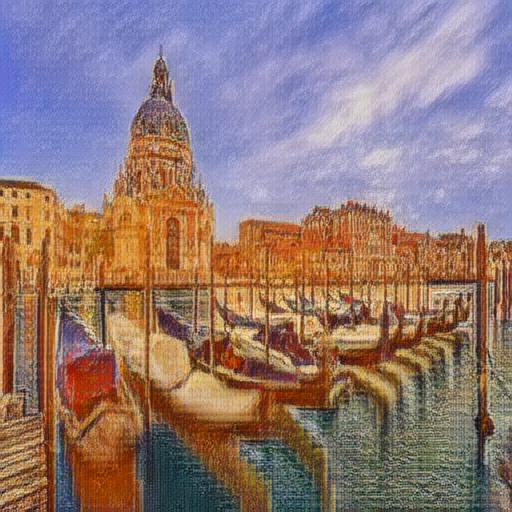} &  \includegraphics[width=0.14\linewidth]{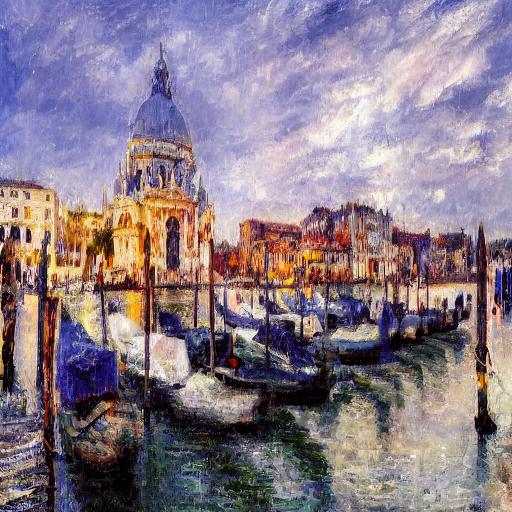} &  \includegraphics[width=0.14\linewidth]{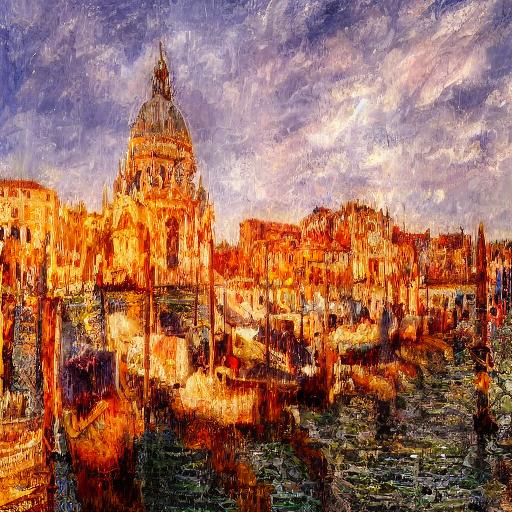} \\
& & & STROTSS & MAST & TR & DIA & GLStyleNet \\
& & & \includegraphics[width=0.14\linewidth]{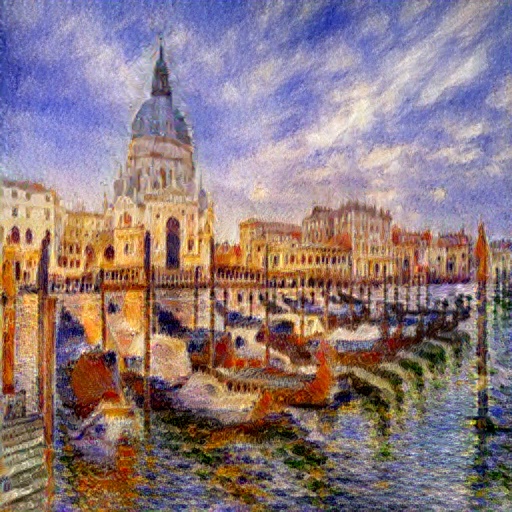} & \includegraphics[width=0.14\linewidth]{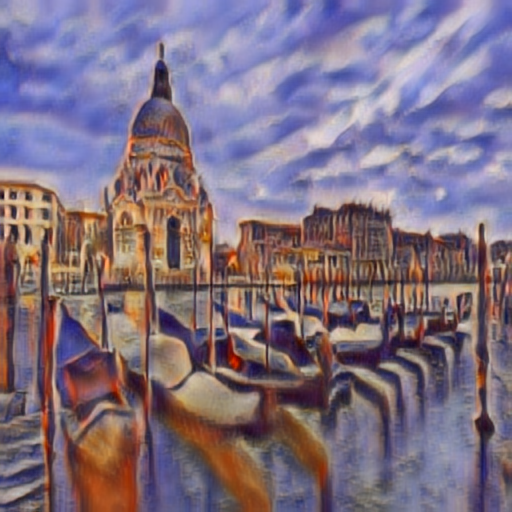} & 
\includegraphics[width=0.14\linewidth]{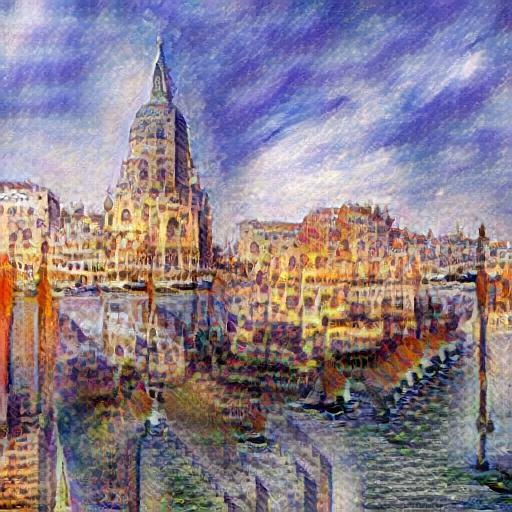} &
\includegraphics[width=0.14\linewidth]{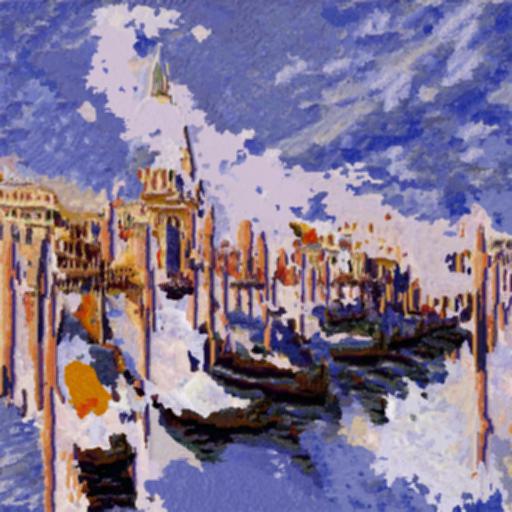} & 
\includegraphics[width=0.14\linewidth]{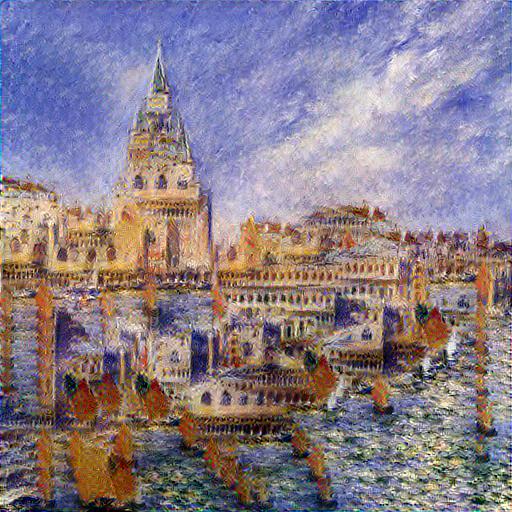} \\

&
\\
 Content & Style & SANet & SANet + SCSA & StyTR$^2$ & StyTR$^2$ + SCSA & StyleID & StyleID + SCSA \\

\includegraphics[width=0.14\linewidth]{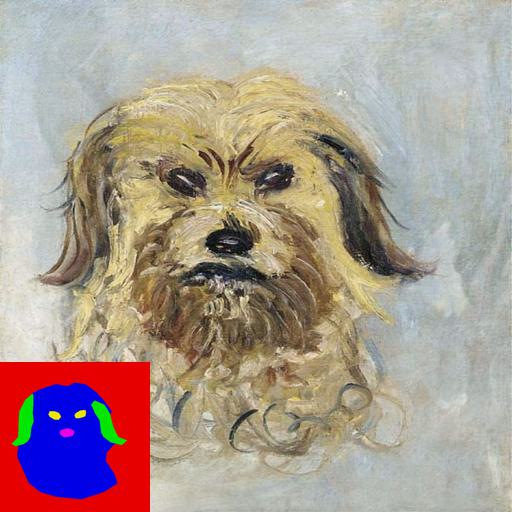} & \includegraphics[width=0.14\linewidth]{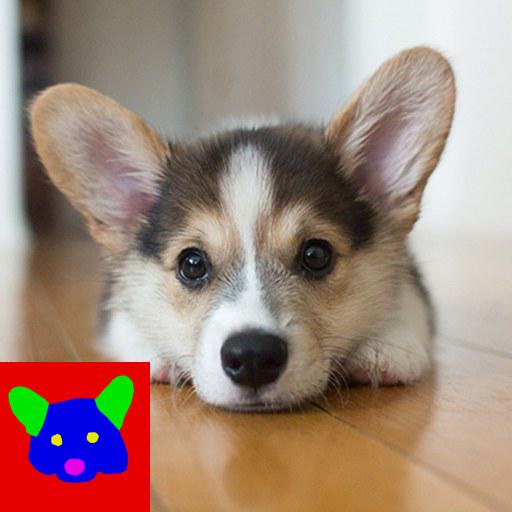}  & \includegraphics[width=0.14\linewidth]{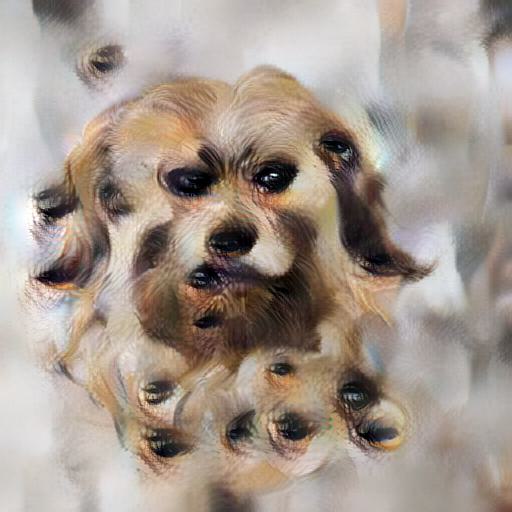}  &
\includegraphics[width=0.14\linewidth]{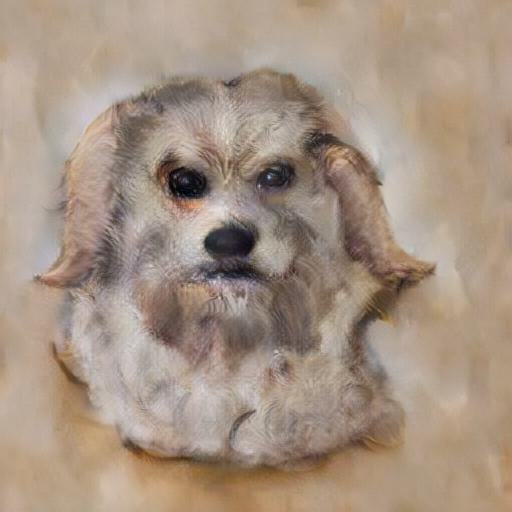}& \includegraphics[width=0.14\linewidth]{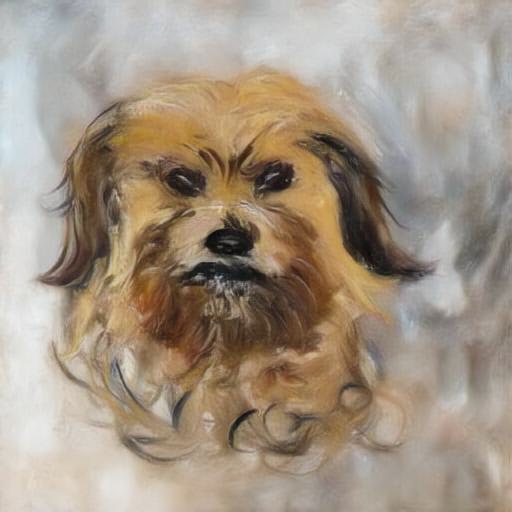} &
\includegraphics[width=0.14\linewidth]{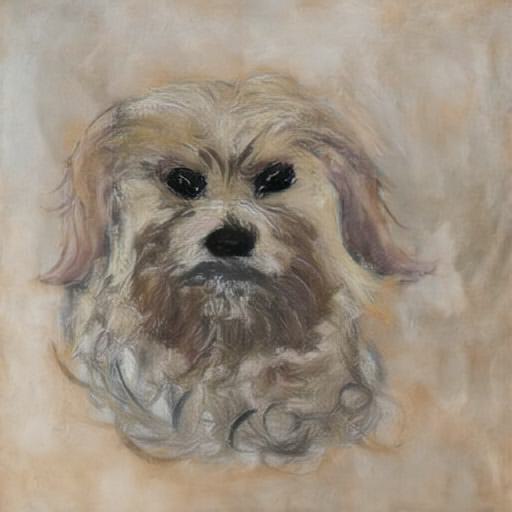} &  \includegraphics[width=0.14\linewidth]{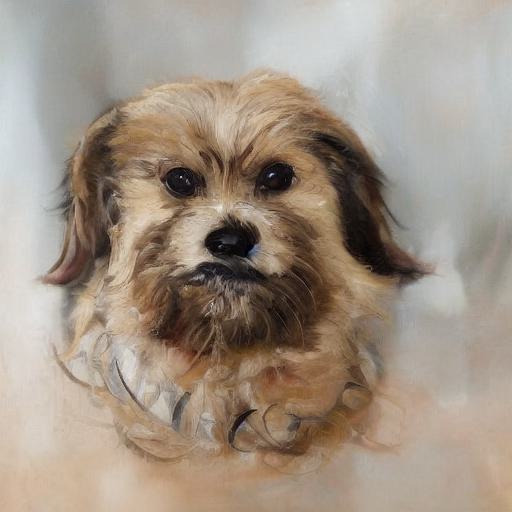} &  \includegraphics[width=0.14\linewidth]{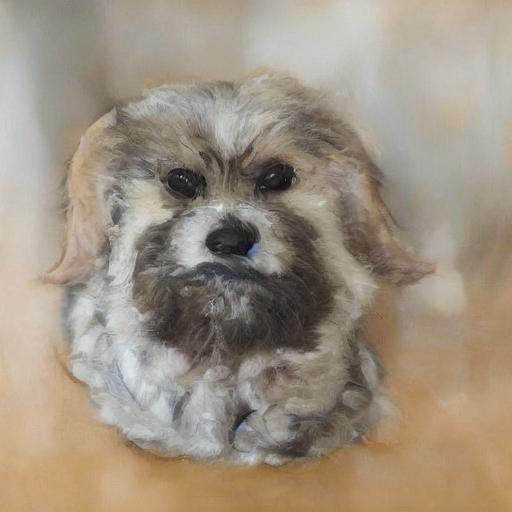} \\
& & & STROTSS & MAST & TR & DIA & GLStyleNet \\
& & & \includegraphics[width=0.14\linewidth]{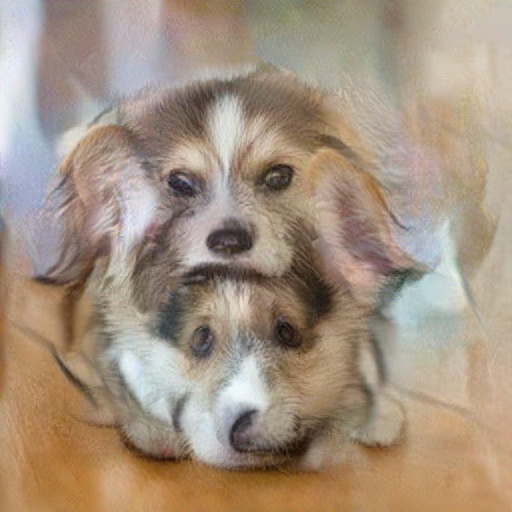} & \includegraphics[width=0.14\linewidth]{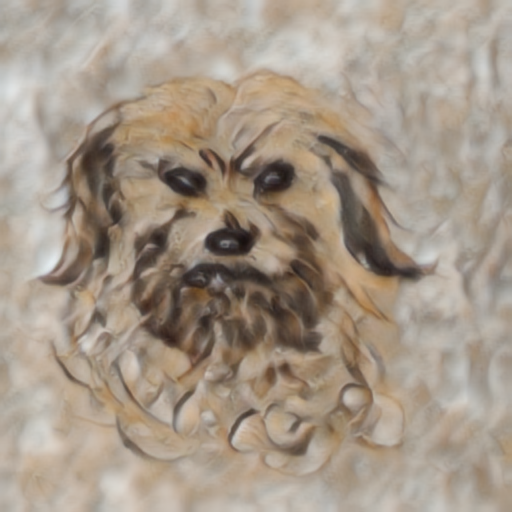} & 
\includegraphics[width=0.14\linewidth]{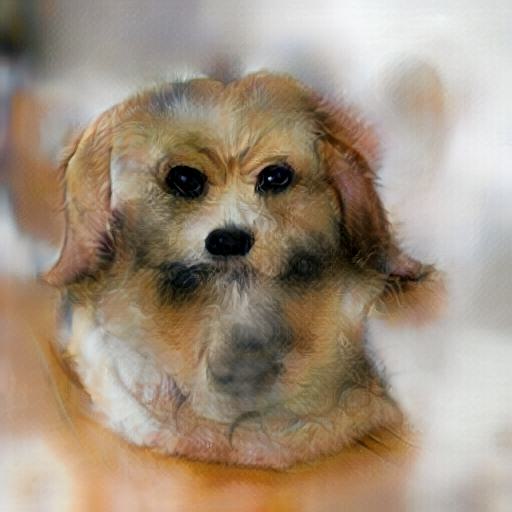} &
\includegraphics[width=0.14\linewidth]{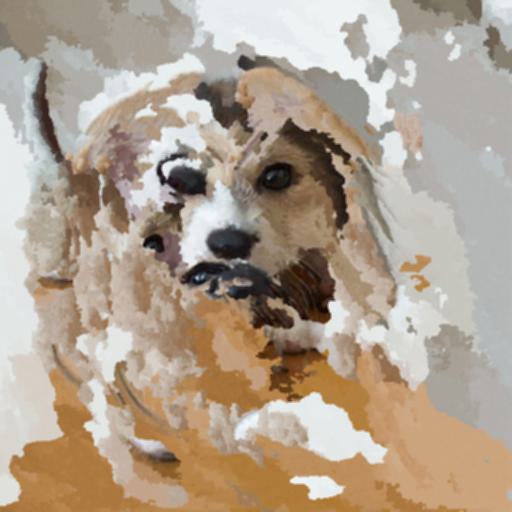} & 
\includegraphics[width=0.14\linewidth]{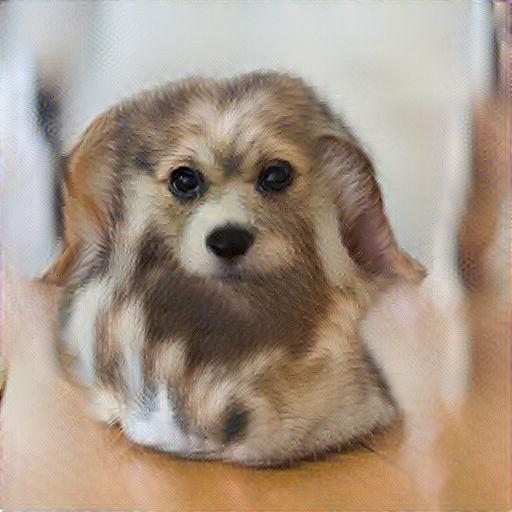} \\

\end{tabular}
}
\caption{Qualitative comparisons among Attn-AST approaches, those with SCSA, and SOTA methods.}
\label{fig:14}
\end{figure*}

\begin{figure*}
\centering
\resizebox{1.0\textwidth}{!}{
\setlength{\tabcolsep}{0.02cm} 
\renewcommand{\arraystretch}{1}  
\begin{tabular}{cccccccc}
 Content & Style & SANet & SANet + SCSA & StyTR$^2$ & StyTR$^2$ + SCSA & StyleID & StyleID + SCSA \\
\includegraphics[width=0.14\linewidth]{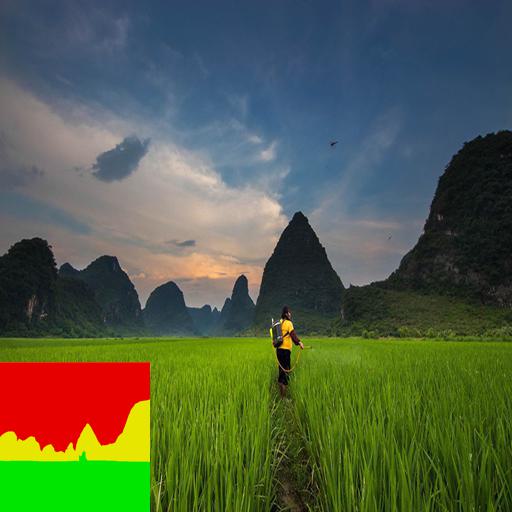} & \includegraphics[width=0.14\linewidth]{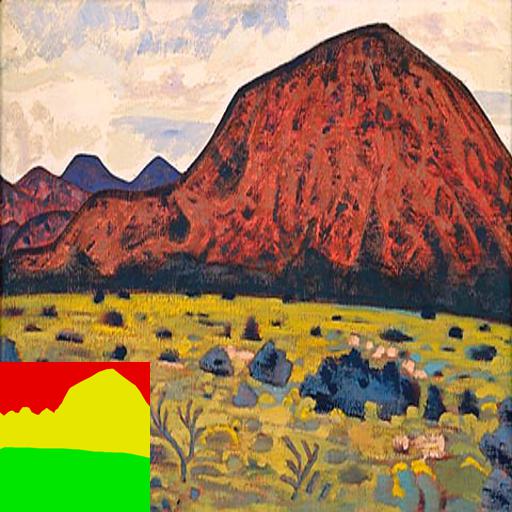}  & \includegraphics[width=0.14\linewidth]{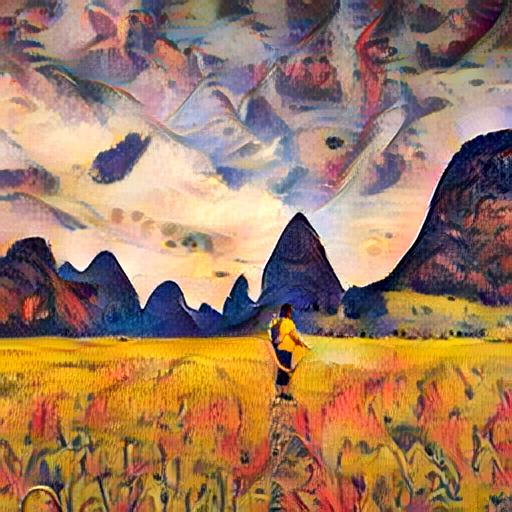}  &
\includegraphics[width=0.14\linewidth]{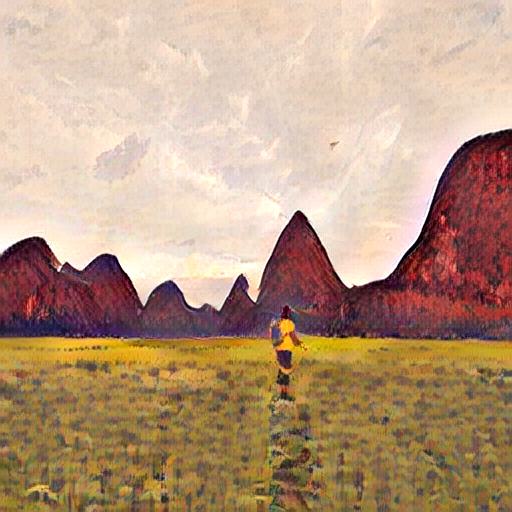}& \includegraphics[width=0.14\linewidth]{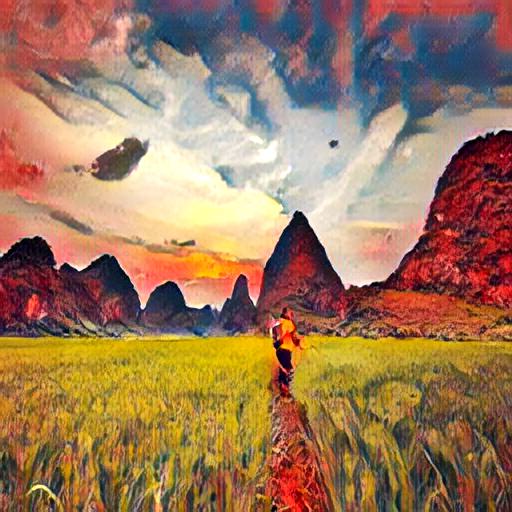} &
\includegraphics[width=0.14\linewidth]{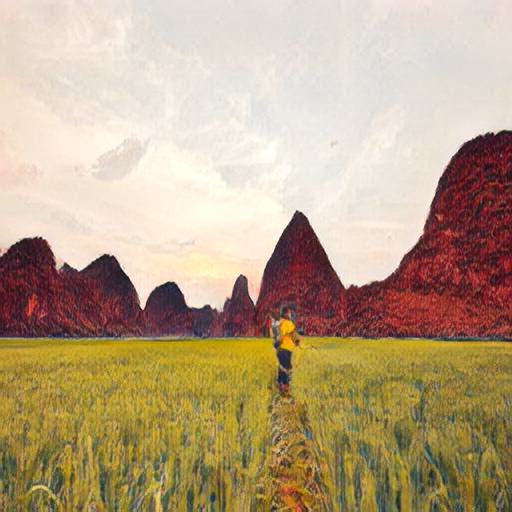} &  \includegraphics[width=0.14\linewidth]{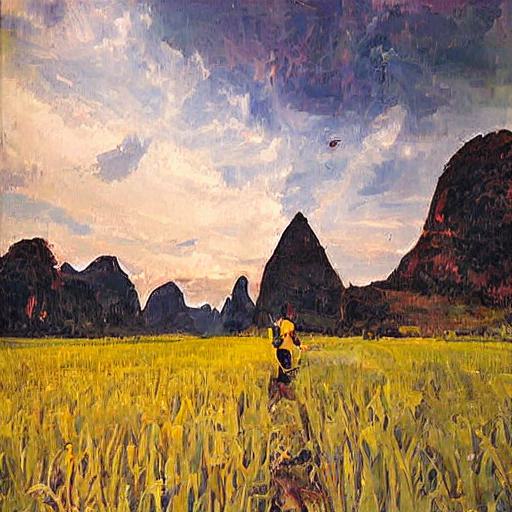} &  \includegraphics[width=0.14\linewidth]{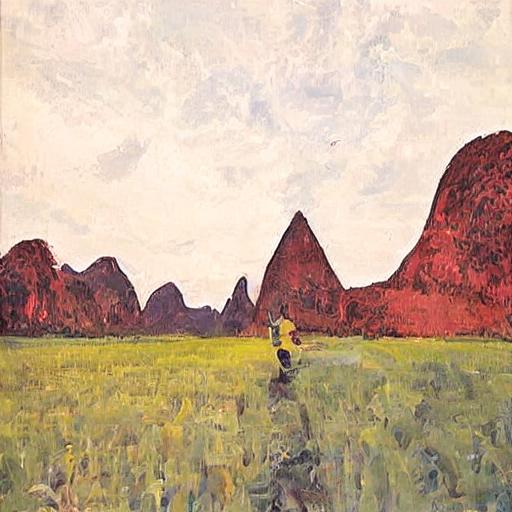} \\
& & & STROTSS & MAST & TR & DIA & GLStyleNet \\
& & & \includegraphics[width=0.14\linewidth]{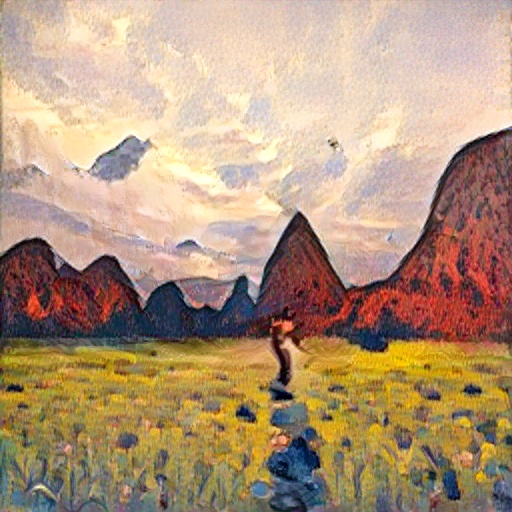} & \includegraphics[width=0.14\linewidth]{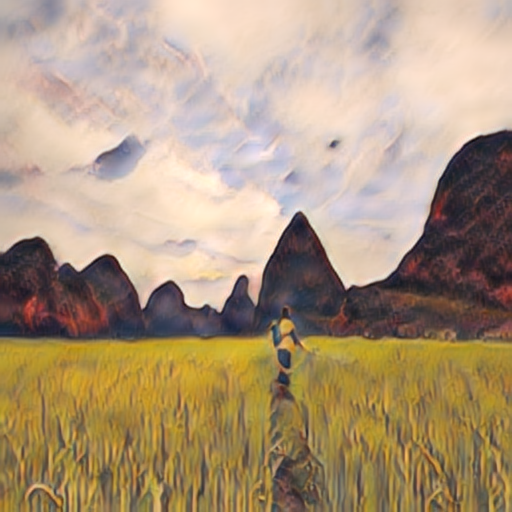} & 
\includegraphics[width=0.14\linewidth]{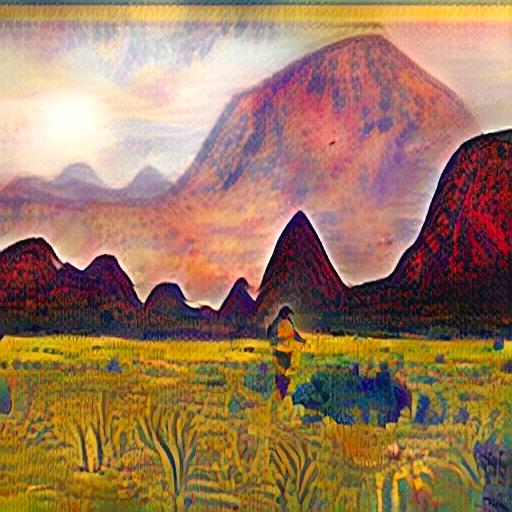} &
\includegraphics[width=0.14\linewidth]{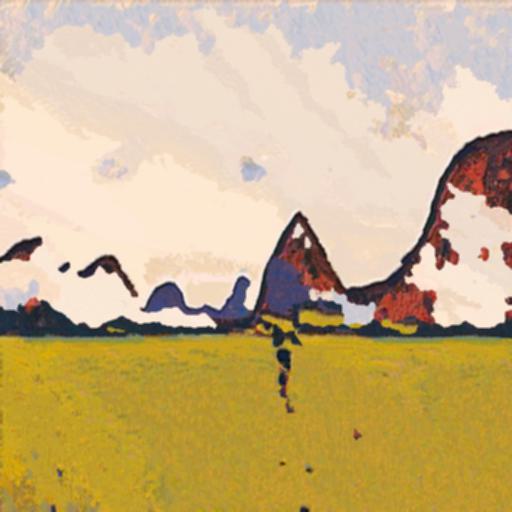} & 
\includegraphics[width=0.14\linewidth]{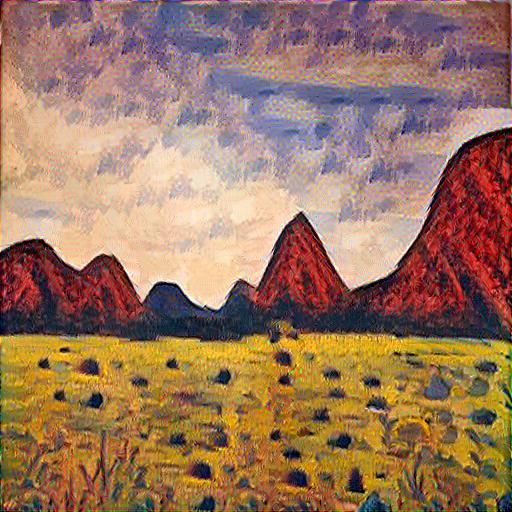} \\
& 
\\
 Content & Style & SANet & SANet + SCSA & StyTR$^2$ & StyTR$^2$ + SCSA & StyleID & StyleID + SCSA \\

\includegraphics[width=0.14\linewidth]{sm/img/12_paint+sem.jpg} & \includegraphics[width=0.14\linewidth]{sm/img/12+sem.jpg}  & \includegraphics[width=0.14\linewidth]{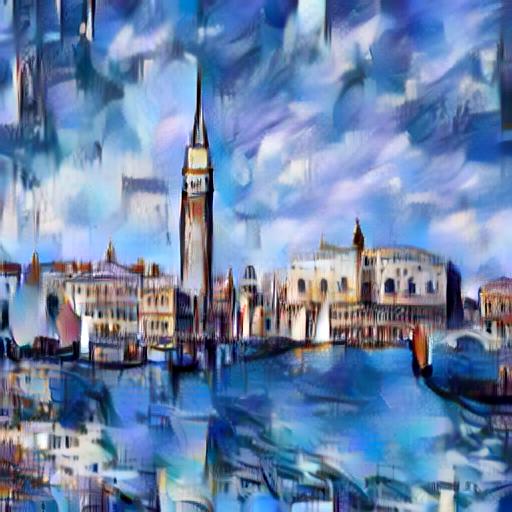}  &
\includegraphics[width=0.14\linewidth]{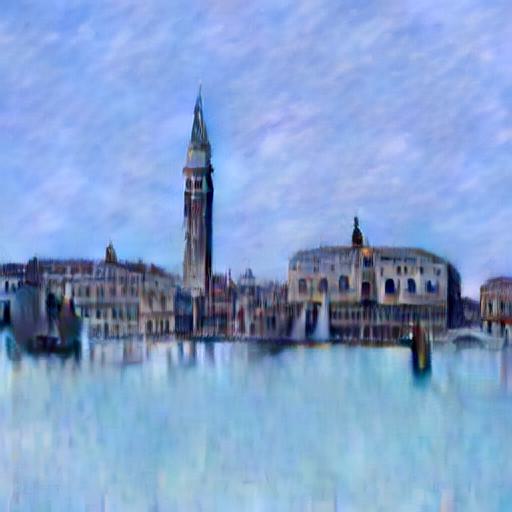}& \includegraphics[width=0.14\linewidth]{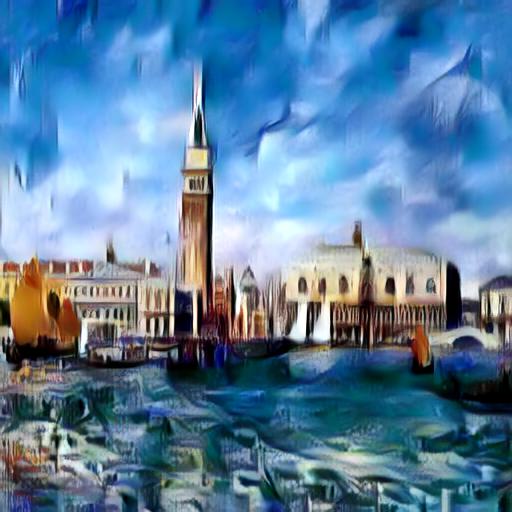} &
\includegraphics[width=0.14\linewidth]{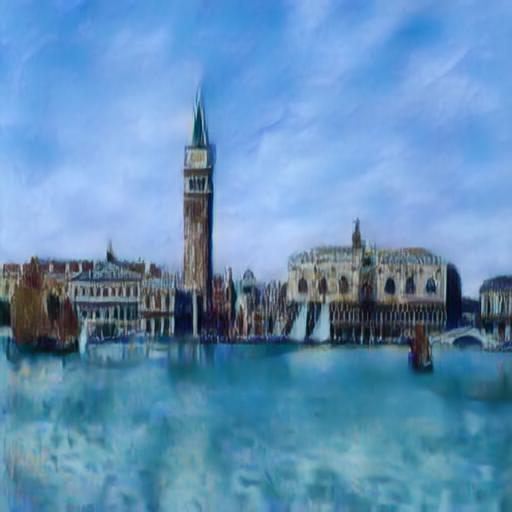} &  \includegraphics[width=0.14\linewidth]{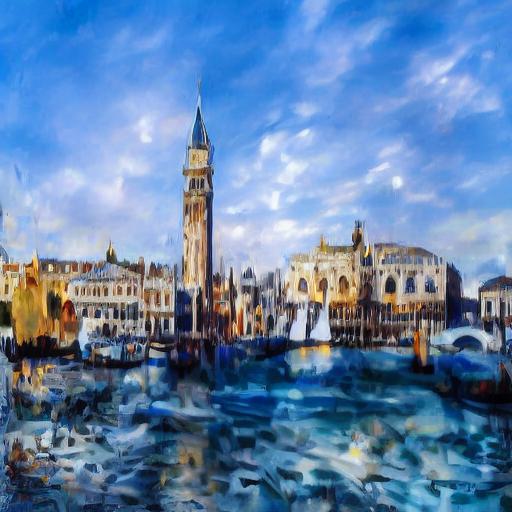} &  \includegraphics[width=0.14\linewidth]{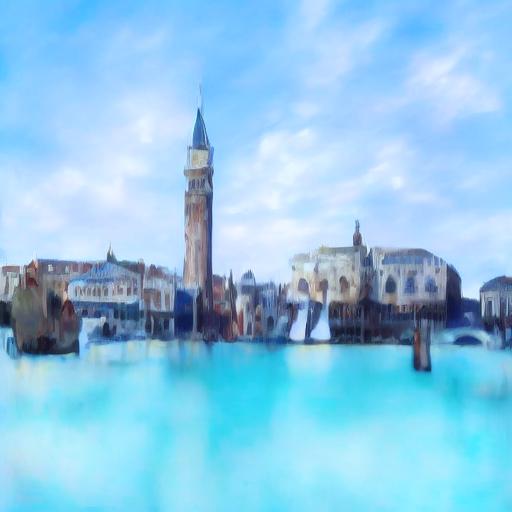} \\
& & & STROTSS & MAST & TR & DIA & GLStyleNet \\
& & & \includegraphics[width=0.14\linewidth]{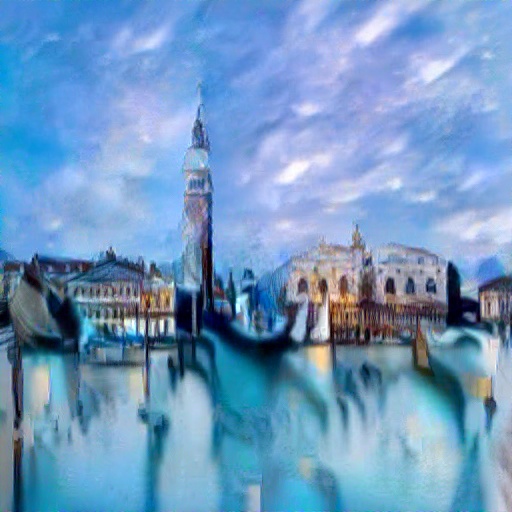} & \includegraphics[width=0.14\linewidth]{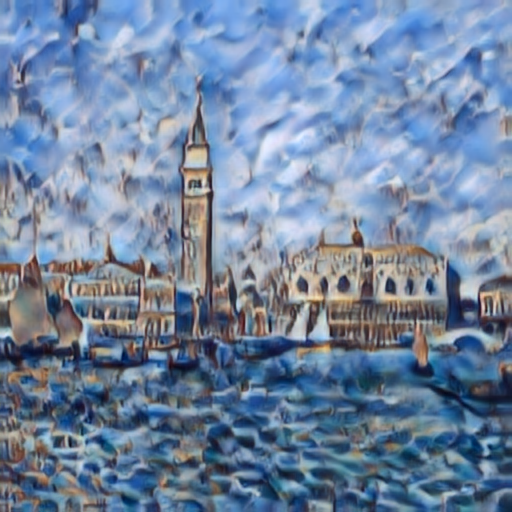} & 
\includegraphics[width=0.14\linewidth]{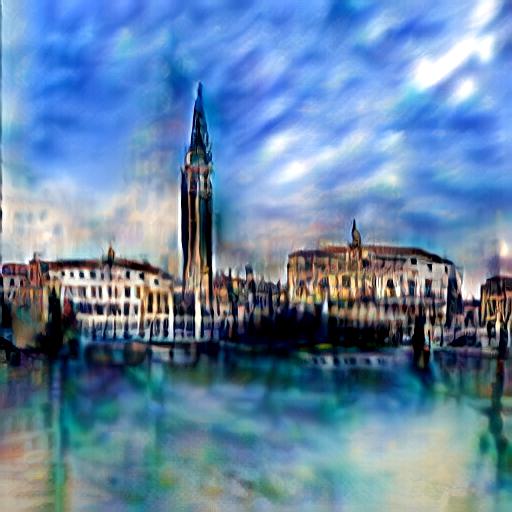} &
\includegraphics[width=0.14\linewidth]{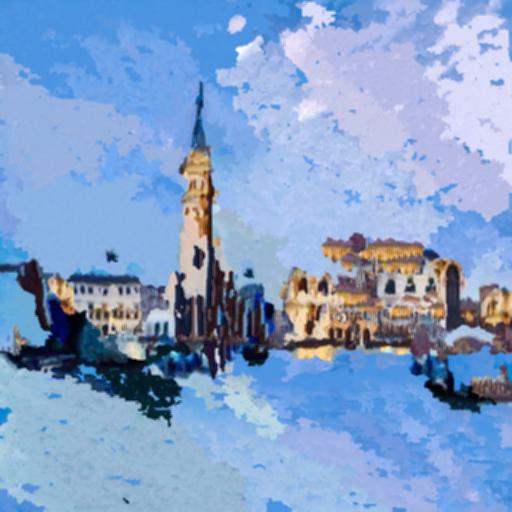} & 
\includegraphics[width=0.14\linewidth]{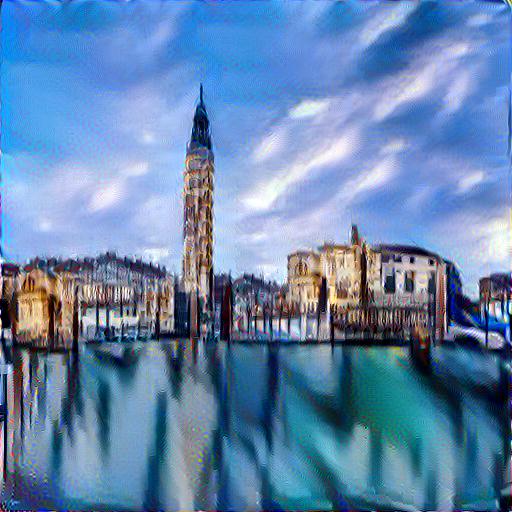} \\
&
\\
 Content & Style & SANet & SANet + SCSA & StyTR$^2$ & StyTR$^2$ + SCSA & StyleID & StyleID + SCSA \\
\includegraphics[width=0.14\linewidth]{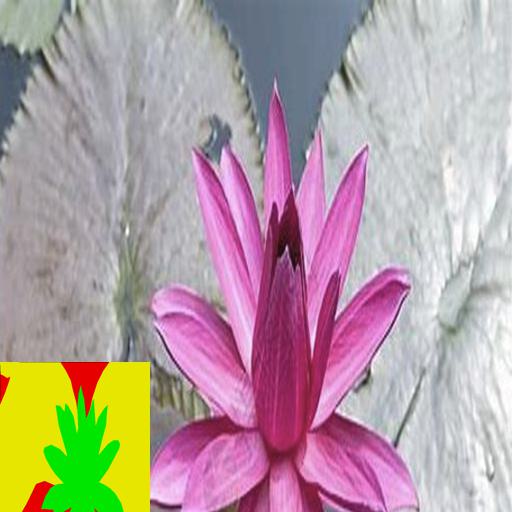} & \includegraphics[width=0.14\linewidth]{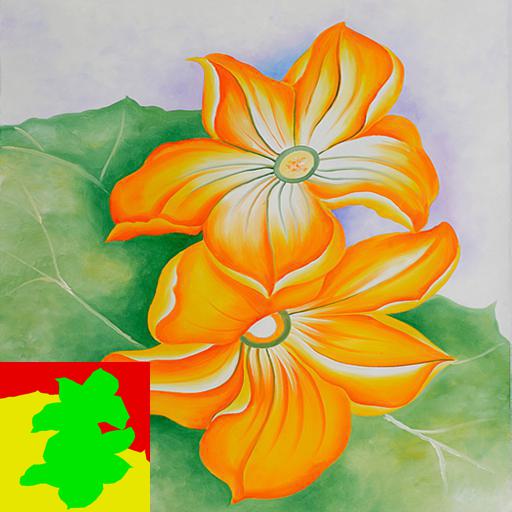}  & \includegraphics[width=0.14\linewidth]{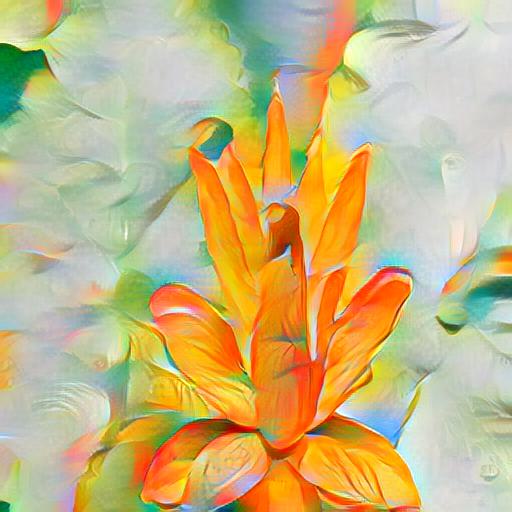}  &
\includegraphics[width=0.14\linewidth]{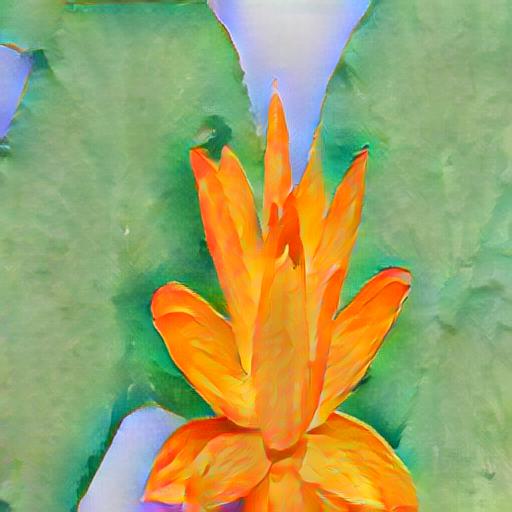}& \includegraphics[width=0.14\linewidth]{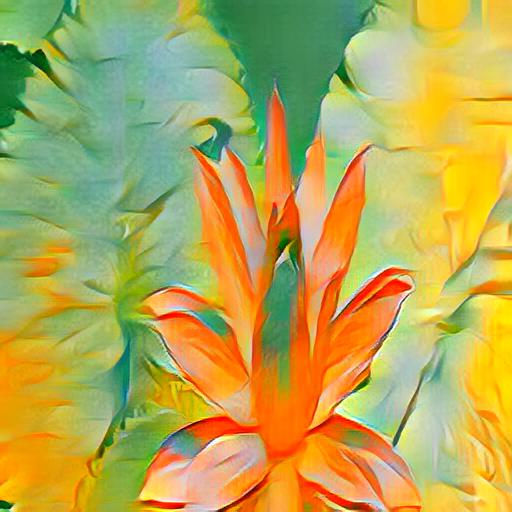} &
\includegraphics[width=0.14\linewidth]{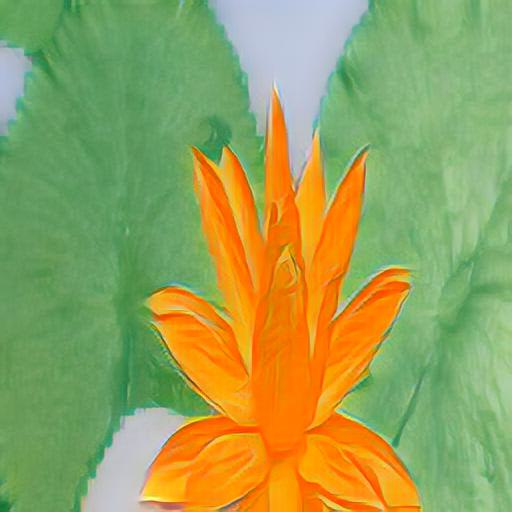} &  \includegraphics[width=0.14\linewidth]{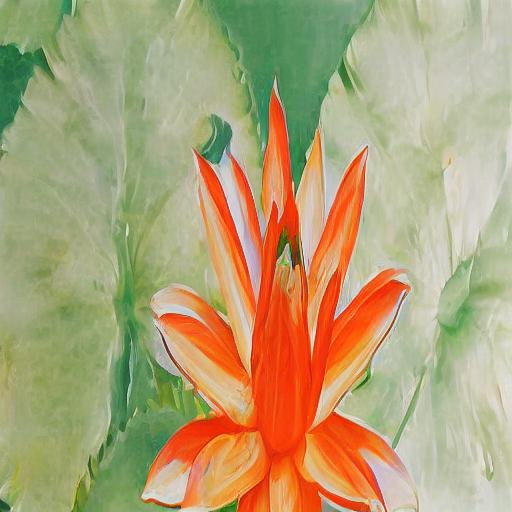} &  \includegraphics[width=0.14\linewidth]{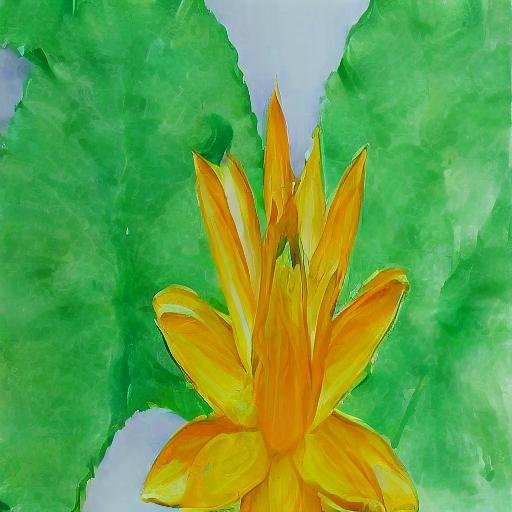} \\
& & & STROTSS & MAST & TR & DIA & GLStyleNet \\
& & & \includegraphics[width=0.14\linewidth]{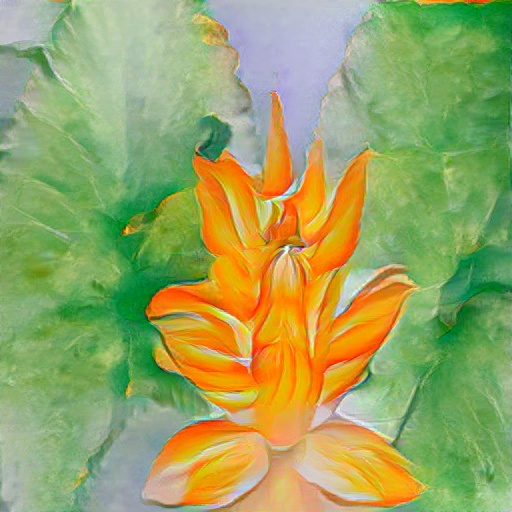} & \includegraphics[width=0.14\linewidth]{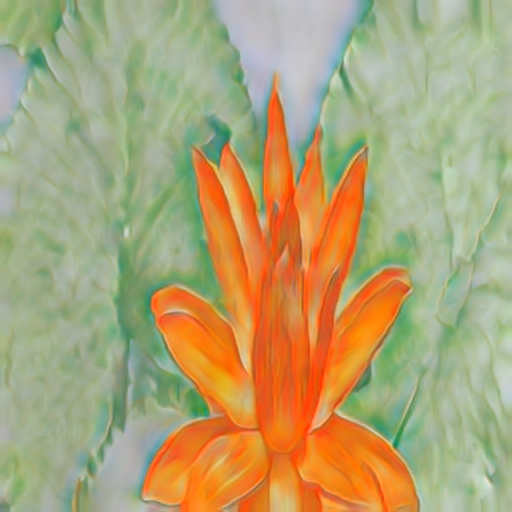} & 
\includegraphics[width=0.14\linewidth]{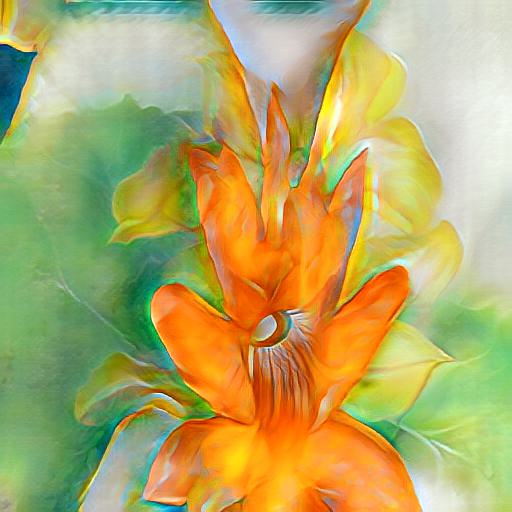} &
\includegraphics[width=0.14\linewidth]{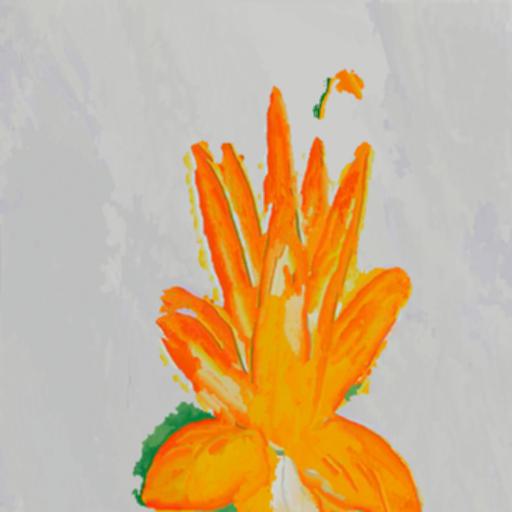} & 
\includegraphics[width=0.14\linewidth]{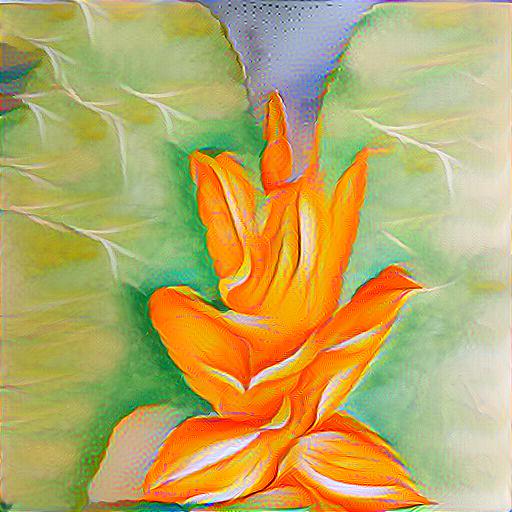} \\

&
\\
 Content & Style & SANet & SANet + SCSA & StyTR$^2$ & StyTR$^2$ + SCSA & StyleID & StyleID + SCSA \\

\includegraphics[width=0.14\linewidth]{sm/img/24_paint+sem.jpg} & \includegraphics[width=0.14\linewidth]{sm/img/24+sem.jpg}  & \includegraphics[width=0.14\linewidth]{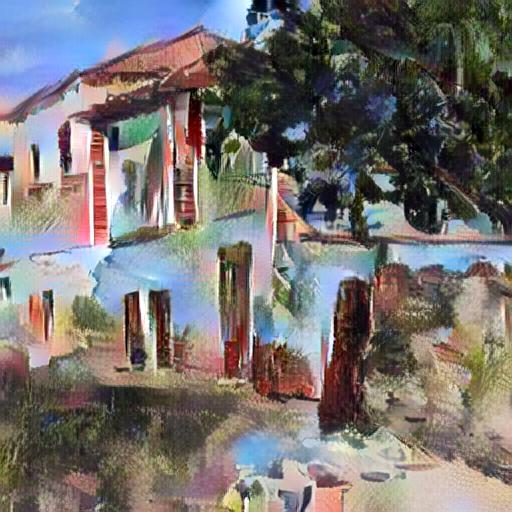}  &
\includegraphics[width=0.14\linewidth]{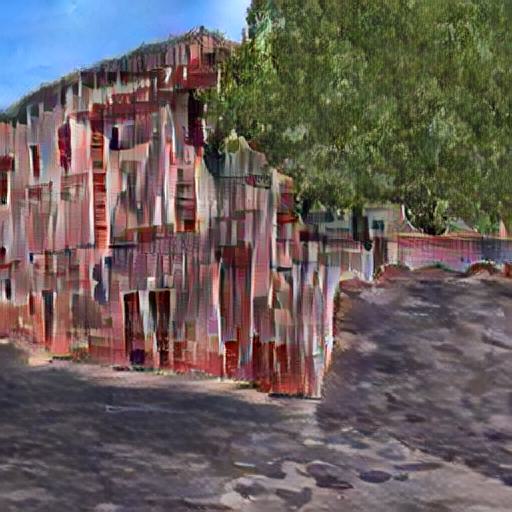}& \includegraphics[width=0.14\linewidth]{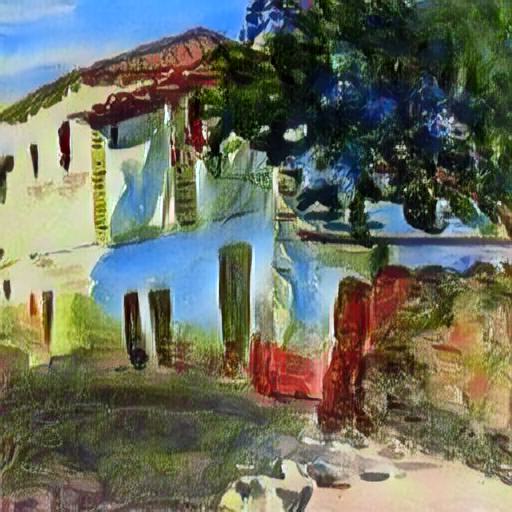} &
\includegraphics[width=0.14\linewidth]{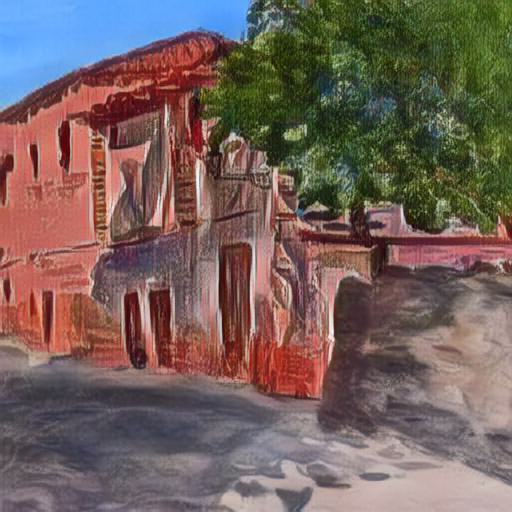} &  \includegraphics[width=0.14\linewidth]{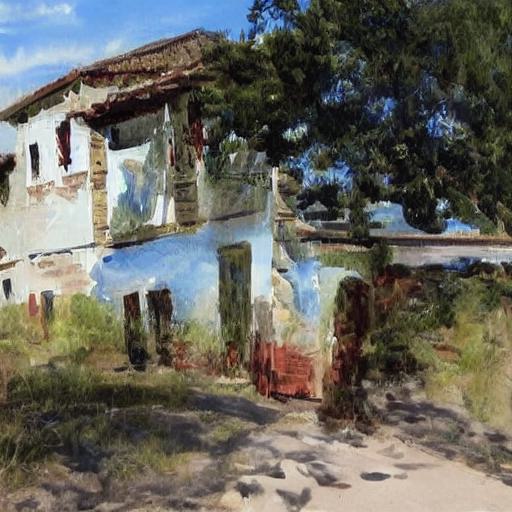} &  \includegraphics[width=0.14\linewidth]{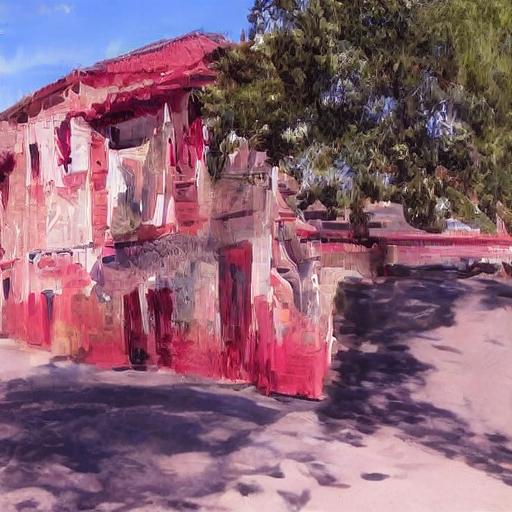} \\
& & & STROTSS & MAST & TR & DIA & GLStyleNet \\
& & & \includegraphics[width=0.14\linewidth]{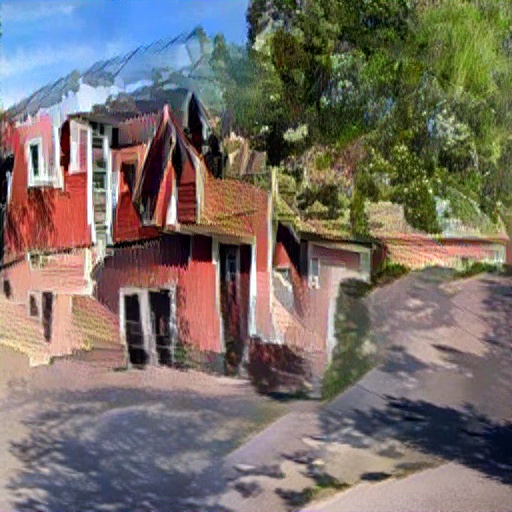} & \includegraphics[width=0.14\linewidth]{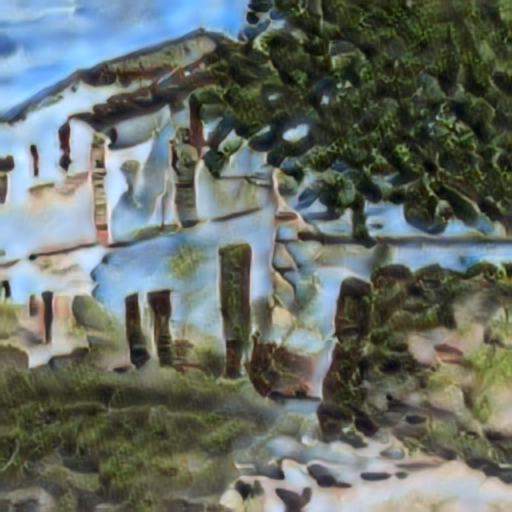} & 
\includegraphics[width=0.14\linewidth]{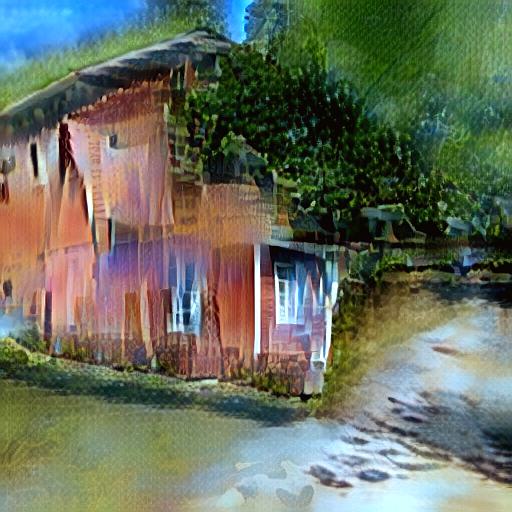} &
\includegraphics[width=0.14\linewidth]{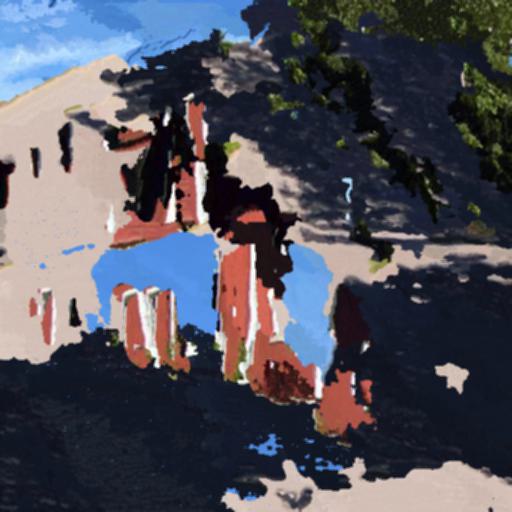} & 
\includegraphics[width=0.14\linewidth]{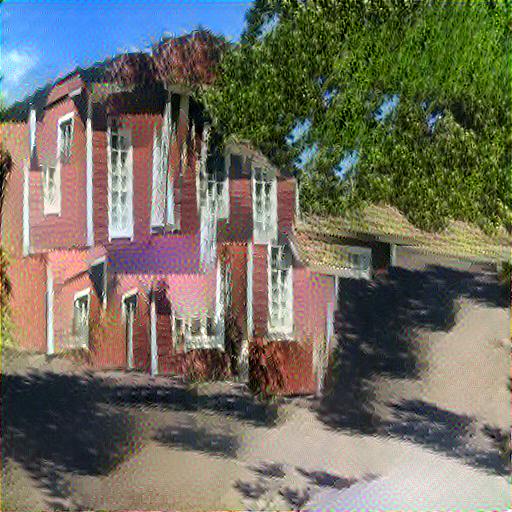} \\

\end{tabular}
}
\caption{Qualitative comparisons among Attn-AST approaches, those with SCSA, and SOTA methods.}
\label{fig:15}
\end{figure*}

\begin{figure*}
\centering
\resizebox{1.0\textwidth}{!}{
\setlength{\tabcolsep}{0.02cm} 
\renewcommand{\arraystretch}{1}  
\begin{tabular}{cccccccc}
 Content & Style & SANet & SANet + SCSA & StyTR$^2$ & StyTR$^2$ + SCSA & StyleID & StyleID + SCSA \\
\includegraphics[width=0.14\linewidth]{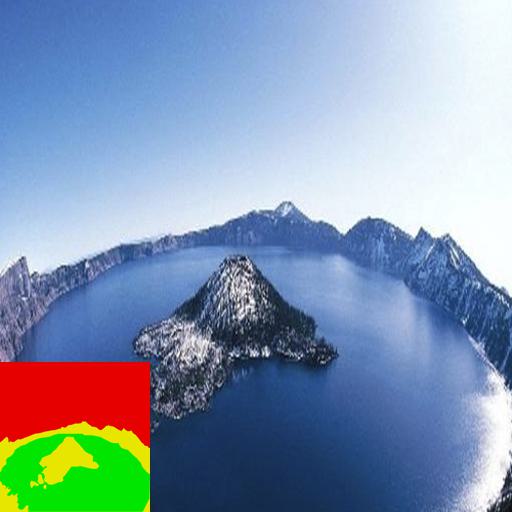} & \includegraphics[width=0.14\linewidth]{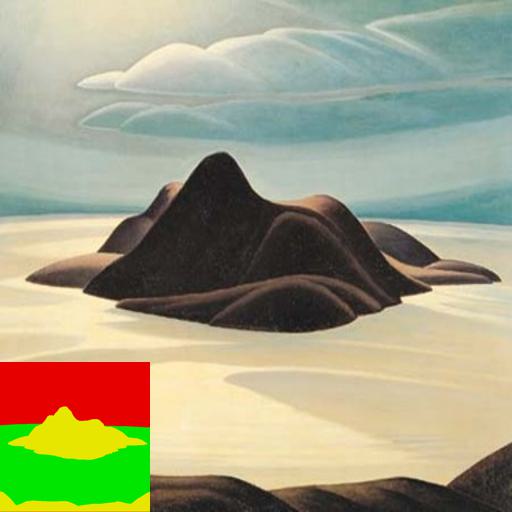} & \includegraphics[width=0.14\linewidth]{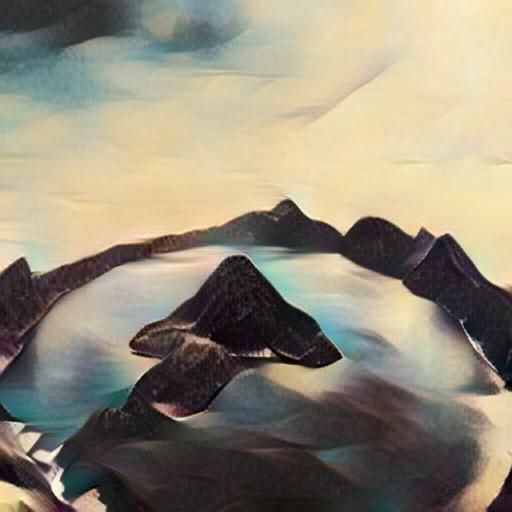}  &
\includegraphics[width=0.14\linewidth]{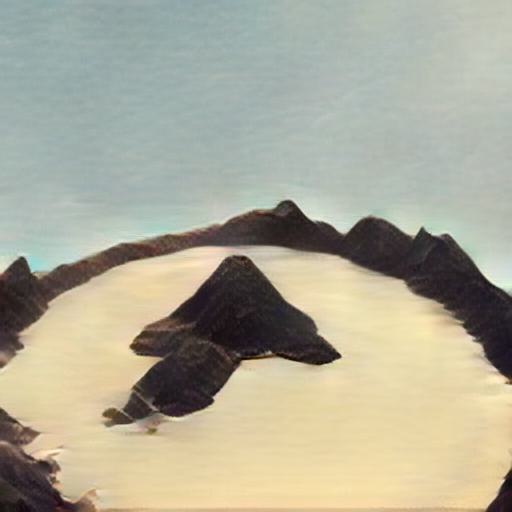}& \includegraphics[width=0.14\linewidth]{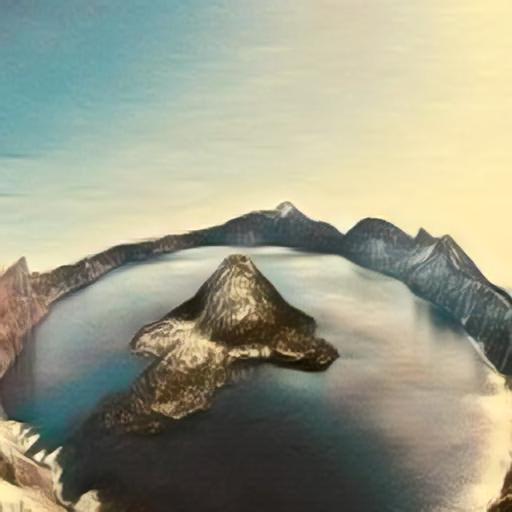} &
\includegraphics[width=0.14\linewidth]{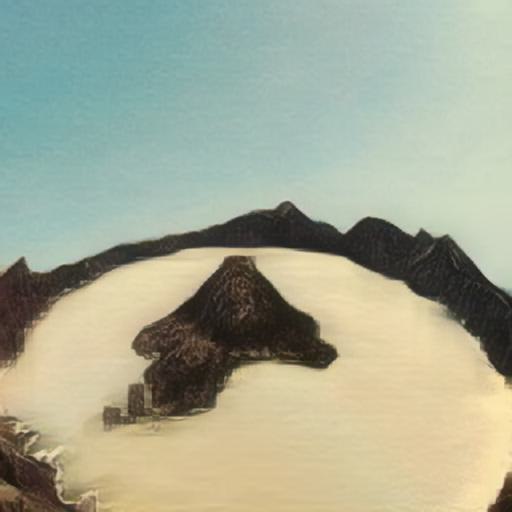} &  \includegraphics[width=0.14\linewidth]{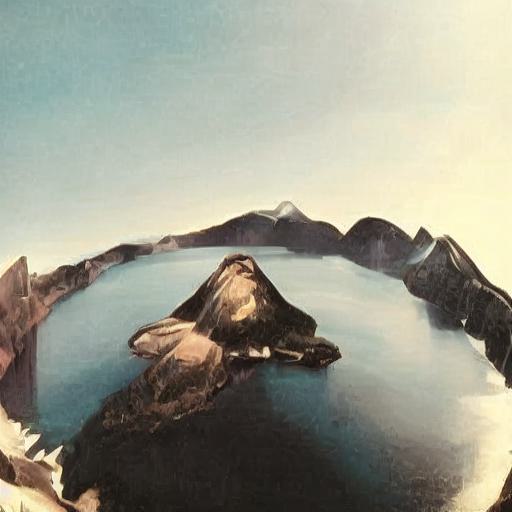} &  \includegraphics[width=0.14\linewidth]{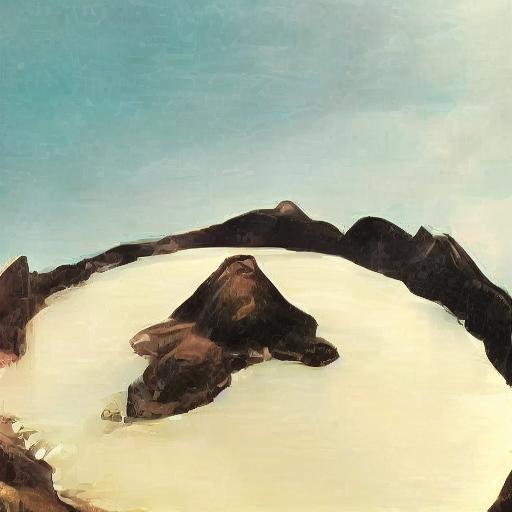} \\
& & & STROTSS & MAST & TR & DIA & GLStyleNet \\
& & & \includegraphics[width=0.14\linewidth]{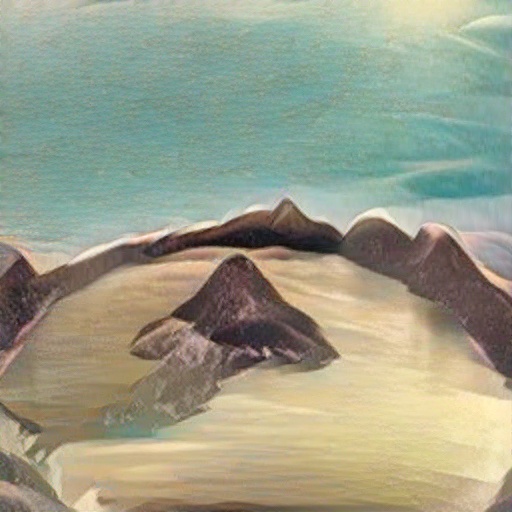} & \includegraphics[width=0.14\linewidth]{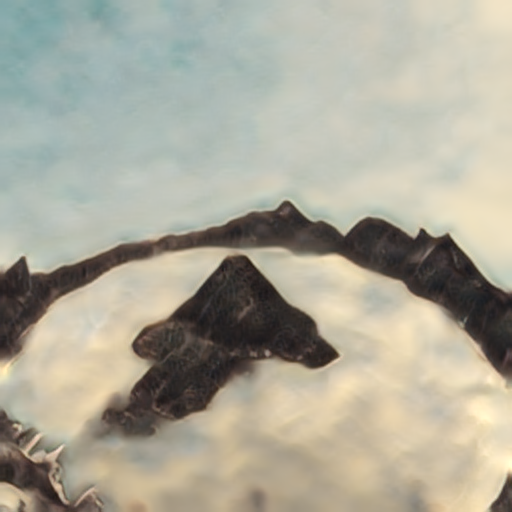} & 
\includegraphics[width=0.14\linewidth]{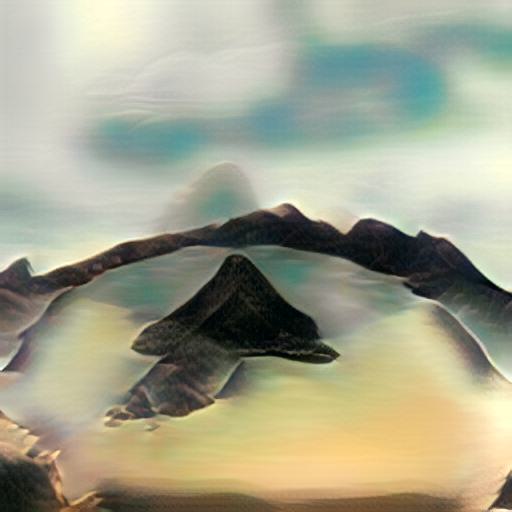} &
\includegraphics[width=0.14\linewidth]{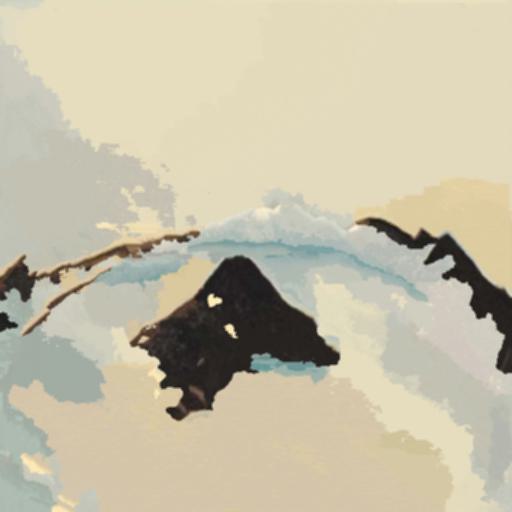} & 
\includegraphics[width=0.14\linewidth]{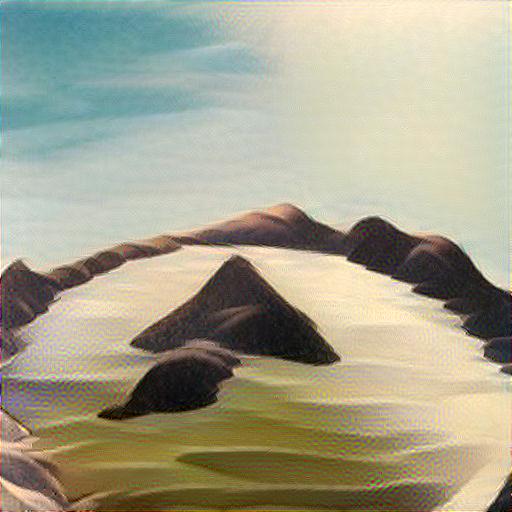} \\
& 
\\
 Content & Style & SANet & SANet + SCSA & StyTR$^2$ & StyTR$^2$ + SCSA & StyleID & StyleID + SCSA \\

\includegraphics[width=0.14\linewidth]{sm/img/35_paint+sem.jpg} & \includegraphics[width=0.14\linewidth]{sm/img/35+sem.jpg}  & \includegraphics[width=0.14\linewidth]{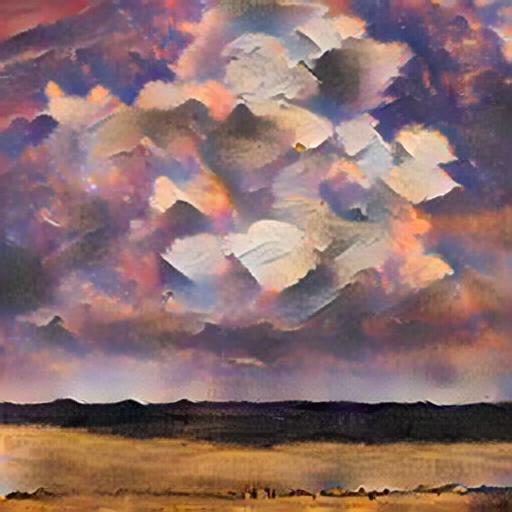}  &
\includegraphics[width=0.14\linewidth]{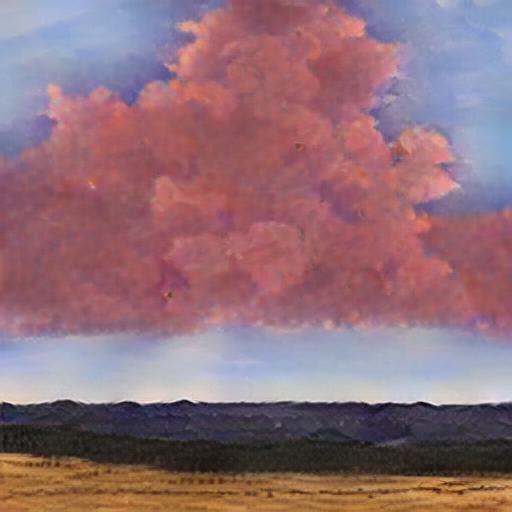}& \includegraphics[width=0.14\linewidth]{sm/img/35_paint_35_StyTR2.jpg} &
\includegraphics[width=0.14\linewidth]{sm/img/35_paint_35_StyTR2_sem.jpg} &  \includegraphics[width=0.14\linewidth]{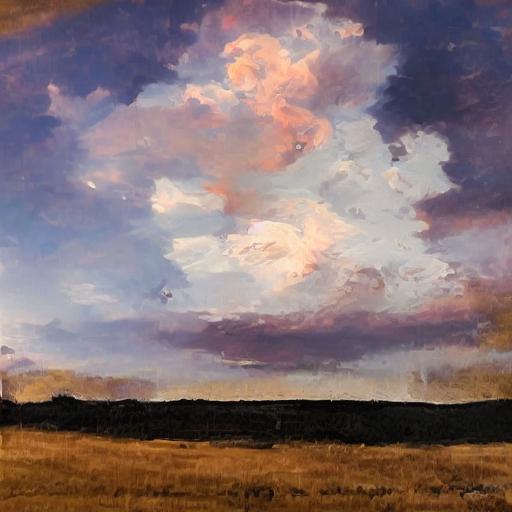} &  \includegraphics[width=0.14\linewidth]{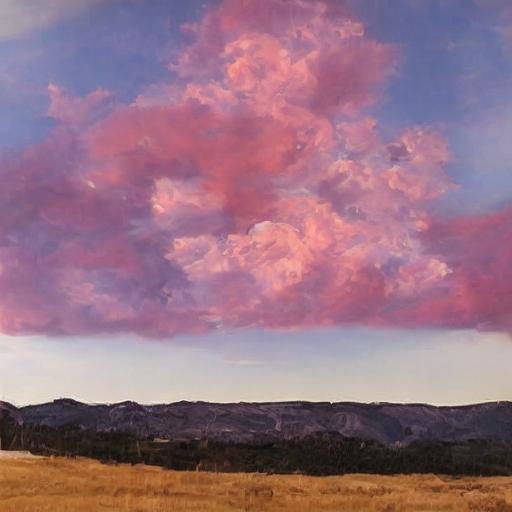} \\
& & & STROTSS & MAST & TR & DIA & GLStyleNet \\
& & & \includegraphics[width=0.14\linewidth]{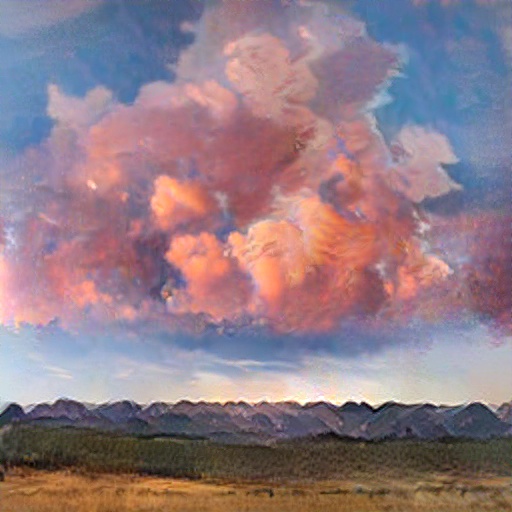} & \includegraphics[width=0.14\linewidth]{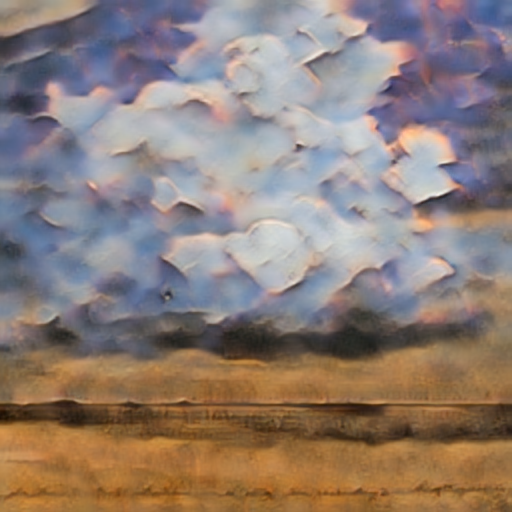} & 
\includegraphics[width=0.14\linewidth]{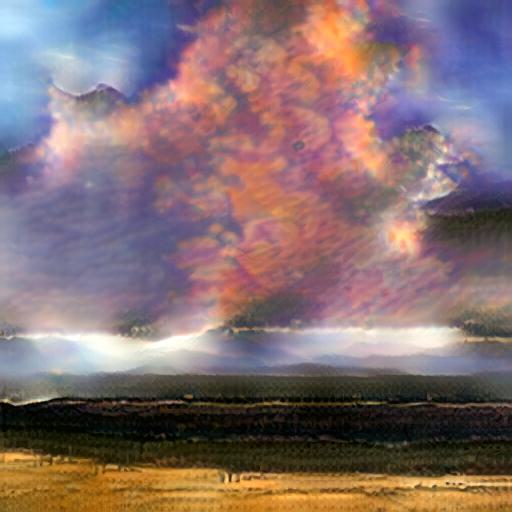} &
\includegraphics[width=0.14\linewidth]{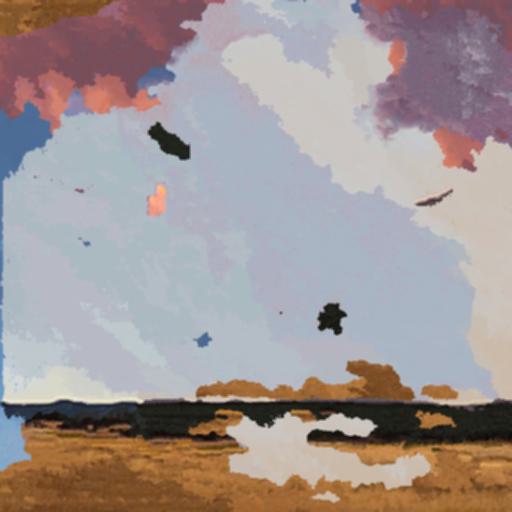} & 
\includegraphics[width=0.14\linewidth]{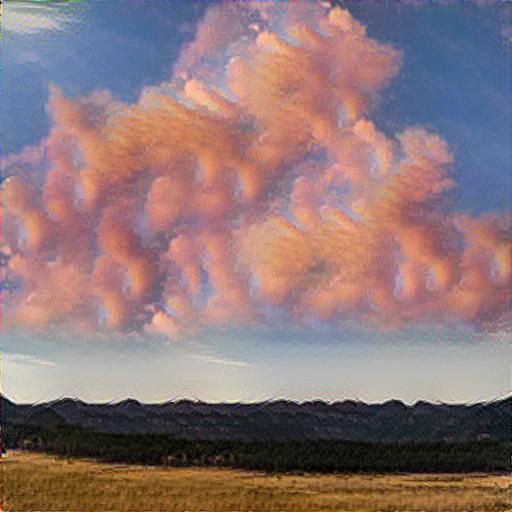} \\
&
\\
 Content & Style & SANet & SANet + SCSA & StyTR$^2$ & StyTR$^2$ + SCSA & StyleID & StyleID + SCSA \\

\includegraphics[width=0.14\linewidth]{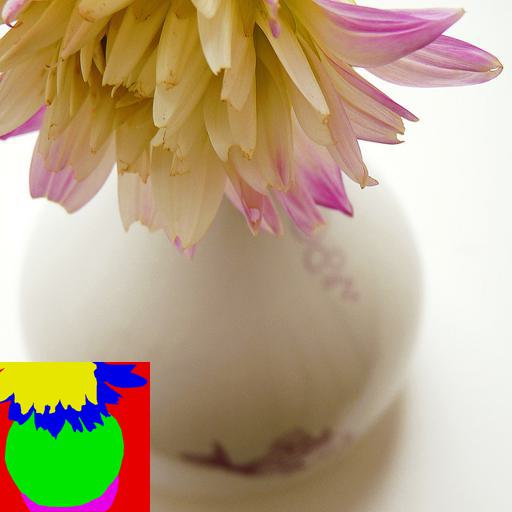} & \includegraphics[width=0.14\linewidth]{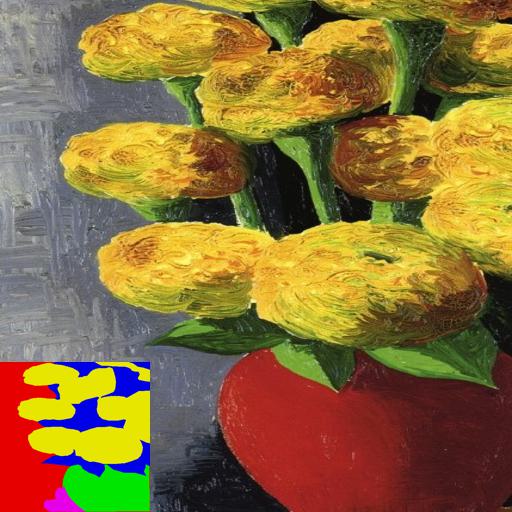}  & \includegraphics[width=0.14\linewidth]{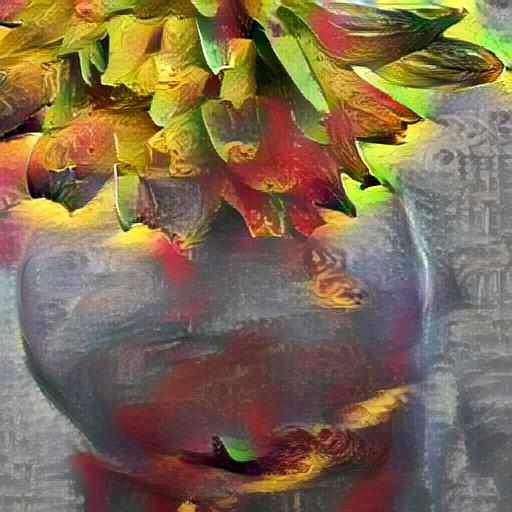}  &
\includegraphics[width=0.14\linewidth]{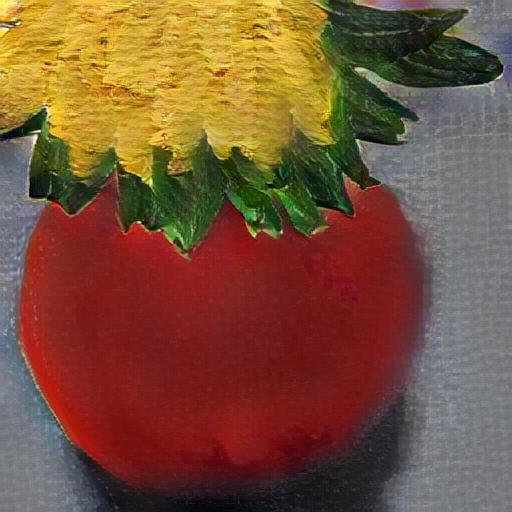}& \includegraphics[width=0.14\linewidth]{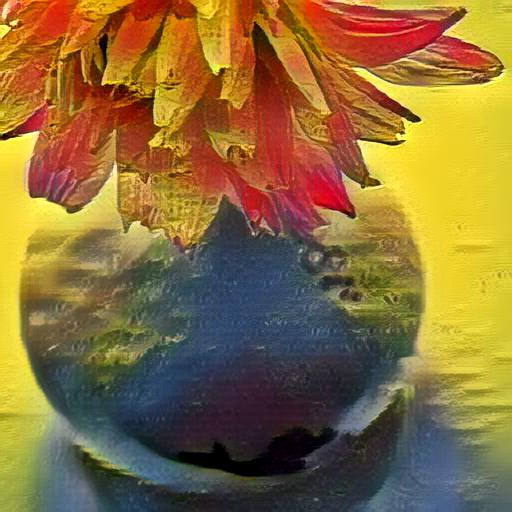} &
\includegraphics[width=0.14\linewidth]{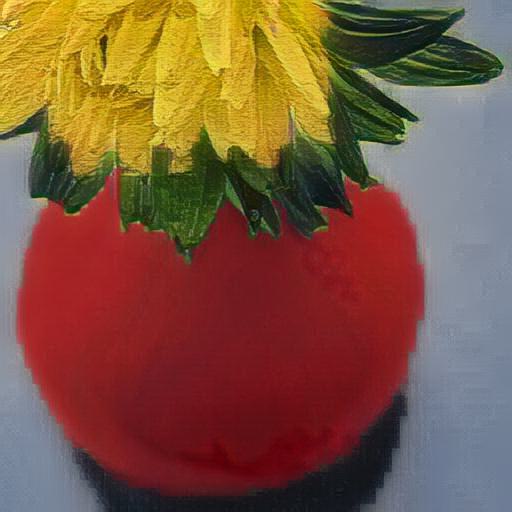} &  \includegraphics[width=0.14\linewidth]{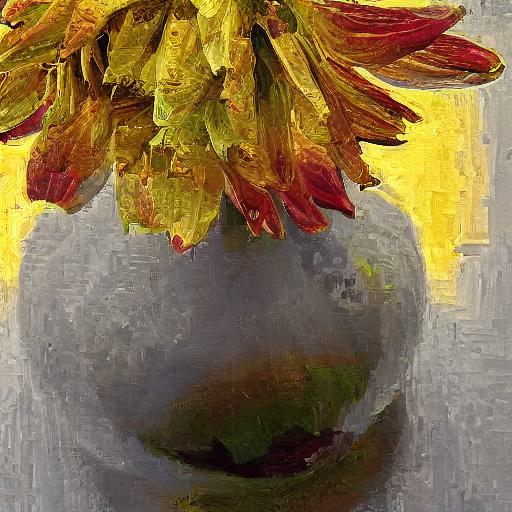} &  \includegraphics[width=0.14\linewidth]{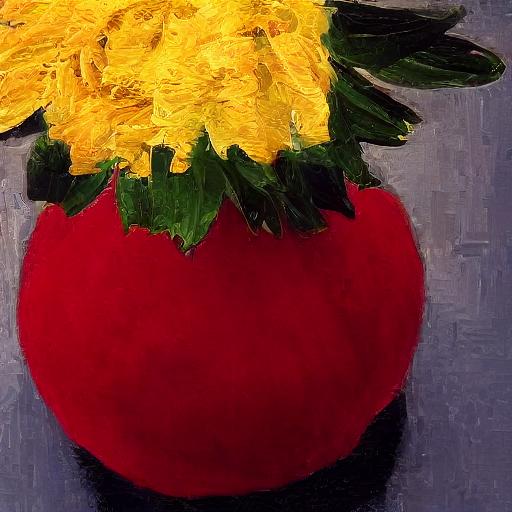} \\
& & & STROTSS & MAST & TR & DIA & GLStyleNet \\
& & & \includegraphics[width=0.14\linewidth]{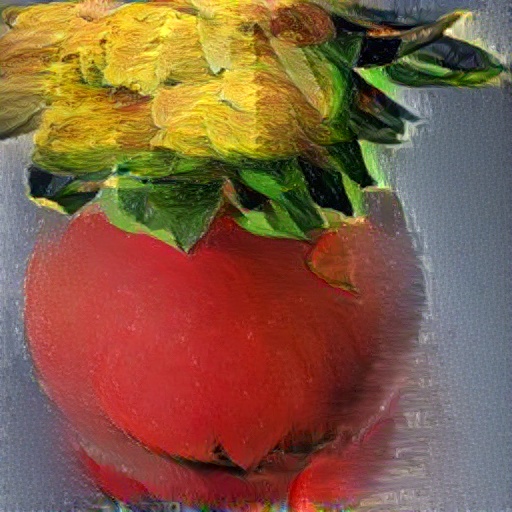} & \includegraphics[width=0.14\linewidth]{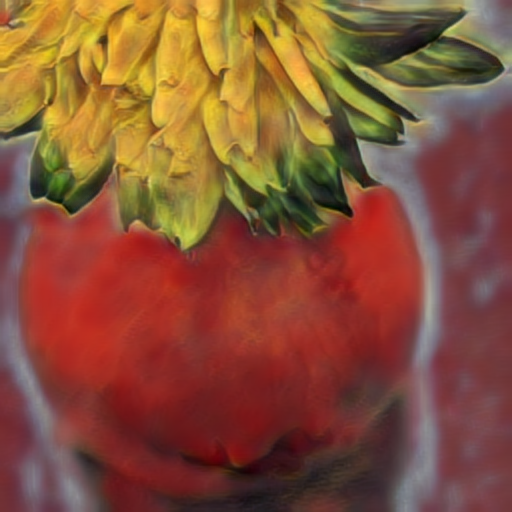} & 
\includegraphics[width=0.14\linewidth]{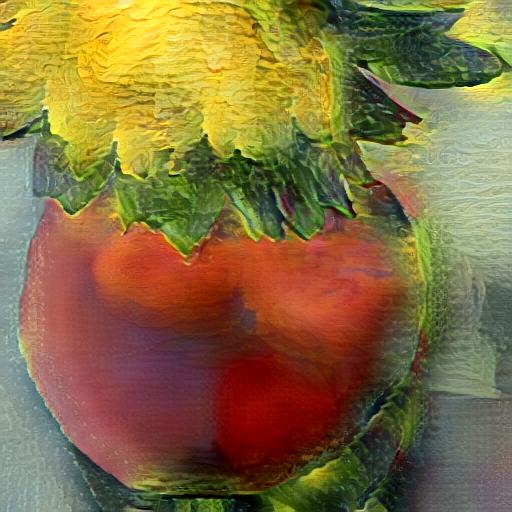} &
\includegraphics[width=0.14\linewidth]{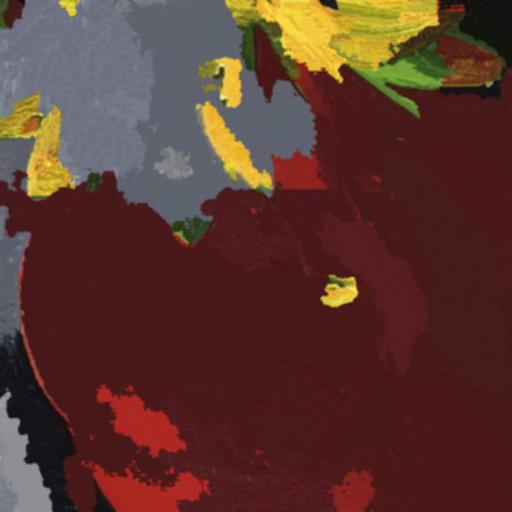} & 
\includegraphics[width=0.14\linewidth]{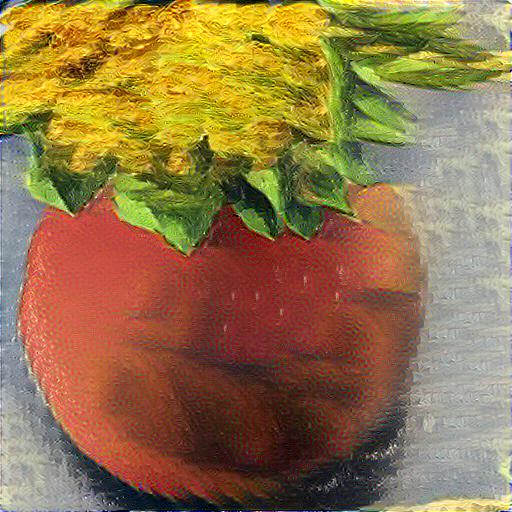} \\

&
\\
 Content & Style & SANet & SANet + SCSA & StyTR$^2$ & StyTR$^2$ + SCSA & StyleID & StyleID + SCSA \\

\includegraphics[width=0.14\linewidth]{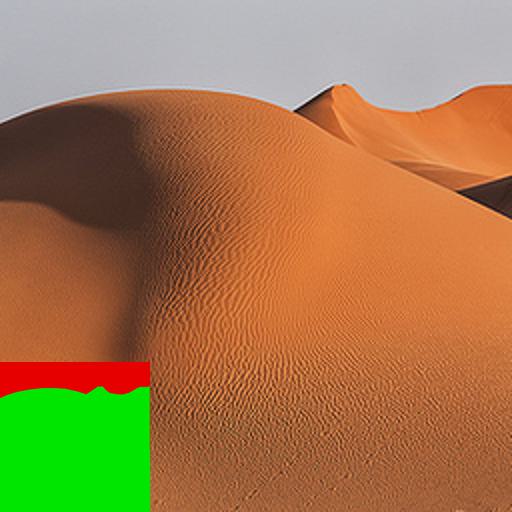} & \includegraphics[width=0.14\linewidth]{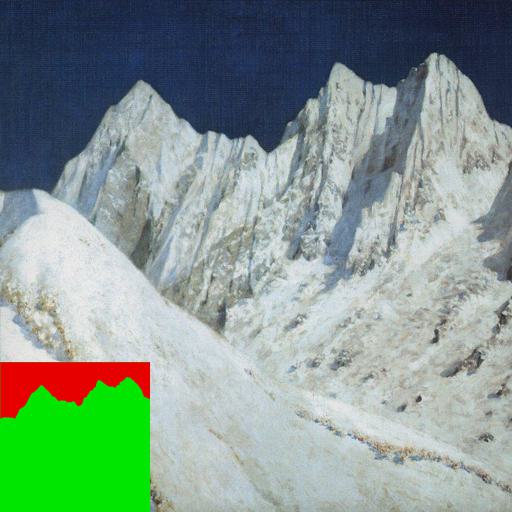}  & \includegraphics[width=0.14\linewidth]{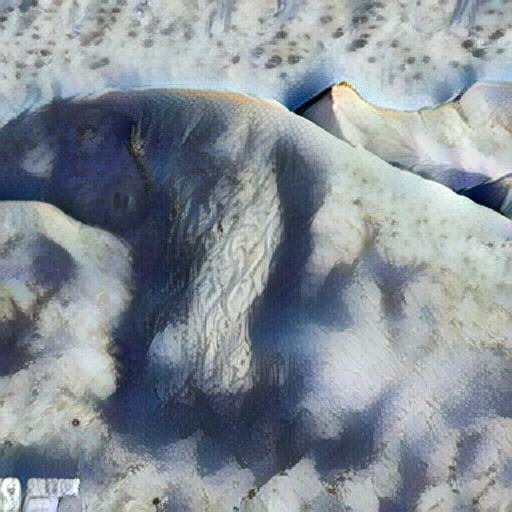}  &
\includegraphics[width=0.14\linewidth]{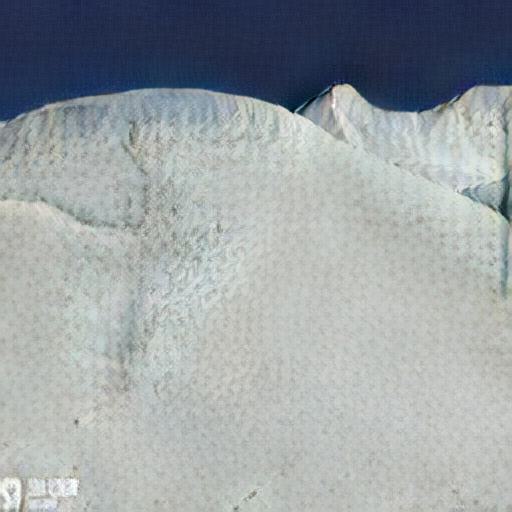}& \includegraphics[width=0.14\linewidth]{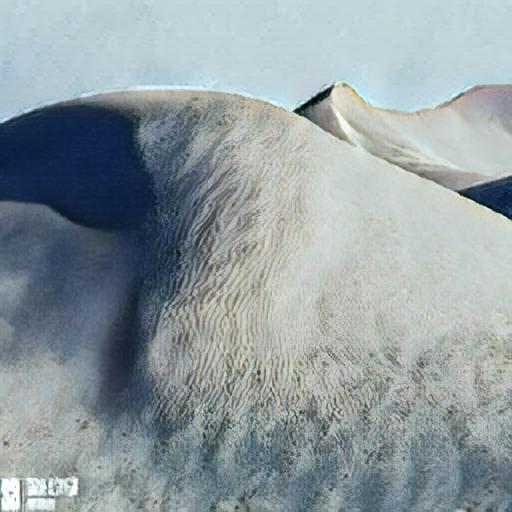} &
\includegraphics[width=0.14\linewidth]{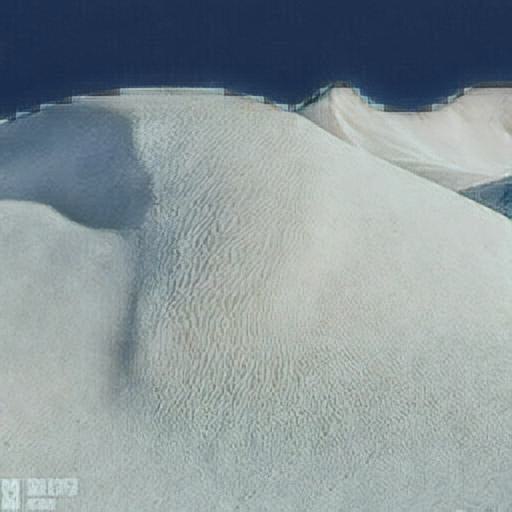} &  \includegraphics[width=0.14\linewidth]{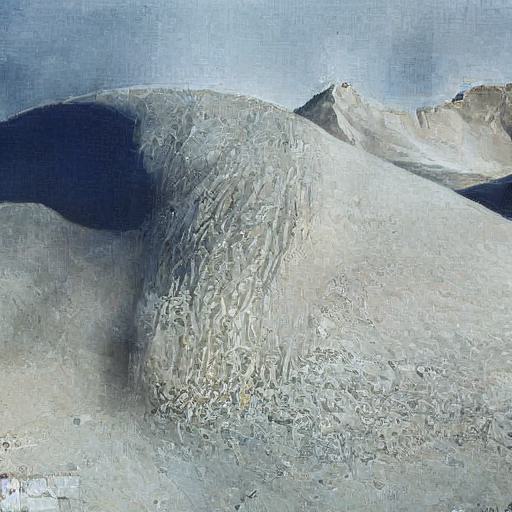} &  \includegraphics[width=0.14\linewidth]{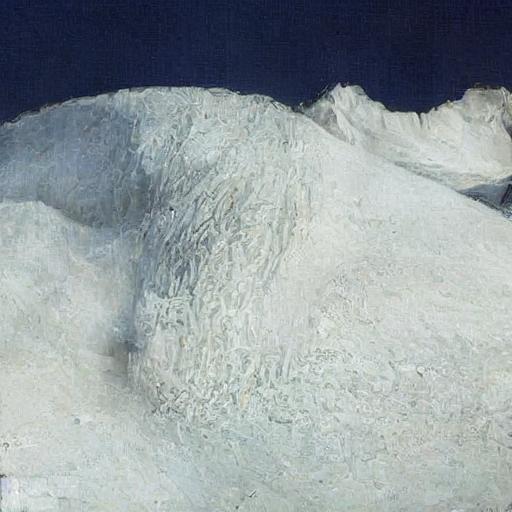} \\
& & & STROTSS & MAST & TR & DIA & GLStyleNet \\
& & & \includegraphics[width=0.14\linewidth]{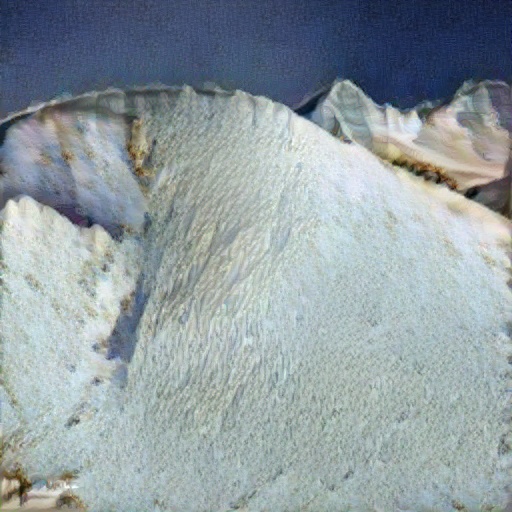} & \includegraphics[width=0.14\linewidth]{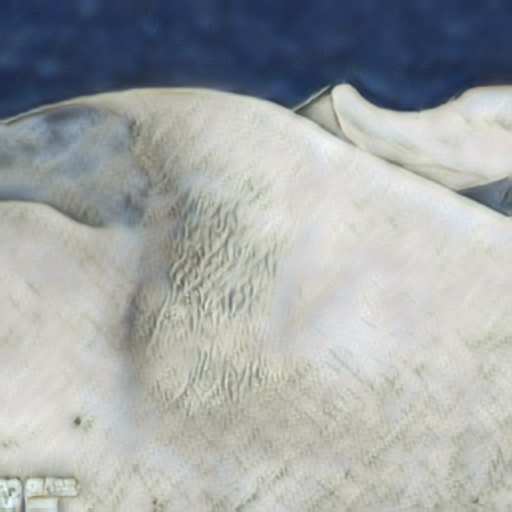} & 
\includegraphics[width=0.14\linewidth]{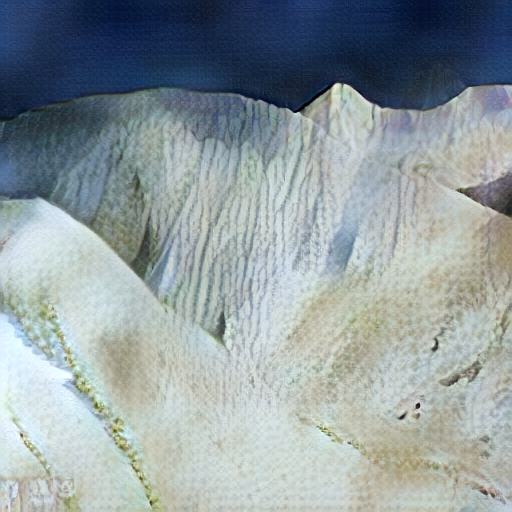} &
\includegraphics[width=0.14\linewidth]{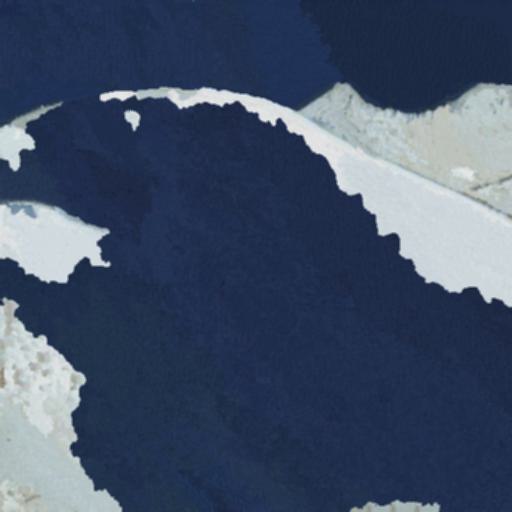} & 
\includegraphics[width=0.14\linewidth]{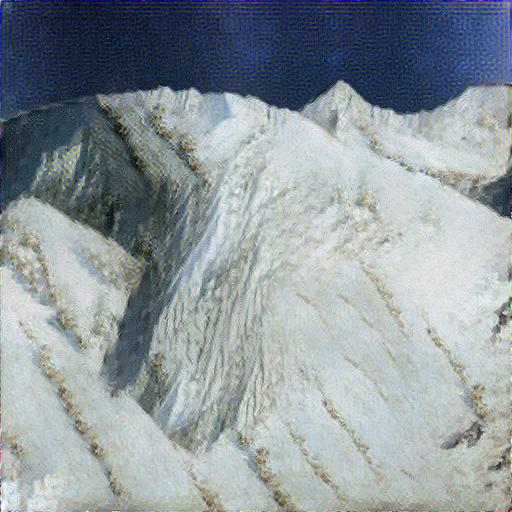} \\

\end{tabular}
}
\caption{Qualitative comparisons among Attn-AST approaches, those with SCSA, and SOTA methods.}
\label{fig:16}
\end{figure*}

\begin{figure*}
\centering
\resizebox{1.0\textwidth}{!}{
\setlength{\tabcolsep}{0.02cm} 
\renewcommand{\arraystretch}{1}  
\begin{tabular}{cccccccc}
 Content & Style & SANet & SANet + SCSA & StyTR$^2$ & StyTR$^2$ + SCSA & StyleID & StyleID + SCSA \\
\includegraphics[width=0.14\linewidth]{sm/img/31_paint+sem.jpg} & \includegraphics[width=0.14\linewidth]{sm/img/31+sem.jpg} & \includegraphics[width=0.14\linewidth]{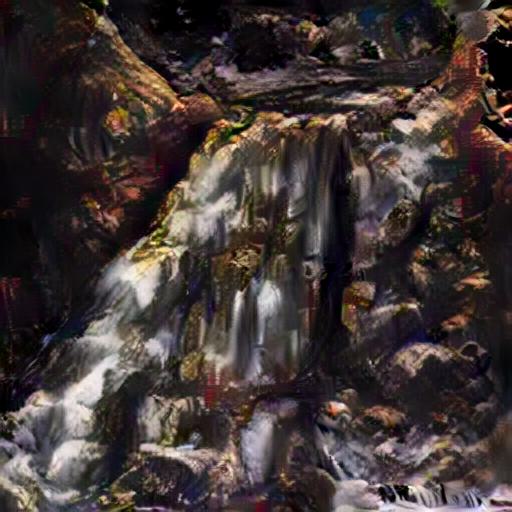}  &
\includegraphics[width=0.14\linewidth]{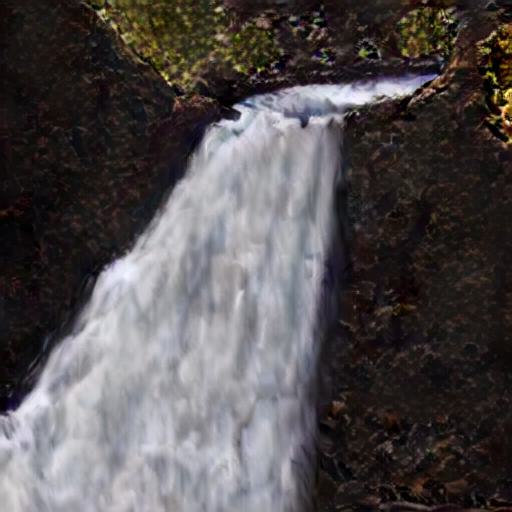}& \includegraphics[width=0.14\linewidth]{sm/img/31_paint_31_StyTR2.jpg} &
\includegraphics[width=0.14\linewidth]{sm/img/31_paint_31_StyTR2_sem.jpg} &  \includegraphics[width=0.14\linewidth]{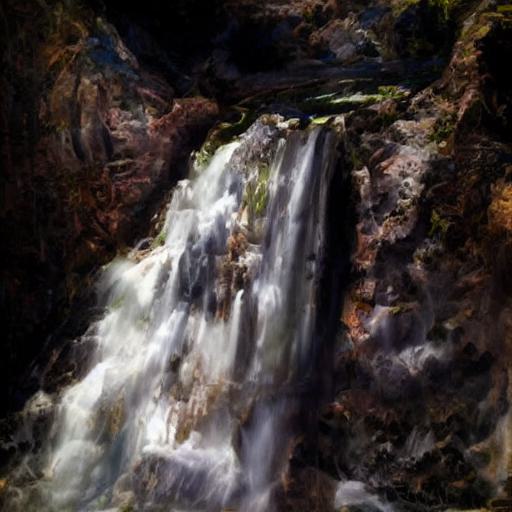} &  \includegraphics[width=0.14\linewidth]{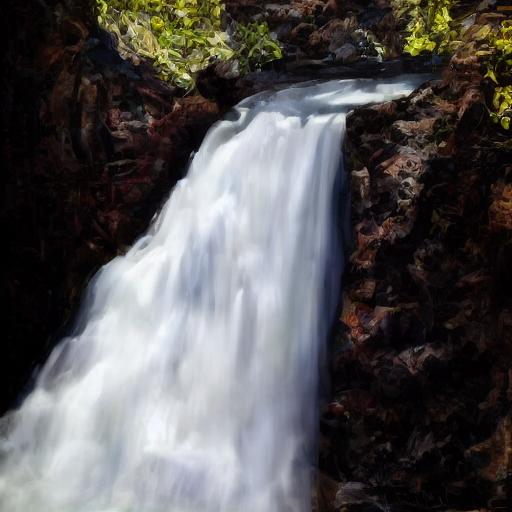} \\
& & & STROTSS & MAST & TR & DIA & GLStyleNet \\
& & & \includegraphics[width=0.14\linewidth]{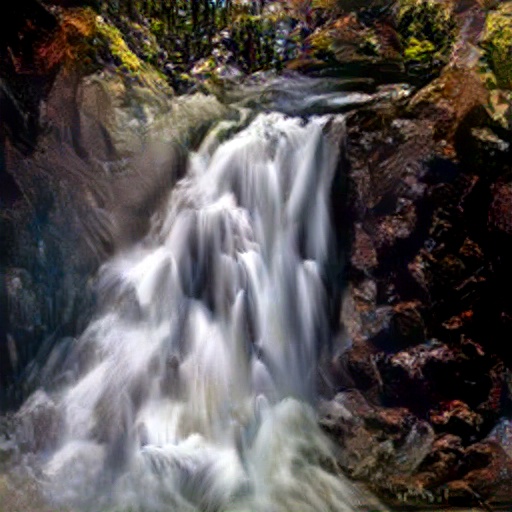} & \includegraphics[width=0.14\linewidth]{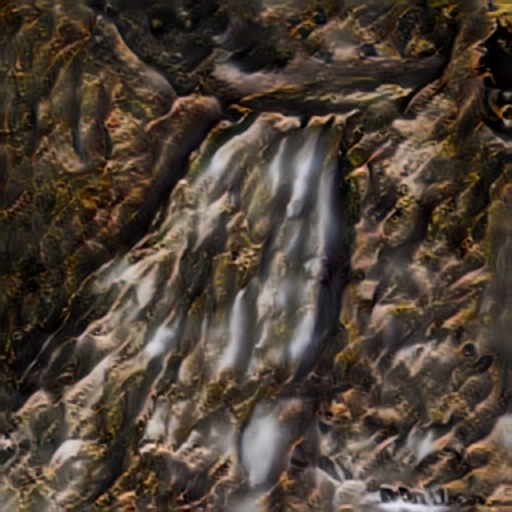} & 
\includegraphics[width=0.14\linewidth]{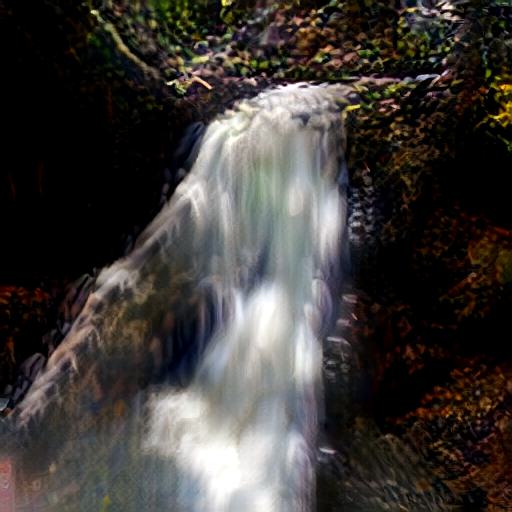} &
\includegraphics[width=0.14\linewidth]{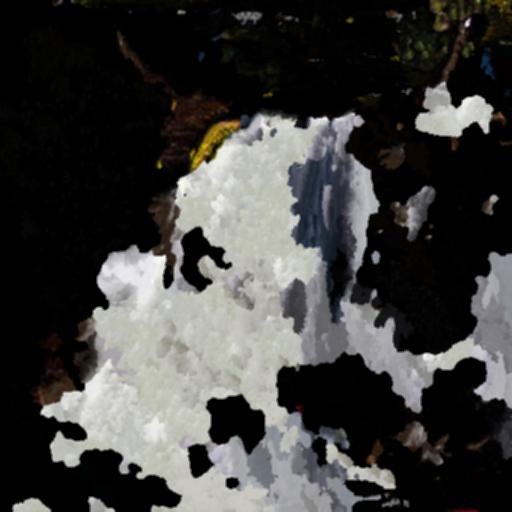} & 
\includegraphics[width=0.14\linewidth]{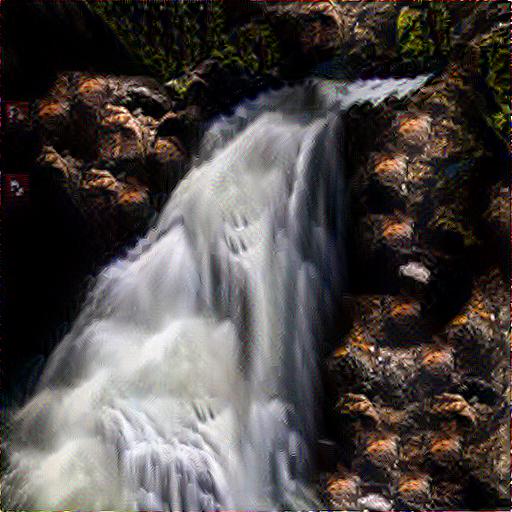} \\
& 
\\
 Content & Style & SANet & SANet + SCSA & StyTR$^2$ & StyTR$^2$ + SCSA & StyleID & StyleID + SCSA \\

\includegraphics[width=0.14\linewidth]{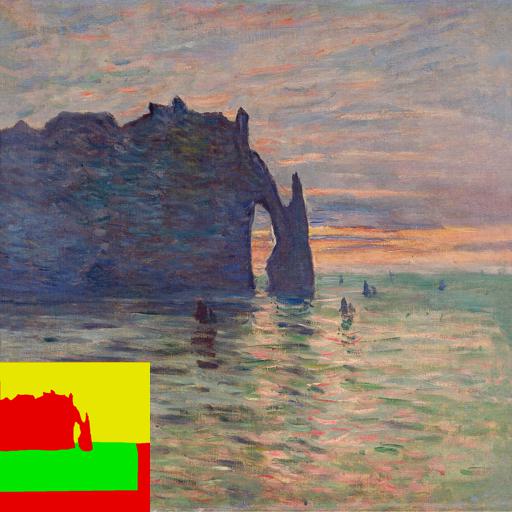} & \includegraphics[width=0.14\linewidth]{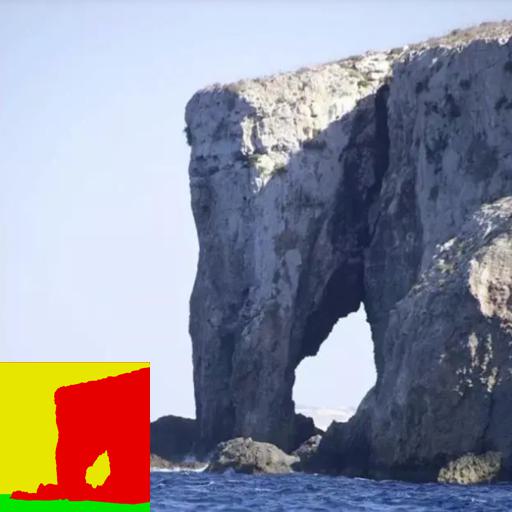}  & \includegraphics[width=0.14\linewidth]{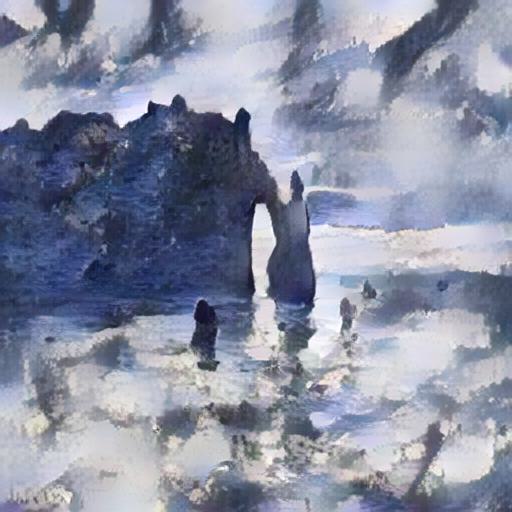}  &
\includegraphics[width=0.14\linewidth]{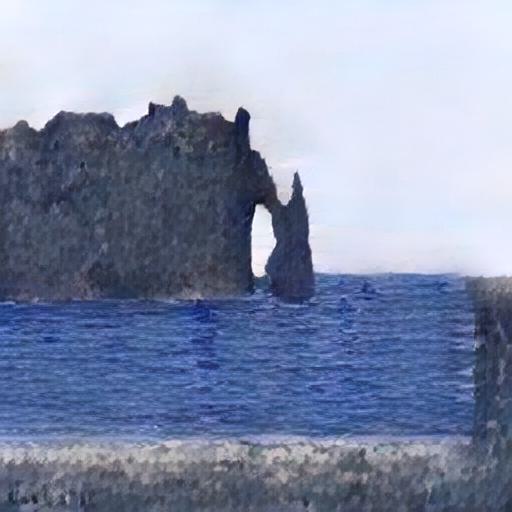}& \includegraphics[width=0.14\linewidth]{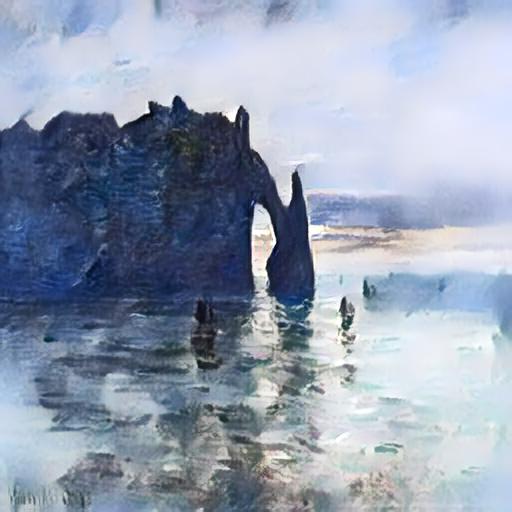} &
\includegraphics[width=0.14\linewidth]{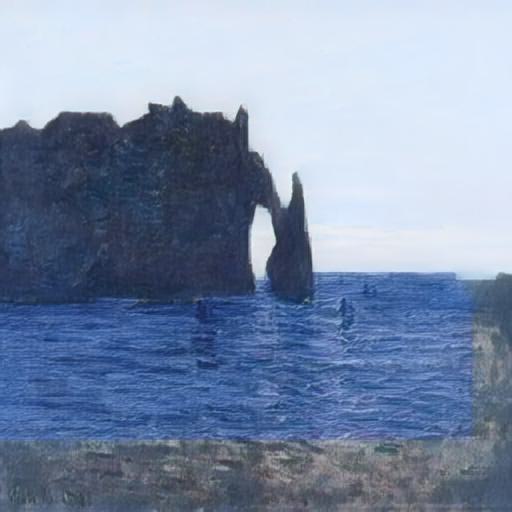} &  \includegraphics[width=0.14\linewidth]{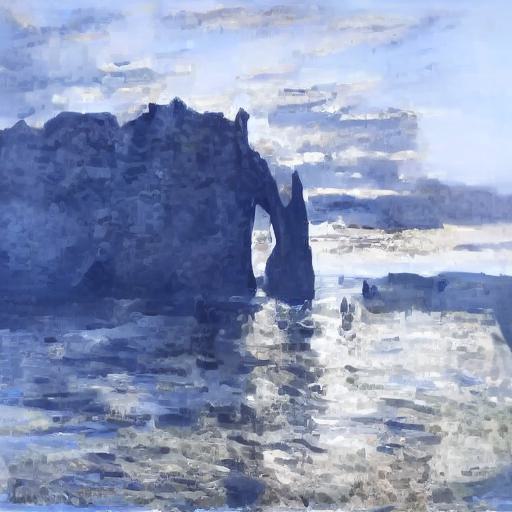} &  \includegraphics[width=0.14\linewidth]{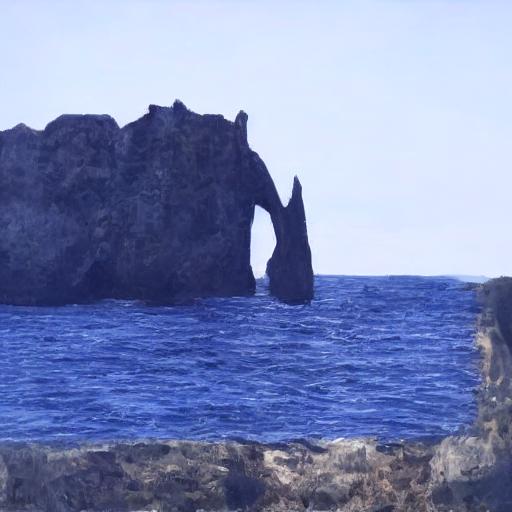} \\
& & & STROTSS & MAST & TR & DIA & GLStyleNet \\
& & & \includegraphics[width=0.14\linewidth]{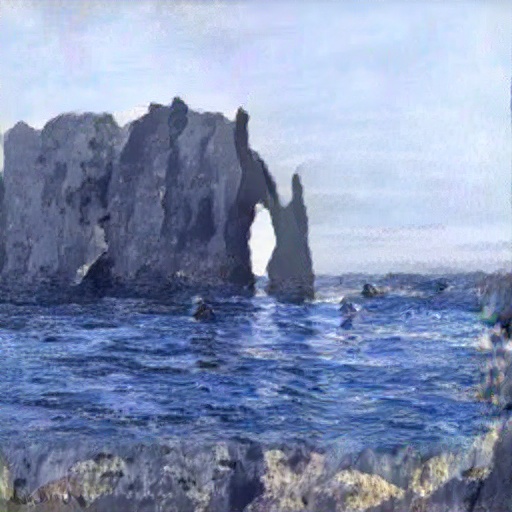} & \includegraphics[width=0.14\linewidth]{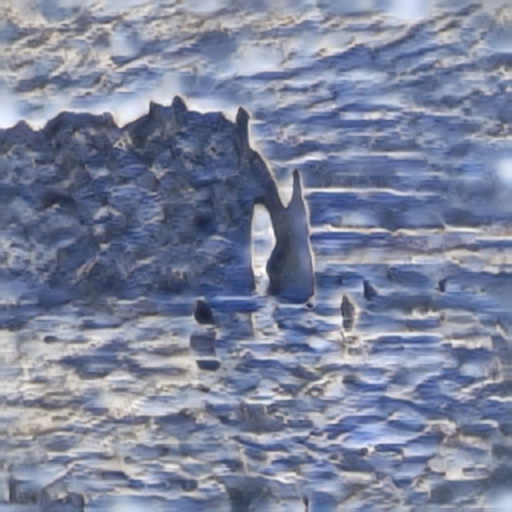} & 
\includegraphics[width=0.14\linewidth]{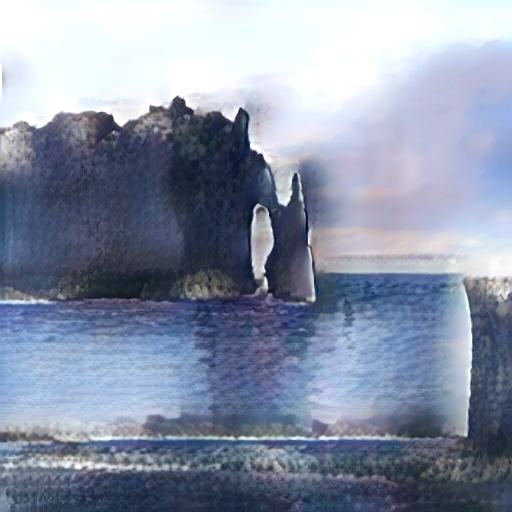} &
\includegraphics[width=0.14\linewidth]{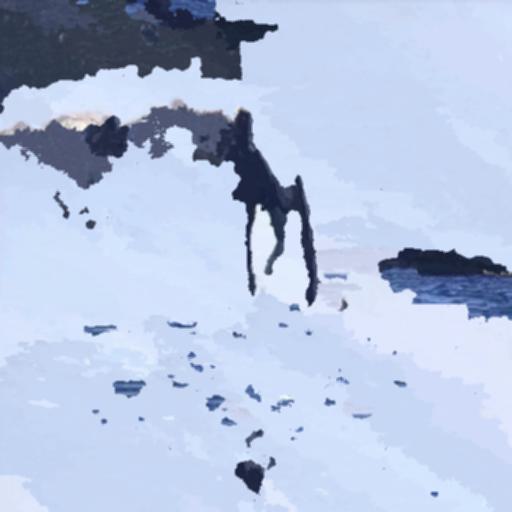} & 
\includegraphics[width=0.14\linewidth]{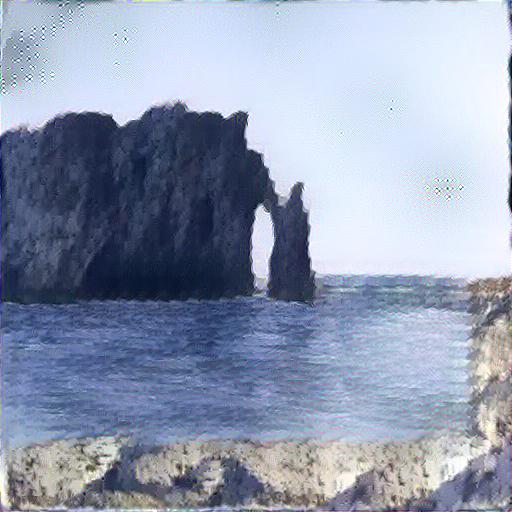} \\
&
\\
 Content & Style & SANet & SANet + SCSA & StyTR$^2$ & StyTR$^2$ + SCSA & StyleID & StyleID + SCSA \\

\includegraphics[width=0.14\linewidth]{sm/img/30+sem.jpg} & \includegraphics[width=0.14\linewidth]{sm/img/30_paint+sem.jpg}  & \includegraphics[width=0.14\linewidth]{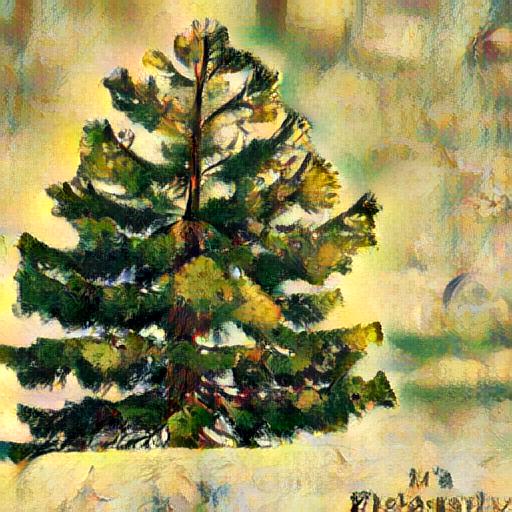}  &
\includegraphics[width=0.14\linewidth]{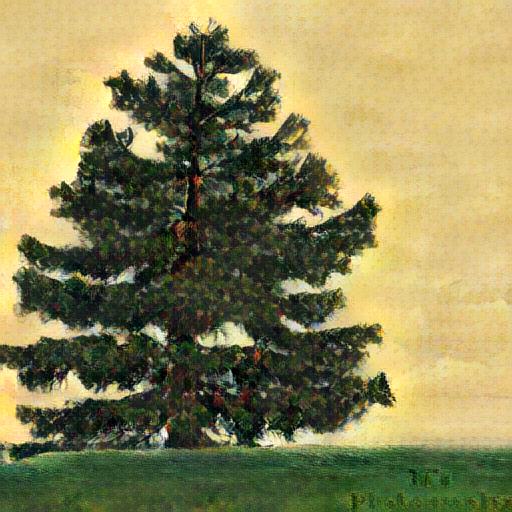}& \includegraphics[width=0.14\linewidth]{sm/img/30_30_paint_StyTR2.jpg} &
\includegraphics[width=0.14\linewidth]{sm/img/30_30_paint_StyTR2_sem.jpg} &  \includegraphics[width=0.14\linewidth]{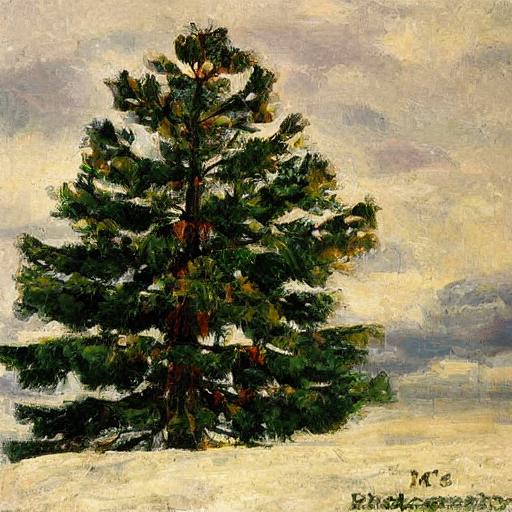} &  \includegraphics[width=0.14\linewidth]{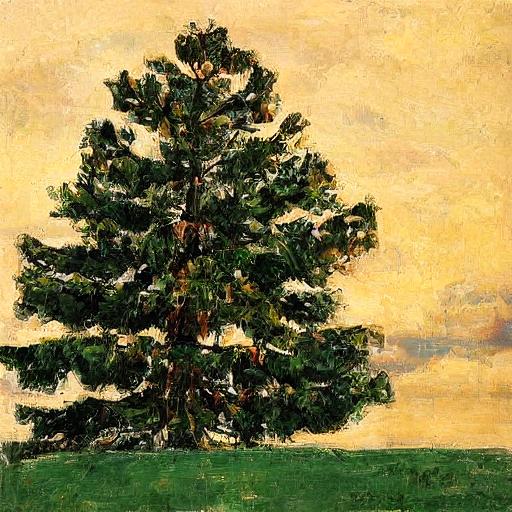} \\
& & & STROTSS & MAST & TR & DIA & GLStyleNet \\
& & & \includegraphics[width=0.14\linewidth]{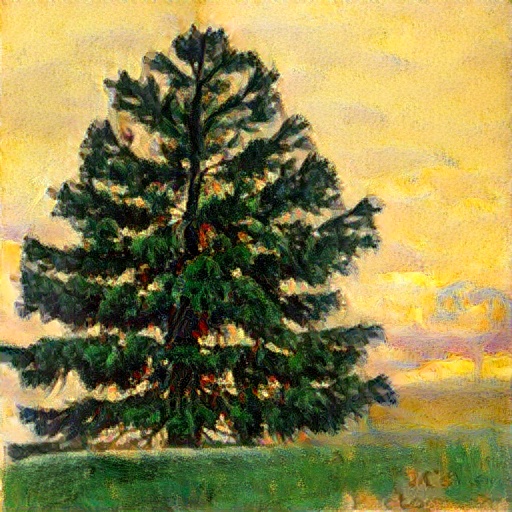} & \includegraphics[width=0.14\linewidth]{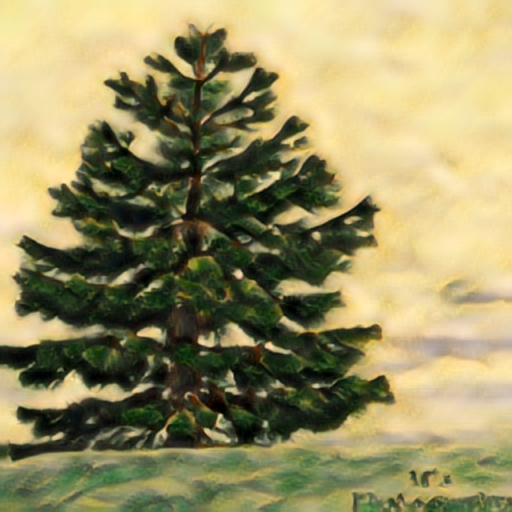} & 
\includegraphics[width=0.14\linewidth]{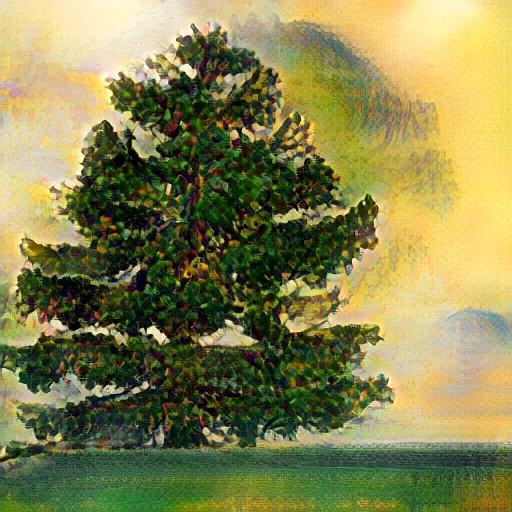} &
\includegraphics[width=0.14\linewidth]{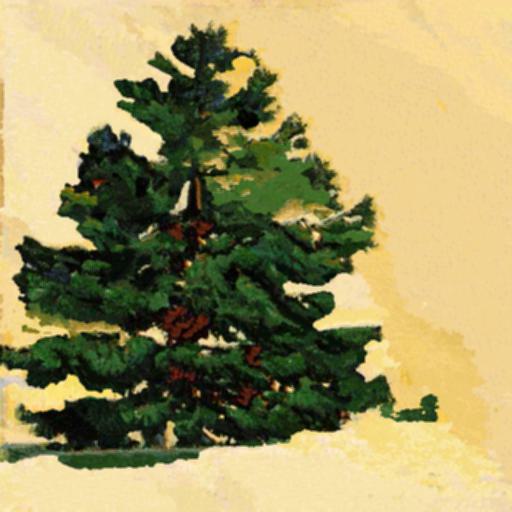} & 
\includegraphics[width=0.14\linewidth]{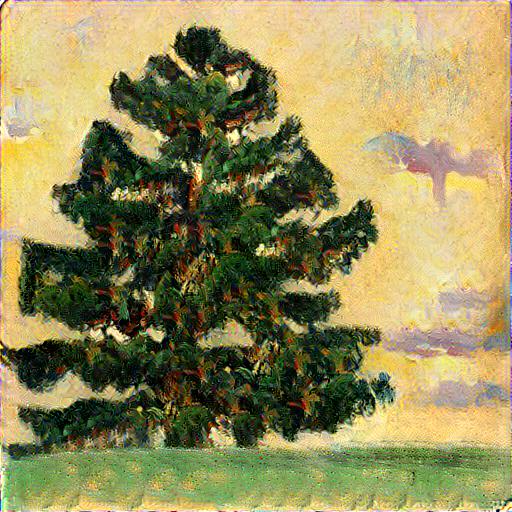} \\

&
\\
 Content & Style & SANet & SANet + SCSA & StyTR$^2$ & StyTR$^2$ + SCSA & StyleID & StyleID + SCSA \\

\includegraphics[width=0.14\linewidth]{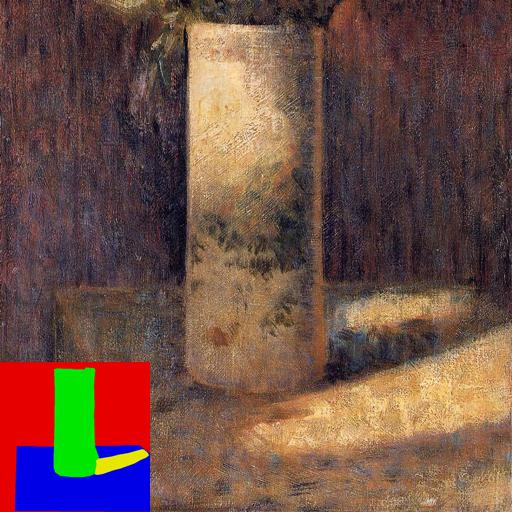} & \includegraphics[width=0.14\linewidth]{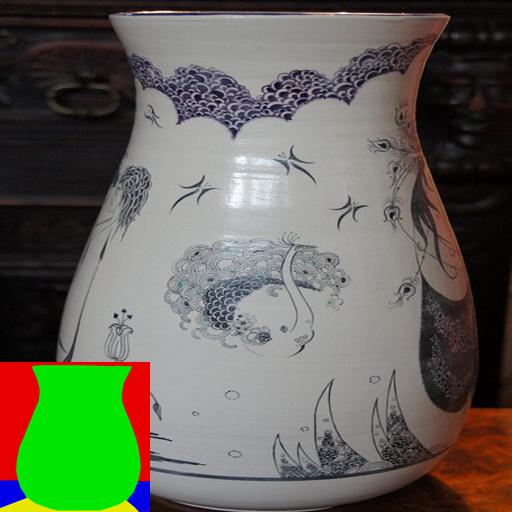}  & \includegraphics[width=0.14\linewidth]{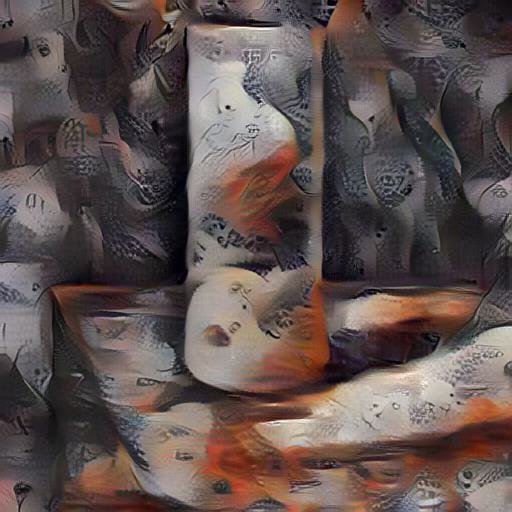}  &
\includegraphics[width=0.14\linewidth]{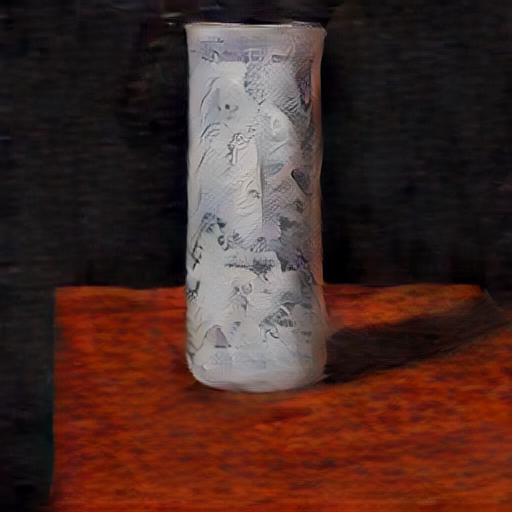}& \includegraphics[width=0.14\linewidth]{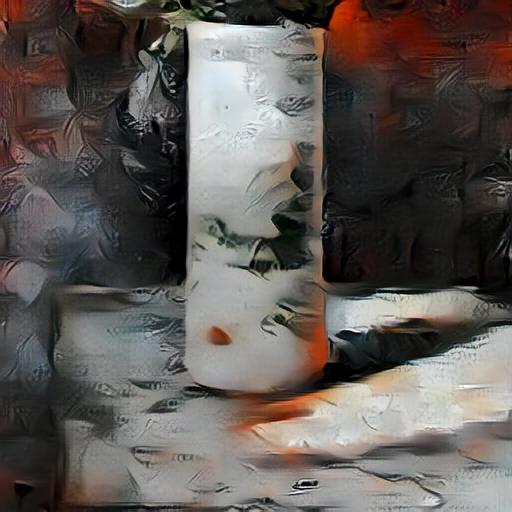} &
\includegraphics[width=0.14\linewidth]{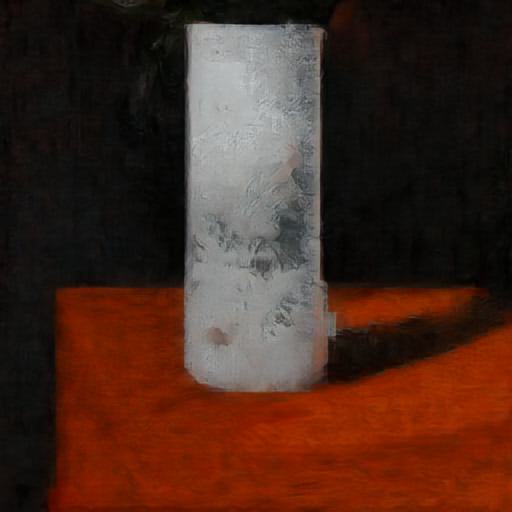} &  \includegraphics[width=0.14\linewidth]{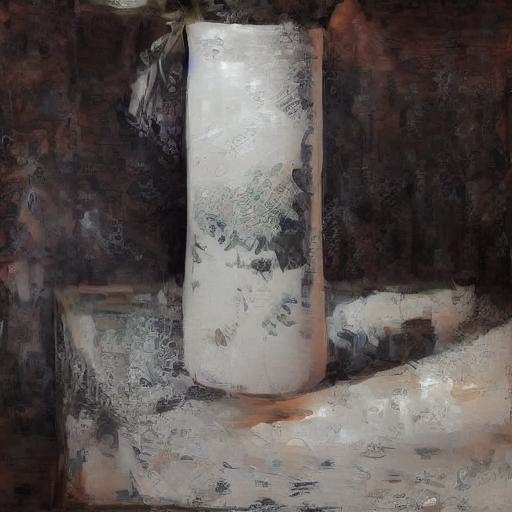} &  \includegraphics[width=0.14\linewidth]{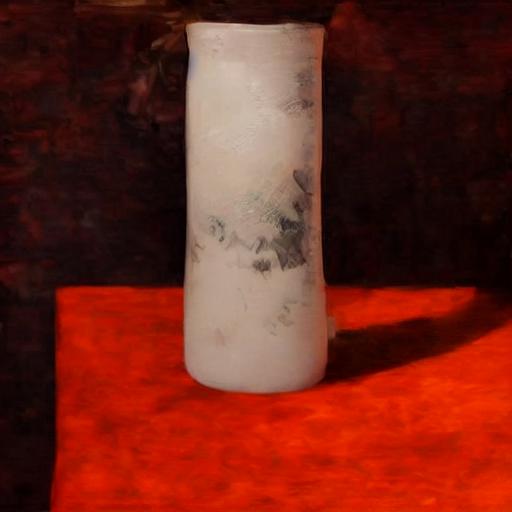} \\
& & & STROTSS & MAST & TR & DIA & GLStyleNet \\
& & & \includegraphics[width=0.14\linewidth]{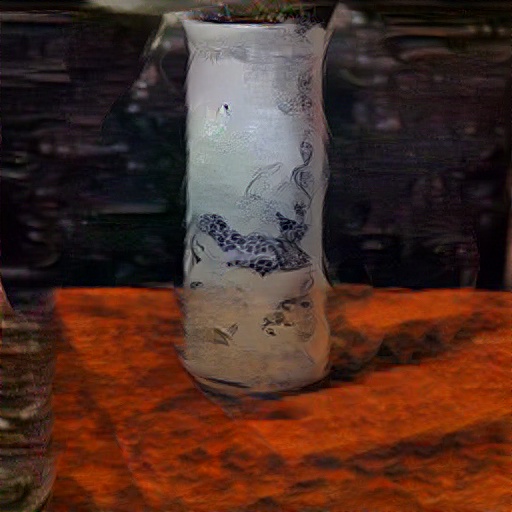} & \includegraphics[width=0.14\linewidth]{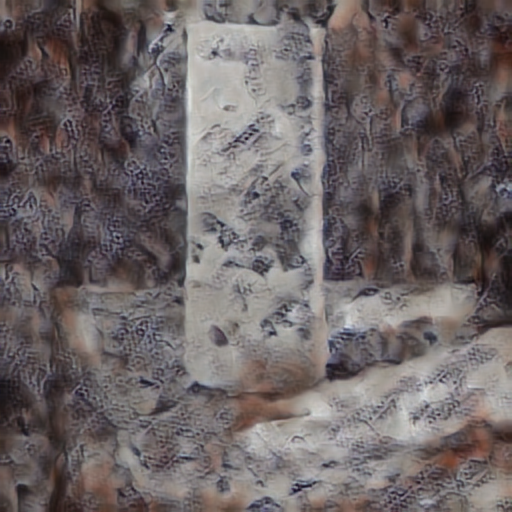} & 
\includegraphics[width=0.14\linewidth]{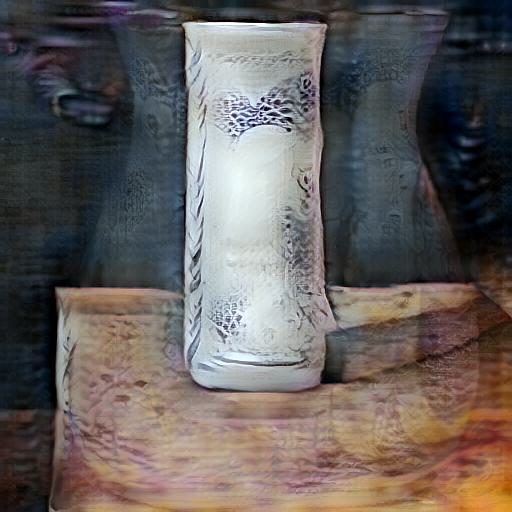} &
\includegraphics[width=0.14\linewidth]{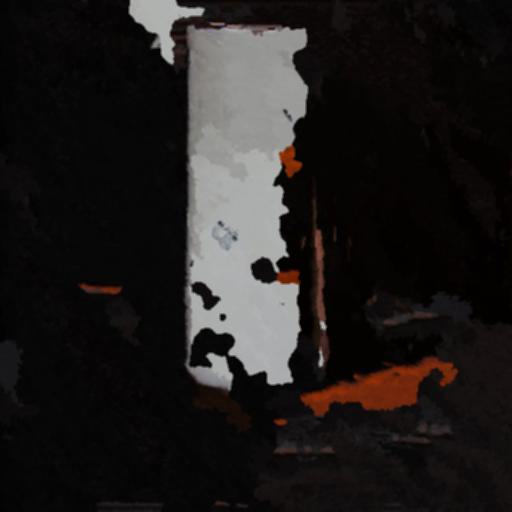} & 
\includegraphics[width=0.14\linewidth]{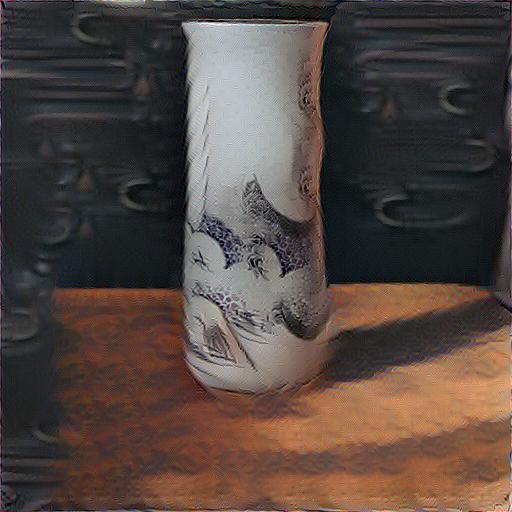} \\

\end{tabular}
}
\caption{Qualitative comparisons among Attn-AST approaches, those with SCSA, and SOTA methods.}
\label{fig:17}
\end{figure*}

\begin{figure*}
\centering
\resizebox{1.0\textwidth}{!}{
\setlength{\tabcolsep}{0.02cm} 
\renewcommand{\arraystretch}{1}  
\begin{tabular}{cccccccc}
 Content & Style & SANet & SANet + SCSA & StyTR$^2$ & StyTR$^2$ + SCSA & StyleID & StyleID + SCSA \\
\includegraphics[width=0.14\linewidth]{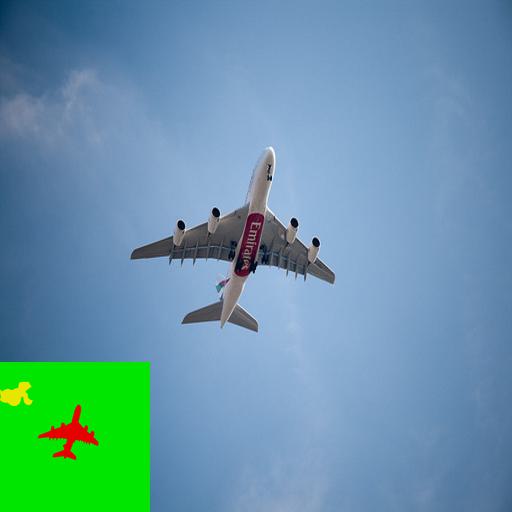} & \includegraphics[width=0.14\linewidth]{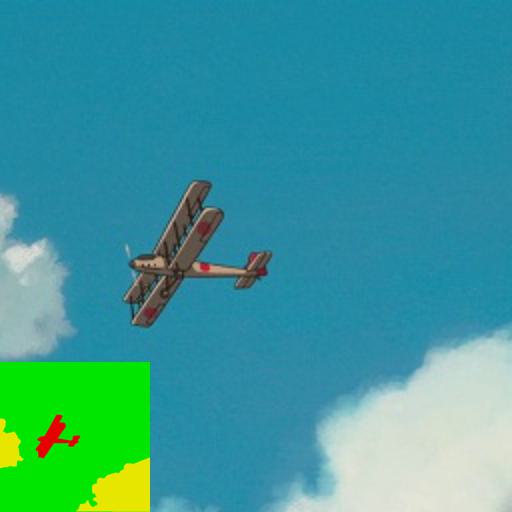} & \includegraphics[width=0.14\linewidth]{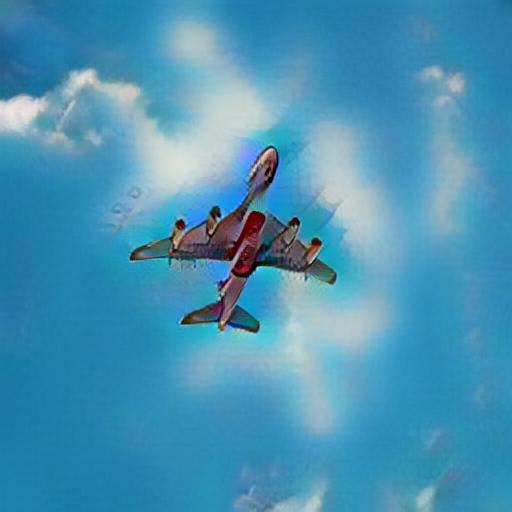}  &
\includegraphics[width=0.14\linewidth]{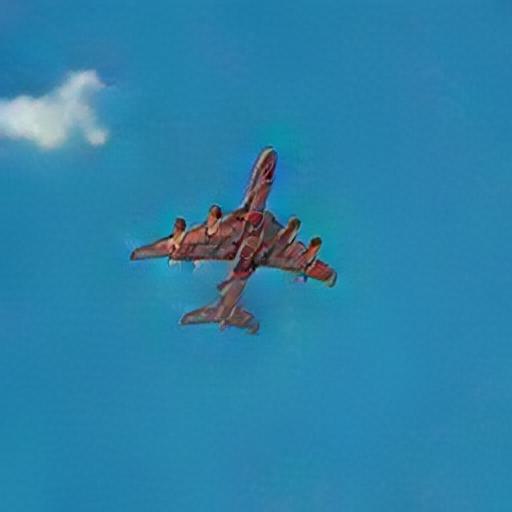}& \includegraphics[width=0.14\linewidth]{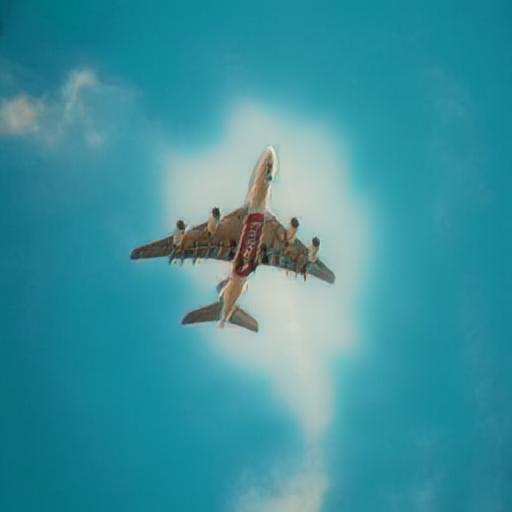} &
\includegraphics[width=0.14\linewidth]{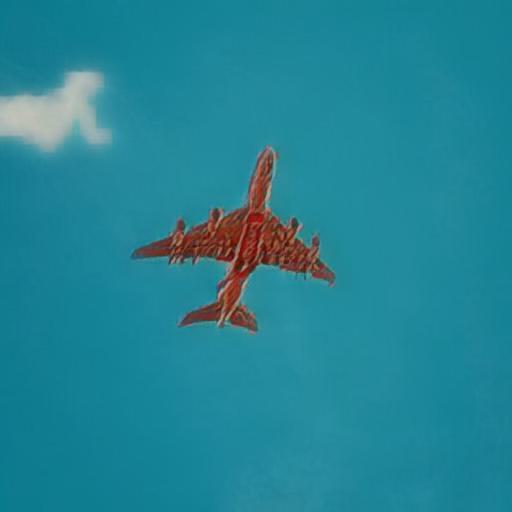} &  \includegraphics[width=0.14\linewidth]{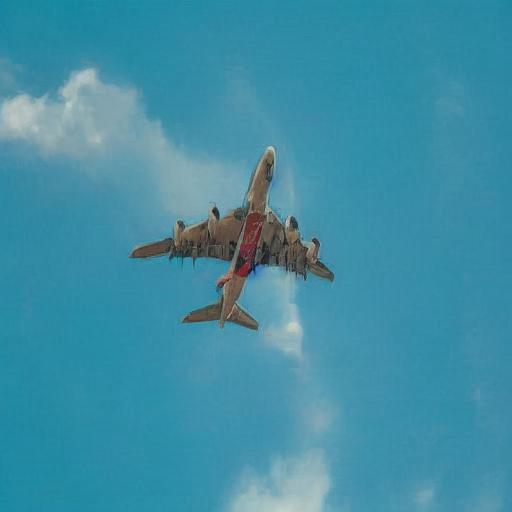} &  \includegraphics[width=0.14\linewidth]{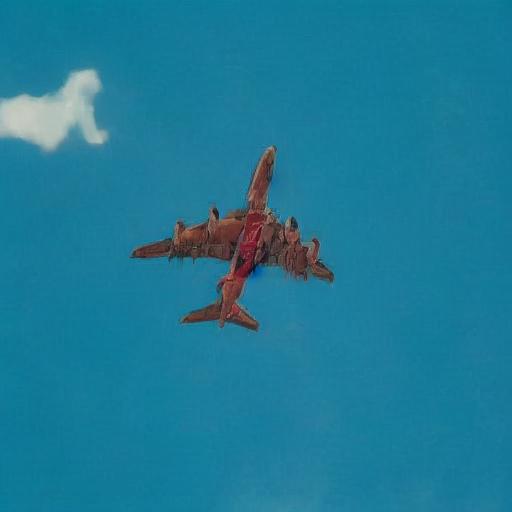} \\
& & & STROTSS & MAST & TR & DIA & GLStyleNet \\
& & & \includegraphics[width=0.14\linewidth]{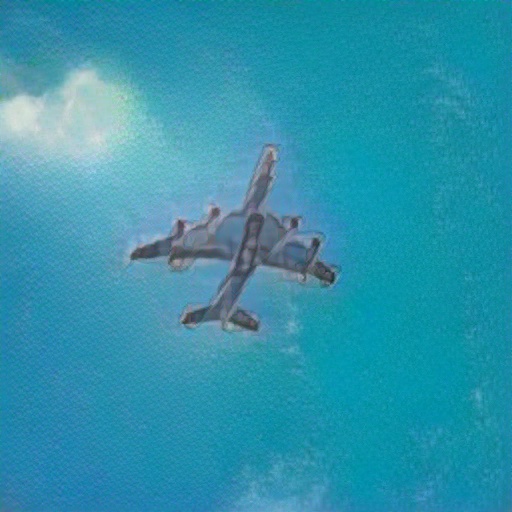} & \includegraphics[width=0.14\linewidth]{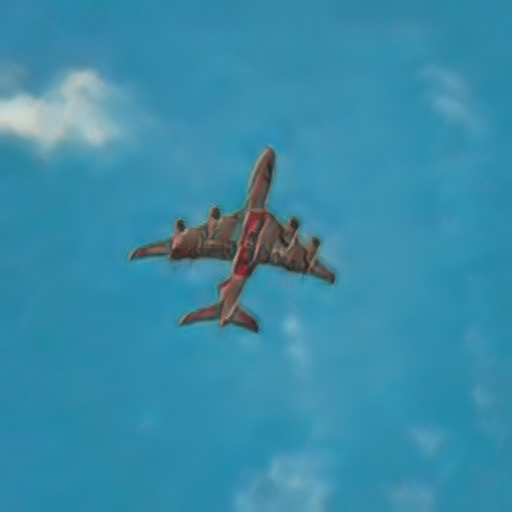} & 
\includegraphics[width=0.14\linewidth]{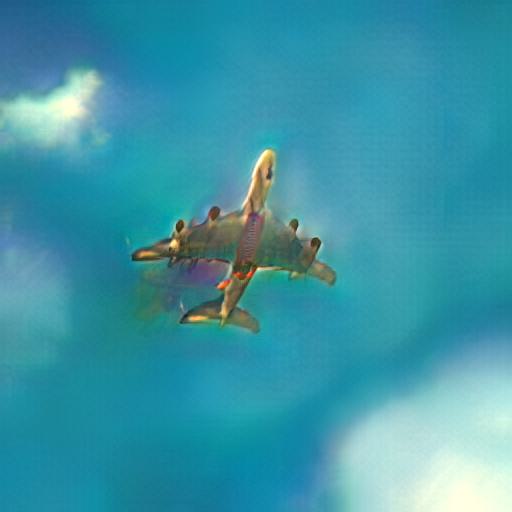} &
\includegraphics[width=0.14\linewidth]{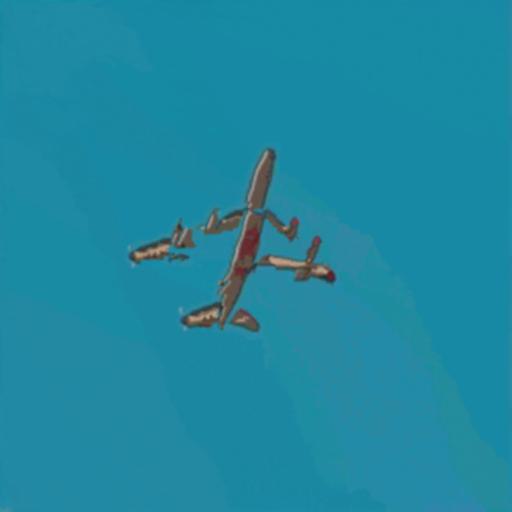} & 
\includegraphics[width=0.14\linewidth]{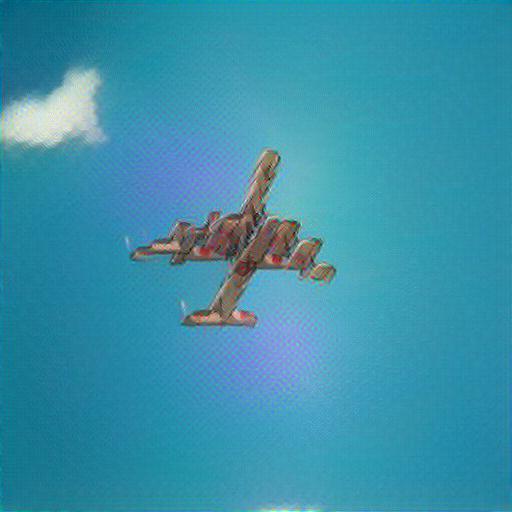} \\
& 
\\
 Content & Style & SANet & SANet + SCSA & StyTR$^2$ & StyTR$^2$ + SCSA & StyleID & StyleID + SCSA \\

\includegraphics[width=0.14\linewidth]{sm/img/36_paint+sem.jpg} & \includegraphics[width=0.14\linewidth]{sm/img/36+sem.jpg}  & \includegraphics[width=0.14\linewidth]{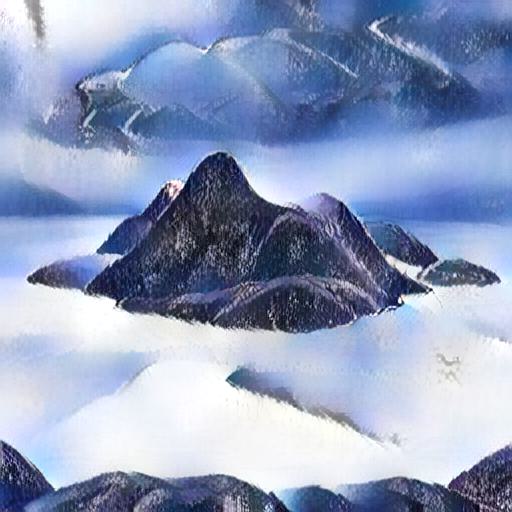}  &
\includegraphics[width=0.14\linewidth]{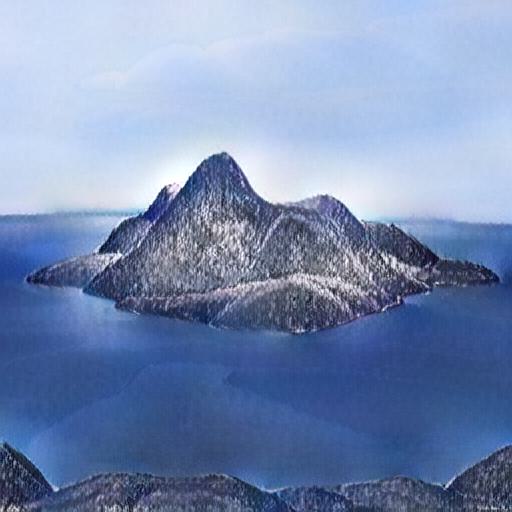}& \includegraphics[width=0.14\linewidth]{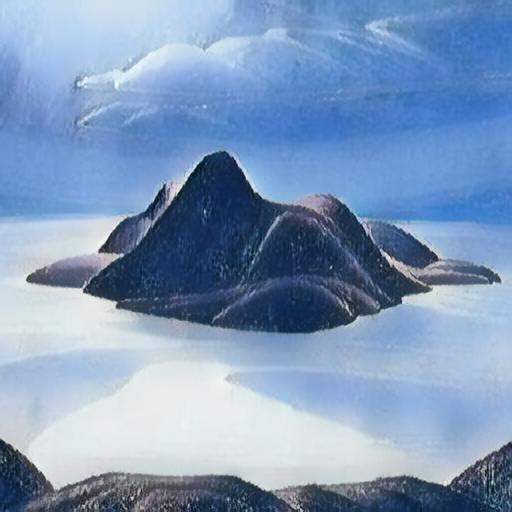} &
\includegraphics[width=0.14\linewidth]{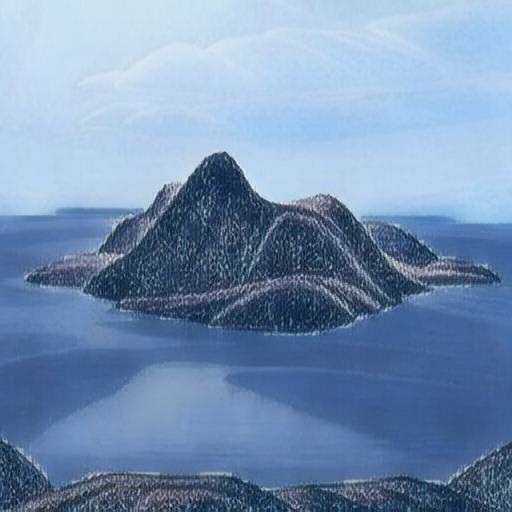} &  \includegraphics[width=0.14\linewidth]{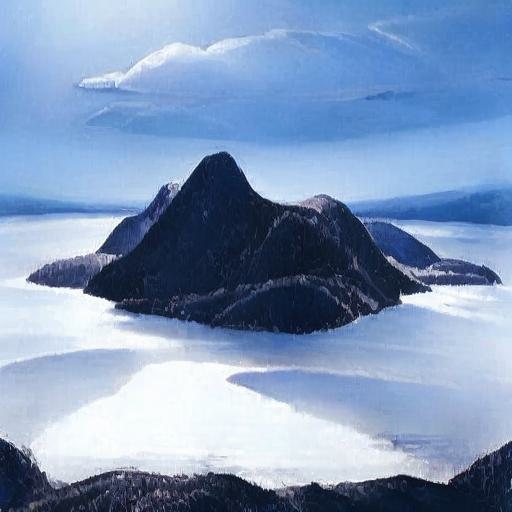} &  \includegraphics[width=0.14\linewidth]{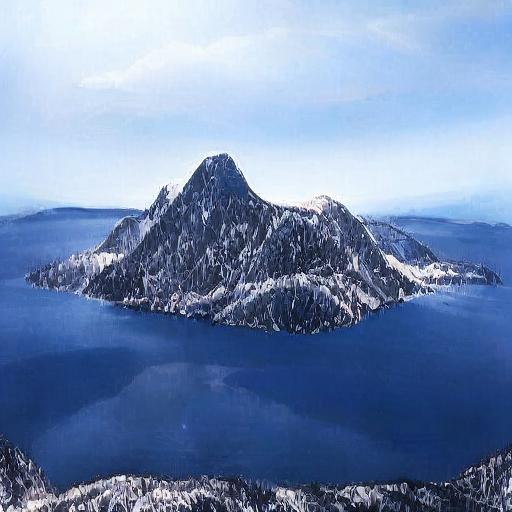} \\
& & & STROTSS & MAST & TR & DIA & GLStyleNet \\
& & & \includegraphics[width=0.14\linewidth]{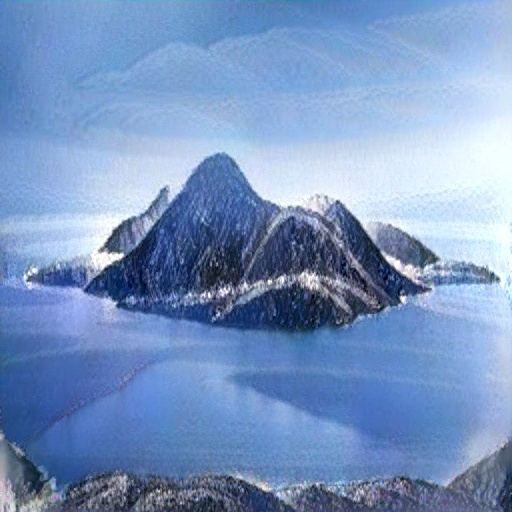} & \includegraphics[width=0.14\linewidth]{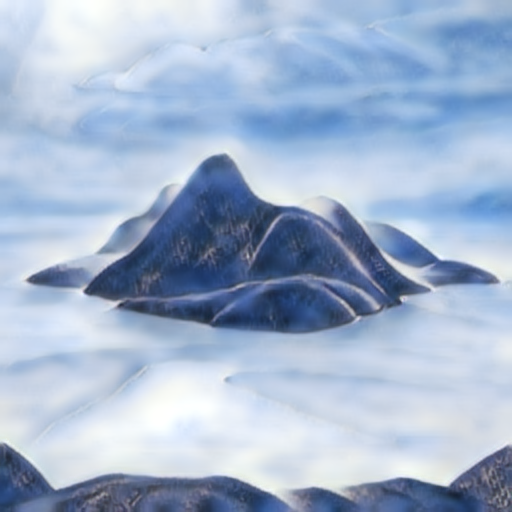} & 
\includegraphics[width=0.14\linewidth]{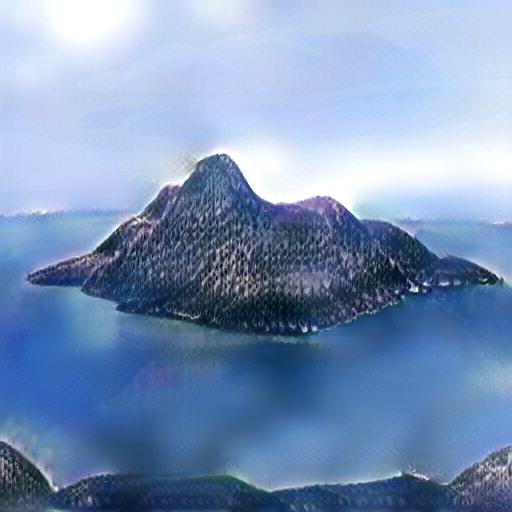} &
\includegraphics[width=0.14\linewidth]{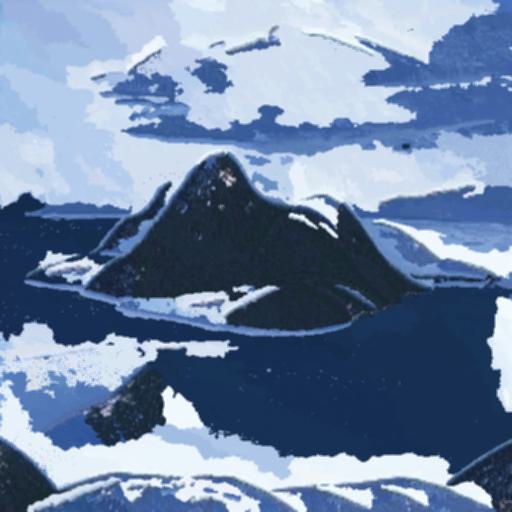} & 
\includegraphics[width=0.14\linewidth]{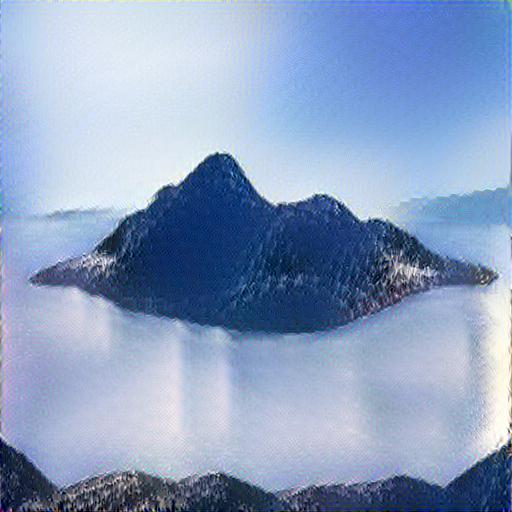} \\
&
\\
 Content & Style & SANet & SANet + SCSA & StyTR$^2$ & StyTR$^2$ + SCSA & StyleID & StyleID + SCSA \\

\includegraphics[width=0.14\linewidth]{sm/img/43_paint+sem.jpg} & \includegraphics[width=0.14\linewidth]{sm/img/43+sem.jpg}  & \includegraphics[width=0.14\linewidth]{sm/img/43_paint_43_SANet.jpg}  &
\includegraphics[width=0.14\linewidth]{sm/img/43_paint_43_SANet_sem.jpg}& \includegraphics[width=0.14\linewidth]{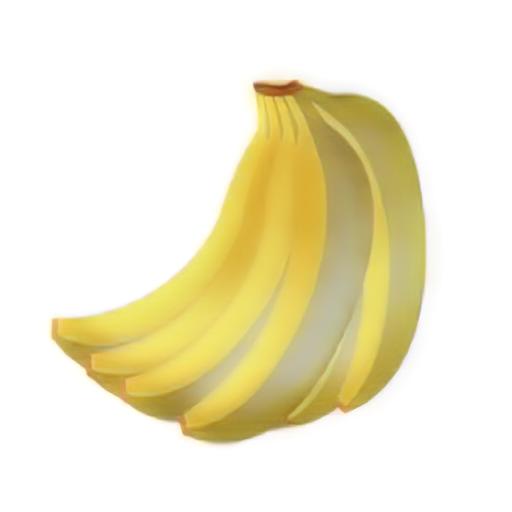} &
\includegraphics[width=0.14\linewidth]{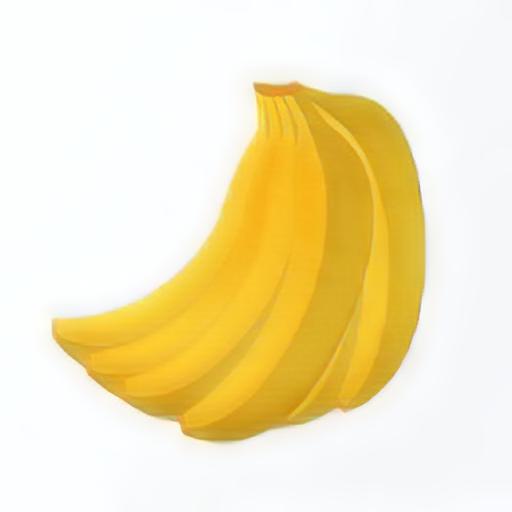} &  \includegraphics[width=0.14\linewidth]{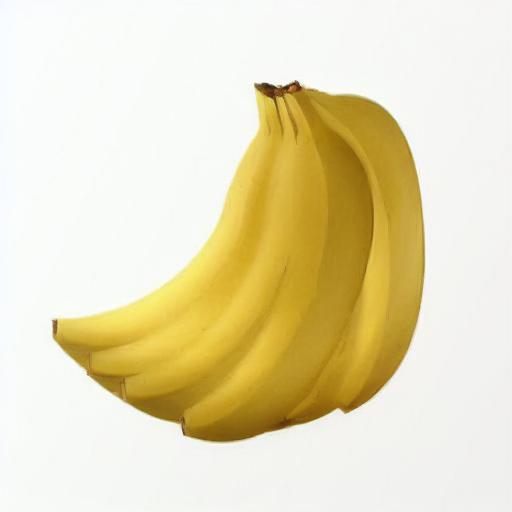} &  \includegraphics[width=0.14\linewidth]{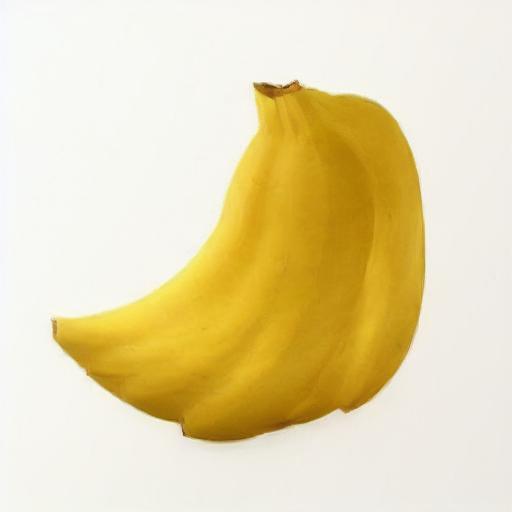} \\
& & & STROTSS & MAST & TR & DIA & GLStyleNet \\
& & & \includegraphics[width=0.14\linewidth]{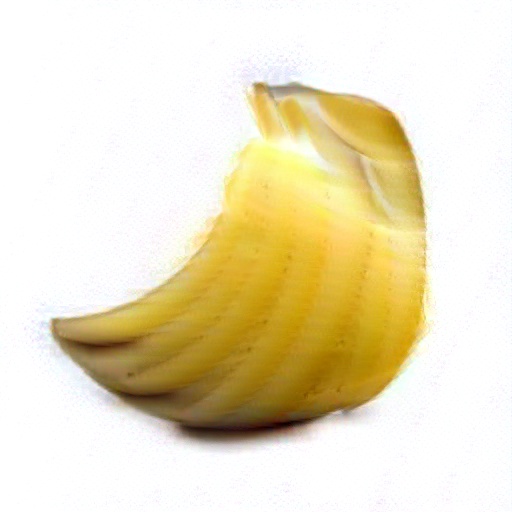} & \includegraphics[width=0.14\linewidth]{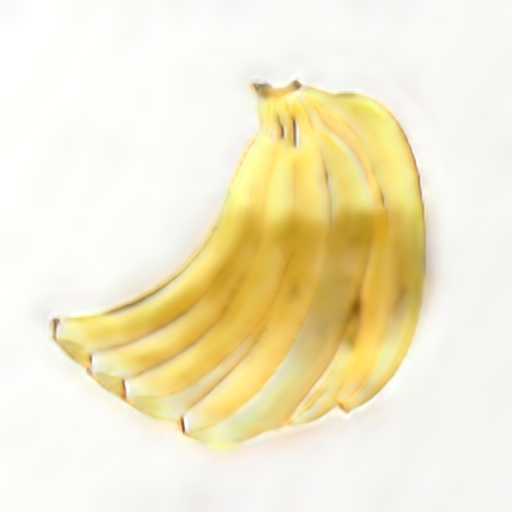} & 
\includegraphics[width=0.14\linewidth]{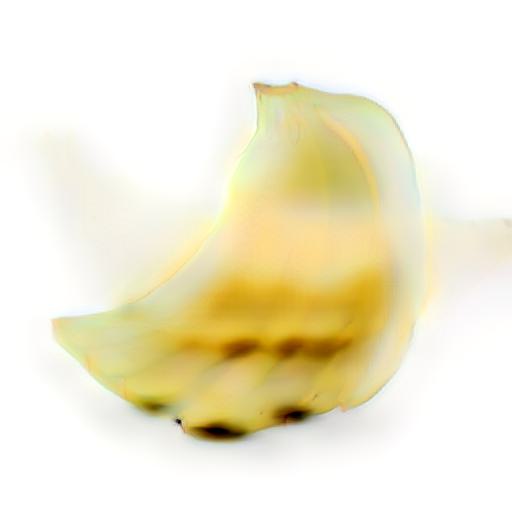} &
\includegraphics[width=0.14\linewidth]{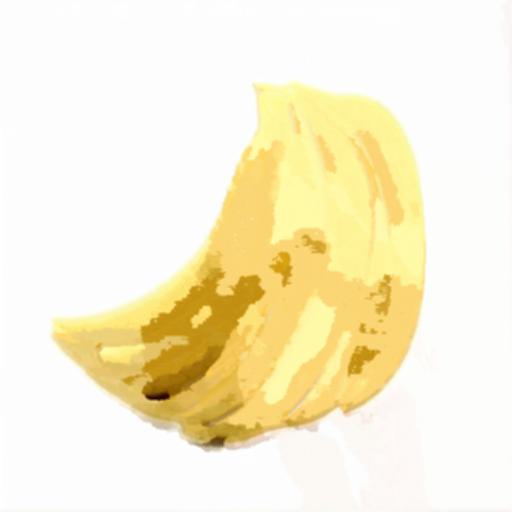} & 
\includegraphics[width=0.14\linewidth]{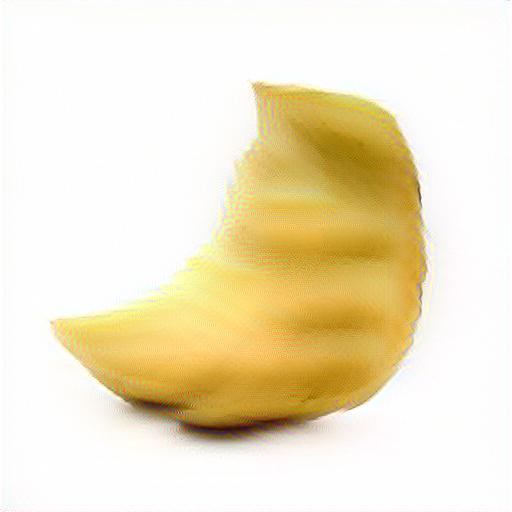} \\

&
\\
 Content & Style & SANet & SANet + SCSA & StyTR$^2$ & StyTR$^2$ + SCSA & StyleID & StyleID + SCSA \\

\includegraphics[width=0.14\linewidth]{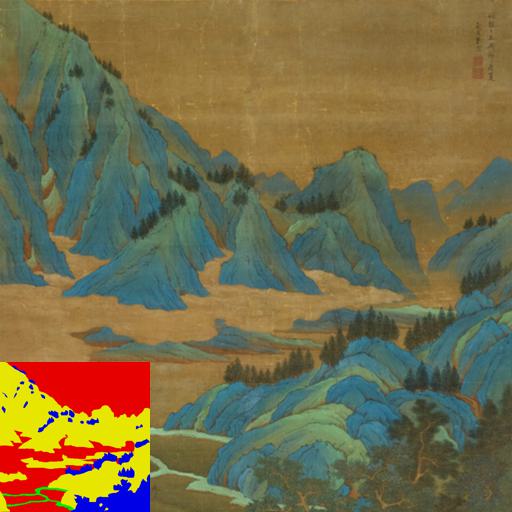} & \includegraphics[width=0.14\linewidth]{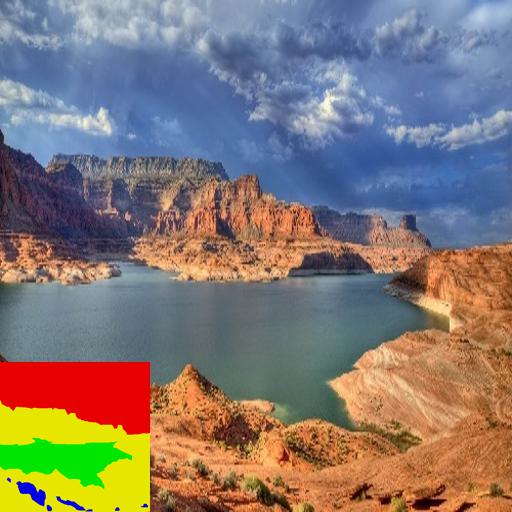}  & \includegraphics[width=0.14\linewidth]{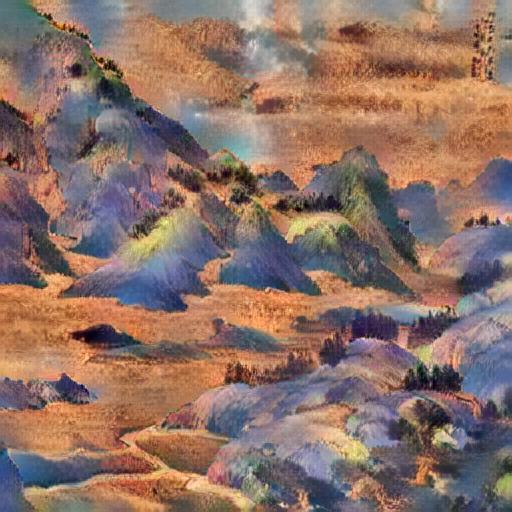}  &
\includegraphics[width=0.14\linewidth]{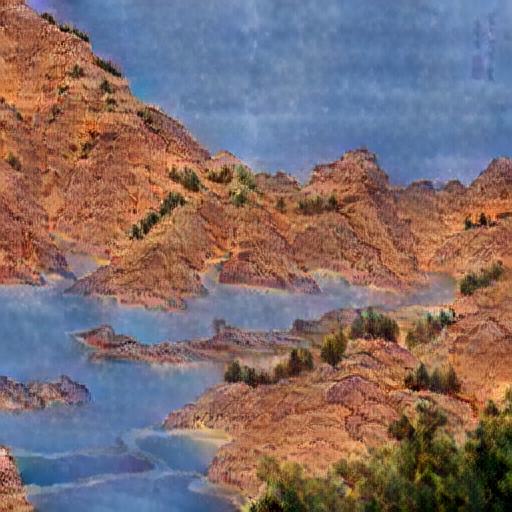}& \includegraphics[width=0.14\linewidth]{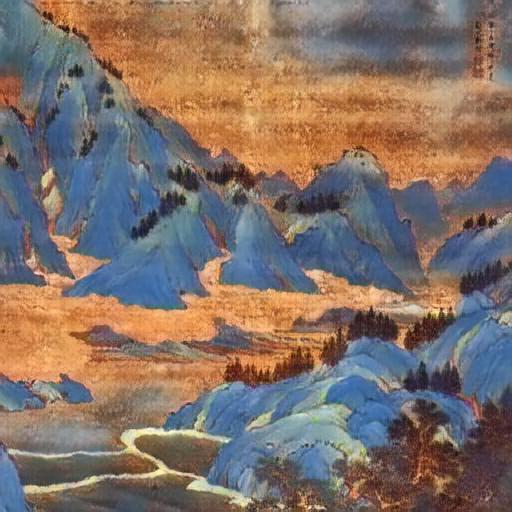} &
\includegraphics[width=0.14\linewidth]{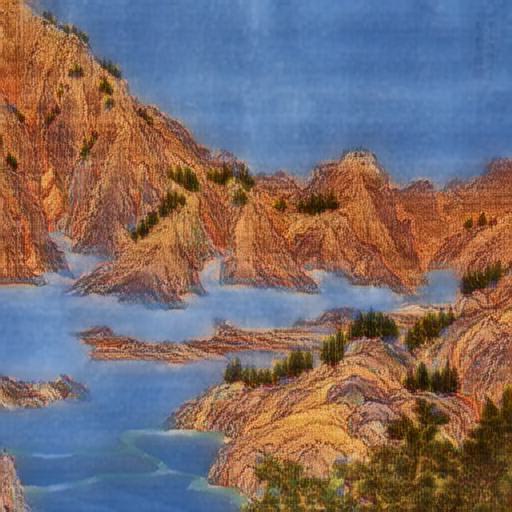} &  \includegraphics[width=0.14\linewidth]{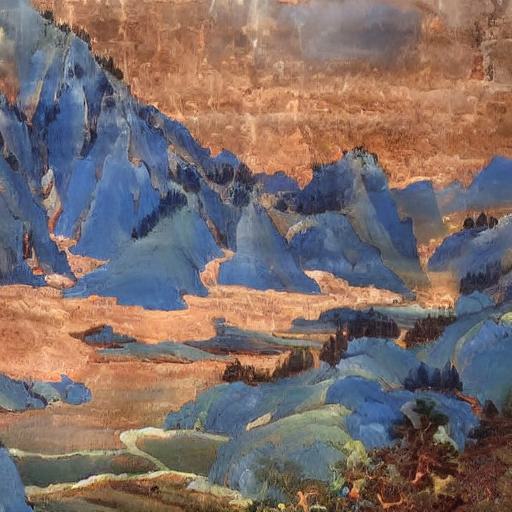} &  \includegraphics[width=0.14\linewidth]{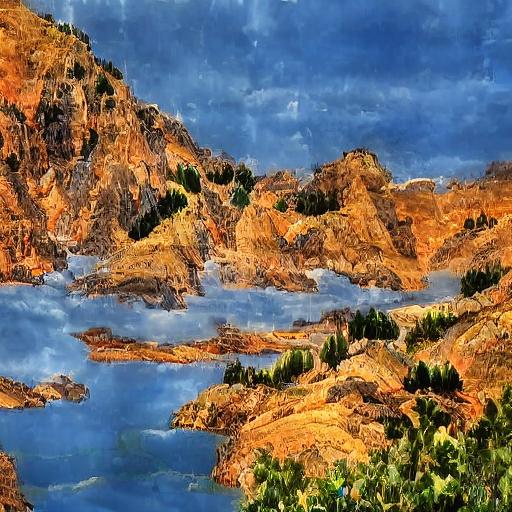} \\
& & & STROTSS & MAST & TR & DIA & GLStyleNet \\
& & & \includegraphics[width=0.14\linewidth]{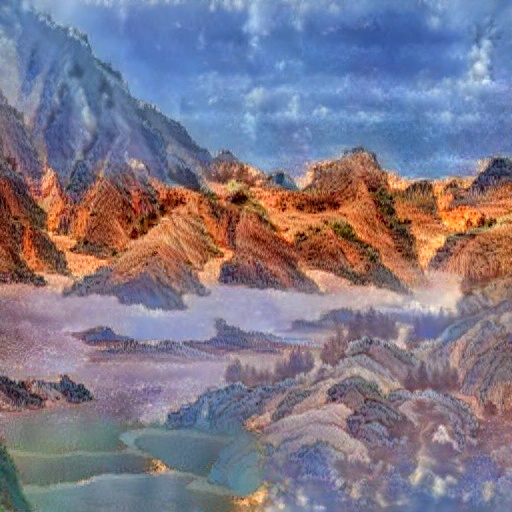} & \includegraphics[width=0.14\linewidth]{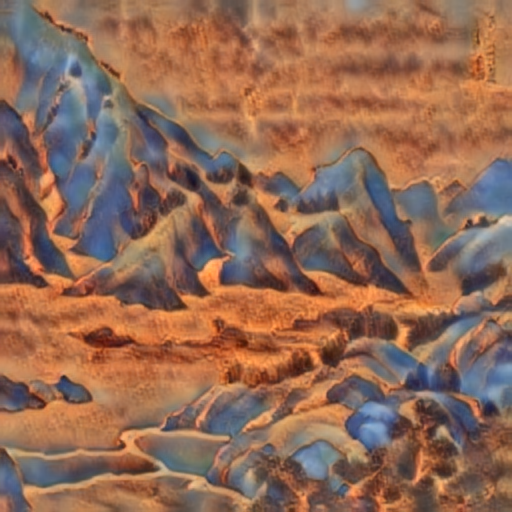} & 
\includegraphics[width=0.14\linewidth]{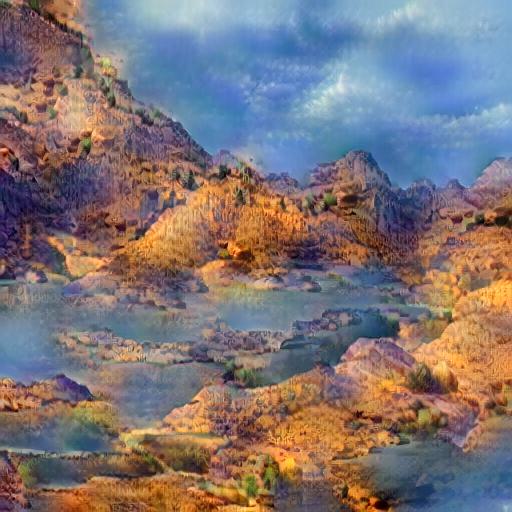} &
\includegraphics[width=0.14\linewidth]{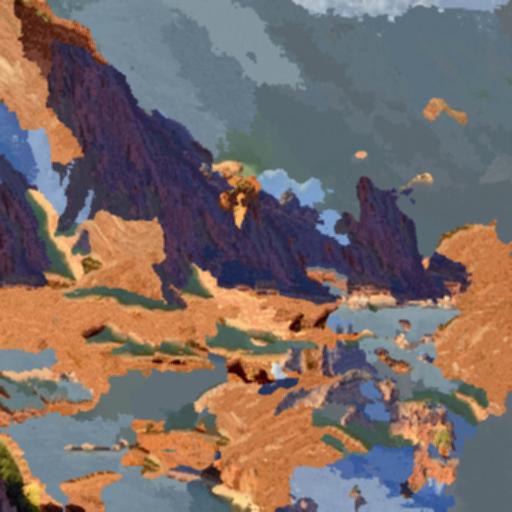} & 
\includegraphics[width=0.14\linewidth]{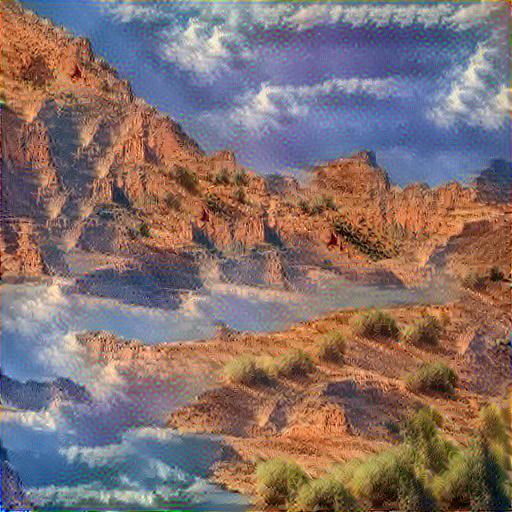} \\

\end{tabular}
}
\caption{Qualitative comparisons among Attn-AST approaches, those with SCSA, and SOTA methods.}
\label{fig:18}
\end{figure*}

\begin{figure*}
\centering
\resizebox{1.0\textwidth}{!}{
\setlength{\tabcolsep}{0.02cm} 
\renewcommand{\arraystretch}{1}  
\begin{tabular}{cccccccc}
 Content & Style & SANet & SANet + SCSA & StyTR$^2$ & StyTR$^2$ + SCSA & StyleID & StyleID + SCSA \\
\includegraphics[width=0.14\linewidth]{sm/img/29+sem.jpg} & \includegraphics[width=0.14\linewidth]{sm/img/29_paint+sem.jpg} & \includegraphics[width=0.14\linewidth]{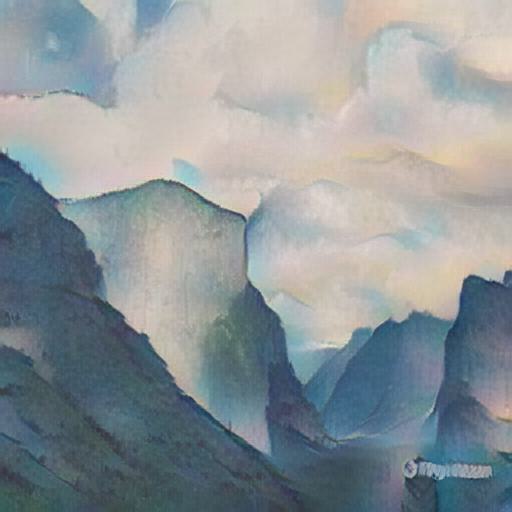}  &
\includegraphics[width=0.14\linewidth]{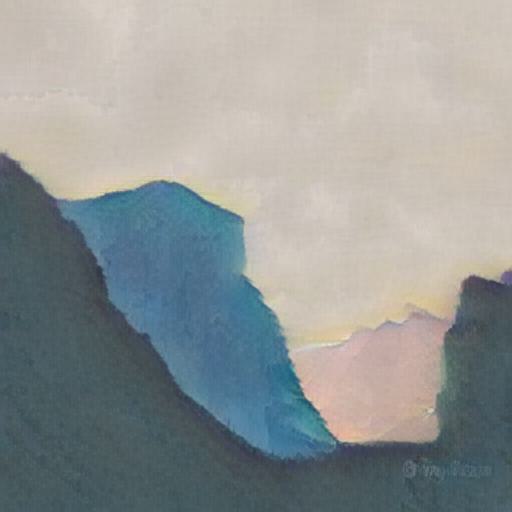}& \includegraphics[width=0.14\linewidth]{sm/img/29_29_paint_StyTR2.jpg} &
\includegraphics[width=0.14\linewidth]{sm/img/29_29_paint_StyTR2_sem.jpg} &  \includegraphics[width=0.14\linewidth]{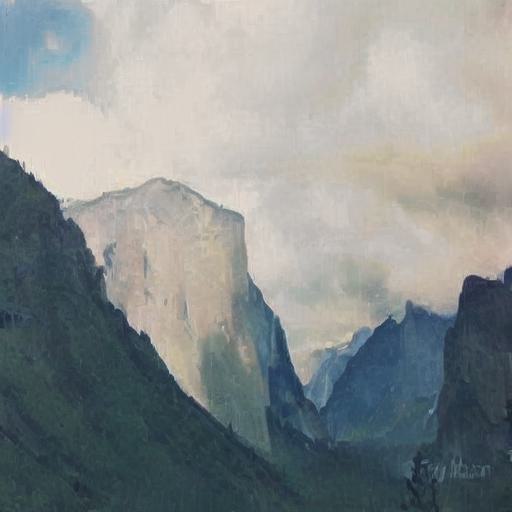} &  \includegraphics[width=0.14\linewidth]{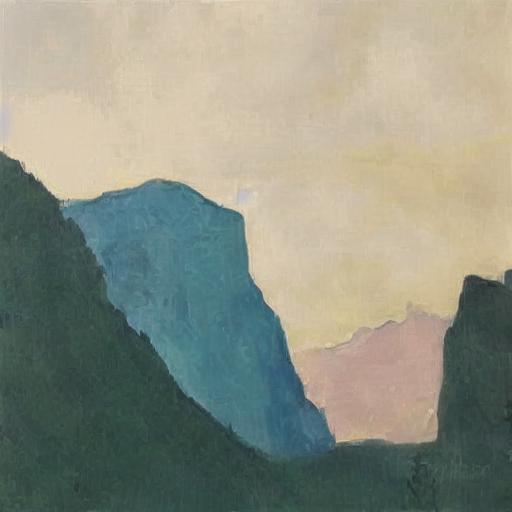} \\
& & & STROTSS & MAST & TR & DIA & GLStyleNet \\
& & & \includegraphics[width=0.14\linewidth]{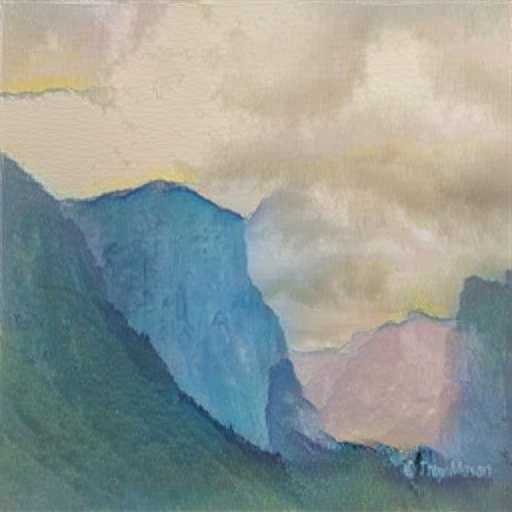} & \includegraphics[width=0.14\linewidth]{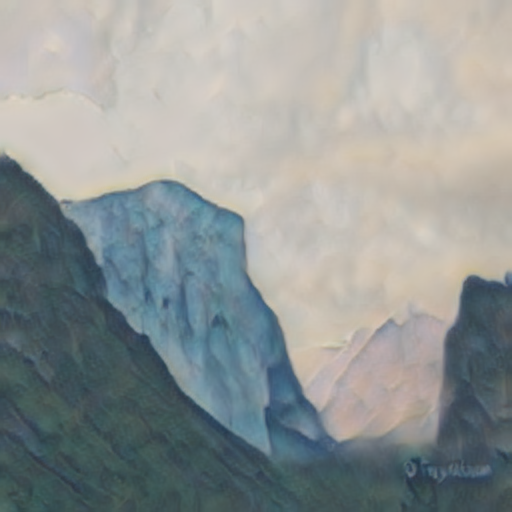} & 
\includegraphics[width=0.14\linewidth]{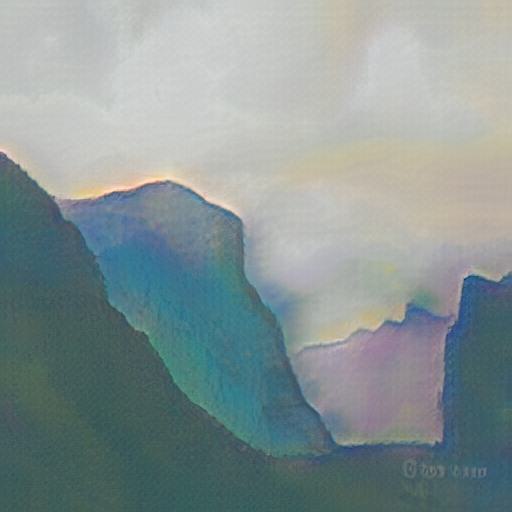} &
\includegraphics[width=0.14\linewidth]{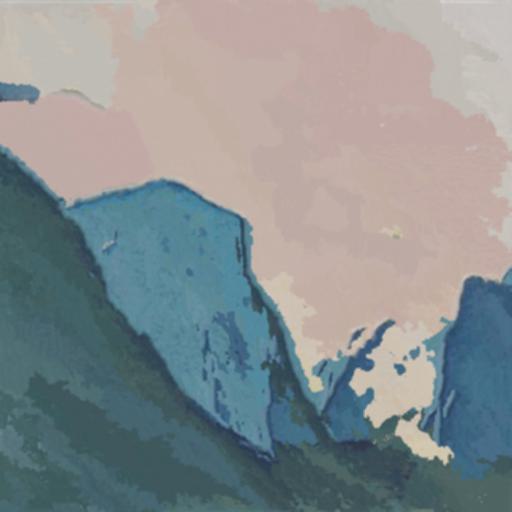} & 
\includegraphics[width=0.14\linewidth]{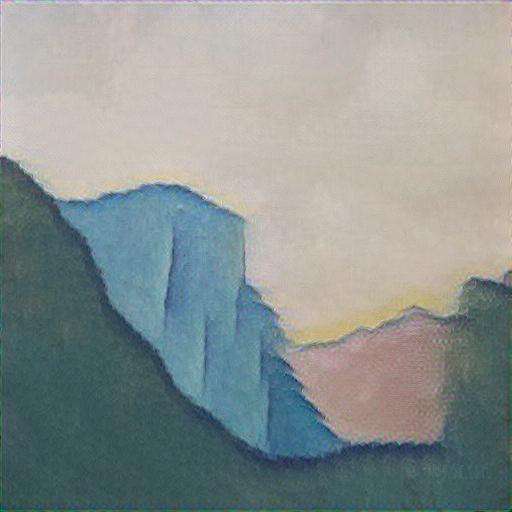} \\
& 
\\
 Content & Style & SANet & SANet + SCSA & StyTR$^2$ & StyTR$^2$ + SCSA & StyleID & StyleID + SCSA \\

\includegraphics[width=0.14\linewidth]{sm/img/30_paint+sem.jpg} & \includegraphics[width=0.14\linewidth]{sm/img/30+sem.jpg}  & \includegraphics[width=0.14\linewidth]{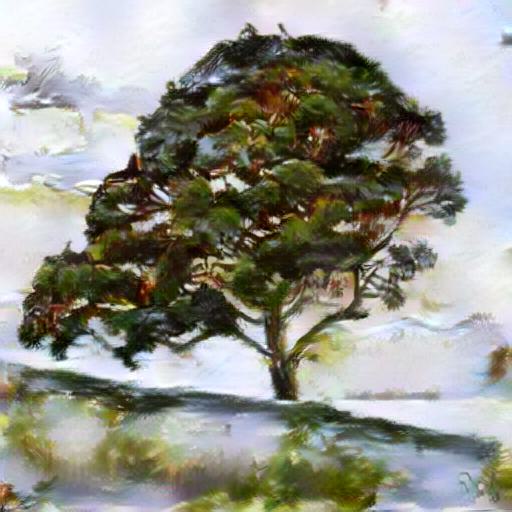}  &
\includegraphics[width=0.14\linewidth]{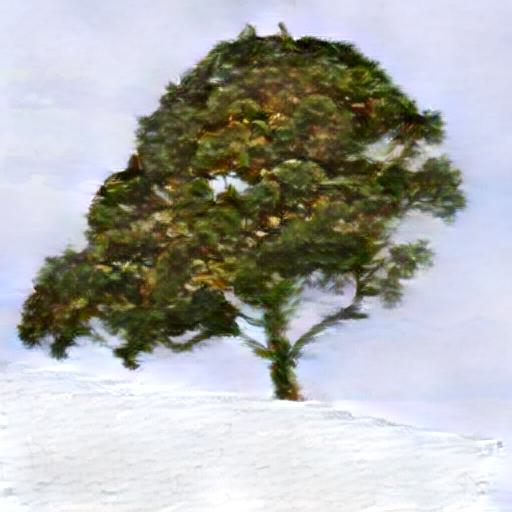}& \includegraphics[width=0.14\linewidth]{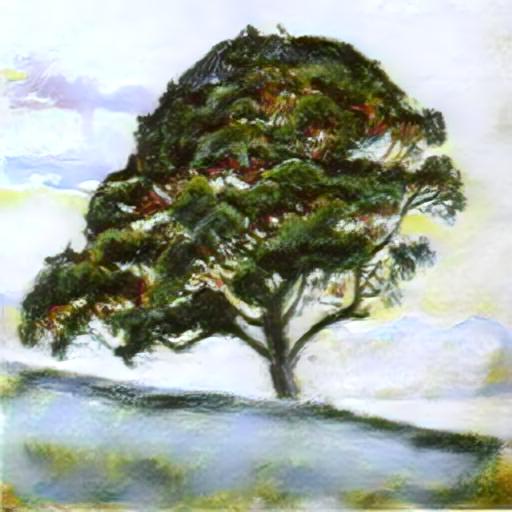} &
\includegraphics[width=0.14\linewidth]{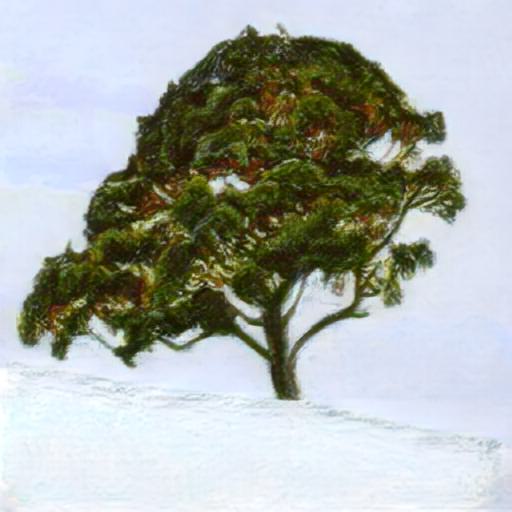} &  \includegraphics[width=0.14\linewidth]{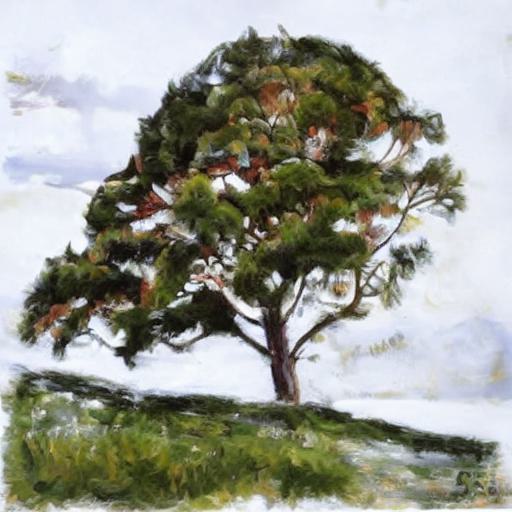} &  \includegraphics[width=0.14\linewidth]{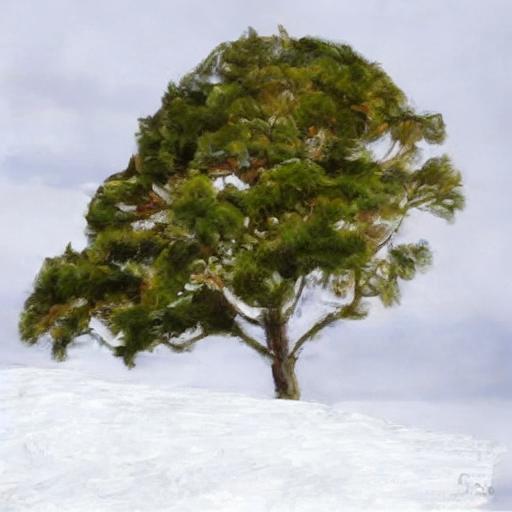} \\
& & & STROTSS & MAST & TR & DIA & GLStyleNet \\
& & & \includegraphics[width=0.14\linewidth]{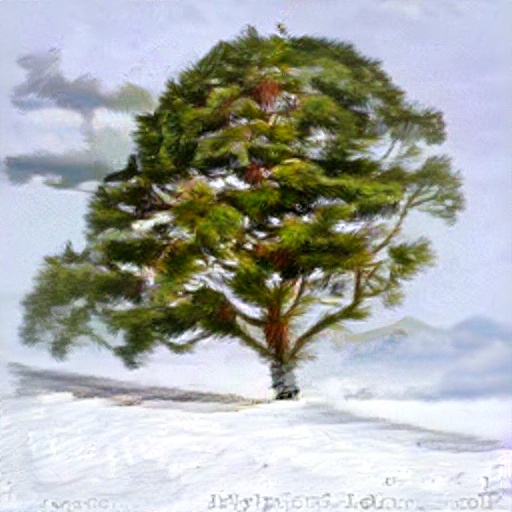} & \includegraphics[width=0.14\linewidth]{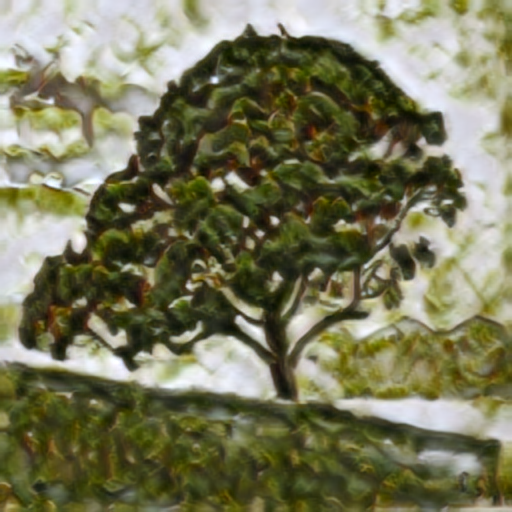} & 
\includegraphics[width=0.14\linewidth]{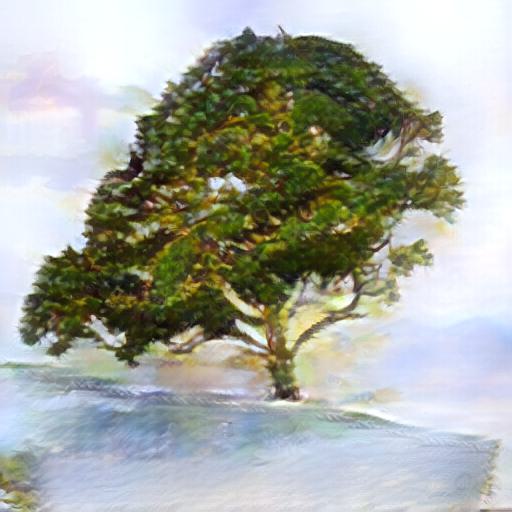} &
\includegraphics[width=0.14\linewidth]{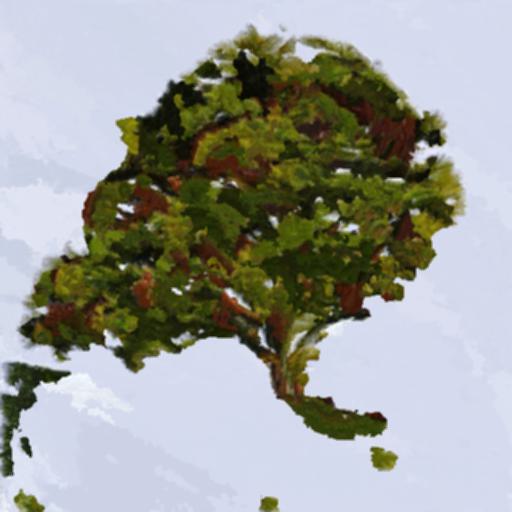} & 
\includegraphics[width=0.14\linewidth]{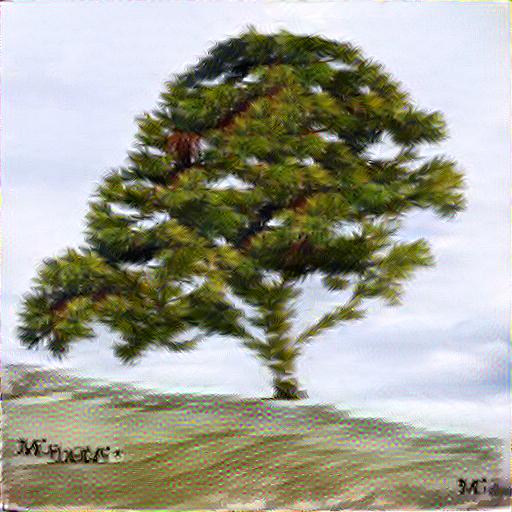} \\
&
\\
 Content & Style & SANet & SANet + SCSA & StyTR$^2$ & StyTR$^2$ + SCSA & StyleID & StyleID + SCSA \\

\includegraphics[width=0.14\linewidth]{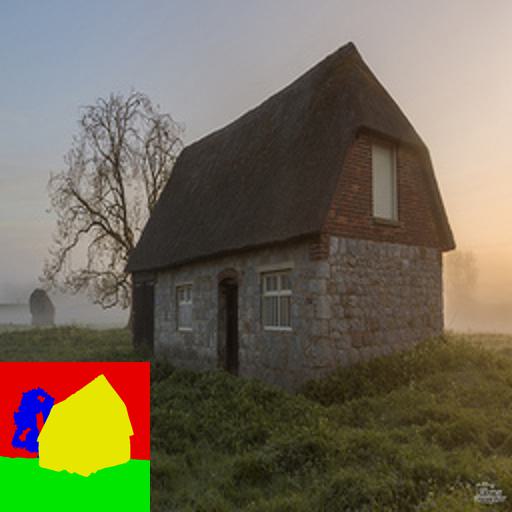} & \includegraphics[width=0.14\linewidth]{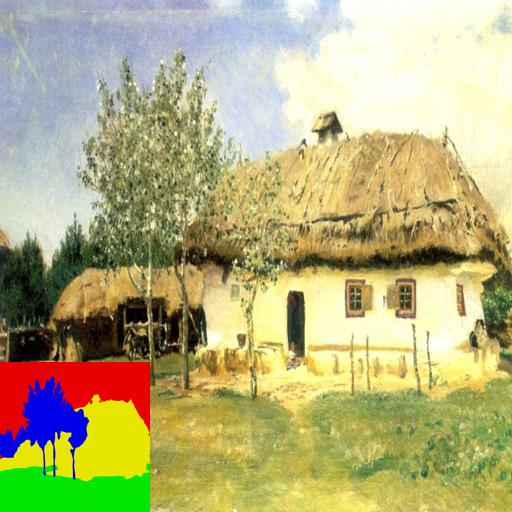}  & \includegraphics[width=0.14\linewidth]{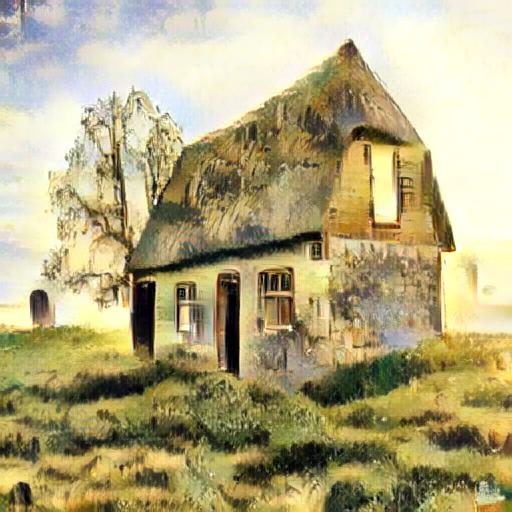}  &
\includegraphics[width=0.14\linewidth]{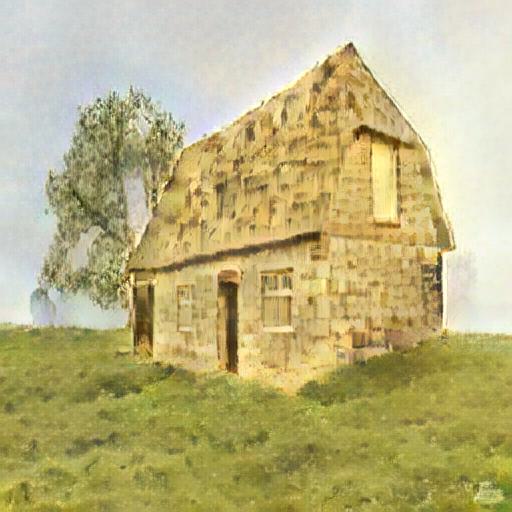}& \includegraphics[width=0.14\linewidth]{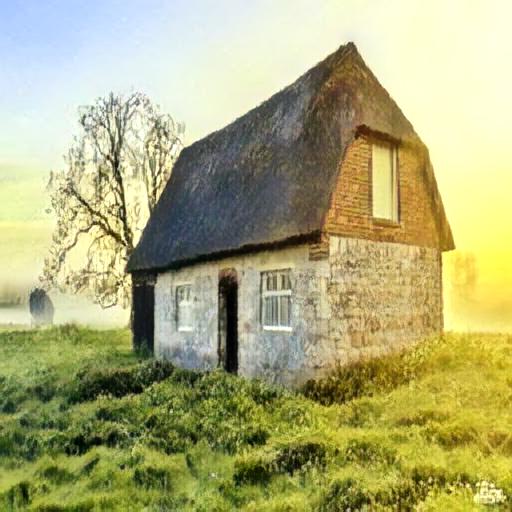} &
\includegraphics[width=0.14\linewidth]{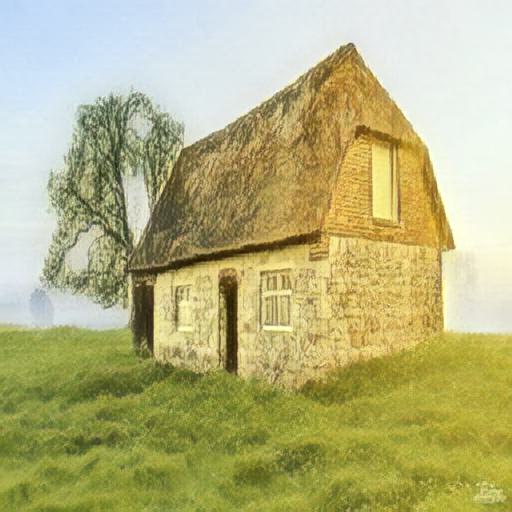} &  \includegraphics[width=0.14\linewidth]{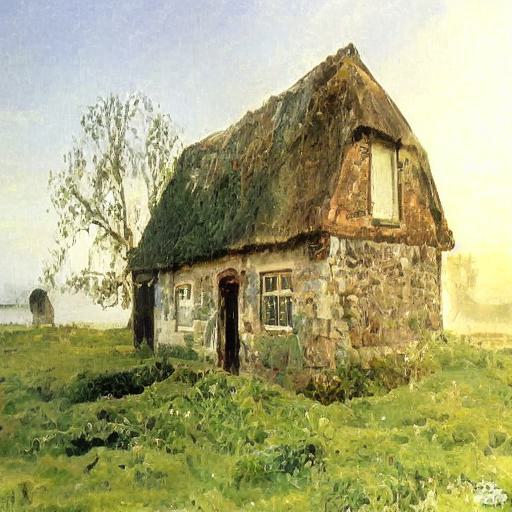} &  \includegraphics[width=0.14\linewidth]{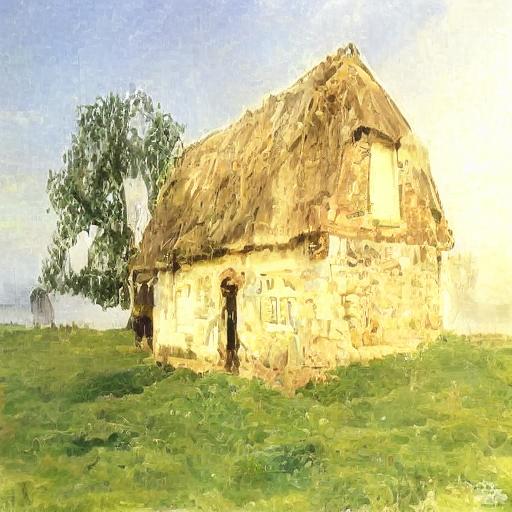} \\
& & & STROTSS & MAST & TR & DIA & GLStyleNet \\
& & & \includegraphics[width=0.14\linewidth]{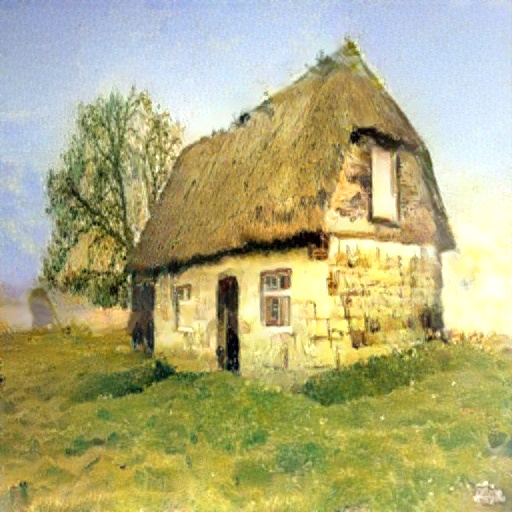} & \includegraphics[width=0.14\linewidth]{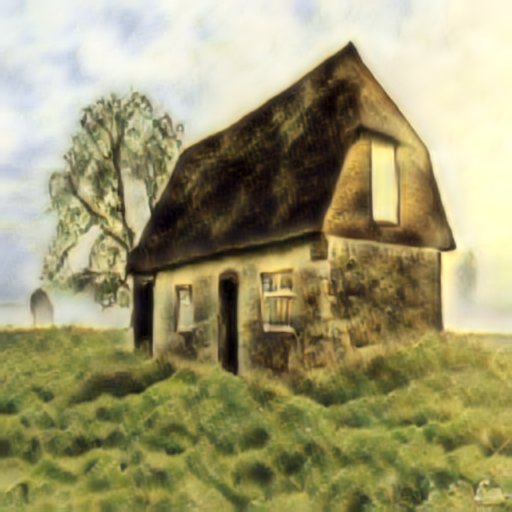} & 
\includegraphics[width=0.14\linewidth]{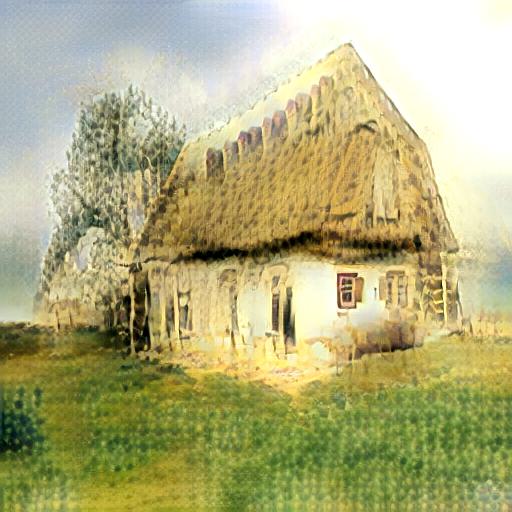} &
\includegraphics[width=0.14\linewidth]{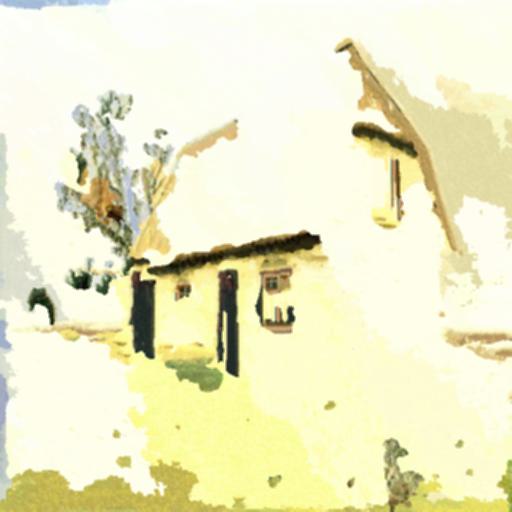} & 
\includegraphics[width=0.14\linewidth]{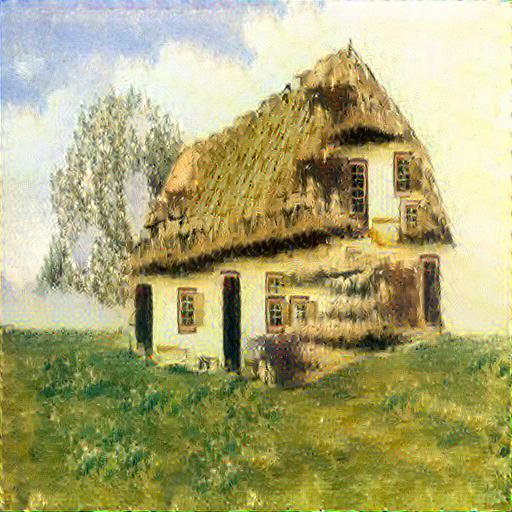} \\

&
\\
 Content & Style & SANet & SANet + SCSA & StyTR$^2$ & StyTR$^2$ + SCSA & StyleID & StyleID + SCSA \\

\includegraphics[width=0.14\linewidth]{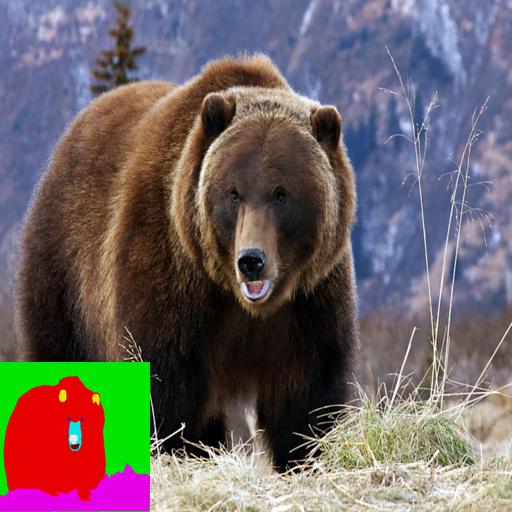} & \includegraphics[width=0.14\linewidth]{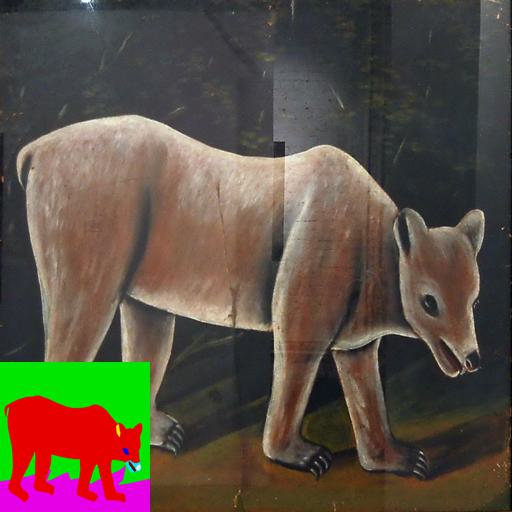}  & \includegraphics[width=0.14\linewidth]{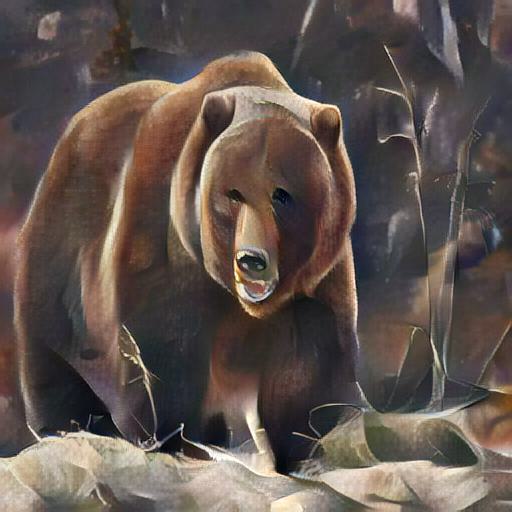}  &
\includegraphics[width=0.14\linewidth]{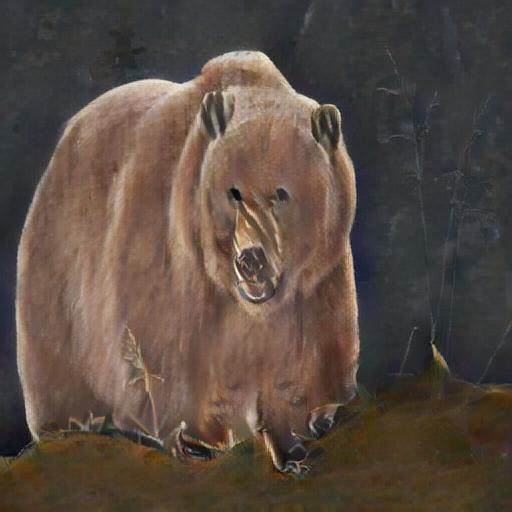}& \includegraphics[width=0.14\linewidth]{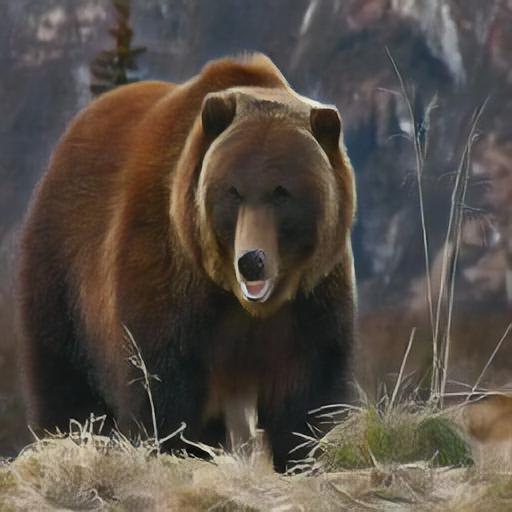} &
\includegraphics[width=0.14\linewidth]{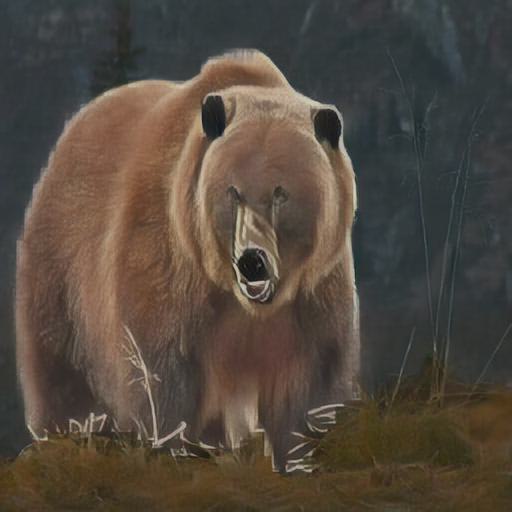} &  \includegraphics[width=0.14\linewidth]{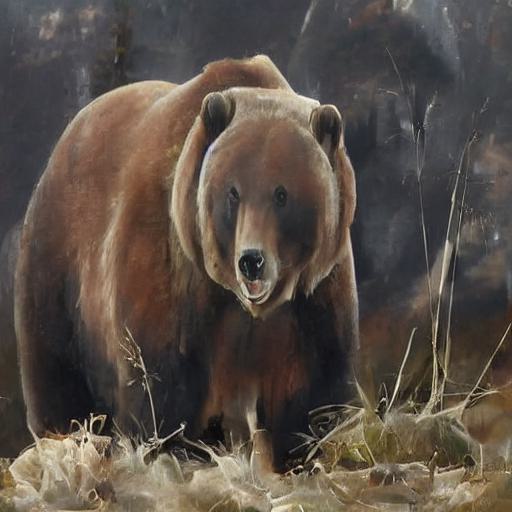} &  \includegraphics[width=0.14\linewidth]{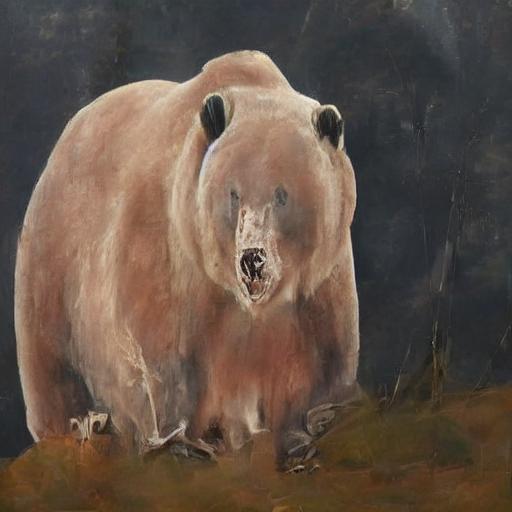} \\
& & & STROTSS & MAST & TR & DIA & GLStyleNet \\
& & & \includegraphics[width=0.14\linewidth]{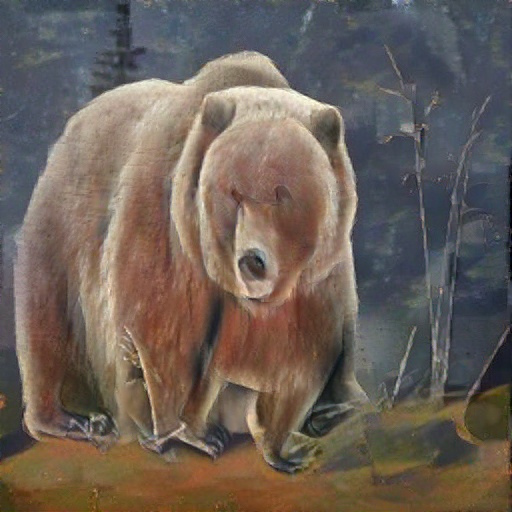} & \includegraphics[width=0.14\linewidth]{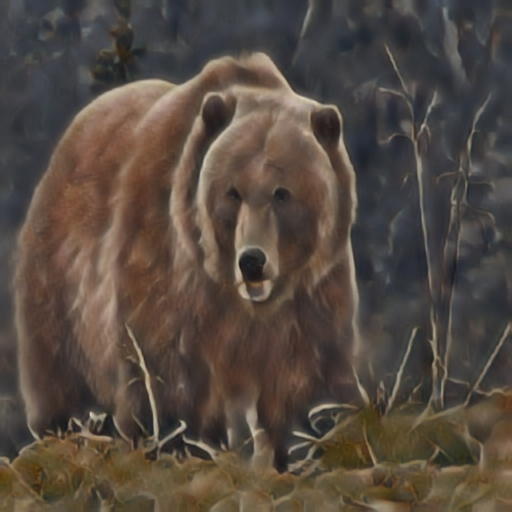} & 
\includegraphics[width=0.14\linewidth]{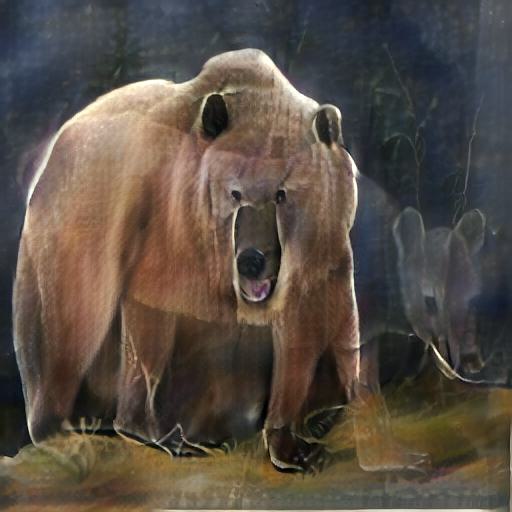} &
\includegraphics[width=0.14\linewidth]{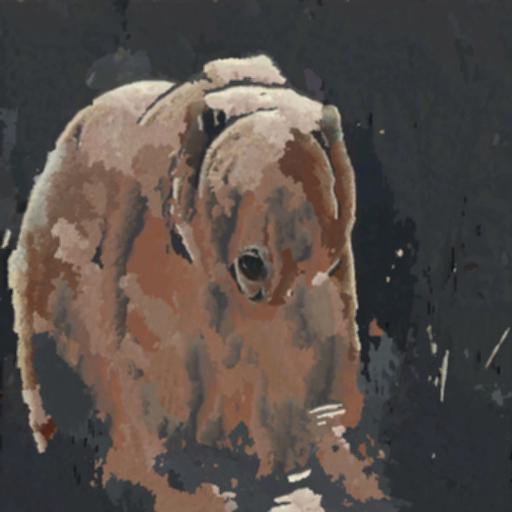} & 
\includegraphics[width=0.14\linewidth]{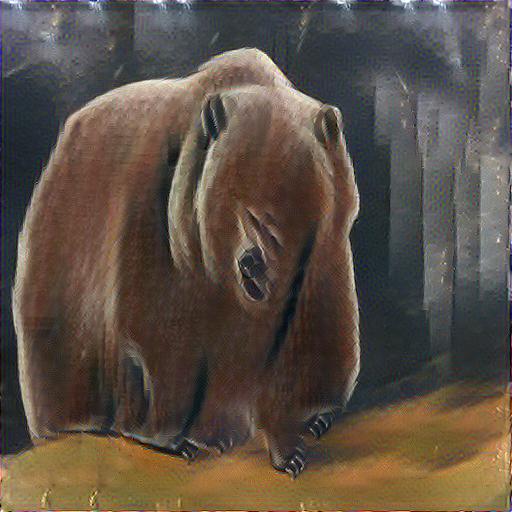} \\

\end{tabular}
}
\caption{Qualitative comparisons among Attn-AST approaches, those with SCSA, and SOTA methods.}
\label{fig:22}
\end{figure*}

As shown in Fig.~\ref{fig:13}, Fig.~\ref{fig:14}, Fig.~\ref{fig:15}, Fig.~\ref{fig:16}, Fig.~\ref{fig:17}, Fig.~\ref{fig:18}, and Fig.~\ref{fig:22}, our SCSA method seamlessly enhances existing arbitrary style transfer techniques, enabling versatile semantic style transfer with performance that exceeds current state-of-the-art methods in both intensity and stability. Specifically, the stylized images produced by DIA exhibit incomplete contents of the content images. The stylized images generated by TR,  STROTSS, and GLStyleNet sometimes incorporate content elements from the style image. MAST occasionally yields semantic stylization results that lack accuracy.

\end{document}